\documentclass[12pt, twoside]{book}
\usepackage[a4paper, includehead, headheight=0.6cm, inner=3cm, outer=2.5cm, top=2.5cm, bottom=2.5cm]{geometry}  

\usepackage[english]{babel} 
\usepackage[utf8]{inputenc} 
\usepackage[T1]{fontenc} 
\usepackage{titlesec}
\usepackage{color, soul}
\let\oldmathcal\mathcal

\usepackage{calrsfs}

\let\mathcal\oldmathcal

\usepackage{longtable}

\usepackage{caption} 
\usepackage{subcaption} 
\usepackage[table,dvipsnames]{xcolor}

\usepackage{float}
\usepackage{placeins}

\usepackage{microtype}
\usepackage{booktabs} 
\usepackage{tabularx} 
\usepackage{arydshln} 
\usepackage{multirow} 
\usepackage{wrapfig} 
\usepackage{capt-of}
\usepackage{comment}

\usepackage{appendix}

\definecolor{darkgreen}{RGB}{0, 100, 0}
\definecolor{lightergray}{gray}{0.95}

\usepackage{dsfont}

\usepackage{algorithm}
\usepackage{algorithmic}
\RequirePackage{algorithm}
\RequirePackage{algorithmic}

\usepackage{afterpage}

\usepackage{amsmath}
\usepackage{amssymb}

\usepackage{amsthm}

\usepackage{enumerate}
\usepackage{enumitem}

\usepackage{graphicx} 

\usepackage{longtable} 
\usepackage{fancyhdr} 
\usepackage[onehalfspacing]{setspace} 

\usepackage{mdframed}

\usepackage{enumitem} 

\usepackage{pdfpages} 

\usepackage{mathtools}

\usepackage{fancyhdr}
\usepackage{etoolbox} 

\usepackage{tocloft}
\usepackage{emptypage} 
\usepackage{eso-pic} 
\usepackage[hyperfootnotes=false]{hyperref} 
\usepackage[most]{tcolorbox}

\newcommand\blfootnote[1]{
    \begingroup
    \renewcommand\thefootnote{}\footnote{#1}
    \addtocounter{footnote}{-1}
    \endgroup
}

\AtBeginEnvironment{subappendices}{%
\chapter*{Appendix}
\addcontentsline{toc}{section}{Appendix}
\counterwithin{figure}{section}
\counterwithin{table}{section}
\counterwithin{equation}{section}
}

\AtEndEnvironment{subappendices}{%
}

\DeclareMathOperator{\argmin}{arg\,min} 

\newcommand{\name}{Konstantin Schürholt} 
\newcommand{\thesistitle}{Hyper-Representations: \\ Learning from Populations of Neural Networks} 
\newcommand{\supervisor}{Prof. Dr. Damian Borth} 
\newcommand{\cosupervisor}{Prof. Michael W. Mahoney, Ph.D.} 
\newcommand{\cosupervisorx}{Prof. Xavier Gir\'{o}-i-Nieto, Ph.D.} 

\usepackage[square, numbers, sort]{natbib} 

\usepackage{amsthm,amsmath,amssymb,amsfonts,bbm} 

\newcommand{\bea}{\begin{eqnarray}} 
\newcommand{\eea}{\end{eqnarray}}

\newtheorem{critere}{Criterion}

\begin{document}

\pagestyle{empty} 

\begin{titlepage}
\newgeometry{left=1cm, right=1cm, bottom=2.5cm, top=3.0cm}
\begin{center}
\vfill
{
    { \par \ }\\[-1.0cm]
	{\large \bfseries \MakeUppercase{\thesistitle} \par \ }\\[1.0cm]
 	{\large DISSERTATION} \\ [0.1cm]
    of the University of St.Gallen, \\
    School of Management, \\ 
    Economics, Law, Social Sciences, \\ 
    International Affairs and Computer Science, \\
    to obtain the title of \\
    Doctor of Philosophy in Computer Science
	\hfill \\[1cm]
	submitted by\\[1.0cm]
	{\bfseries\name}\\ [0.5cm]
    from \\ [0.5cm]
    Germany \\ [0.5cm]
	\hfill \\ [-0.0cm]
    Approved on the application of \\ [0.5cm]
	{
    	\vfill \textbf{\supervisor}\\ [0.5cm]
        and \\ 
    	\vfill \textbf{\cosupervisor} \\ [0.5cm]
    	\vfill \textbf{\cosupervisorx}
    } \\ 
	\vspace{1cm}
	\vfill
    Dissertation no. 5459 \\
    \vspace{1.0cm}
    Digitaldruckhaus GmbH, Konstanz 2024
}
\end{center}
\restoregeometry
\end{titlepage}
\makeatletter
\renewcommand{\chaptermark}[1]{\markboth{#1}{}}
\makeatother

\pagestyle{fancy} 
\renewcommand{\chaptermark}[1]{\markboth{#1}{}} 
\renewcommand{\sectionmark}[1]{\markright{\thesection\; #1}} 
\fancyhf{} 
\fancyhead[LE]{\nouppercase{\leftmark}}
\fancyhead[RO]{\nouppercase{\rightmark}}
\fancyfoot[C]{\thepage}
\renewcommand{\headrulewidth}{0.4pt}
\renewcommand{\footrulewidth}{0pt}

\setcounter{chapter}{0}

\fancypagestyle{plain}{%
  \fancyhf{} 
  \fancyfoot[C]{\thepage} 
  \renewcommand{\headrulewidth}{0pt} 
  \renewcommand{\footrulewidth}{0pt} 
}

\fancypagestyle{empty}{%
  \fancyhf{}%
  \fancyhead[LO]{\nouppercase{\leftmark}}
  \fancyhead[RE]{\nouppercase{\rightmark}}
  \fancyfoot[C]{\thepage} 
  \renewcommand{\headrulewidth}{0.5pt}
  \renewcommand{\footrulewidth}{0pt}
}

\fancypagestyle{emptythesis}{%
  \fancyhf{} 
  \fancyfoot[C]{\thepage} 
  \renewcommand{\headrulewidth}{0pt} 
  \renewcommand{\footrulewidth}{0pt} 
}

\fancypagestyle{emptythesis}{%
  \fancyhf{} 
  \fancyfoot[C]{\thepage} 
  \renewcommand{\headrulewidth}{0.5pt} 
  \renewcommand{\footrulewidth}{0pt} 
}

\fancypagestyle{emptythesistransition}{%
  \fancyhf{} 
  \fancyfoot[C]{xv} 
  \renewcommand{\headrulewidth}{0.5pt} 
  \renewcommand{\footrulewidth}{0pt} 
}
\fancypagestyle{emptychaptertransition}{%
  \fancyhf{} 
  \fancyfoot[C]{\thepage} 
  \renewcommand{\headrulewidth}{0.5pt} 
  \renewcommand{\footrulewidth}{0pt} 
}

\fancypagestyle{emptyempty}{%
  \fancyhf{} 
  \renewcommand{\headrulewidth}{0pt} 
  \renewcommand{\footrulewidth}{0pt} 
}


\patchcmd{\chapter}{\thispagestyle{plain}}{\thispagestyle{plain}}{}{}




\addtocontents{toc}{\vspace{-0.5cm}} 
\renewcommand{\contentsname}{Table of Contents}

\raggedbottom

\pagenumbering{roman}

\afterpage{
    \thispagestyle{emptythesis}
    \vspace*{\fill}
    \newpage
    \thispagestyle{plain}
}

\clearpage

{\small

\noindent The University of St.Gallen, School of Management, Economics, Law, Social Sciences, \\ International Affairs, and Computer Science, hereby consents to the printing of the present dissertation, without hereby expressing any opinion on the views herein expressed.

\vspace{1cm}

\noindent St.Gallen, May 22\textsuperscript{th}, 2024

\vspace{1cm}

\begin{flushright}
  \parbox{6.0cm}{
    The President: \hspace{2cm} \\ \\ \\ \\
    Prof. Dr. Manuel Ammann
  }
\end{flushright}

}

\newpage

\afterpage{
    \thispagestyle{emptythesis}
    \vspace*{\fill}
    \newpage
    \thispagestyle{plain}
}

\linepenalty=5000
\linepenalty=5000

\clearpage

\subsection*{}
\thispagestyle{plain}

{
\vspace{3cm}
\centering Für Leonie. 





}


\newpage

\afterpage{
    \thispagestyle{emptythesis}
    \vspace*{\fill}
    \newpage
    \thispagestyle{plain}
}

\clearpage

\subsection*{Acknowledgements}
\thispagestyle{plain}

\vspace{8pt}

{\footnotesize
\hspace*{0.5cm} I would like to express my deepest gratitude for the invaluable contributions and support I received throughout my PhD journey.

\vspace{4pt}

Firstly, my heartfelt thanks to Professor Damian Borth for his guidance and belief in this project from its inception. His support, countless hours spent discussing and refining research questions, and deep commitment to the project have been pivotal for my career. His mentorship provided me with invaluable opportunities to work in machine learning, explore new research avenues, and gain international experience, which significantly broadened my academic horizons. I am profoundly grateful for his role in opening doors that have shaped my professional journey.

\vspace{4pt}

I am deeply grateful to my co-advisor, Professor Xavier Giro-i-Nieto, for his openness to new ideas, invaluable guidance, and critical feedback at a crucial stage in my thesis. His insights and support were instrumental in shaping the direction of my research.

\vspace{4pt}

My sincere thanks go to my co-advisor, Professor Michael Mahoney, for the opportunity to visit his lab at Berkeley. Immersing myself in the vibrant research culture there was inspiring and transformative. Our collaboration provided me with new perspectives and significantly enriched my work. His critical questions and fresh insights were invaluable in refining my research.

\vspace{4pt}

I am also thankful to Professor Michael Mommert for his academic and moral support, as well as his ability to bring humor and lightness to our interactions. His balanced approach to work and recreation, including our bouldering sessions, was refreshing and supportive.

\vspace{4pt}

I appreciate the companionship and supportive environment created by my colleagues Alex Bogun, Marco Schreyer, Shijun Wang, Hamed Hemati, Joelle Hanna, Linus Scheibenreif, and Leo Meynent. Their contributions to our many discussions, their academic and personal support, were invaluable.

\vspace{4pt}

I would like to acknowledge the contributions of the students I got to work with: Pol Caselles Rico, Diyar Taskiran, Thomas Fey, Dominik Honegger, Kris Reynisson, Alex Lontke, Julius Lautz, and Damian Falk. Their motivation, hard work, and curiosity significantly contributed to the research in this thesis and to my growth as an academic.

\vspace{4pt}

My thanks go to Martin Eigenmann, Marcel Cahenzli, and Angelika Graefingholt, who enabled me to pursue this research. Martin’s patience and IT support were indispensable to my work, and the technical assistance and knowledge I gained from him were invaluable. Marcel and Angelika provided crucial administrative support, helping me navigate the rules and regulations that allowed me to pursue my PhD efficiently and graduate on time. 

\vspace{4pt}

A special thanks to Christian Nauck and Carlos Iglesias-Aguirre for their enduring friendship, patience in engaging in any academic discussion, and willingness to ask thought-provoking questions that broadened my research horizons.

\vspace{4pt}

Finally, a large debt of gratitude goes to my partner, Leonie, for her patience, unwavering support, and belief in me. Her encouragement was a constant source of strength and motivation, enabling me to persevere through the challenges of this journey.

\vspace{24pt}

Saint Gallen in June 2024 \hspace{7.0cm} Konstantin Schürholt

}

\newpage

\afterpage{
    \thispagestyle{emptythesis}
    \vspace*{\fill}
    \newpage
    \thispagestyle{plain}
}

\clearpage

\tableofcontents

\newpage

\afterpage{
    \thispagestyle{emptythesis}
    \vspace*{\fill}
    \newpage
    \thispagestyle{plain}
}

\clearpage

\listoffigures
\addcontentsline{toc}{chapter}{List of Figures}

\newpage

\clearpage

\listoftables
\addcontentsline{toc}{chapter}{List of Tables}

\newpage

\clearpage

\subsection*{Abstract}
\thispagestyle{plain}
\addcontentsline{toc}{chapter}{Abstract}


Neural Networks (NNs) have undergone a remarkable evolution, transitioning from academic labs to key technologies across various domains. This proliferation underscores their capability and versatility. As these models become integral to critical decision-making processes, the demand for methods to understand their inner workings likewise becomes more pronounced. This thesis addresses the challenge of understanding NNs through the lens of their most fundamental component: the weights, which encapsulate the learned information and determine the model behavior.

The NN weight space contains complex local and global structures which makes it a challenging domain. Addressing these challenges, this thesis develops innovative representation learning methods for the domain of weight spaces. The proposed methods embed and disentangle model weights in a representation space. The representation space allows not only to analyze existing models but also to generate new models with specified characteristics. Such an analysis builds on populations of models, to develop a nuanced understanding of the structure of NN weights.

At the core of this thesis is a fundamental question: Can we learn general, task-agnostic representations from populations of Neural Network models? The key contribution of this thesis to answer that question are \textit{hyper-representations}, a self-supervised method to learn representations of NN weights. Work in this thesis finds that trained NN models indeed occupy meaningful structures in the weight space, that can be learned and used. Through extensive experiments, this thesis demonstrates that \textit{hyper-representations} uncover model properties, such as their performance, state of training, or hyperparameters.

Moreover, the identification of regions with specific properties in \textit{hyper-representation} space allows to sample and generate model weights with targeted properties. This thesis demonstrates applications for fine-tuning, and transfer learning to great success. Lastly, it presents methods that allow \textit{hyper-representations} to generalize beyond model sizes, architectures, and tasks. The practical implications of that are profound, as it opens the door to foundation models of Neural Networks, which aggregate and instantiate their knowledge across models and architectures. 

Ultimately, this thesis contributes to the deeper understanding of Neural Networks by investigating structures in their weights which leads to more interpretable, efficient, and adaptable models. By laying the groundwork for representation learning of NN weights, this research demonstrates the potential to change the way Neural Networks are developed, analyzed, and used.

\newpage

\afterpage{
    \thispagestyle{emptythesis}
    \vspace*{\fill}
    \newpage
    \thispagestyle{plain}
}

\clearpage

\subsection*{Zusammenfassung}
\thispagestyle{plain}

\addcontentsline{toc}{chapter}{Zusammenfassung}

Neuronale Netze (NNs) haben eine bemerkenswerte Evolution durchlaufen, von akademischen Laboren zu Schlüsseltechnologien in verschiedenen Bereichen. Diese Ausbreitung unterstreicht ihre Vielseitigkeit und Fähigkeiten. Da diese Modelle integraler Bestandteil kritischer Prozesse werden, wird auch die Nachfrage nach Methoden, ihre inneren Abläufe zu verstehen, immer ausgeprägter. Diese Dissertation adressiert die Herausforderung, NNs durch die Linse ihrer grundlegendsten Komponente zu verstehen: die Gewichte, welche die gelernte Information beinhalten und das Modellverhalten bestimmen.

Angesichts der hohen Informationsmenge und der herausfordernden Struktur der NN-Gewichte entwickelt diese Arbeit innovative Methoden des Repräsentationslernens für diesen Domäne. Diese Methoden betten Modellgewichte in einen aussagekräftigen Repräsentationsraum ein. Solche Repräsentationen ermöglichen nicht nur die Analyse bestehender Modelle, sondern auch die Generierung neuer Modelle mit spezifizierten Eigenschaften. Eine solche Analyse baut auf Populationen von Modellen auf, um ein nuanciertes Verständnis der Gewichtsstruktur zu entwickeln.

Im Kern dieser Dissertation steht eine grundlegende Frage: Können allgemeine, aufgabenagnostische Repräsentationen aus Populationen von neuronalen Netzwerkmodellen gelernt werden? Der Schlüsselbeitrag dieser Dissertation zur Beantwortung dieser Frage sind \textit{hyper-representations}, eine selbstüberwachte Methode, um Strukturen innerhalb der NN-Gewichte zu lernen. Beiträge in dieser Dissertation finden heraus, dass trainierte NN-Modelle tatsächlich bedeutungsvolle Strukturen im Gewichtsraum besetzen, die gelernt und genutzt werden können. Durch umfangreiche Experimente demonstriert diese Dissertation, dass \textit{hyper-representations} Modellcharakteristiken, wie ihre Leistung, ihren Lernfortschritt oder Hyperparameter, aufdecken.

Darüber hinaus ermöglicht die Identifikation von Regionen mit spezifischen Eigenschaften im \textit{hyper-representation} Raum das Generieren von Modellgewichten mit gezielten Eigenschaften. Diese Dissertation zeigt erfolgreiche Anwendungen für Feinabstimmung und Transferlernen auf. Zuletzt präsentiert sie Methoden, die es \textit{hyper-representations} ermöglichen, über Modellgrössen, Architekturen und Aufgaben hinaus zu generalisieren. Dadurch sind erstmals Grundlagenmodellen von NNs möglich, die Wissen über Modelle und Architekturen aggregieren und instanziieren können.

Letztendlich trägt diese Dissertation zum tieferen Verständnis Neuronaler Netze bei, indem sie Strukturen in ihren Gewichten untersucht, was zu interpretierbareren, effizienteren und anpassungsfähigeren Modellen führt. Indem sie die Grundlage für das Repräsentationslernen von NN-Gewichten legt, zeigt diese Forschung das Potenzial, die Art zu verändern, wie Neuronale Netze entwickelt, analysiert und genutzt werden.

\newpage

\afterpage{
    \thispagestyle{emptythesistransition}
    \vspace*{\fill}
    \newpage
    \thispagestyle{plain}
}


\pagenumbering{arabic}
\setcounter{page}{0}

\let\oldrightmark\rightmark
\let\oldleftmark\leftmark
\let\oldtitleformat\titleformat

\renewcommand{\rightmark}{\oldrightmark}
\renewcommand{\leftmark}{\oldleftmark}

\chapter{Introduction}
\label{chap::introduction}



Over the past years, Neural Networks (NNs) have transitioned from experimental tools in laboratory environments to cornerstone technologies in production systems across the globe. Their applications range from enabling autonomous vehicles to detect and navigate through complex environments to curating personalized news feeds in traditional and social media platforms. NNs can digest and generate content across various mediums—text, audio, images, and video—and even determine insurance policies. As NNs become increasingly integrated into applications with significant societal impact, the demand for their trustworthiness and safety becomes paramount.\looseness-1

Trustworthiness in NNs is multifaceted. It is not solely about achieving high performance and making accurate decisions in challenging conditions. Trustworthiness also hinges on transparency and explainability. A core part of this is understanding how and why decisions are made, especially in high-stakes applications. Furthermore, accountability is integral to trust. This requires implementing robust mechanisms for identity and version control of models, ensuring that modifications and deployments are traceable and justifiable. Fundamentally, achieving this level of trust, transparency, and accountability necessitates a profound understanding of NN models themselves.

\begin{figure}[h!]
\vspace{-18pt}
\centering
\includegraphics[width=0.6\columnwidth]{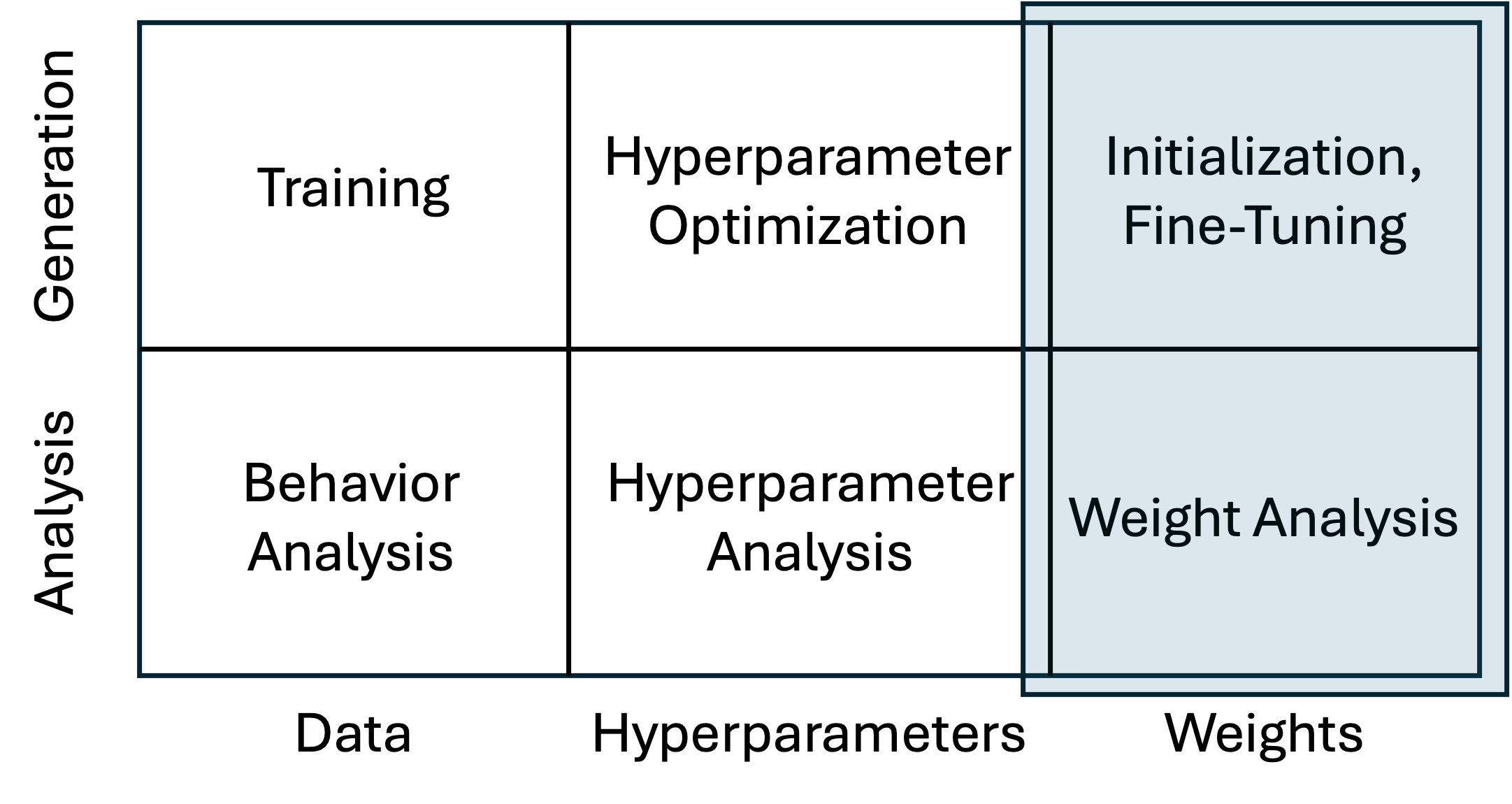}
\caption[Schematic Overview of the Neural Network Application Landscape]{Overview of the landscape spanned by applications for and perspectives on Neural Network models. Due to the rich information and application potential, the focus of this thesis lies on NN weights for model analysis and generation.}
\label{fig:landscape}
\vspace{-8pt}
\end{figure}

Consequently, there is a need for technical solutions to understand the inner workings of NN models. An improved understanding also helps guide NN training and thus improve the performance of models. 
On a high level, improved NN model understanding can serve two purposes: \textbf{(i)} model analysis and \textbf{(ii)} model generation. 
These two purposes make up one axis on the landscape considered for this thesis, see Figure \ref{fig:landscape}.
For both, different perspectives can be considered, which make up the other axis of the landscape. Along the machine learning (ML) pipeline, three elements can be used to operate on models: \textbf{(a)} their behavior when confronted with data; \textbf{(b)} their generating factors, i.e., their hyperparameters, as proxies for model behavior; and \textbf{(c)} the trainable weights of NN models. \\

Data is used for model generation in regular gradient-based training to minimize the loss. More complex hyper-networks generate weights but likewise use gradient signals from data~\citep{haHyperNetworks2017}. Data-based analysis is realized by evaluating model behavior on holdout and test data. The advantage of using data is that it is close to real-world usage. On the other hand, it can be limited by the availability and expressiveness of data as well as suitable evaluation metrics. 
Hyperparameter analysis predicts model performance based on hyperparameters. For model generation, hyperparameters are used as inputs in hyperparameter optimization\citep{hutterAutomatedMachineLearning2019}. These techniques implicitly connect hyperparameters with model behavior for both analysis and generation. While widely used, the connection between hyperparameters and behavior is indirect and nonlinear, which makes modeling and using it a challenging task.
Lastly, weight initialization, fine-tuning, or transfer learning from pre-trained models are an inherent part of NN training. 
The weights can also be used directly as inputs for model analysis \citep{unterthinerPredictingNeuralNetwork2020,eilertsenClassifyingClassifierDissecting2020,martinPredictingTrendsQuality2021} or model generation \citep{wortsmanModelSoupsAveraging2022,ainsworthGitReBasinMerging2022}. NN model weights are the immediate outcome of NN training, and as such determine model behavior. However, they also encode training information like model accuracy, epoch, and hyperparameters. 
Therefore, this thesis focuses on using weights of trained NNs for both model analysis and model generation. 

The relation between behavior and weights has complex local and global structures.
To make statements beyond one model and improve understanding of NN models more broadly therefore requires sufficient coverage of these structures. Hence, research in this domain requires analyzing not just single models, but a broad spectrum of trained NNs. 
By varying the generating factors such as hyperparameters, datasets, and architectures and training multiple models, diverse populations of trained NN models can be generated. 
Studying diverse populations of models allows the findings to generalize to real-world models. Therefore, the scope of this thesis is on model populations.\looseness-1

\newpage
Prior to the work on this thesis, a unified, task-agnostic representation learning method on populations of NN weights that is suitable for both model analysis and model generation did not exist.\\
\vspace{12pt}

\noindent
\begin{tcolorbox}[colback=gray!10,
                  colframe=black,
                  width=1.00\linewidth,
                  arc=0mm, auto outer arc,
                  boxrule=2pt,
                  box align=center,
                  halign=center,
                  valign=center,
                  enhanced,
                 ]
    \vspace{4pt}
    \textit{\textbf{The goal of this thesis is therefore to develop suitable representation learning methods for Neural Network weight spaces, which can be used to analyze models in a model representation space, as well as generate new weights with targeted properties.}}
    \vspace{4pt}
\end{tcolorbox}
\vspace{12pt}

%
The following paragraphs discuss the complexities in NN training, structures, and challenges in NN weight spaces to identify the research gap and research questions. Subsequently, the contribution of this thesis to address these questions is summarized.
\vspace{12pt}

\subsection*{Neural Network Weight Spaces}
\paragraph*{The Unreasonable Success of Neural Networks}
Over the past decade, Neural Networks (NNs) have been improved tremendously, and are the state of the art across many domains, such as computer vision \citep{heDeepResidualLearning2016,dosovitskiyImageWorth16x162021}, natural language processing \citep{devlinBERTPretrainingDeep2018,brownLanguageModelsAre2020,touvronLlamaOpenFoundation2023}, text-to-speech systems \citep{radfordRobustSpeechRecognition2022}, or even video generation \citep{videoworldsimulators2024}.
The success of NNs is impressive, considering that the training of NNs is a hard optimization problem.
NN training is NP-complete \citep{blumTraining3NodeNeural1988}. Further, the loss surface and optimization problem are highly non-convex \citep{dauphinIdentifyingAttackingSaddle2014,goodfellowQualitativelyCharacterizingNeural2015,lecunDeepLearning2015}. This makes navigating the loss surface to a global minimum a more challenging task. 
Due to the non-convexity, NN models with different random initialization or hyperparameters may end up in different local minima on the loss surface and therefore also have different model weights.
With recent growing model sizes, NN training is also increasingly high dimensional, which requires the tuning of ever more optimization parameters. 
These properties of the optimization problem not only make training difficult. The trained model as the outcome of the optimization is also sensitive to hyperparameter choices \citep{haninHowStartTraining2018,liVisualizingLossLandscape2018,yangTaxonomizingLocalGlobal2021}.
Recent work investigates the mode connectivity of different training outcomes \citep{garipovLossSurfacesMode2018,draxlerEssentiallyNoBarriers2018,nguyenConnectedSublevelSets2019,frankleLinearModeConnectivity2019,bentonLossSurfaceSimplexes2021,ainsworthGitReBasinMerging2022}. Nonetheless, it remains an open question, whether trained models with different model weights learn qualitatively different representations. \\

\newpage
The gaps in understanding the relation between the trained NN weights and the model behavior create the two major challenges for NNs identified earlier:
\textbf{(i) predicting model behavior} to diagnose trained models ~\citep{corneanuComputingTestingError2020,martinImplicitSelfregularizationDeep2021,yangTaxonomizingLocalGlobal2021}, for hyperparameter optimization~\citep{bergstraAlgorithmsHyperParameterOptimization2011,chenLearningUniversalHyperparameter2022}, or neural architecture search\citep{elskenNeuralArchitectureSearch2019};
and \textbf{(ii) generating weights} with desirable properties as initializations~\citep{glorotUnderstandingDifﬁcultyTraining2010,heDelvingDeepRectifiers2015,dauphinMetaInitInitializingLearning2019,zhu2021gradinit}, for fine-tuning or transfer-learning ~\citep{yosinskiHowTransferableAre2014,mensinkFactorsInfluenceTransfer2021}.\looseness-1

\paragraph*{Structures in Neural Network Weights}
This thesis attempts to address both challenges by using the weights only. This assumes sufficient structure in the weights of populations of NN models, and that such structure encodes latent factors of the models. 
From an information theory standpoint, NNs learn structured information in the data, during which the weights also become structured. Indeed, weight matrix entropy as a proxy for disorder is reduced during training \citep{martinImplicitSelfregularizationDeep2021}.\\

Formally, NN training is a combination of a dataset $\mathcal{D}$, architecture $\mathcal{A}$, task $\mathcal{T}$, loss $\mathcal{L}$ and training hyperparameters $\lambda$, all of which are structured.
Concretely, the dataset $\mathcal{D}$ contains samples $x$ which are structured, i.e. images. Datasets for supervised learning tasks further contain labels $y$ for each sample. 
The training task $\mathcal{T}$ and corresponding loss $\mathcal{L}$ determine what signal from the data is learned. 
The NN architecture $\mathcal{A}$ with training hyperparameters $\lambda$ imposes structure in how data and weights are processed.
NN training finds optimal weights $\mathbf{W^*}$ by minimizing the training loss as $\mathbf{W^*} = \argmin_{\mathbf{W}} \mathcal{L}(\mathbf{W},\mathcal{D},\mathcal{A},\lambda)$. Through this optimization, the information from the data is encoded in the model weights $\mathbf{W}$, imposing structure on $\mathbf{W}$.
Consequently, the structure in the weights $\mathbf{W^*}$ reflects their latent generating factors $\mathcal{D}$, $\mathcal{L}$, $\lambda$ and $\mathcal{A}$.
Since model behavior and performance are also a consequence of $\{ \mathcal{D}, \mathcal{L},   \lambda,  \mathcal{A}\}$, they, too, are reflected in the NN weights. 
The perspective on structure in a single model can be extended to structure in the \textit{weight space}, the space spanned by the individual weight dimensions, in which a single model is one point.
Extending the earlier thoughts, by induction, the notion of structure in weights of individual models also implies structure in populations of trained models. \\
 
\vspace{4pt}
\noindent
This leads to the main hypothesis for this thesis:  
\begin{flushleft}
    \begin{itemize}
    \item [\textbf{(i)}] Neural Network models populate a structure in weight space;
    \item [\textbf{(ii)}] These structures encode properties and generating factors of models.
    \item [\textbf{(iii)}] Such structures can be exploited for discriminative and generative tasks.
    \end{itemize}
\end{flushleft}

\newpage
\subsection*{Challenges, Thesis Objectives, and Contributions}
\paragraph*{Challenges of Neural Network Weight Spaces}
Operating in weight space to identify such structure poses several challenges. 
First, with growing models, the size of the parameter space also grows. The larger space requires more samples to have sufficient coverage. The correspondingly increasing cost of training those models exacerbates the curse of dimensionality of weight spaces.
Secondly, changes to the architecture affect the parameter count and thus the dimensionality of the space, complicating the comparison of models.
In addition, NN weights contain invariances and equivariances that translate to symmetries in weight space. For example, changing the order of two neurons in a layer with all ingoing and outgoing connections does not change the underlying function of the model, but it does change the position in weight space \citep{hecht-nielsenALGEBRAICSTRUCTUREFEEDFORWARD1990}. 
Similarly, piece-wise linear activation functions allow for linear up and down scaling in subsequent layers which results in unchanged model behavior\citep{dinhSharpMinimaCan2017a}. 
Further symmetries can arise if a model has excess capacity \citep{grigsbyHiddenSymmetriesReLU2023}. 
Due to these symmetries, there is a large finite number of equivalent versions of every model in weight space. The number of equivalent versions grows with the factorial of layer width, and can even be infinite for continuous equivalence classes. 
This property of the NN weight space forms complex local and global structures which complicate working with weights to identify structure and renders notions of neighborhood or distance murky.

\paragraph*{Learning Representations of Neural Network Weights}
To address these challenges, previous work extracts robust features \citep{eilertsenClassifyingClassifierDissecting2020, unterthinerPredictingNeuralNetwork2020,martinPredictingTrendsQuality2021}, aligns models in weight space \citep{ainsworthGitReBasinMerging2022}, uses permutation invariant or equivariant architectures \citep{navonEquivariantArchitecturesLearning2023,andreisSetbasedNeuralNetwork2023,zhouPermutationEquivariantNeural2023,zhouUniversalNeuralFunctionals2024}, or prioritizes single modes in weight generation \citep{haHyperNetworks2017,zhangGraphHyperNetworksNeural2019,knyazevParameterPredictionUnseen2021,zhmoginovHyperTransformerModelGeneration2022,knyazevCanWeScale2023,wortsmanModelSoupsAveraging2022,wortsmanRobustFinetuningZeroshot2022}.
These individual approaches are designed to either a) extract features for model analysis, or to b) generate weights. However, there may be synergies between the two. Generating weights may profit from an understanding of beneficial structures. Likewise, generative capabilities may provide more generalizing features to use for analysis. 
Similar synergies have been demonstrated on other domains \citep{radfordImprovingLanguageUnderstanding2018, chenGenerativePretrainingPixels2020,zhangGenerativeTablePretraining2023}
Prior to the work in this thesis, no general representations of NN models existed that are suitable for both model analysis and model generation downstream tasks.

\newpage
\noindent
Therefore, work in this thesis fundamentally is concerned with the following question.\\
\vspace{12pt}
\noindent
\begin{tcolorbox}[colback=gray!10,
                  colframe=black,
                  width=1.00\linewidth,
                  arc=0mm, auto outer arc,
                  boxrule=2pt,
                  box align=center,
                  halign=center,
                  valign=center,
                  enhanced,
                 ]
    \vspace{4pt}
    \textit{\textbf{Can general, task-agnostic representations be learned from populations of Neural Network models?}}
    \vspace{4pt}
\end{tcolorbox}
\vspace{28pt}

\noindent
This overarching question can be broken down into smaller parts: 
\vspace{8pt}
\begin{flushleft}
    \begin{itemize}
    \item [\textbf{(i)}] Do trained NN models populate a structure in weight space?
    \item [\textbf{(ii)}] How can diverse populations of neural networks be created?
    \item [\textbf{(iii)}] What are suitable methods to learn model structures in weight space?
    \item [\textbf{(iv)}] Are model structures in weight space predictive of model properties, such as model performance or latent generating factors?
    \item [\textbf{(v)}] Can models be generated by sampling from structures in weight space?
    \item [\textbf{(vi)}] What are suitable specific downstream tasks to test discriminative (iv) and generative (v) applications?
    \end{itemize}
\end{flushleft}

\vspace{32pt}

\subsection*{Contributions of this Thesis}
\begin{figure}[ht!]
\centering
\includegraphics[width=1.03\columnwidth]{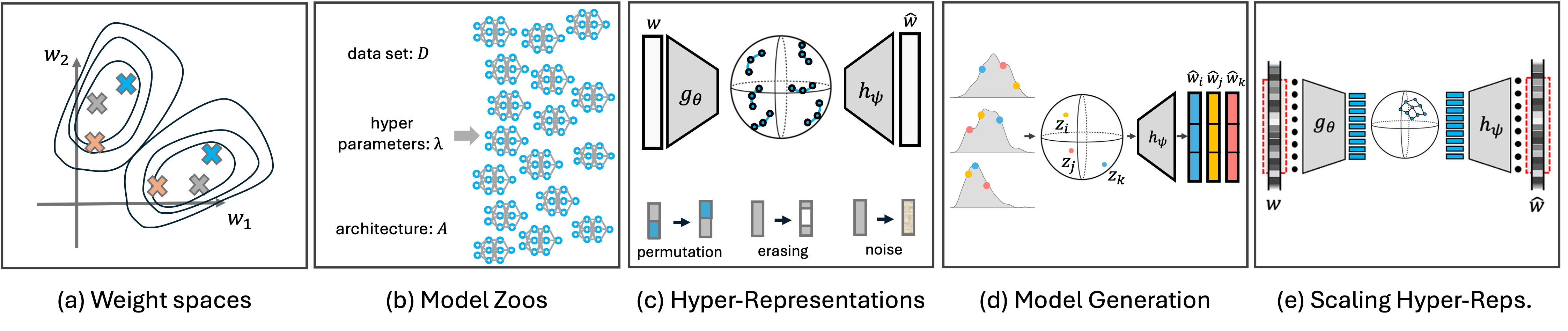}
\caption[Schematic Overview of the Contribution]{Overview of the contribution of the thesis. 
\textbf{(a)}: Chapter~\ref{chap::weight_space} investigates local and global structure in weight spaces as well the potential and challenges of operations on NN weights. 
\textbf{(b)}: Chapter~\ref{chap::model_zoos} proposes a blueprint for diverse populations of NNs as a dataset for the work in this thesis. 
\textbf{(c)}: Chapter~\ref{chap::hyper_reps} introduces \textit{hyper-representations} as a self-supervised representation learning method on NN weights, as well as NN weight augmentation methods.
\textbf{(d)}: Chapter~\ref{chap::generative_hyper_reps} extends \textit{hyper-representations} for model generation.
\textbf{(e)}: Chapter~\ref{chap::scalable_hyper_reps} proposes methods to scale \textit{hyper-representations} to large models and diverse architectures.
}
\label{fig:overview}
\end{figure}

The work collected in this thesis addresses these questions directly as outlined in Figure \ref{fig:overview}. Fundamentally, it establishes that trained NN models populate meaningful structures in weight space. It proposes \textit{hyper-representations} as a self-supervised method to learn these structures from NN weights. Experiments demonstrate that such representations of structure in weight space encode information on latent model properties, such as performance, or hyper-parameters. Further, sampling \textit{hyper-representations} generates model weights that are competitive in fine-tuning and transfer-learning, and generalize to new architectures and tasks.
The individual contributions are organized as follows.\\


\newpage
\subsubsection*{Chapter \ref{chap::weight_space} Challenges in Neural Network Weight Spaces} 
The contributions of \textbf{Chapter \ref{chap::weight_space}} are centered around establishing properties of the domain of NN weight spaces.
It builds on existing work, it describes the complex global and local structure in NN weight spaces.
For this work, the primary task is to evaluate the impact of these global and local structures on operations in weight space.

\vspace{7pt}

\textbf{Proposed Methods:} The research presented in Chapter~\ref{chap::weight_space} is based on the proposition, implementation, and evaluation of the following methods:

\begin{enumerate}[itemsep=-4pt]

\item \textit{Augmentations} of NN models to explore local and global structure in weight spaces. 
\item \textit{Similarity of Behavior} to assess similarity in weight space. 
\item \textit{Analysis and Generation} to evaluate the robustness of weight operations.
 
\end{enumerate}

The empirical evaluation of the NN weight spaces validates the notion of structure through training. It demonstrates the existence of local and global structures. Further, it is shown that the operations on weights can become unstable when these complex relations are not considered. 

\vspace{7pt}

\textbf{Conclusion:} This chapter establishes fundamental properties of the NN weight space domain for this thesis. It shows the potential for analysis and generation of NN weights, but also the need for feature extractors that consider the complex local and global structure of NN weights.

The work in this chapter has originally appeared as
\underline{Konstantin Schürholt}; \textit{Challenges in Neural Network Weight Spaces}; 2024. \\
\subsubsection*{Chapter \ref{chap::model_zoos}: Model Zoos} 
The contributions of \textbf{Chapter \ref{chap::model_zoos}} are centered around the generation of diverse datasets of trained NN models.
In the absence of suitable model populations, this chapter proposes a blueprint for model zoo creation, as well as different diversity metrics. 
For this work, the primary task is to identify generating factors of model zoos that result in sufficient diverse model populations.

\vspace{7pt}

\textbf{Proposed Methods:} The research presented in Chapter~\ref{chap::model_zoos} is based on the proposition, implementation, and evaluation of the following methods:

\begin{enumerate}[itemsep=-4pt]

\item \textit{Supervised Training} for populations of NN models. 
\item \textit{Diversity Metrics} to assess the spread of populations in different spaces. 
\item \textit{Application Examples and Baselines} for model zoos.
 
\end{enumerate}

The empirical evaluation of the model zoo approach demonstrates the ability to control different aspects of diversity in model populations. It is shown that the diversity properties of model zoos generalize to different tasks and architectures. The established baselines in model analysis indicate nontrivial structures in the model zoos.

\vspace{7pt}

\textbf{Conclusion:} This chapter contributes the model zoo datasets for all other work in this thesis. It also outlines the potential for applications of model zoos beyond \textit{hyper-representations} and provides a starting point for future research in this domain.

The work in this chapter has originally appeared as
\underline{Konstantin Schürholt}, Diyar Taskiran, Boris Knyazev, Xavier Giró-i-Nieto, Damian Borth; \textit{Model Zoo: A Dataset of Diverse Populations of Neural Network Models}; Conference on Neural Information Processing Systems (NeurIPS), Datasets and Benchmarks Track, 2022. \\

\subsubsection*{Chapter \ref{chap::hyper_reps}: Hyper-Representations} 
The contributions of \textbf{Chapter \ref{chap::hyper_reps}} are centered around method development and application of self-supervised representation learning on weights of trained NNs. 
Going beyond previous supervised work with hand-crafted feature extractors, this chapter proposes \textit{hyper-representations} with corresponding self-supervised learning task, architecture, and augmentations for NN weights. 
For this work, the primary task is to establish structure in NN weights and uncover latent generating factors of the NN models.

\vspace{7pt}

\textbf{Proposed Methods:} The research presented in Chapter~\ref{chap::hyper_reps} is based on the proposition, implementation, and evaluation of the following methods:

\begin{enumerate}[itemsep=-4pt]

\item \textit{Self-Supervised Learning Task} for representation learning on NN weights. 
\item \textit{Data Augmentation} to increase sample efficiency and include inductive biases. 
\item \textit{Transformer Architecture} with tokenization scheme for NN Weights.
 
\end{enumerate}

The empirical evaluation of the implemented hyper-representation approach demonstrates the ability to learn task-agnostic, lower-dimensional representations of NN weights. It is shown that these representations are highly predictive of model properties such as accuracy or hyperparameters, and generalize to new tasks and models. Furthermore, the proposed augmentations for NN weights improve generalization.

\vspace{7pt}

\textbf{Conclusion:} This chapter contributes to the core of this thesis by demonstrating the feasibility, utility, and effectiveness of \textit{Representation Learning} on NN weights. It also outlines the potential for applications of hyper-representations and lays the groundwork for future research in this domain.

The work in this chapter has originally appeared as
\underline{Konstantin Schürholt}, Dimche Kostadinov, Damian Borth; \textit{Self-Supervised Representation Learning on Neural Network Weights for Model Characteristic Prediction}; Conference on Neural Information Processing Systems (NeurIPS), 2021. \\

\subsubsection*{Chapter \ref{chap::generative_hyper_reps}: Generative Hyper-Representations} 
The contributions of \textbf{Chapter \ref{chap::generative_hyper_reps}} extends \textit{hyper-representations} for generative tasks to allow for NN weight generation. 
Compared to previous \textit{hyper-representations}, this chapter adjusts the self-supervised learning task to improve reconstruction quality, robustness, and smoothness. 
Further, it proposes sampling methods to target specific model properties.
For this work, the primary task is to identify the distribution of targeted models in latent space.

\vspace{7pt}

\textbf{Proposed Methods:} The research presented in Chapter~\ref{chap::generative_hyper_reps} is based on the proposition, implementation, and evaluation of the following methods:

\begin{enumerate}[itemsep=-4pt]

\item \textit{Layer-Wise Normalization} of NN weights to improve reconstruction. 
\item \textit{Sampling Schemes} to identify distributions of targeted properties. 
\item \textit{Fine-tuning and Transfer Learning} of sampled models.
 
\end{enumerate}

The empirical evaluation of the generative \textit{hyper-representation} approach demonstrates the ability to improve reconstruction quality. It is shown that sampling methods can target the distributions of specific properties in latent space. Furthermore, the sampled models match or outperform baselines in fine-tuning and transfer learning.

\vspace{7pt}

\textbf{Conclusion:} This chapter builds on previous work by demonstrating the feasibility, utility, and effectiveness of generative models on NN weights. It shows that a task-agnostic representation learning method is suitable for analysis and generation tasks.\looseness-1

The work in this chapter has originally appeared as
\underline{Konstantin Schürholt}, Boris Knyazev, Xavier Giró-i-Nieto, Damian Borth; \textit{Hyper-Representations as Generative Models: Sampling Unseen Neural Network Weights}; Conference on Neural Information Processing Systems (NeurIPS), 2022. \\

\subsubsection*{Chapter \ref{chap::scalable_hyper_reps}: Scalable Hyper-Representations} 
The contributions of \textbf{Chapter \ref{chap::scalable_hyper_reps}} proposes methods for scaling \textit{hyper-representations} to much larger models of varying architectures. 
This work extends previous \textit{hyper-representations} in data pre-processing and representation, representation learning architecture, and sampling methods. 
For this work, the primary task is to decouple the representation learning model from the architecture and size of the model population by consistent tokenization of models.

\vspace{7pt}

\textbf{Proposed Methods:} The research presented in Chapter~\ref{chap::scalable_hyper_reps} is based on the proposition, implementation, and evaluation of the following methods:

\begin{enumerate}[itemsep=-4pt]

\item \textit{Representation Learning} on sequences to scale to large and varying architectures.\looseness-1
\item \textit{Model Pre-pocessing} by aligning, standardization and sequentialization. 
\item \textit{Unified Models} for discriminative and generative tasks on large models.
\item \textit{Novel Sampling Schemes} to target new architectures and tasks.\looseness-1
 
\end{enumerate}

The empirical evaluation of the scalable \textit{hyper-representation} approach demonstrates the ability to extend representation learning on NN weights to large models and new architectures. It is shown that significant trends in models are preserved in embedding space. Furthermore, the predictive power as well as model sampling generalizes to large models and unseen architectures.

\vspace{7pt}

\textbf{Conclusion:} This chapter builds on previous work by demonstrating the feasibility, utility, and effectiveness of \textit{hyper-representations} on ResNet-18 models and beyond. It shows that a task-agnostic representation learning method is suitable across model sizes for both analysis and generation tasks.

The work in this chapter has originally appeared as
\underline{Konstantin Schürholt}, Michael Mahoney, Damian Borth; \textit{Towards Scalable and Versatile Weight Space Learning}; International Conference on Machine Learning (ICML) 2024. \\

\subsection*{Relevance and Potential Future Impact}
Over the past years, \textit{weight space learning} has become a dynamic research topic and has been recognized by the Machine Learning community.
The project that this thesis is part of has received a Google Research Scholar Award and a HSG Impact Award.
In the context of this thesis, work on populations of models has continued, extending populations towards sparsified models \citep{honeggerSparsifiedModelZoo2023} and models on remote sensing data \citep{honeggerEurosatModelZoo2023}. 
Recently, work done in another lab extends \textit{hyper-representations} to detect backdoors \citep{langoscoDetectingBackdoorsMetaModels2023}.
In parallel, other groups, too, have begun to work on \textit{weight space learning} similar to the work presented in this thesis. 
\citet{berardiLearningSpaceDeep2022} train auto-encoders on weights of CNN models. \citet{peeblesLearningLearnGenerative2022} propose a conditional diffusion approach to generate weights. Several approaches have been proposed to encode implicit neural representations or neural radiance field models \citep{ashkenaziNeRNLearningNeural2022,navonEquivariantArchitecturesLearning2023,deluigiDeepLearningImplicit2023,zhouPermutationEquivariantNeural2023,andreisSetbasedNeuralNetwork2023,zhangNeuralNetworksAre2023} or on RNNs \citep{herrmannLearningUsefulRepresentations2024}. \\

Work on \textit{weight space learning} contributes to a deeper understanding of Neural Networks. Understanding and identifying structure in NN weight spaces can help analyze models to select ideal candidates and reveal weaknesses or backdoors in models. It may lead to meaningful notions of provenance for model versioning and intellectual property protection, and a meaningful similarity metric for model governance and certification.  
The general trend of large foundation models for individual domains can arguably be extended to the domain of NN weights. 
With the work in Chapter \ref{chap::scalable_hyper_reps}, \textit{Hyper-representations} trained on the corpus of publicly available models can become \textit{foundation models of Neural Networks} that incorporate their collective knowledge. 
Such models could be used to generate initializations for new datasets or tasks, meta-learn in latent space, or manipulate model properties like robustness or sparsity.
The future holds the promise of advancements in this area which may change the way NNs are trained and used. I hope that the research provided in this thesis can spark new and exciting work in this direction and provide a foundation in the promising area of \textit{weight space learning}.


\thispagestyle{empty}
\hbox{}
\afterpage{
    \thispagestyle{emptychaptertransition}
    \vspace*{\fill}
    \newpage
    \thispagestyle{plain}
}

\chapter[Challenges in Neural Network Weight Spaces]{Challenges in \\Neural Network Weight Spaces}
\label{chap::weight_space}

%
\vspace{-8pt}
\section*{Abstract}
The success of Neural Networks (NNs) raises the demand for robustness and analysis of models.
Among the available perspectives on trained models, their weights are especially interesting as they contain the information the models have learned. 
We argue, summarizing previous work, that the weights become structured during training, and that this structure encodes latent information of the models.
However, we also identify challenges of operating in NN weight spaces. Their high dimensionality, lack of interoperability, and symmetries form complex local and global relations between weights and model behavior. 
We perform experiments to evaluate the effect of these challenges and demonstrate that they affect model analysis and generation in weight space so much that it can render results random. 
These results indicate that operating directly in weight space is inadvisable. Instead, we call for robust methods that consider the properties of NN weight spaces.\looseness-1

\section{Introduction}
\label{introduction}
In recent years, Neural Networks have become state of the art for complex challenges across a multitude of fields, demonstrating remarkable success in areas such as natural language processing with advancements like GPT-3~\citep{brownLanguageModelsAre2020}, computer vision through breakthroughs in image recognition~\citep{heDeepResidualLearning2016,dosovitskiyImageWorth16x162021} and generative models~\citep{daiScalableDeepGenerative2020}, and reinforcement learning~\citep{silverMasteringGameGo2016, jumperHighlyAccurateProtein2021,driessPaLMEEmbodiedMultimodal2023}.

As the deployment of Neural Networks increases, the need to understand their inner workings intensifies. Achieving a better understanding of these models is necessary for ensuring their reliability, improving their interpretability, and thus enabling trust in their applications. To this end, there are three principal avenues through which we can gain insights into Neural Network models: 
(i) examining their behavior, e.g, by computing the prediction error on test data; 
(ii) understanding their generating factors, e.g., by optimizing the training hyperparameters; 
and (iii) analyzing the weights as the outcome of their training, e.g., evaluating the distribution of weights to determine overfitting. 
These perspectives offer a holistic framework for navigating the intricacies of Neural Networks, providing a structured approach to dissecting and comprehending their functionality.\looseness-1

\textbf{Model Behavior:} 
Exploring model behavior offers direct insights into how a Neural Network interacts with data, revealing its strengths and predictive capabilities across diverse scenarios. Formally, model behavior describes the output $y = f(x)$ of a model $f$ confronted with data $x$, but can extend to intermediate internal representations of the data within the model. Using model behavior is grounded in empirical evidence, as it assesses the model's outputs against real-world or synthetic data. 
However, this perspective is inherently limited by the quality and diversity of the data used for evaluation. If the datasets do not fully represent the complexity of real-world applications or omit critical edge cases, the analysis cannot uncover significant model limitations. \looseness-1

\textbf{Training Hyperparameters:} 
There is abundant work trying to connect the generating factors of models, i.e. their hyperparameters, to model behavior~\citep{hutterAutomatedMachineLearning2019,chenLearningUniversalHyperparameter2022}.
However, the relationship between hyperparameters, architectural choices, and model performance is complex. The search space is vast and often requires substantial computational resources to navigate effectively. Moreover, the relation between generating factors and the performance of models is highly non-linear, indirect, and incomplete, which makes modeling the relation a challenging task.

\textbf{Neural Network Weights:} 
The weights of a Neural Network are the outcome of the learning process. As such, they encode information on the generating factors, learned features, and training progress~\citep{eilertsenClassifyingClassifierDissecting2020,unterthinerPredictingNeuralNetwork2020,martinPredictingTrendsQuality2021,schurholtSelfSupervisedRepresentationLearning2021}. 
Despite their informative potential, the weight space of Neural Networks poses significant challenges. Navigating the weight space of Neural Networks presents a difficult challenge due to its inherent high dimensionality, architectural incompatibilities, as well as complex local and global structures within. \looseness-1

In this work, we explore the potential and pitfalls of working directly with weights. We begin by establishing that Neural Network weight spaces become structured during training via the entropy of weight matrices in Section \ref{sec:weight_space:structure}. Given that weights are structured and contain latent information on models, we then identify specific challenges of weight spaces in Section \ref{sec:weight_space:challenges}. Subsequently, we investigate the impact of these challenges on the relation between similarity in weights and behavior, the usefulness of weights as inputs to predict model properties, and on aggregation of weights in Sections \ref{sec:weight_space:experiments} and \ref{sec:weight_space:results}. Our experiments demonstrate that there are complex local and global relations between weights and model behavior, which can lead to unintended behavior in downstream tasks.
Consequently, we argue against direct operations in the raw weight space. Our results highlight the need for robust methods that consider the global and local structure of weight spaces. 

\section{Related Work}
\paragraph*{Structure in Neural Network Weights.} 
Different strands of work explicitly or implicitly identify structure in the weights of trained Neural Networks. 
\citet{fortLargeScaleStructure2019} identify wedge-shaped substructures, while \citet{garipovLossSurfacesMode2018, draxlerEssentiallyNoBarriers2018, bentonLossSurfaceSimplexes2021} identify simplexes of connected low-loss regions. Along similar lines, other work identifies local or global connected substructures in weight space with homogeneous properties~\citep{wortsmanLearningNeuralNetwork2021,yangTaxonomizingLocalGlobal2021}.
\citet{martinImplicitSelfregularizationDeep2021} investigate the eigenvalue spectrum of weight matrices from a random matrix theory perspective. They describe the evolution from random to heavy-tailed spectra and apply matrix entropy to describe increasing order in weights.
Different work generates weights for Neural Network from some latent factors, which implies structure in those weights~\citep{haHyperNetworks2017,knyazevParameterPredictionUnseen2021,knyazevCanWeScale2023,zhmoginovHyperTransformerModelGeneration2022,wangRecurrentParameterGenerators2021,schurholtHyperRepresentationsGenerativeModels2022,peeblesLearningLearnGenerative2022}.

\paragraph*{Symmetries in Neural Network Weights.}
Permutation and sign symmetries in Neural Networks have been known for a long time~\citep{hansenNeuralNetworkEnsembles1990,bishopPatternRecognitionMachine2006}. Later work added more continuous equivariances caused by piece-wise linear activation functions~\citep{dinhSharpMinimaCan2017a,grigsbyHiddenSymmetriesReLU2023}
Methods have been proposed to align models in weight space and map Neural Network to a pseudo-canonical form~\citep{ainsworthGitReBasinMerging2022}.
Some of these symmetries have also been used as data augmentation and as inductive bias~\citep{schurholtSelfSupervisedRepresentationLearning2021,peeblesLearningLearnGenerative2022}.\looseness-1

\paragraph*{Feature extractors.}
Several approaches have been proposed to extract features from trained weights. \citet{unterthinerPredictingNeuralNetwork2020, eilertsenClassifyingClassifierDissecting2020, corneanuComputingTestingError2020} use weights, weight-statistics, or derived features to predict properties,  the last of which are invariant to permutation symmetries. \citet{martinPredictingTrendsQuality2021} extract norm-based and eigenvalue-based features from the weight matrices, which are likewise invariant to weight permutations. 
Another line of work learns representations of weights, which are either equivariant or invariant by architecture~\citep{navonEquivariantArchitecturesLearning2023,zhangNeuralNetworksAre2023,zhouUniversalNeuralFunctionals2024} or approximately invariant via contrastive learning~\citep{schurholtSelfSupervisedRepresentationLearning2021}.

\section{Structure in Neural Network Weights}
\label{sec:weight_space:structure}
Using the weights of trained Neural Networks as inputs for downstream tasks implies that there is information in the weights of converged models. In this context, a necessary condition for information is some degree of order in these weights. In this section, we argue that during training, models become structured. Subsequently, we empirically evaluate the order in weights during training using matrix entropy.

Training a Neural Network involves iteratively adjusting its weights, $\theta$, to minimize a loss function $L(y,f_\theta(x))$, where $f_{\theta}(x)$ denotes the network's output for input $x$, and $y$ is the true output. This process inherently imposes a structure on the weights $\theta$, reflecting the patterns in the training data.

One way of thinking about the structuring of weights during training is the evolution of weight entropy. Entropy, represented as $H(\Theta)$, quantifies the level of disorder within the weight distribution of a network, where $\Theta$ is the distribution of weights. Entropy is defined as $ H (\Theta)=-\sum_i P(\theta_i) \log P(\theta_i) $, with $P(\theta_i)$ being the probability of a specific weight configuration $\theta_i$. However, in the context of Neural Networks, directly computing $H(\Theta)$ is impractical since modeling $P(\theta_i)$ is challenging.

\begin{figure}[h!]
\centering
\includegraphics[width=0.6\columnwidth]{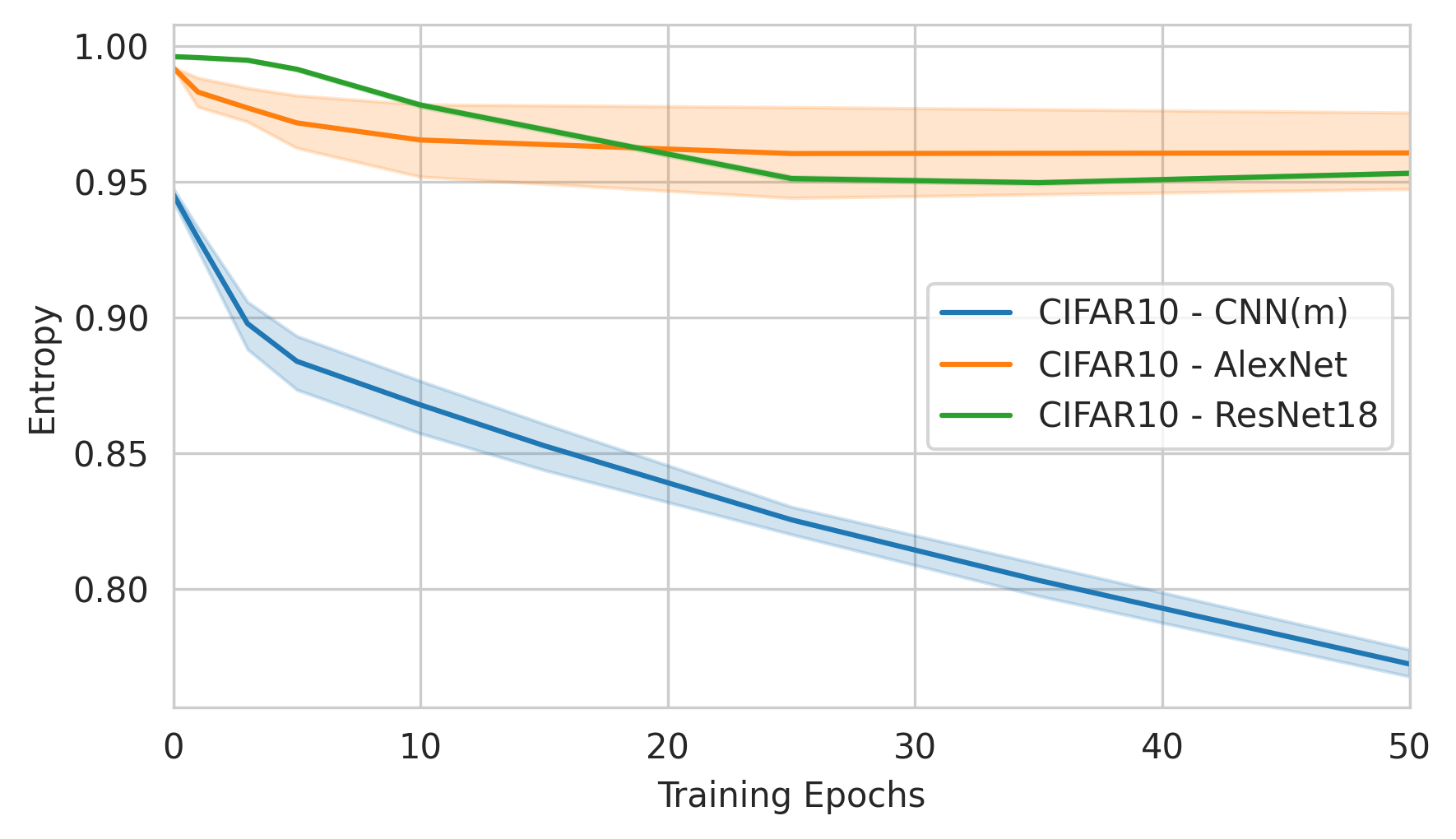}
\caption[Weight Matrix Entropy over Training Epochs]{Weight entropy over training epochs for populations of CNNs (with $\sim 12k$ parameters), AlexNet and ResNet-18 models trained on CIFAR10. Weight entropy is approximated via the empirical spectral density of the weight matrices. During training, the entropy of weight matrices decreases as order is induced in the weights.}
\label{fig:weight_space:entropy}
\end{figure}
However, the evolution of the eigenvalue spectrum of weight matrices offers another perspective on order in weights.
Specifically, analyzing the eigenvalue distribution $\rho(\lambda)$ of a weight matrix offers insights into the network's internal structure. A structured weight matrix will have a characteristic spectral density that differs significantly from that of a random matrix. For instance, a more sharply peaked $\rho(\lambda)$ suggests a higher degree of organization within the weights, acting as a proxy for lower entropy. This spectral analysis serves as an indirect measure of the weight space's organization, reflecting how training imprints structure on the network's parameters. We follow \citet{martinImplicitSelfregularizationDeep2021} in computing the matrix entropy $\mathcal{S}$ which we use as a proxy for the model's entropy.
To get a strong signal, we considered the largest fully connected layer, which is usually the (pen-) ultimate layer. 
The empirical evaluation of that approach indeed shows decreasing entropy during training, see Figure \ref{fig:weight_space:entropy}. The effect varies between model sizes. Models with higher capacity relative to the task appear to go through more entropy reduction than smaller models. 

The experiments show that indeed the weights of models become more structured during training. By induction, the weights of populations of models are also structured, and shaped by the individual generating factors. That structure can be used to analyze models or generate new weights. However, the weight space also poses challenges, to which we turn in the next section.

\section{Challenges in Neural Network Weight Spaces}
In the previous section, we argue that training imposes structure on Neural Network weights, which contains information on latent generating factors of the models and encodes their properties.
However, using that structure by analyzing and manipulating Neural Network weights presents significant challenges. The following paragraphs discuss some of these challenges, the impact of which we experimentally evaluate in the following section.

\paragraph*{Weight Space Dimension.}
The dimensionality and size of Neural Network weight spaces grow with network complexity. 
To reasonably identify structures in that space, more and more samples are required - a "curse of dimensionality" situation. 
Beyond the sheer vastness of this space, larger models incur higher computational costs for training and evaluation, exacerbating the challenge. This dual issue of increased dimensionality and computational expense makes optimizing and even comprehensively understanding the weight space increasingly difficult as models scale.
Higher dimensional spaces also defeat the intuition of lower dimensional spaces~\citep{dauphinIdentifyingAttackingSaddle2014}. By the law of large numbers, samples from Gaussian distributions focus almost all of their probability mass on the shell of a high-dimensional sphere. Further, the distance between points drawn from Gaussian distributions is almost all equally large~\citep{blumFoundationsDataScience2020}. Since Neural Networks are commonly initialized with Gaussian distributions, these findings carry over to high dimensional weight spaces and challenge the concept of distance between weights.

\paragraph*{Architectural Incompatibility.}
The specificity of weights to their network architectures complicates the direct transfer of learned weights between models. For two architectures, $A$ and $B$, with their respective weight spaces $\Theta_A$ and $\Theta_B$, and a weight vector $\theta_A \in \Theta_A$, attempting to apply $\theta_A$ directly to $B$ without addressing architectural disparities usually fails. Even if that is not the case, the mismatch affects the interpretation. Confronted with the same input $x$, the output $f(x)$ generally varies between parameters $\theta_A$ and $\theta_B$: $ f_{\theta_A}(x) \neq f_{\theta_B}(x)$. This limitation highlights the challenge of leveraging knowledge across different models, necessitating architecture-invariant strategies for weight analysis and transfer.

\paragraph*{Weight Space Symmetries.}
\label{sec:weight_space:challenges}
In addition to their scale, weight spaces exhibit multiple forms of symmetries and invariances, which complicate their interpretation. First, most layers contain permutation invariances: For any layer $l$, swapping the order of neurons (and their corresponding weights) does not change the network's output~\citep{hansenNeuralNetworkEnsembles1990}. If $\mathbf{P}$ is a permutation matrix, then for weight matrix $\mathbf{W}$, $\mathbf{W}'=\mathbf{P}\mathbf{W}$ (or $\mathbf{W}'=\mathbf{W}\mathbf{P}^T$) maintains the same function: $f_{\mathbf{W}}(x)=f_{\mathbf{W'}}(x)$. If $P$ is of shape $d_p \times d_p$, there are $d_p!$ unique permutation matrices. That is, the number of distinct symmetric versions of the same function grows with the factorial of a model's width. These multitudes of replications of the loss landscape over weight space make identifying any structure a very hard problem, as they introduce a complex global relation between distance and behavior. Point-symmetric activations like the Sigmoid function introduce additional sign-change symmetries.
In networks with piece-wise linear or point-symmetric activations (e.g., ReLU), there are further, continuous invariances. Scaling the weights and biases in one layer can be compensated by inverse scaling in the subsequent layer, leaving the overall network function unchanged~\citep{dinhSharpMinimaCan2017a}.
Lastly, in highly complex networks, certain weights or combinations thereof may not significantly contribute to the network's function, leading to redundancy. This excess capacity means that multiple, significantly different weight configurations can produce the same output, obscuring the direct relationship between specific weights and network behavior~\citep{grigsbyHiddenSymmetriesReLU2023}.
These symmetries in weight space create a global structure and cause highly nontrivial relations between distance and behavior.  

\paragraph*{Sensitivity to Perturbations.}
Training of Neural Network is an inherently noisy process, where noise can be contributed from the data, the parameter updates, or explicitly as regularization~\citep{martinRethinkingGeneralizationRequires2019,yangTaxonomizingLocalGlobal2021}. From a weight space perspective, adding perturbations to weights explores the local relation between weight and behavior. The effect of perturbations on the weight has been studied as the shape of the Neural Network loss landscape~\citep{liVisualizingLossLandscape2018}. It is characterized by the function $L(\theta)$ and illustrates the model's sensitivity to weight perturbations. This landscape features regions with sharp minima, where small deviations in weights can lead to substantial increases in loss, indicating high sensitivity to noise. Conversely, flat minima represent regions where the model exhibits greater robustness to changes in weights~\citep{dinhSharpMinimaCan2017a}. The presence of these diverse topological features in the loss landscape underlines the variable impact of noise on model performance and stability, making the identification of structure and relating it to properties a difficult task.
\section{Experiments}
\label{sec:weight_space:experiments}
In the previous sections, we argued that the weight space contains latent information on the models, but also complex local and global structure that poses challenges for operations in weight space. Here, we test how these challenges affect applications in weight space in a set of experiments. With these experiments, we test the underlying properties of weight space and evaluate how suitable operations in a local or global scope are. 
Further, we perform experiments as proxies for two general types of applications in weight space introduced in the introduction: (i) predicting model properties from weights as one way of model analysis, and (ii) aggregating the weights of several pre-trained models into one to evaluate generative applications of weights. 
\vspace{-4pt}

\paragraph*{Correlating Weight Similarity and Behavioral Similarity.}
To perform operations in weight space, there is often an implicit or explicit assumption on the relation between changes in weight space and changes in behavior of models, e.g.~\citep{ilharcoEditingModelsTask2022}. In task arithmetic and model souping, changes in weights are expected to translate to proportional changes in behavior \citep{wortsmanRobustFinetuningZeroshot2022}. 
We test this assumption by examining how similarities in model behavior correlate with their similarities in weight space. We measure behavioral similarity using Centered Kernel Alignment (CKA)~\citep{kornblithSimilarityNeuralNetwork2019} and weight space similarity through cosine or $l_2$ similarity metrics. CKA correlates the activations of models at intermediate layers and is permutation invariant, which makes it ideal to compare the behavior of Neural Networks processing the same data. The cosine similarity is computed on the vectorized weights $\theta_A$ and $\mathbf{\theta_B}$ as $sim_{cos} = \frac{\theta_A \mathbf{\theta_B}}{\| \theta_A \|  \|\mathbf{\theta_B}\|}$. We compute $l_2$ similarity as $sim_{l_2} = \exp(-\| \theta_A - \theta_B \|_2^2)$.
\vspace{-4pt}

\paragraph*{Predicting Neural Network Accuracy from Weights.}
As a second experiment, we evaluate the suitability of raw weights as input to infer model properties. Previous work has demonstrated that such weights can be used to predict model properties like test accuracy, generalization gap, or hyperparameters~\citep{unterthinerPredictingNeuralNetwork2020, eilertsenClassifyingClassifierDissecting2020, martinPredictingTrendsQuality2021, schurholtSelfSupervisedRepresentationLearning2021}.
To evaluate the usefulness and sensitivity of weights for model analysis, we linear probe from weights for model test-accuracy following previous work~\citep{unterthinerPredictingNeuralNetwork2020, eilertsenClassifyingClassifierDissecting2020, schurholtSelfSupervisedRepresentationLearning2021}. In these experiments, we fit the linear probe to a train set of aligned models and evaluate how performance varies under changes to the test set. 
\vspace{-4pt}

\paragraph*{Merging Weights of Pretrained Models.}
In the last set of experiments, we evaluate the sensitivity of methods that merge weights of models. 
Re-using weights of pre-trained models for continued training is a common strategy.
In conventional fine-tuning or transfer learning, the weights of one model are re-used directly~\citep{yosinskiHowTransferableAre2014}. Various factors affect the success of transfer learning, such as the domain overlap, dataset and model size, and complexity~\citep{mensinkFactorsInfluenceTransfer2021}.
Recently, several methods have attempted to break up the 1-to-1 match between models and combine the weights of several source models into one target model. Among those, there are learned re-combinations of weights such as zoo-tuning~\citep{shuZooTuningAdaptiveTransfer2021} or knowledge flow~\citep{liuKnowledgeFlowImprove2019}, but also much simpler interpolating or averaging of weights~\citep{wortsmanLearningNeuralNetwork2021, wortsmanRobustFinetuningZeroshot2022, wortsmanModelSoupsAveraging2022}. Averaging model weights, often called \textit{model soup}, has been successfully applied to improve model performance, robustness or combine task knowledge~\citep{ilharcoEditingModelsTask2022,ainsworthGitReBasinMerging2022,chitaleTaskArithmeticLoRA2023}.
With similar goals in mind, other work learns latent representations of Neural Network weights~\citep{schurholtHyperRepresentationsGenerativeModels2022}. Experimental evaluation shows sampling from latent distributions of models generates weights with high zero-shot performance. That raises the question if learning latent representations is necessary, or if similar methods cannot be employed in weight space directly.
We therefore experiment with weight averaging (model soup) and weight sampling similar to~\citep{schurholtHyperRepresentationsGenerativeModels2022} and evaluate the sensitivity of both methods to variations in weights. Following~\citet{schurholtHyperRepresentationsGenerativeModels2022}, weight sampling models the weight distribution per weight dimension via the Kernel Density Estimation of base models, and then draws samples from the estimated weight distribution.\looseness-1 
\vspace{-4pt}

\paragraph*{Base Models and Weight Variations.}
In all three experiments, we evaluate sets of trained models as a proxy for real-world models. These models are taken from the modelzoo repository~\citep{schurholtModelZoosDataset2022}. We use models of small and medium CNNs as well as ResNet18 trained on common computer vision datasets. Within architecture and task, the models are varied only in seed, which causes them to have similar performance but different weights.
We evaluate how the results of the experiments change under specific changes in the weights of the models. Specifically, we 
i) change the number of base models, which extends the global coverage of the weight space; 
ii) align or permute models as proposed by \citet{ainsworthGitReBasinMerging2022} following the methodology of~\citep{schurholtSelfSupervisedRepresentationLearning2021}, which explores local or global structure of the weight space; 
iii) add noise to the weights $\mathbf{\tilde{W}} = \mathbf{W} + r*\mathcal{N}(0,1)$ where $r$ denotes relative noise ratio; adding noise explores the local structure around models; and 
iv) change in model and task complexity to the impact of weight space dimensionality and its relation to the shape of the loss surface. \\

%
\vspace{-40pt}
\section{Results}
\label{sec:weight_space:results}
In this section, the results of the three sets of experiments are presented and discussed.
\vspace{-6pt}
\subsection{Relation Between Weight Distance and Behavior}
\begin{wrapfigure}{r}{0.6\linewidth}
\vspace{-14pt}
\centering
\includegraphics[width=0.95\linewidth]{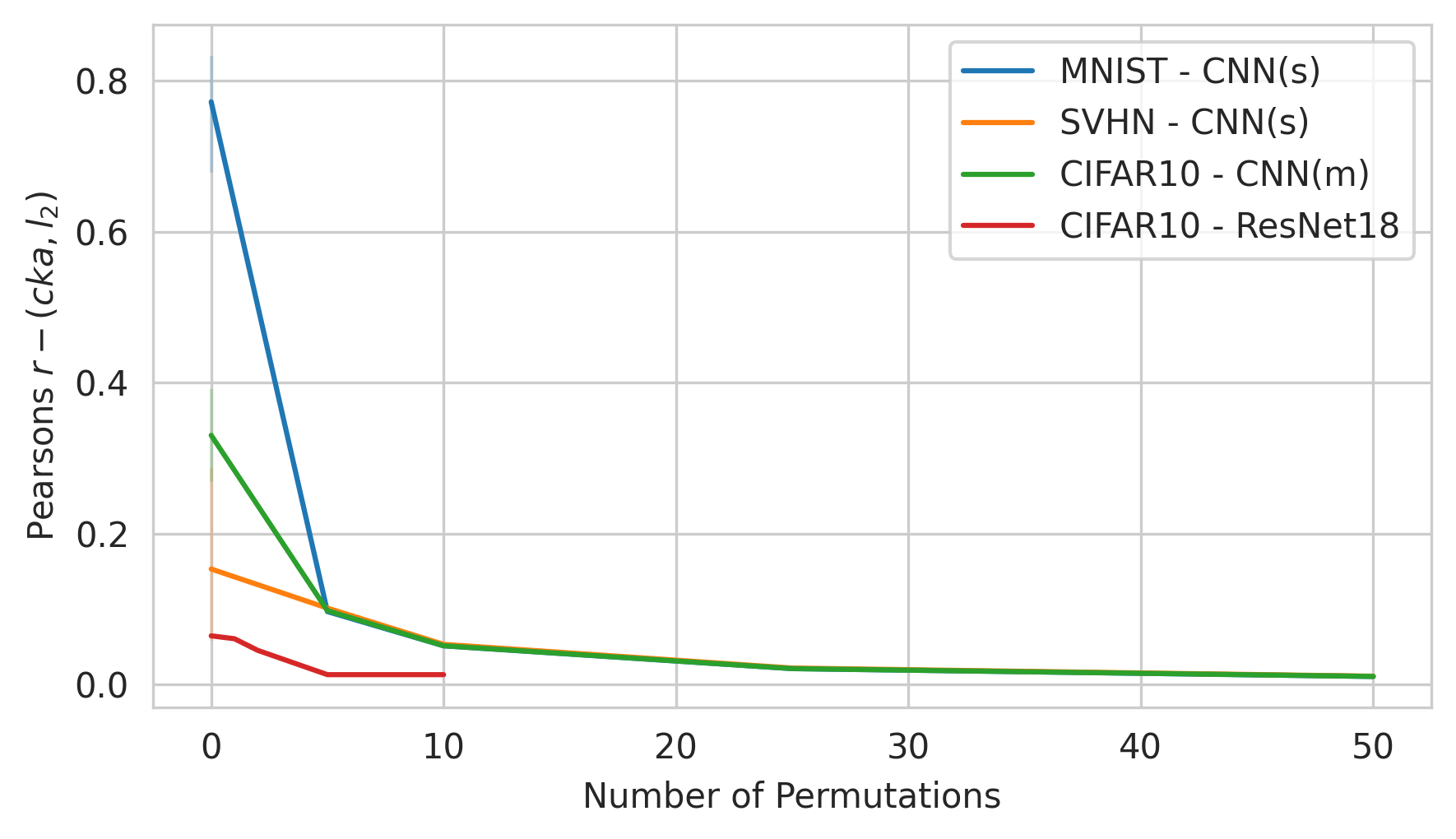}
\vspace{-14pt}
\captionof{figure}[Correlation between CKA and $l_2$ Similarity over Permutations]{Absolute correlation coefficient between pairwise CKA and $l_2$ similarity scores over number of permutations of 15 models. Increasing the number of permutations increases the number of pairs and the global coverage of the weight space.}
\label{fig:weight_space:cka_l2_permutations}
\vspace{-10pt}
\end{wrapfigure}
In the first set of experiments, we compute pairwise similarities in weights and behavior for sets of trained models. We compute the correlation between weight similarity and behavioral similarity to evaluate how much changes in weights correspond to changes in behavior. We vary the number of permutations, the number of models, and the amount of noise added. Due to the nonlinear interaction of weights and behavior, an increasing number of pairs that mix local and global relationships through more models, permutations, or noise, the correlation between similarities should decrease. Vice-versa, aligning models should simplify global relations and thus increase the correlation between weight and behavior similarity. \\

\begin{wrapfigure}{r}{0.6\linewidth}
\vspace{-14pt}
\centering
\includegraphics[width=0.95\linewidth]{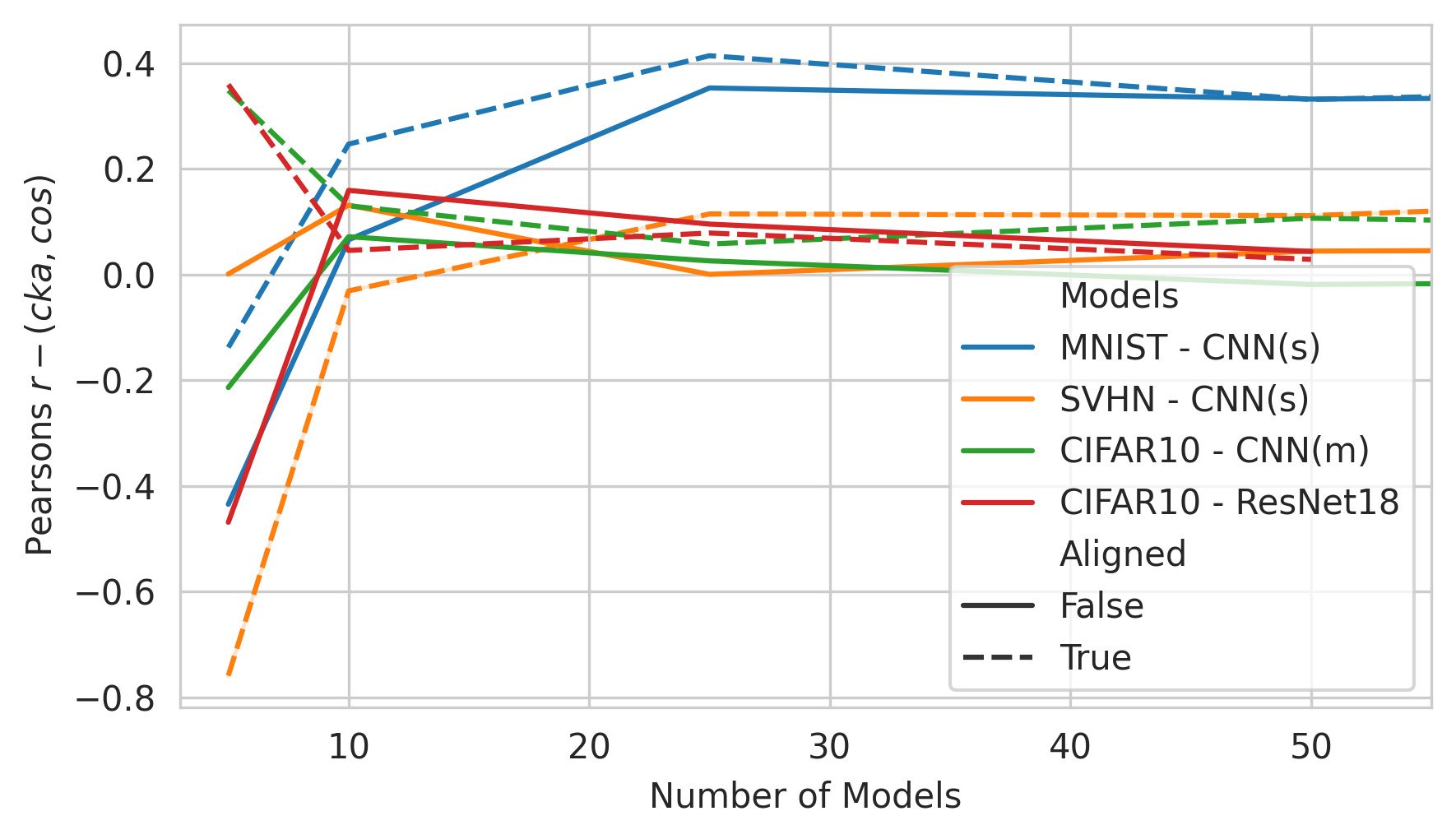}
\vspace{-14pt}
\captionof{figure}[Correlation between CKA and $l_2$ Similarity over Model Number]{Correlation between pairwise CKA and $cos$ similarity over the number of models. Increasing the number of models increases the global coverage of the weight space.}
\label{fig:weight_space:cka_cos_models}
\vspace{-10pt}
\end{wrapfigure}
Empirical evaluations largely support these expectations. In populations of fully aligned models, we observe absolute correlation coefficients in the range of $0.1$ and $0.78$ between behavioral similarity and $l_2$ similarity, see Figure \ref{fig:weight_space:cka_l2_permutations}. This correlation strength diminishes as model and task complexity increase. Introducing more permutations significantly lowers the absolute correlation between behavioral and Euclidean similarities to near zero. These results demonstrate the complex global structures in unaligned weight space. \\

\begin{wrapfigure}{r}{0.6\linewidth}
\centering
\includegraphics[width=0.95\linewidth]{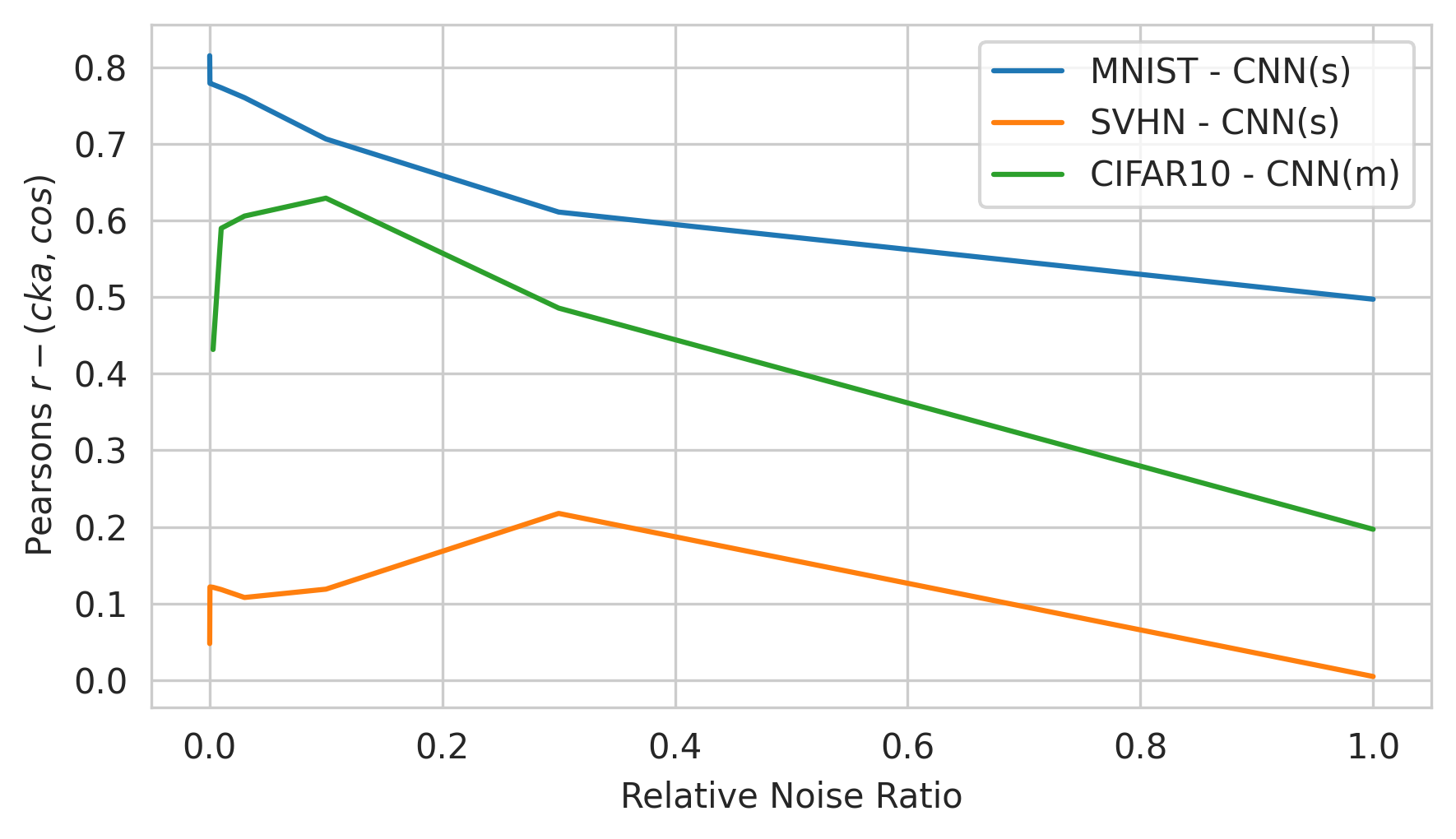}
\vspace{-14pt}
\captionof{figure}[Correlation between CKA and $cos$ Similarity over Weight Noise]{Correlation between pairwise CKA and $cos$ similarity over relative noise added to model weights. Increasing the noise increases the local coverage of the weight space.}
\label{fig:weight_space:cka_cos_noise}
\vspace{-10pt}
\end{wrapfigure}
Similar patterns are observed with an increasing number of models (Figure \ref{fig:weight_space:cka_cos_models}), which likewise affects the global relation between models. Aligning models generally leads to stronger correlations. Local changes by adding noise to the weights have a similar decreasing effect on correlation, but it is less abrupt than increasing permutations (Figure \ref{fig:weight_space:cka_cos_noise}). These findings highlight the complex global and local relationship between models in weight space and their behavior, emphasizing the need for careful consideration of underlying weight structures and changes. Specifically, proportional relations between weight changes and behavior changes only seem to be supported in local neighborhoods around models, while global relations are far more complex.

\subsection{Weights for Model Analysis}

\begin{wrapfigure}{r}{0.6\linewidth}
\vspace{-10pt}
\centering
\includegraphics[width=0.95\linewidth]{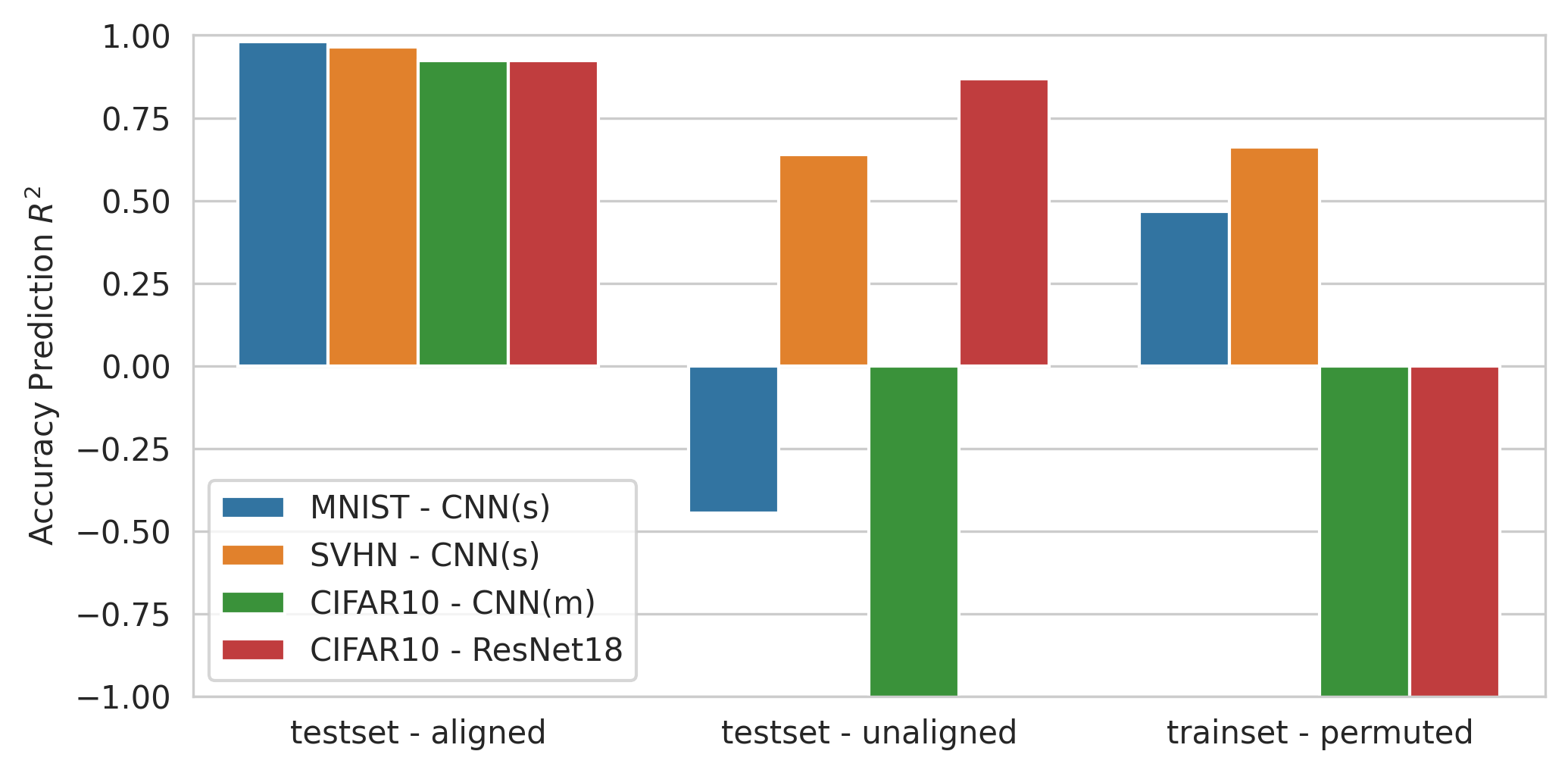}
\vspace{-14pt}
\captionof{figure}[Linear Probing for Accuracy under Variations to the Test Set]{Linear probing for model accuracy performance in $R^2$ for different model and task complexities. The linear probes are fitted to aligned train sets and evaluated on different test sets. Deviations to aligned test sets increase global coverage of the weight space.}
\label{fig:weight_space:discr_size_testset}
\vspace{-10pt}
\end{wrapfigure}
In the second set of experiments, we predict model accuracy using linear probes from the weights. We fit linear probes to aligned train sets and evaluate on different test sets to test the effects of variations. 
As a baseline, the linear probe is evaluated on aligned test sets, where linear probes achieve regression-$R^2$ of above 90\%, see Figure \ref{fig:weight_space:discr_size_testset}. Harder tasks and larger models reduce the performance.
However, if the test sets are not aligned or if linear probes are evaluated on permuted versions of the train set, both of which increase the global coverage of the weight space, the performance is significantly reduced, in many cases to $R^2$ far below zero. 
These results again indicate the complex global structure in weight space.

\begin{wrapfigure}{r}{0.6\linewidth}
\centering
\includegraphics[width=0.95\linewidth]{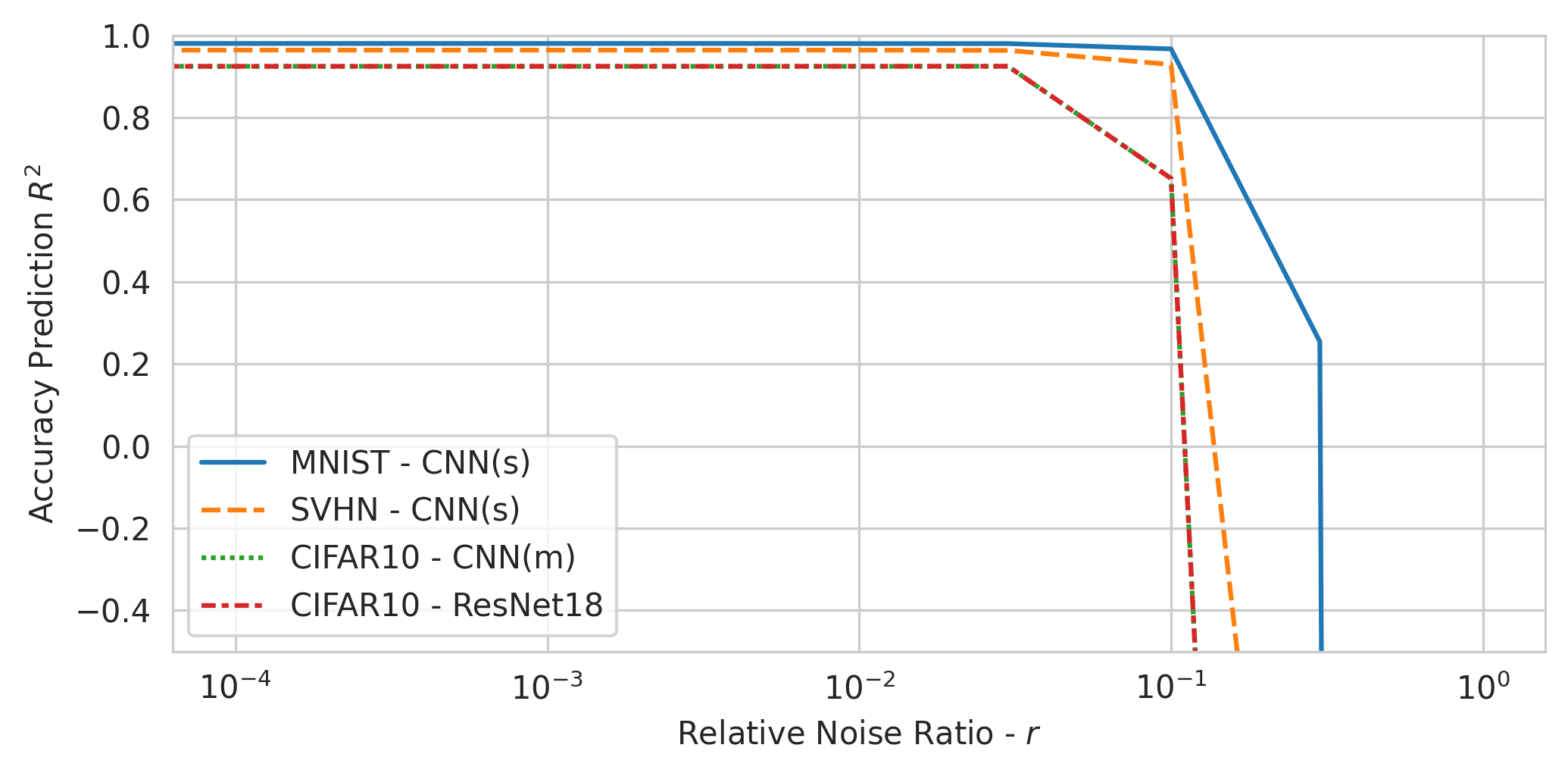}
\vspace{-14pt}
\captionof{figure}[Linear Probing for Accuracy under Weight Noise]{Linear probing for model accuracy performance in $R^2$ for different model and task complexities over relative noise added to model weights. The linear probes are fitted to aligned train sets and evaluated on aligned test sets. Noise increases the local coverage of the weight space.}
\label{fig:weight_space:discr_noise}
\vspace{-10pt}
\end{wrapfigure}
Adding noise to models explores the local neighborhood in weight space. Experimental evaluations show that predicting accuracy for such models collapses at a certain threshold for $r$, see Figure \ref{fig:weight_space:discr_noise}. While the collapse happens at relatively large levels of relative noise, it decreases with model and task complexity.

Fitting predictors on relatively homogeneous, fully aligned populations is not implausible in practice. Such settings can occur if random seeds are shared, or models are fine-tuned from only a few pre-trained models. The experiments show that making predictions on weights based on such populations breaks very quickly for models that are not aligned or explore local regions, which can not always be ruled out. On the other hand, methods that are more robust to changes, such as weight statistics~\citep{unterthinerPredictingNeuralNetwork2020,eilertsenClassifyingClassifierDissecting2020}, the eigenvalue spectrum~\citep{martinPredictingTrendsQuality2021} or learned representations~\citep{schurholtSelfSupervisedRepresentationLearning2021} are also more reliable for downstream predictions of model properties, see Table \ref{tab:weight_space:discr_baselines}.

\begin{table}[h]
\centering
\captionof{table}[Linear Probing Baseline using Weight Statistics under Weight Variations]{
$R^2$ of linear probing for test accuracy using layer-wise weight statistics. Using aligned trainsets and variations of the testset. Weight statistics are invariant to permutations and therefore robust to lack of alignment or random permutations, but are sensitive to noise. 
}
\label{tab:weight_space:discr_baselines}
\small
\begin{center}
\setlength{\tabcolsep}{6pt}
\begin{tabularx}{0.95\linewidth}{lccccc}
\toprule
& \multicolumn{5}{c}{Testset}                 \\
\cmidrule(l){2-6}
                 & aligend & not aligned & permuted & noise $r=0.1$ & noise $r=0.3$  \\
\cmidrule(r){1-1}  \cmidrule(l){2-2} \cmidrule(l){3-3} \cmidrule(l){4-4} \cmidrule(l){5-6} 
MNIST - CNN(s)     & 0.989   & 0.989       & 0.988    & 0.972     & -0.119    \\
SVHN - CNN(s)      & 0.988   & 0.988       & 0.988    & 0.955     & -2.302    \\
CIFAR10 - CNN(m)   & 0.970   & 0.970       & 0.970    & 0.546     & -9.811    \\
CIFAR10 - ResNet18 & 0.970   & 0.970       & 0.963    & 0.546     & -9.811   \\
\\
\bottomrule
\end{tabularx}
\end{center}
\end{table} 

\newpage
\subsection{Combining Model Weights}

\begin{wrapfigure}{r}{0.6\linewidth}
\vspace{-10pt}
\centering
\includegraphics[width=0.95\linewidth]{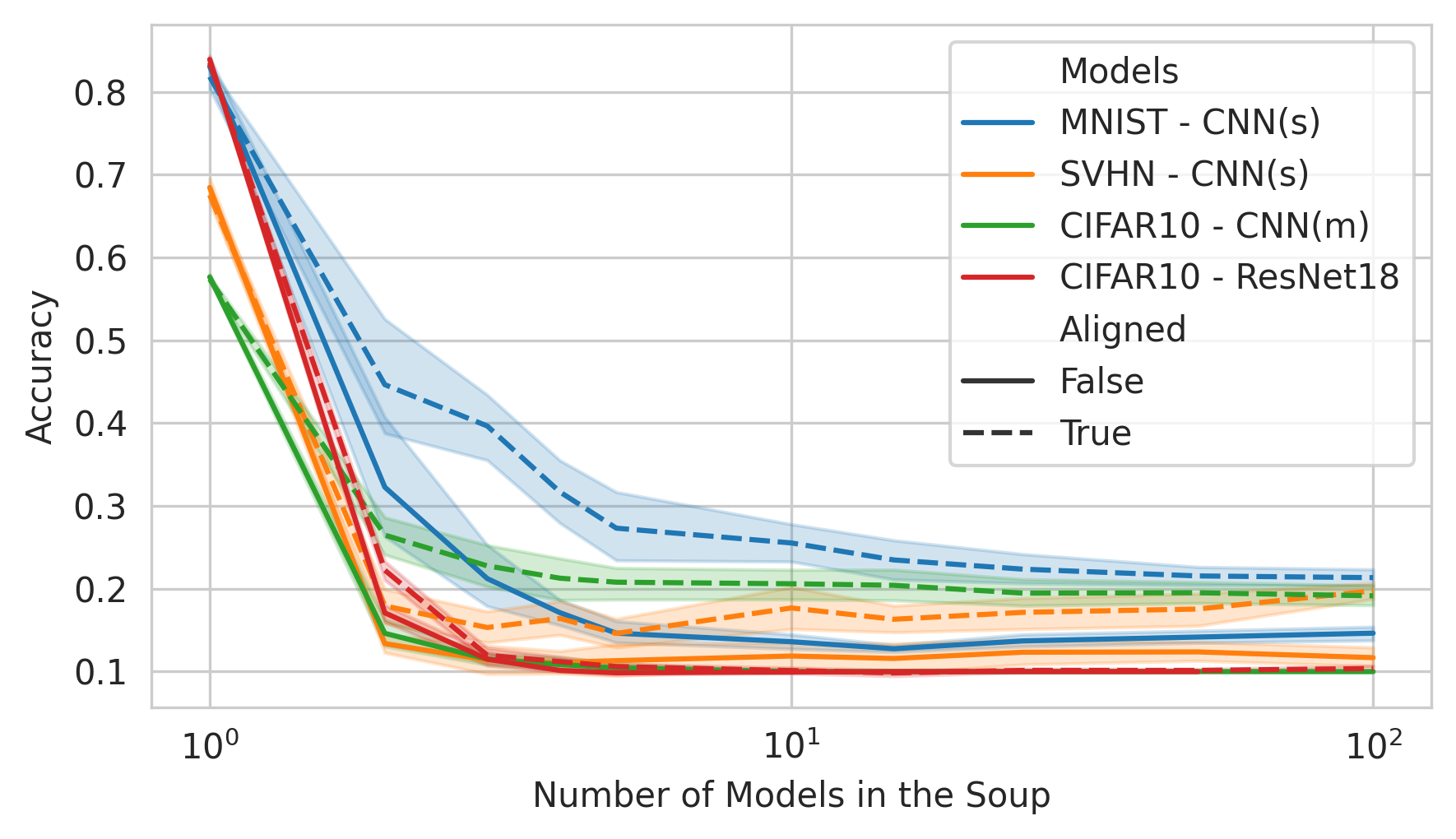}
\vspace{-14pt}
\captionof{figure}[Model Soup Accuracy over Number of Models]{Model soup test accuracy over number of averaged models. Increasing the number of models increases the global coverage of the weight space.}
\label{fig:weight_space:generative_soup_models}
\vspace{-10pt}
\end{wrapfigure}
In this last set of experiments, we evaluate the sensitivity of weight averaging and weight sampling to variations in weights. Specifically, we evaluate the effects of alignment, permutations, and the number of base models. Since both methods operate in raw weight space to achieve a specific behavior, we expect effects similar to those of the previous experiments. That is, combining models with a clearer signal improves performance; with more models and permutations the signal in weights becomes less clear and performance decreases. \\

\begin{wrapfigure}{r}{0.6\linewidth}
\vspace{-10pt}
\centering
\includegraphics[width=0.95\linewidth]{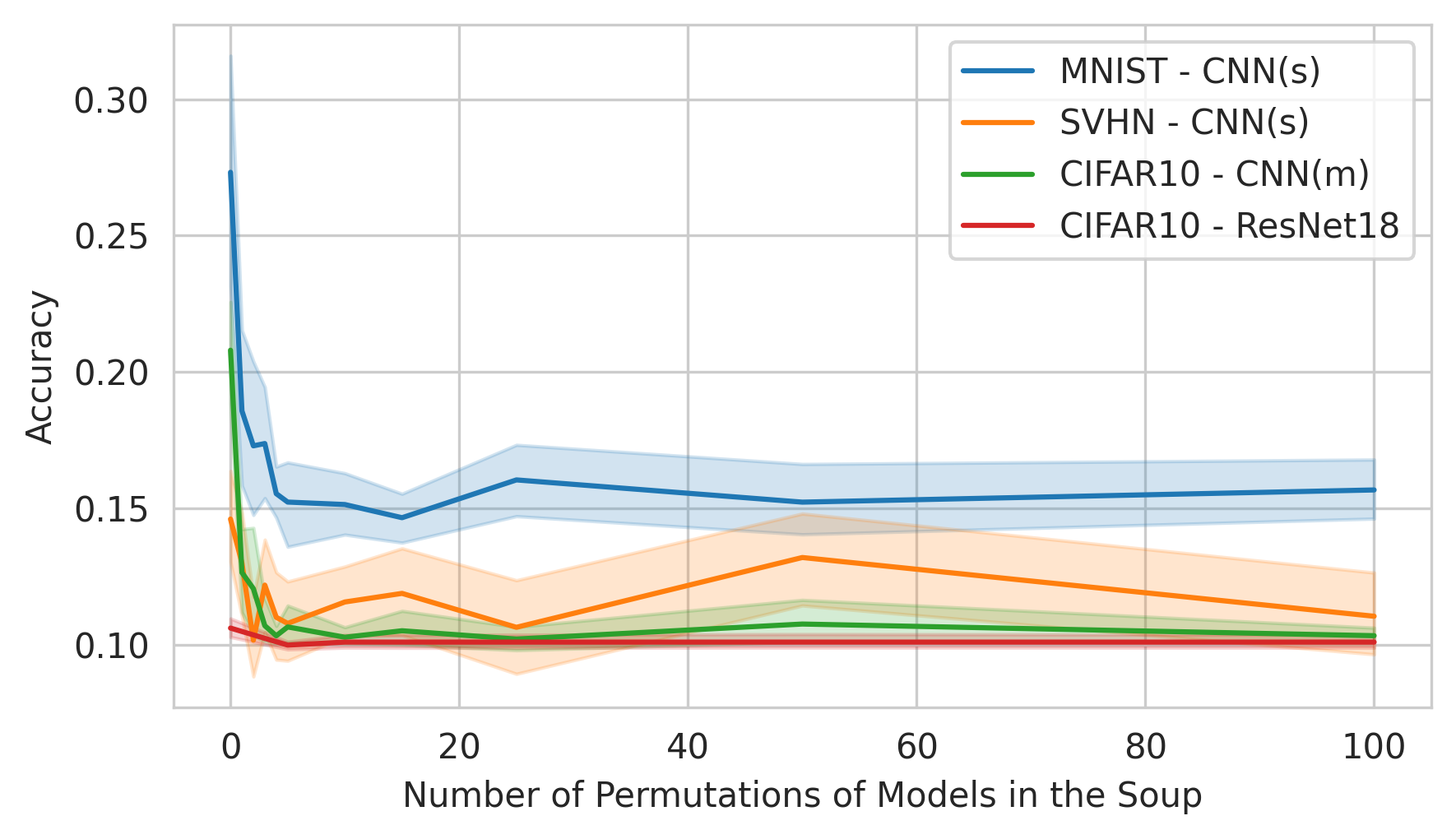}
\vspace{-14pt}
\captionof{figure}[Model Soup Accuracy over Number of Permutations]{Test accuracy of soup of 5 models over number of permutations. Increasing the number of permutations increases the global coverage of the weight space.}
\label{fig:weight_space:generative_soup_permutations}
\vspace{-10pt}
\end{wrapfigure}
\begin{figure}[h!]
\centering
\includegraphics[width=0.7\linewidth]{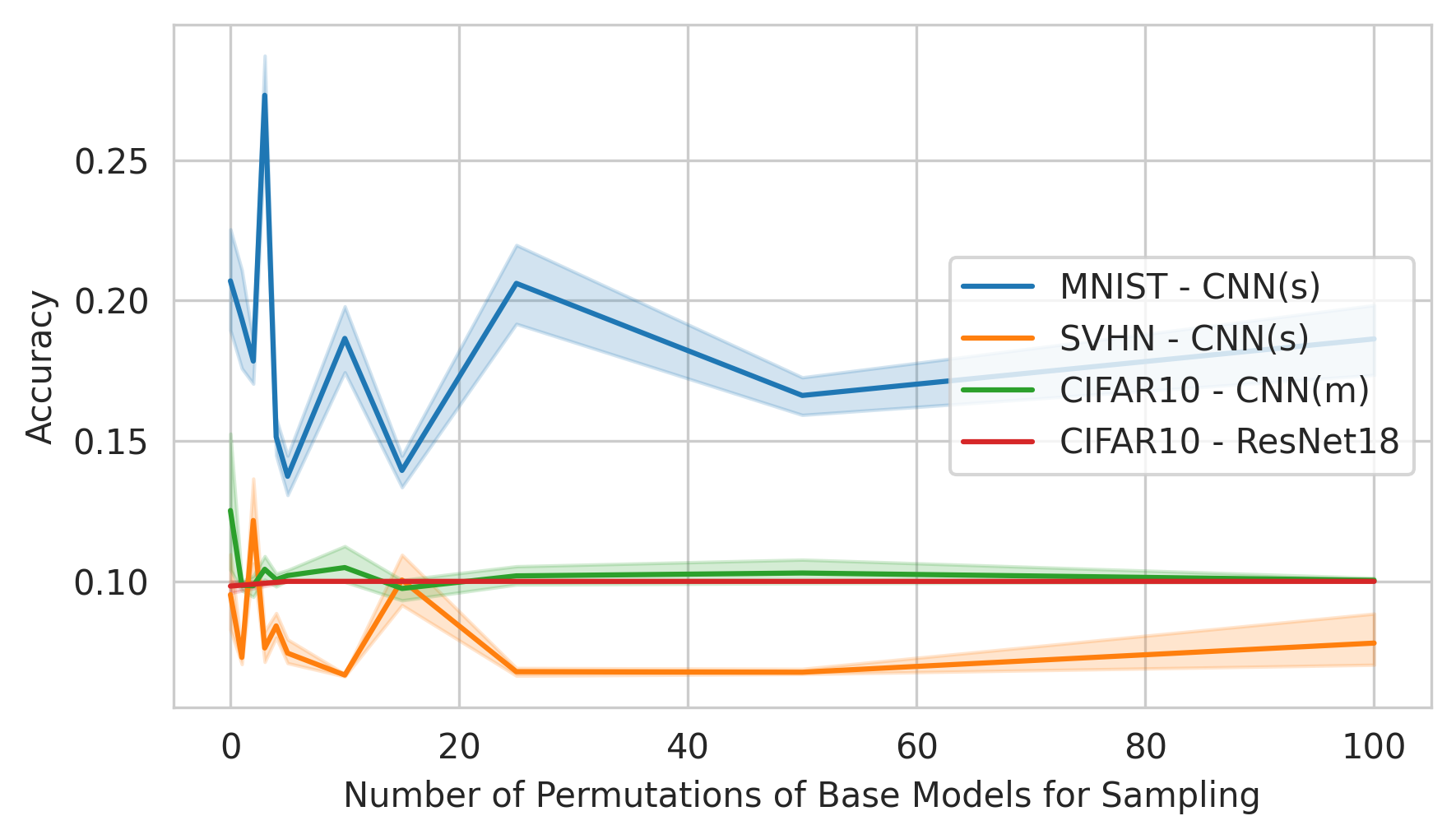}
\caption[Model Soup Accuracy over Number of Permutations]{Test accuracy for sampled models over number permutations of the base models used to model the weight distribution. Increasing the number of permutations increases the global coverage of the weight space.}
\label{fig:weight_space:generative_sampling_permutations}
\end{figure}

Experimental evaluations on model soups with averaged weights largely support those expectations, see Figure \ref{fig:weight_space:generative_soup_models} and \ref{fig:weight_space:generative_sampling_permutations}. The performance of individual models is demonstrated as the performance of soups with 1 model. Combining more than one model decreases the performance in all our experiments. Aligning models improves performance over non-aligned source models, adding permutations decreases performance. Similarly, using more source models to combine into a single target model generally hurts performance, more base models seem to make the target weights noisier. 
Further, performance decreases with task and model complexity. Notably, even averaging aligned models decreases performance over the base population. This indicates that averaging weights of models that are not very close to each other does not generally improve performance.  \\

\begin{wrapfigure}{r}{0.6\linewidth}
\centering
\includegraphics[width=0.95\linewidth]{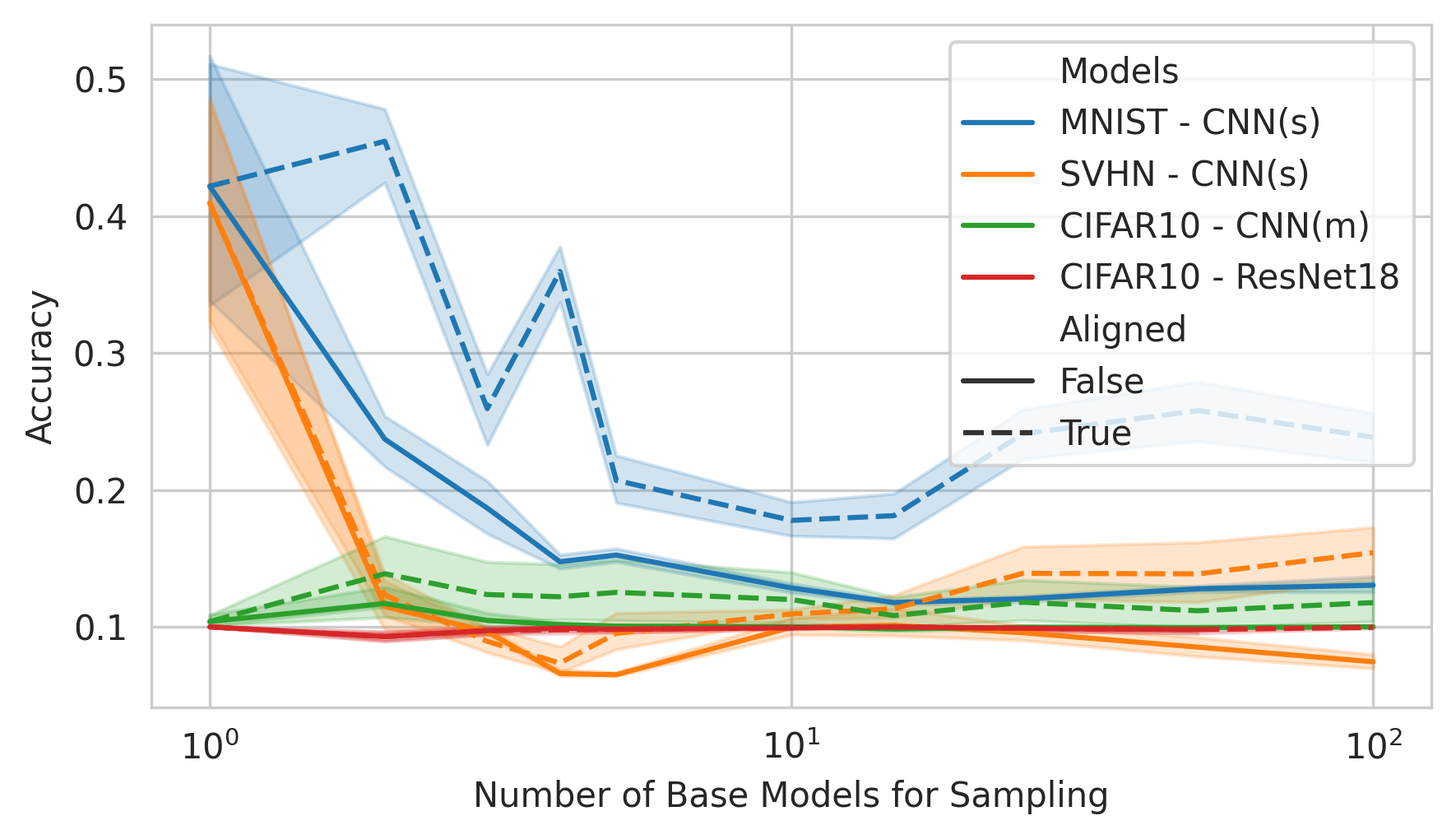}
\vspace{-14pt}
\captionof{figure}[Weight Sampling Accuracy over Number of Models]{Test accuracy for sampled models over number of base models. Increasing the number of base models increases the global coverage of the weight space.}
\label{fig:weight_space:generative_sampling_models}
\vspace{-10pt}
\end{wrapfigure}
Interestingly, experiments sampling in weight space directly shows quite different behavior, see Figures \ref{fig:weight_space:generative_sampling_models} and \ref{fig:weight_space:generative_sampling_permutations}. On the simpler MNIST and SVHN datasets, performance is higher than that of comparable model soups. On the other hand, with increasing task and model complexity (CIFAR10 CNN or ResNet-18), sampled models default to random guessing. Further, there does not appear to be any significant impact of aligning, permutations, or number of base models.  More work is necessary to gain a better understanding of the mechanics, as the sampling method and hyper-parameters may overshadow the source model impact.
Also, the overall performance of models sampled in weight space is significantly lower than models sampled in a learned representation space~\citep{schurholtHyperRepresentationsGenerativeModels2022}, which indicates again that the weight space may not be suitable for such operations, and they instead may benefit from an abstract (learned) feature space. \\

\section{Discussion}
The weights of Neural Networks are not only the byproduct of training but can be used for model analysis, control training, or the generation of new models. The training process structures the weights and embeds information of the model, which makes them suitable for model evaluation and the exploration of novel training strategies.\looseness-1

This work confirms that under ideal circumstances, weights are a good input for model analysis and model generation. Yet, we also identify challenges in dealing with Neural Network weights rooted in local and global relations between weights and behavior of models. 
We observe that even minor deviations from optimal conditions can introduce instability. 
Effects are noticeable for lack of model alignment, varying model size, and local variations similar to training noise.
In real-world scenarios, such deviations cannot be ruled out and can make operations in weight space hard to control.\looseness-1

Our results highlight the need for methods that focus on extracting robust features from weights like weight statistics or learned abstract representations. These features should be resilient to the inherent local and global structure within weight spaces, ensuring stable and reliable outcomes. \looseness-1

\FloatBarrier
\newpage
\clearpage

\newpage
\thispagestyle{empty}
\hbox{}
\afterpage{
    \thispagestyle{emptychaptertransition}
    \vspace*{\fill}
    \newpage
    \thispagestyle{plain}
}

\chapter[Model Zoos: A Dataset of Diverse Populations of Neural Networks]{Model Zoos: A Dataset of Diverse Populations of Neural Network Models}
\label{chap::model_zoos}
\blfootnote{This work originally was accepted for publication at NeurIPS 2022 \citep{schurholtModelZoosDataset2022} }
\vspace{-8pt}
\section*{Abstract}
In the last years, neural networks (NN) have evolved from laboratory environments to the state-of-the-art for many real-world problems. 
It was shown that NN models (i.e., their weights and biases) evolve on unique trajectories in weight space during training. Following, a population of such neural network models (referred to as \textit{model zoo}) would form structures in weight space. We think that the geometry, curvature and smoothness of these structures contain information about the state of training and can reveal latent properties of individual models.
With such \textit{model zoos}, one could investigate novel approaches for (i) model analysis, (ii) discover unknown learning dynamics, (iii) learn rich representations of such populations, or (iv) exploit the \textit{model zoos} for generative modelling of NN weights and biases.
Unfortunately, the lack of standardized \textit{model zoos} and available benchmarks significantly increases the friction for further research about populations of NNs. With this work, we publish a novel dataset of \textit{model zoos} containing systematically generated and diverse populations of NN models for further research. In total the proposed model zoo dataset is based on eight image datasets, consists of 27 \textit{model zoos} trained with varying hyperparameter combinations and includes 50'360 unique NN models as well as their sparsified twins, resulting in over 3’844’360 collected model states. 
Additionally, to the model zoo data we provide an in-depth analysis of the zoos and provide benchmarks for multiple downstream tasks.
The dataset can be found at \href{www.modelzoos.cc}{www.modelzoos.cc}.\looseness-1

\section{Introduction}
%
%
%
The success of Neural Networks (NN) is surprising, considering the hard optimization problem to 
be solved during training of NNs. Specifically, NN training is NP-complete~\citep{blumTraining3NodeNeural1988}, 
the loss surface and optimization problem are non-convex~\citep{dauphinIdentifyingAttackingSaddle2014,goodfellowQualitativelyCharacterizingNeural2015, lecunDeepLearning2015} and the parameter space to fit during training is high dimensional~\citep{brownLanguageModelsAre2020}. 
Additionally, NN training is sensitive to random initialization and hyperparameter selection~\citep{haninHowStartTraining2018,liVisualizingLossLandscape2018}.
%
%
Together, this leads to an interesting characteristic of NN training: given a dataset and an architecture, different random initializations or hyperparameters lead to different minima on the loss surface and therefore result in different model parameters (i.e., weights and biases). Consequently, multiple training results in different NN models.
The resulting population of NN (referred to as \textit{model zoo})
is an interesting object to study: Do individual models of the model zoo have something in common? 
Do they form structures in weight space?
What can we infer from such structures?
Can we learn representations of them?
Lastly, can such structures be exploited to generate new models with controllable properties?
\looseness-1


%
%

These questions have been partially answered in prior work.
Theoretical and empirical work demonstrates increasingly well-behaved loss surfaces for growing number of parameters \citep{goodfellowQualitativelyCharacterizingNeural2015,dauphinMetaInitInitializingLearning2019,liVisualizingLossLandscape2018}. The shape of the loss surface and the starting point is determined by hyperparameters and the initialization, respectively~\citep{liVisualizingLossLandscape2018}. NN training navigates the loss surface with iterative, gradient-based update schemes smoothed by momentum. The step length along a trajectory as well as the curvature are determined by the change of the loss as well as how aligned the subsequent updates are \citep{cazenavetteDatasetDistillationMatching2022,schurholtInvestigationWeightSpace2021}. Together, these findings suggest that populations of NN models evolve on unique and smooth trajectories in weight space.
Related work has empirically confirmed the existence of such structures in NNs~\citep{denilPredictingParametersDeep2013}, demonstrated the feasibility to learn representations of them, showed that they encode information on model properties ~\citep{unterthinerPredictingNeuralNetwork2020,eilertsenClassifyingClassifierDissecting2020,schurholtSelfSupervisedRepresentationLearning2021} and can be used to generate unseen models with desirable properties~\citep{schurholtHyperRepresentationsPreTrainingTransfer2022,schurholtHyperRepresentationsGenerativeModels2022,zhmoginovHyperTransformerModelGeneration2022,knyazevParameterPredictionUnseen2021}
%
%
%
To thoroughly answer the questions above, a large and systematically created dataset of model weights is necessary. 

\begin{figure*}[t]
\begin{minipage}[t]{1.0\textwidth}
\begin{center}
\includegraphics[trim=0in 0in 0in 0in, clip, width=1.0\linewidth]{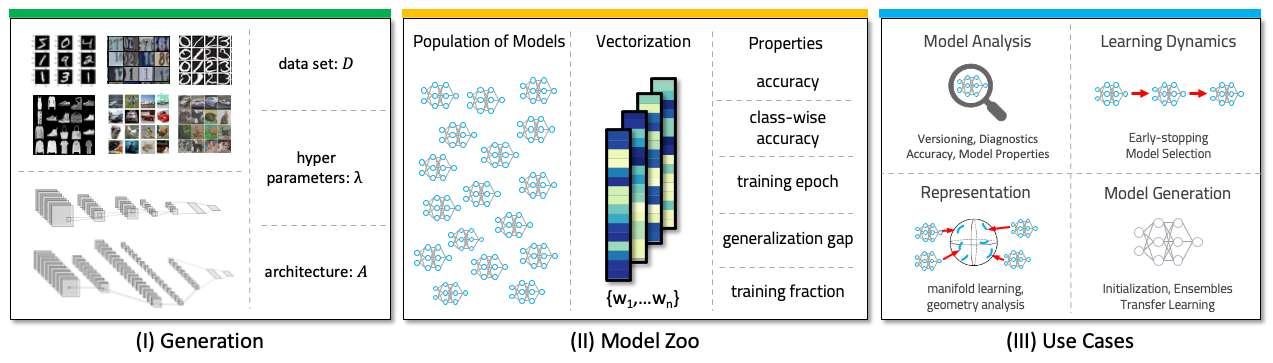}
\caption[Model Zoo Overview of the Approach]{\small The proposed dataset of model zoos is trained on several image datasets with two CNN architectures and multiple configurations of hyperparameters. The resulting population of neural network models is vectorized and made available with all meta-data such as the generating factors of the model zoo as well as the model properties such as accuracy, generalization gap, and others. Potential use cases are (a) model property prediction, (b) inference of learning dynamics, (c) representation learning, or (d) model generation.
}
\label{fig:model_zoos:overview}    
\end{center}
\end{minipage}
\end{figure*}

%
%
Unfortunately, so far only few model zoos with specific properties have been published~\citep{unterthinerPredictingNeuralNetwork2020,eilertsenClassifyingClassifierDissecting2020, suchAtariModelZoo2019, schurholtSelfSupervisedRepresentationLearning2021}. While many machine learning domains have standardized datasets, there is no model zoo nor a benchmark to evaluate and compare against.
The lack of a standardized model zoos has three significant disadvantages: 
(i), existing model zoos are usually designed for a specific purpose and of limited general utility. Their design space is rather sparse, covering only small portions of all available hyperparameter combinations. 
Moreover, some existing zoos are generated on synthetic tasks and are small, containing only a small population of models; 
(ii), researchers have to choose between using an existing zoo or generating a new one for each new experiment, weighing the disadvantages of existing zoos against the effort and computational resources required to generate a new zoo; 
(iii), a new model zoo causes subsequent work to lose comparability to existing research. Therefore, the lack of a benchmark model zoo significantly increases the friction for new research.

%
%
\paragraph*{Our contributions:}
To study the behaviour of populations of NNs, we publish a large-scale model zoo of diverse populations of neural network models with controlled generating factors of model training.  
Special care has been taken in their design and the used protocols for training. To do so, we have defined and restricted the generating factors of model zoo training to achieve desired zoo characteristics.\looseness-1

%
%
The zoos are trained on eight standard image classification datasets, with a broad range of hyperparameters and contain thousands of configurations. Further, we add sparsified \textit{model zoo twins} to each of these zoos.
%
Alltogether, the zoos include a total of 50'360 unique image classification NNs, resulting in over 3’844’360 collected model states.\looseness-1

%
%
Potential use cases for the model zoo include (a) model analysis for reliability, bias, fairness, or adversarial vulnerability, (b) inference of learning dynamics for efficiency gain, model selection or early stopping, (c) representation learning of such populations, or (d) model generation.
%
%
Additionally, we present an analysis of the model zoos and a set of experimental setups for benchmarks on these use cases and initial results as foundation for evaluation and comparison.

\newpage
%
%
With this work we provide a standardized dataset of diverse model zoos connected to popular image datasets, its corresponding meta-data and performance evaluations to the machine learning research community. All data is made publicly available to foster community building around the topic and to provide a ground for use beyond the defined benchmark tasks. 
An overview of the proposed dataset and benchmark as well as potential use cases can be found in Fig.~\ref{fig:model_zoos:overview}

\section{Existing Populations of Neural Networks Models}
\vspace{-1pt}
With the increase in usage of neural networks, requirements for evaluation, testing and certification have grown. 
Methods to analyze NN models may attempt to visualize salient features for a given class~\citep{zeilerVisualizingUnderstandingConvolutional2014,karpathyVisualizingUnderstandingRecurrent2015,yosinskiUnderstandingNeuralNetworks2015}, investigate the robustness of models to specific types of noise \citep{zintgrafVisualizingDeepNeural2017,dabkowskiRealTimeImage2017}, predict model properties from model features \citep{yakTaskArchitectureIndependentGeneralization2018, jiangPredictingGeneralizationGap2019,corneanuComputingTestingError2020} or compare models based on their activations \citep{raghuSVCCASingularVector2017,morcosInsightsRepresentationalSimilarity2018,nguyenWideDeepNetworks2020}
However, while most of these methods rely on common (image) datasets to train and evaluate their models, there is no common dataset of neural network models to compare the evaluation methods on.
Model zoos as common evaluation datasets can be a step up to evaluate the evaluation methods.

There are only few publications who use model zoos.
In \citep{liuKnowledgeFlowImprove2019}, zoos of pre-trained models are used as teacher models to train a target model. 
Similarly, \citep{shuZooTuningAdaptiveTransfer2021} propose a method to learn a combination of the weights of models from a zoo for a new task. 
\citep{zhouJittorGANFasttrainingGenerative2021} uses a zoo of GAN models trained with different methods to accelerate GAN training.
To facilitate continual learning, \citep{rameshModelZooGrowing2022} propose to generate zoos of models trained on different tasks or experiences, and to ensemble them for future tasks.\looseness-1

Larger model zoos containing a few thousand models are used in \citep{unterthinerPredictingNeuralNetwork2020} to predict the accuracy of the models from their weights. Similarly, \citep{eilertsenClassifyingClassifierDissecting2020} use zoos of larger models to predict hyperparameters from the weights.
In \citep{gavrikovCNNFilterDB2022}, a large collection of 3x3 convolutional filters trained on different datasets is presented and analysed.
Other work identifies structures in the form of subspaces with beneficial properties \citep{lucasAnalyzingMonotonicLinear2021,wortsmanLearningNeuralNetwork2021,bentonLossSurfaceSimplexes2021}.
\citep{schurholtSelfSupervisedRepresentationLearning2021} use zoos to learn self-supervised representations on the weights of the models in the zoo. The authors demonstrate that the learned representations have high predictive capabilities for model properties such as accuracy, generalization gap, epoch and various hyperparameters. Further, they investigate the impact of the generating factors of model zoos on their properties. 
\citep{schurholtHyperRepresentationsPreTrainingTransfer2022,schurholtHyperRepresentationsGenerativeModels2022} demonstrate that learned representations can be instantiated in new models, as initialization for fine-tuning or transfer learning. 
This work systematically extends their zoos to more datasets and architectures.\looseness-1
%

\section{Model Zoo Generation}
\vspace{-1pt}
The proposed model zoo datasets contain systematically generated and diverse populations of neural networks.
Since the applicability of the model zoos for downstream tasks largely depends on the composition and properties 
of the zoos, special care has to be taken in their design and the used protocol for training. 
The entire procedure can be considered as defining and restricting the generating factors of
model zoo training with respect to their latent relation of desired zoo characteristics. 
The described procedure and protocol could be also used as a general blueprint for the generation of model zoos.\looseness-1

In our paper, the term architecture means the structure of an NN, i.e., a set of operations and their connectivity. 
We use 'model' to denote an instantiating of an architecture with weights over all stages of training, 'model state' to denote the model with the specific state of weights at a specific training epoch, and the weights ${\textbf{w}}$ to denote all trainable parameters (weights and biases).\looseness-1

%
%
\subsection{Model Zoo Design}
\label{sec:model_zoos:generation}

\vspace{-3pt}
\paragraph*{Generating Factors}
Following~\citep{unterthinerPredictingNeuralNetwork2020}, we define the tuple $\{ \mathcal{D}, \lambda,  \mathcal{A}\}$ as a configuration of a model 
zoo's generating factors. We denote the dataset of image samples with their corresponding labels as $\mathcal{D}$. 
The NN architecture is denoted by $\mathcal{A}$.
We denote the set of hyperparameters used for training, (\textit{e.g.}, loss function, optimizer, learning rate, weight initialization, seed, batch-size, epochs) as $\lambda$. 
While dataset $\mathcal{D}$ and architecture $\mathcal{A}$ are fixed for a model zoo, $\lambda$ provides not only the set of hyperparameters but also configures the ranges for individual hyperparameter such as learning rate for model zoo generation.
Training with such differing configurations $\{\mathcal{D}, \lambda,  \mathcal{A}\}$ results 
in a population of NN models i.e., the model zoo. We convert the weights and biases of each model 
to a vectorized form. In the resulting model zoo $\mathcal{W} = \{ {\textbf{w}}_1, ...., {\textbf{w}}_M \}$, ${\textbf{w}}_i$ 
denotes the flattened vector of the weights and biases of one trained 
NN model from the set of $M$ models of the zoo. \looseness-1

\vspace{-3pt}
\paragraph*{Configurations \& Diversity}
The model zoos have to be representative of real-world models, but also diverse and span an interesting range of properties.
The definition of the diversity of model zoos, as well as the choice of how much diversity to include, 
is as difficult as in image datasets, e.g.~\citep{dengImageNetLargeScaleHierarchical2009,feiConstructionAnalysisLarge2009}.
Model zoos can be diverse in their properties (i.e., performance) as well as in their generating factors $\lambda$, or in their weights $\textbf{w}$. 
We aim to generate model zoos with a rich set of models and diversity in these aspects.
As these zoo properties are effects of the generating factors, we tune the diversity of the generating factors 
and evaluate the diversity in Section \ref{sec:model_zoos:analysis}.\\

%
\begin{table}[ht]
{
\scriptsize
\caption[Model Zoo Generating Factors]{Generating factors of the model zoos. Several values for each parameter define the grid.
\texttt{Arch} denotes the architecture: CNN (s) - small CNN architecture, CNN (m) - medium CNN architecture, RN-18 - ResNet-18.
\texttt{Init} denotes the initialization methods: U - uniform, N - normal, KU - Kaiming Uniform, KN - Kaiming Normal.
\texttt{Activation} denotes the activation function: T - Tanh, S - Sigmoid, R - ReLU, G - GeLU.
\texttt{Optim} denotes the optimizer: AD - Adam, SGD - Stochastic Gradient Descent.
Models with learning rates denoted with * have been trained with a one-cycle LR scheduler, the listed LR is the maximum value.
}
\label{tab:model_zoos:generating_factors}
{
\centering
\setlength{\tabcolsep}{3.4pt}
\begin{tabular}{@{}lllccccccc@{}}
\toprule
Dataset                        & Arch & Config & \texttt{Init}           & \texttt{Activation}     & \texttt{Otpim}     & \texttt{LR}        & \texttt{WD}              & \texttt{Dropout}       & \texttt{Seed}   \\
\cmidrule(r){1-1} \cmidrule(rl){2-2}  \cmidrule(rl){3-3} \cmidrule(rl){4-4} \cmidrule(rl){5-5}  \cmidrule(rl){6-6}  \cmidrule(rl){7-7}  \cmidrule(rl){8-8}  \cmidrule(rl){9-9} \cmidrule(rl){10-10} 
\multirow{3}{*}{MNIST}         & CNN (s) & Seed          & U              & T              & AD          & 3e-4      & 0             & 0           & 1-1000 \\
                               & CNN (s) & Hyp-10-r   & U, N, KU, KN & T, S, R, G     & AD, SGD     & 1e-3,1e-4 & 1e-3, 1e-4      & 0, 0.5      & $\sim 10$     \\
                               & CNN (s) & Hyp-10-f    & U, N, KU, KN & T, S, R, G     & AD, SGD     & 1e-3,1e-4 & 1e-3, 1e-4      & 0, 0.5      & 1-10   \\
\cmidrule(r){1-1} \cmidrule(rl){2-2}  \cmidrule(rl){3-3} \cmidrule(rl){4-4} \cmidrule(rl){5-5}  \cmidrule(rl){6-6}  \cmidrule(rl){7-7}  \cmidrule(rl){8-8}  \cmidrule(rl){9-9} \cmidrule(rl){10-10} 
\multirow{3}{*}{F-MNIST}       & CNN (s) & Seed          & U              & T              & AD          & 3e-4      & 0             & 0           & 1-1000 \\
                               & CNN (s) & Hyp-10-r   & U, N, KU, KN & T, S, R, G     & AD, SGD     & 1e-3,1e-4 & 1e-3, 1e-4      & 0, 0.5      & $\sim 10$        \\
                               & CNN (s) & Hyp-10-f    & U, N, KU, KN & T, S, R, G     & AD, SGD     & 1e-3,1e-4 & 1e-3, 1e-4      & 0, 0.5      & 1-10   \\
\cmidrule(r){1-1} \cmidrule(rl){2-2}  \cmidrule(rl){3-3} \cmidrule(rl){4-4} \cmidrule(rl){5-5}  \cmidrule(rl){6-6}  \cmidrule(rl){7-7}  \cmidrule(rl){8-8}  \cmidrule(rl){9-9} \cmidrule(rl){10-10} 
\multirow{3}{*}{SVHN}          & CNN (s) & Seed          & U              & T              & AD          & 3e-3      & 0             & 0           & 1-1000 \\
                               & CNN (s) & Hyp-10-r   & U, N, KU, KN & T, S, R, G     & AD, SGD     & 1e-3,1e-4 & 1e-3, 1e-4, 0 & 0, 0.3, 0.5 & $\sim 10$        \\
                               & CNN (s) & Hyp-10-f    & U, N, KU, KN & T, S, R, G     & AD, SGD     & 1e-3,1e-4 & 1e-3, 1e-4, 0 & 0, 0.3, 0.5 & 1-10   \\
\cmidrule(r){1-1} \cmidrule(rl){2-2}  \cmidrule(rl){3-3} \cmidrule(rl){4-4} \cmidrule(rl){5-5}  \cmidrule(rl){6-6}  \cmidrule(rl){7-7}  \cmidrule(rl){8-8}  \cmidrule(rl){9-9} \cmidrule(rl){10-10} 
\multirow{3}{*}{USPS}          & CNN (s) & Seed          & U              & T              & AD          & 3e-4      & 1e-3            & 0           & 1-1000 \\
                               & CNN (s) & Hyp-10-r   & U, N, KU, KN & T, S, R, G     & AD, SGD     & 1e-3,1e-4 & 1e-2, 1e-3      & 0, 0.5      & $\sim 10$        \\
                               & CNN (s) & Hyp-10-f    & U, N, KU, KN & T, S, R, G     & AD, SGD     & 1e-3,1e-4 & 1e-2, 1e-3      & 0, 0.5      & 1-10   \\
\cmidrule(r){1-1} \cmidrule(rl){2-2}  \cmidrule(rl){3-3} \cmidrule(rl){4-4} \cmidrule(rl){5-5}  \cmidrule(rl){6-6}  \cmidrule(rl){7-7}  \cmidrule(rl){8-8}  \cmidrule(rl){9-9} \cmidrule(rl){10-10} 
\multirow{3}{*}{CIFAR10}       & CNN (s) & Seed          & KU            & G              & AD          & 1e-4      & 1e-2            & 0           & 1-1000 \\
                               & CNN (s) & Hyp-10-r   & U, N, KU, KN & T, S, R, G     & AD, SGD     & 1e-3      & 1e-2, 1e-3      & 0, 0.5      & $\sim 10$        \\
                               & CNN (s) & Hyp-10-f    & U, N, KU, KN & T, S, R, G     & AD, SGD     & 1e-3      & 1e-2, 1e-3      & 0, 0.5      & 1-10   \\
\cmidrule(r){1-1} \cmidrule(rl){2-2}  \cmidrule(rl){3-3} \cmidrule(rl){4-4} \cmidrule(rl){5-5}  \cmidrule(rl){6-6}  \cmidrule(rl){7-7}  \cmidrule(rl){8-8}  \cmidrule(rl){9-9} \cmidrule(rl){10-10} 
\multirow{3}{*}{CIFAR10}       & CNN (m) & Seed          & KU            & G              & AD          & 1e-4      & 1e-2            & 0           & 1-1000 \\
                               & CNN (m) & Hyp-10-r   & U, N, KU, KN & T, S, R, G     & AD, SGD     & 1e-3      & 1e-2, 1e-3      & 0, 0.5      & $\sim 10$        \\
                               & CNN (m) & Hyp-10-f    & U, N, KU, KN & T, S, R, G     & AD, SGD     & 1e-3      & 1e-2, 1e-3      & 0, 0.5      & 1-10   \\
\cmidrule(r){1-1} \cmidrule(rl){2-2}  \cmidrule(rl){3-3} \cmidrule(rl){4-4} \cmidrule(rl){5-5}  \cmidrule(rl){6-6}  \cmidrule(rl){7-7}  \cmidrule(rl){8-8}  \cmidrule(rl){9-9} \cmidrule(rl){10-10} 
\multirow{3}{*}{STL (s)}       & CNN (s) & Seed          & KU            & T              & AD          & 1e-4      & 1e-3            & 0           & 1-1000 \\
                               & CNN (s) & Hyp-10-r   & U, N, KU, KN & T, S, R, G     & AD, SGD     & 1e-3,1e-4 & 1e-2, 1e-3      & 0, 0.5      & $\sim 10$        \\
                               & CNN (s) & Hyp-10-f    & U, N, KU, KN & T, S, R, G     & AD, SGD     & 1e-3,1e-4 & 1e-2, 1e-3      & 0, 0.5      & 1-10   \\
\cmidrule(r){1-1} \cmidrule(rl){2-2}  \cmidrule(rl){3-3} \cmidrule(rl){4-4} \cmidrule(rl){5-5}  \cmidrule(rl){6-6}  \cmidrule(rl){7-7}  \cmidrule(rl){8-8}  \cmidrule(rl){9-9} \cmidrule(rl){10-10} 
\multirow{3}{*}{STL}       & CNN (m) & Seed          & KU            & T              & AD          & 1e-4      & 1e-3            & 0           & 1-1000 \\
                               & CNN (m) & Hyp-10-r   & U, N, KU, KN & T, S, R, G     & AD, SGD     & 1e-3,1e-4 & 1e-2, 1e-3      & 0, 0.5      & $\sim 10$        \\
                               & CNN (m) & Hyp-10-f    & U, N, KU, KN & T, S, R, G & AD, SGD & 1e-3,1e-4 & 1e-2, 1e-3      & 0, 0.5      & 1-10   \\ 
\cmidrule(r){1-1} \cmidrule(rl){2-2}  \cmidrule(rl){3-3} \cmidrule(rl){4-4} \cmidrule(rl){5-5}  \cmidrule(rl){6-6}  \cmidrule(rl){7-7}  \cmidrule(rl){8-8}  \cmidrule(rl){9-9} \cmidrule(rl){10-10} 
CIFAR10                         & RN-18 & Seed          & KU            & R              & SGD          & 1e-4*      & 5e-4            & 0           & 1-1000 \\
CIFAR100                         & RN-18 & Seed          & KU            & R              & SGD          & 1e-4*      & 5e-4            & 0           & 1-1000 \\
t-Imagenet                         & RN-18 & Seed          & KU            & R              & SGD          & 1e-4*      & 5e-4            & 0           & 1-1000 \\

\bottomrule
\end{tabular}
}
}
\end{table}

%
%
Prior work discusses the impact of random seeds on the properties of model zoos. While~\citep{yakTaskArchitectureIndependentGeneralization2018} 
use multiple random seeds for the same hyperparameter configuration, ~\citep{unterthinerPredictingNeuralNetwork2020} 
explicitly argues against that to prevent information leakage between models from train to test set. 
To achieve diverse model zoos and disentangle the generating factors (seeds and hyperparameters), we train model zoos in three different configurations, some with random seeds, others with fixed seeds.

\paragraph*{\texttt{Random Seeds:}} The first configuration, denoted as \texttt{Hyp-10-rand}, varies a broad range of hyperparameters to define a grid of hyperparameters. 
To include the effect of different random initializations, each of the hyperparameter nodes in the grid is repeated with ten randomly drawn seeds. 
One model is configured with the combination of hyperparameters and seed, with a total of ten models per hyperparameter node. 
It is very unlikely for two models in the zoo share the same random seed.
With this, we achieve the highest amount of diversity in properties, generating factors and weights. 

\paragraph*{\texttt{Fixed Seeds:}}
The second configuration, denoted as \texttt{Hyp-10-fix}, uses the same hyperparameter grid as but repeats each node with ten fixed seeds $[1,2,...,10]$. Fixing the seeds allows evaluation methods to control for the seed, isolate the influence of hyperparameter choices, and still get robust results over 10 repetitions. A side effect of the (desired) isolation of factors of influence is, that fixing the seeds leads to repetitions of the starting point in weight space for models with the same seed and initialization methods. At the beginning of the training, these models may have similar trajectories.

\paragraph*{\texttt{Fixed Hyperparameters:}}
For the third configuration, denoted as \texttt{Seed}, we fix one set of hyperparameters and repeat that with 1000 different seeds. With that, we achieve zoos that are very diverse in weights and covers a broad range in weight space. These zoos can be used to evaluate the impact of weights and their starting point on model performance. The hyperparameters for the \texttt{Seed} zoos are chosen such that there is still a level of diversity in model performance.\looseness-1

%
%
\subsection{Specification of Generating Factors for Model Zoos}
This section describes the systematic specification of the trained model zoos. Multiple generating factors define a configuration $\{\mathcal{D}, \lambda, \mathcal{A}\}$ for the model zoo generation, detailed in Table \ref{tab:model_zoos:generating_factors}.

\vspace{10pt}
\paragraph*{Datasets $\mathcal{D}$:}
We generate model zoos for the following image classification datasets: \texttt{MNIST}~\citep{lecunGradientbasedLearningApplied1998}, \texttt{Fashion-MNIST}~\citep{xiaoFashionMNISTNovelImage2017}, \texttt{SVHN}~\citep{netzerReadingDigitsNatural2011}, \texttt{CIFAR-10}~\citep{krizhevskyLearningMultipleLayers2009},
\texttt{STL-10}~\citep{coatesAnalysisSingleLayerNetworks2011}, \texttt{USPS}~\citep{hullDatabaseHandwrittenText1994}, \texttt{CIFAR-100} \citep{krizhevskyLearningMultipleLayers2009} and \texttt{Tiny Imagenet} \citep{leTinyImageNetVisual2015}.
%
%
%
%

\paragraph*{Hyperparameter $\lambda$:}
varied hyperparameters to train models in zoos are: (1) \texttt{seed}, (2) \texttt{initialization method}, (3) \texttt{activation function}, 
(4) \texttt{dropout}, (4) \texttt{optimization algorithm}, (5) \texttt{learning rate}, and (6) \texttt{weight decay}. The batch size and number of training epochs are kept constant within zoos.
%

%
%
%
\paragraph*{Architecture $\mathcal{A}$:}
To preserve the comparability within a model zoo, each zoo is generated using a single neural 
network architecture. One of three standard architectures is used to generate each zoo. 
Our intention with this dataset is similar to research communities such as Neural Architecture Search (NAS), Meta-Learning or Continual Learning (CL), where initial work started small scale
\citep{zhmoginovHyperTransformerModelGeneration2022,rameshModelZooGrowing2022}. 
Hence, the first two architectures are a small and a slightly larger Convolutional Neural Network (CNN), both have three convolutional and two fully-connected layers, but different numbers of channels (details in Appendix \ref{app:zoo_generation}). The third architecture is a standard ResNet-18 \citep{heDeepResidualLearning2016}.
The (1) \texttt{small CNN} has a total of $2'464$-$2'864$ 
parameters, the (2) \texttt{medium CNN} has $10'853$ parameters, the (3) ResNet-18 has 11.2M-11.3M parameters. 

Compared to (1), the medium architecture (2) provides additional diversity to the collection of model zoos and performs significantly better on more complex datasets \texttt{CIFAR-10} and \texttt{STL-10}. 
These architectures are similar to the one used  in~\citep{schurholtHyperRepresentationsGenerativeModels2022}. 
The ResNet-18 architecture is included to apply the model zoo blueprint to models of the widely used ResNet family and so facilitate research on populations of real-world-sized models.\looseness-1

\begin{figure}[]
\begin{minipage}[t]{1.00\textwidth}
\begin{center}
\includegraphics[trim=0in 0in 0in 0in, width=0.85\linewidth]{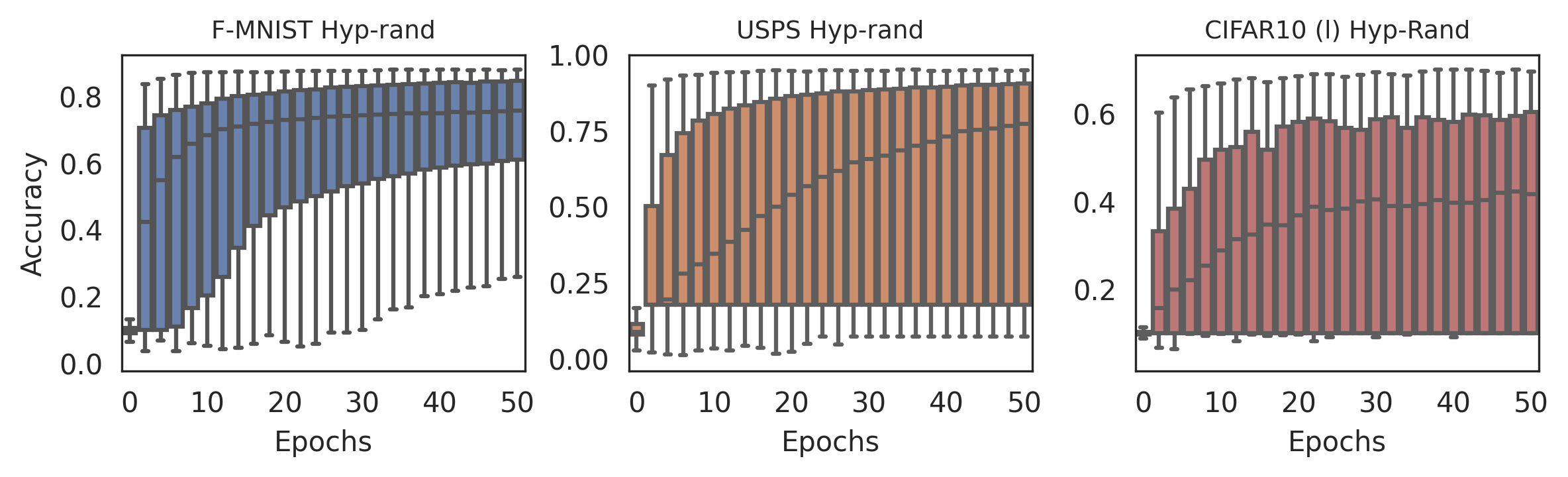}
\vspace{-8pt}
\caption[Model Zoos Accuracy over Training Epochs]{
Accuracy distribution over epochs for the \texttt{F-MNIST Hyp-rand}, \texttt{USPS Hyp-rand}, and \texttt{CIFAR Hyp-rand} zoos. All zoos show training progress and considerable performance diversity.\looseness-1
}

\label{fig:model_zoos:boxplots}    
\end{center}
\end{minipage}
\end{figure}

%
%
\subsection{Training of Model Zoos}
Neural network models are trained from the previously defined three configurations 
$\{\mathcal{D}, \lambda, \mathcal{A}\}$ (Seed, Hyp-10-rand, Hyp-10-fix, see Sec 2.1). With the \textit{8 image datasets} and the three configurations, 
this results in \textit{27 model zoos}.
%
%
The zoos include a total of around \textit{50'360} unique neural network models.\looseness-1

\paragraph*{Training Protocol:}
Every model in the collection of zoos is trained according to the same protocol. 
We keep the same train, validation, and test splits for each zoo, and train each model for 50 epochs with gradient descent methods (SGD+momentum or ADAM). 
At every epoch, the model checkpoint as well as accuracy and loss of all splits are recorded. 
Validation and test performance are also recorded before the first training epoch.
This makes 51 checkpoints per model training trajectory including the starting checkpoint 
representing the model initialization before training starts. 
The ResNet-18 zoos on CIFAR100 and Tiny Imagenet require more updates and are trained for 60 epochs.
In total, this results in a set of \textit{2’585’360} collected model states.\looseness-1

\paragraph*{Splits:}
To enable comparability, this set of models is split into \texttt{training} (70\%), \texttt{validation} (15\%), and \texttt{test} (15\%) subsets. 
This split is done such that all individual checkpoints of one model training (i.e., the 51 checkpoints per training) 
is entirely in either \texttt{training}, \texttt{validation}, or \texttt{test} and therefore no information is leaked between these subsets.\looseness-1

\paragraph*{Sparsified Model Zoo Twins:}
Model sparsification is an effective method to reduce the computational cost of models. 
However, methods to sparsify models to a high degree while preserving the performance are still actively researched~\citep{hoeflerSparsityDeepLearning2021}. 
In order to allow systematic studies of sparsification, we are extending the model zoos with sparsified \textit{model zoo twins} serving as counterparts to existing zoos in the dataset. Using Variational Dropout (VD)~\citep{molchanovVariationalDropoutSparsifies2017}, we sparsify the trained models from existing model zoos. VD generates a sparsification trajectory for each model, along which we track the performance, degree of sparsity, and the sparsified checkpoint. With 25 sparsification epochs, this yields 1'259'000 sparsification model states.

%
%
%
%
%
%
%
\subsection{Data Management and Accessibility of Model Zoos}
\label{sec:model_zoos:data_management}
The model zoos are made publicly available in an accessible, standardized, and well-documented way to the research community under the Creative Commons Attribution 4.0 license (CC-BY 4.0). 
We ensure the technical accessibility of the data by hosting it on Zenodo, where the data will be hosted for at least 20 years.
Further, we take steps to reduce access barriers by providing code for data loading and preprocessing, to reduce the friction associated with analyzing the raw zoo files.
All code can be found on the model zoo website \href{www.modelzoos.cc}{www.modelzoos.cc}.
To ensure conceptional accessibility, we include detailed insights, visualizations, and the analysis of the model zoo (Sec. \ref{sec:model_zoos:analysis}) with each zoo.
Further details can be found in Appendix \ref{app:data_management}.\looseness-1 
%

%
%
%


\section{Model Zoo Analysis}
\label{sec:model_zoos:analysis}
\begin{table}[ht]
{
\centering
\scriptsize
\caption[Model Zoo Diversity Analysis]{Analysis of the diversity of our 27 model zoos (one row per zoo). Mean (std) values in \% per zoo, computed on the last epoch. Agreement is computed using samples from the test split of the image dataset pairwise over the entire zoo. Higher agreement values indicate more uniform behavior and less behavioral diversity. Distance in weight space are computed pairwise over the entire zoo. Higher distance values indicate larger diversity in weight space.}
\label{tab:model_zoos:analysis}
{
\begin{center}
\centering
\setlength{\tabcolsep}{3.4pt}
\begin{tabular}{@{}lllcccccc@{}}
\toprule
                               &               &               & Performance & \multicolumn{2}{c}{Agreement}    & \multicolumn{3}{c}{Weights}               \\ 
\cmidrule(rl){4-4} \cmidrule(rl){5-6} \cmidrule(rl){7-9} \\
Dataset                        & Architecture  & Config & Accuracy    & $\kappa_{aggr}$ & $\kappa_{cka}$ & $\mathbf{w}$           & l2-dist       & cos dist    \\
\cmidrule(r){1-1} \cmidrule(rl){2-2}  \cmidrule(rl){3-3} \cmidrule(rl){4-4} \cmidrule(rl){5-6} \cmidrule(rl){7-9}
\multirow{3}{*}{MNIST}         & CNN (s)   & Seed          & 91.1 (0.9)  & 88.5 (1.3)      & 77.2 (5.2)     & 18.9 (58.4) & 124.1 (4.9)   & 77.1 (4.1)  \\
                               & CNN (s)   & Hyp-10-r   & 79.9 (30.7) & 67.7 (35.5)     & 58.6 (25.9)    & 0.4 (46.5)  & 150.6 (66.5)  & 98.8 (7.2)  \\
                               & CNN (s)   & Hyp-10-f    & 80.3 (30.3) & 68.3 (35.3)     & 58.8 (25.7)    & 0.3 (46.7)  & 149.7 (66.8)  & 97.7 (10.0) \\
\cmidrule(r){1-1} \cmidrule(rl){2-2}  \cmidrule(rl){3-3} \cmidrule(rl){4-4} \cmidrule(rl){5-6} \cmidrule(rl){7-9}
\multirow{3}{*}{F-MNIST}       & CNN (s)   & Seed          & 72.7 (1.0)  & 79.8 (2.6)      & 82.3 (12.6)    & 22.6 (55.6) & 122.0 (4.9)   & 74.5 (4.4)  \\
                               & CNN (s)   & Hyp-10-r   & 68.4 (23.7) & 59.9 (29.1)     & 64.6 (23.5)    & 1.0 (46.0)  & 149.6 (62.2)  & 99.2 (6.8)  \\
                               & CNN (s)   & Hyp-10-f    & 68.7 (23.4) & 60.4 (28.7)     & 64.6 (22.7)    & 0.9 (46.3)  & 148.5 (61.9)  & 97.9 (9.9)  \\
\cmidrule(r){1-1} \cmidrule(rl){2-2}  \cmidrule(rl){3-3} \cmidrule(rl){4-4} \cmidrule(rl){5-6} \cmidrule(rl){7-9}
\multirow{3}{*}{SVHN}          & CNN (s)   & Seed          & 71.1 (8.0)  & 67.2 (10.3)     & 67.7 (15.7)    & 7.1 (113.7) & 137.6 (8.3)   & 94.5 (5.1)  \\
                               & CNN (s)   & Hyp-10-r   & 35.9 (24.3) & 61.6 (35.9)     & 17.8 (28.0)    & 1.4 (42.2)  & 170.5 (149.4) & 83.6 (30.4) \\
                               & CNN (s)   & Hyp-10-f    & 36.0 (24.4) & 61.4 (36.0)     & 18.1 (27.9)    & 1.3 (42.2)  & 170.0 (149.0) & 83.2 (30.7) \\
\cmidrule(r){1-1} \cmidrule(rl){2-2}  \cmidrule(rl){3-3} \cmidrule(rl){4-4} \cmidrule(rl){5-6} \cmidrule(rl){7-9}
\multirow{3}{*}{USPS}          & CNN (s)   & Seed          & 87.0 (1.7)  & 87.3 (2.2)      & 86.7 (6.3)     & 8.2 (26.9)  & 123.1 (5.2)   & 75.9 (5.0)  \\
                               & CNN (s)   & Hyp-10-r   & 64.7 (30.8) & 55.3 (31.4)     & 50.9 (30.5)    & 2.1 (39.6)  & 155.5 (92.6)  & 99.1 (8.9)  \\
                               & CNN (s)   & Hyp-10-f    & 65.0 (30.7) & 55.4 (31.3)     & 50.4 (30.4)    & 1.9 (40.1)  & 154.2 (93.1)  & 97.3 (13.7) \\
\cmidrule(r){1-1} \cmidrule(rl){2-2}  \cmidrule(rl){3-3} \cmidrule(rl){4-4} \cmidrule(rl){5-6} \cmidrule(rl){7-9}
\multirow{3}{*}{CIFAR10}      & CNN (s)   & Seed          & 48.7 (1.4)  & 65.7 (3.1)      & 72.9 (11.3)    & 1.1 (11.0)  & 138.7 (5.6)   & 96.3 (5.1)  \\
                               & CNN (s)   & Hyp-10-r   & 35.1 (16.3) & 33.3 (22.9)     & 47.5 (34.0)    & -0.2 (17.0) & 155.6 (71.0)  & 97.5 (10.8) \\
                               & CNN (s)   & Hyp-10-f    & 35.1 (16.2) & 33.3 (22.8)     & 47.3 (34.2)    & -0.2 (16.9) & 155.3 (70.0)  & 97.2 (11.1) \\
\cmidrule(r){1-1} \cmidrule(rl){2-2}  \cmidrule(rl){3-3} \cmidrule(rl){4-4} \cmidrule(rl){5-6} \cmidrule(rl){7-9}
\multirow{3}{*}{CIFAR10}      & CNN (m)   & Seed          & 61.5 (0.7)  & 76.0 (1.6)      & 92.4 (1.7)     & 0.1 (18.2)  & 137.0 (7.9)   & 94.1 (9.2)  \\
                               & CNN (m)   & Hyp-10-r   & 39.6 (21.8) & 34.5 (27.1)     & 43.2 (36.5)    & -0.4 (23.0) & 158.9 (79.9)  & 98.6 (12.2) \\
                               & CNN (m)   & Hyp-10-f    & 39.6 (21.7) & 34.4 (26.7)     & 42.8 (37.8)    & -0.4 (22.9) & 158.1 (77.2)  & 98.0 (13.1) \\
\cmidrule(r){1-1} \cmidrule(rl){2-2}  \cmidrule(rl){3-3} \cmidrule(rl){4-4} \cmidrule(rl){5-6} \cmidrule(rl){7-9}
\multirow{3}{*}{STL}           & CNN (s)   & Seed          & 39.0 (1.0)  & 48.4 (3.0)      & 81.5 (3.9)     & -0.1 (19.1) & 141.2 (5.0)   & 99.8 (4.2)  \\
                               & CNN (s)   & Hyp-10-r   & 23.1 (12.3) & 23.4 (20.9)     & 39.0 (30.7)    & 3.0 (40.0)  & 158.7 (107.3) & 98.7 (10.9) \\
                               & CNN (s)   & Hyp-10-f    & 23.0 (12.2) & 23.3 (21.1)     & 38.1 (30.0)    & 3.0 (39.8)  & 157.1 (107.2) & 96.8 (16.3) \\
\cmidrule(r){1-1} \cmidrule(rl){2-2}  \cmidrule(rl){3-3} \cmidrule(rl){4-4} \cmidrule(rl){5-6} \cmidrule(rl){7-9}
\multirow{3}{*}{STL}           & CNN (m)   & Seed          & 47.4 (0.9)  & 53.9 (2.2)      & 83.3 (2.3)     & 0.1 (26.6)  & 141.3 (6.0)   & 99.9 (5.8)  \\
                               & CNN (m)   & Hyp-10-r   & 24.3 (14.7) & 23.2 (24.2)     & 34.1 (30.0)    & 2.3 (45.7)  & 159.3 (103.0) & 99.1 (12.5) \\
                               & CNN (m)   & Hyp-10-f    & 24.4 (14.7) & 23.7 (24.5)     & 34.6 (30.3)    & 2.3 (46.5)  & 157.4 (104.1) & 97.6 (20.1) \\
\cmidrule(r){1-1} \cmidrule(rl){2-2}  \cmidrule(rl){3-3} \cmidrule(rl){4-4} \cmidrule(rl){5-6} \cmidrule(rl){7-9}
CIFAR10                       & ResNet-18     & Seed          & 92.1 (0.2)  & 93.4 (0.7)      & --.- (-.-)     & -0.01 (1.7)  & 122.1 (3.9)   & 72.2 (2.3)  \\
CIFAR100                      & ResNet-18     & Seed          & 74.2 (0.3)  & 77.6 (1.2) & --.- (-.-)  &  -0.1 (1.6) & 130.8 (4.1) & 83.1 (2.6)  \\
Tiny ImageNet                  & ResNet-18     & Seed         & 63.9 (0.7)  & 66.1 (1.9) &  --.- (-.-)  &  -0.1  (1.9)  & 125.4 (4.9) & 77.1 (3.0)  \\
\bottomrule
\end{tabular}
\end{center}
}
}

\end{table}

The model zoos have been created aiming at diversity in generating factors, weights, and performance. In this section, we analyze the zoos and their properties. Zoo cards with key values and visualizations are provided along with the zoos online. We consider models at their last epoch for the analysis.
For all later analyses, non-viable checkpoints are excluded from each zoo. This includes the removal of every checkpoint with NaN values or values beyond a threshold. The threshold value is set for each zoo, such that it only excludes diverging models.


\paragraph*{Performance}
To investigate the performance diversity, we consider the accuracy of the models in the zoo, see Table \ref{tab:model_zoos:analysis} and Figure \ref{fig:model_zoos:boxplots}.
As expected, the zoos with variation only in the seed show the smallest variation in performance. Changing the hyperparameters induces a broader range of variation. Changing (Hyper-10-rand) or fixing  (Hyper-10-fix) the seeds does not affect the accuracy distribution.\looseness-1

\paragraph*{Model Agreement}
To get more in-depth insights into the diversity of model behavior, we investigate their pairwise agreement, see Table \ref{tab:model_zoos:analysis}. 
To that end, we compute the rate of agreement of class prediction between two models as $\kappa_{aggr} = \frac{1}{N}\sum_{1=1}^{N} \delta_{y_i}$. Here $y_i^k, y_i^l$ are the predictions of models $k,l$ for sample $i$ of $N$ samples. Further, $\delta_{y_i}=1$ if $y_i^k=y_i^l$ and otherwise $\delta_{y_i}=0$.
Further, we compute the pairwise centered kernel alignment (cka) score between intermediate and last layer outputs and denote it as $\kappa_{cka}$. The cka score evaluates the correlation of activations, compensating for equivariances typical for neural networks ~\citep{nguyenWideDeepNetworks2020}. 
In empirical evaluations, we found the cka score robust for a relatively small number of image samples, and compute the score using 50 images to reduce the computational load. 
Both agreement metrics confirm the expectation and performance results. Zoos with higher overall performance naturally have a higher agreement on average, as there are fewer samples on which to disagree. 
Zoos with varying hyperparameters(Hyp-10-rand and Hyp-10-fix) agree less on average than zoos with changes in seed only (Seed). What is more, the distribution of $\kappa_{aggr}$ and $\kappa_{cka}$ in the \texttt{Seed} zoos is unimodal and approximately Gaussian. In the \texttt{Hyp-10} zoos, the distributions are bi-modal, with one mode around 0.1 (0.0) and the other around 0.9 (0.75) in hard agreement (cka score). In these zoos, models agree to a rather high degree with some models and disagree with others.

\begin{figure}[]
\begin{minipage}[t]{1.0\textwidth}
\begin{center}
\includegraphics[trim=0in 0in 0in 0in, width=0.85\linewidth]{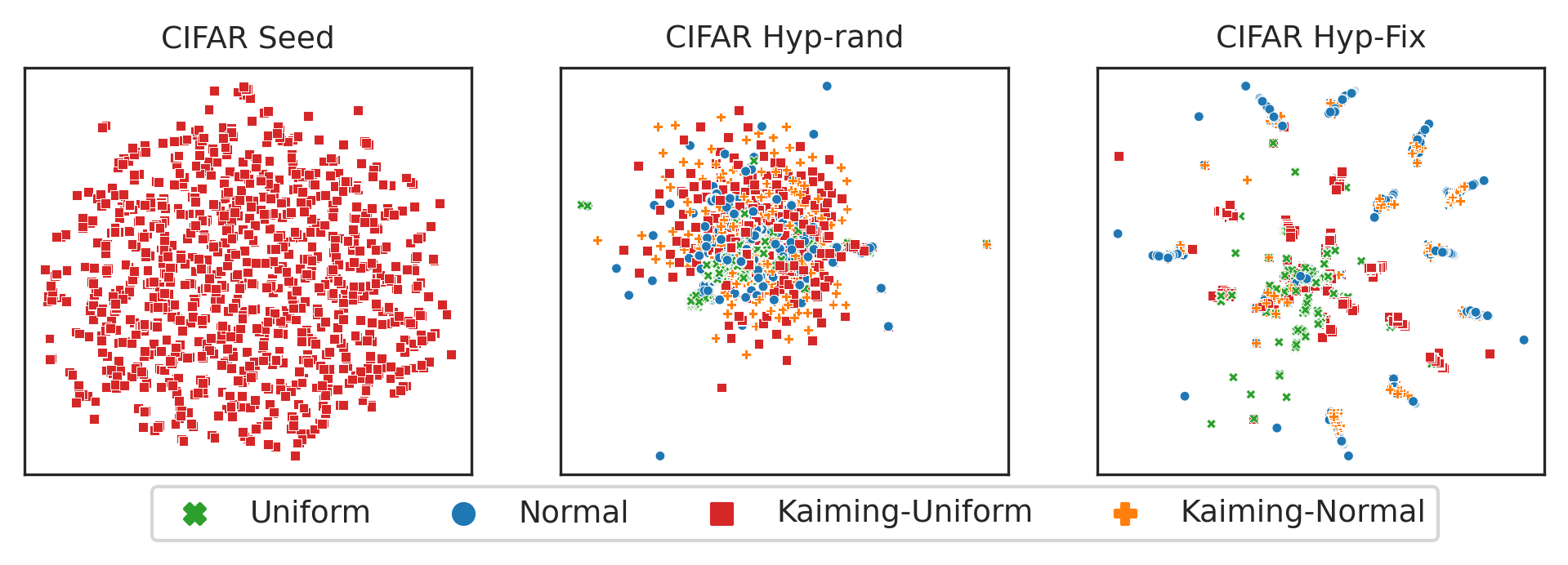}

\caption[UMAP Weights Visualizations of CIFAR CNN Model Zoos]{
Visualization of the weights of the large CIFAR model zoos in different configurations. The weights are reduced to 2d using UMAP, preserving both local and global structure.
In the \texttt{Seed} configuration, the UMAP reduction contains little structure. The \texttt{Hyp-rand} is equally little structured.
In contrast, \texttt{Hyp-fix} contains visible clusters of initialization methods.\looseness-1
}
\label{fig:model_zoos:umap_visualizations}    
\end{center}
\end{minipage}
\end{figure}

\vspace{16pt}
\paragraph*{Weights}
Lastly, we investigate the diversity of the model zoos in weight space, see again Table \ref{tab:model_zoos:analysis}. 
By design, the mean weight value of the zoos varying only in the seed is larger than in the other zoos, while the standard deviation does not differ greatly (Table \ref{tab:model_zoos:analysis}, column w). 
To get a better intuition in the distribution of models in weight space, we compute the pairwise $\ell_2(\mathbf{w_k},\mathbf{w_l})=\frac{\|\mathbf{w_k}-\mathbf{w_l}\|_2^2}{1/N \sum_{n=1}^N\| \mathbf{w_n} \|_2^2}$ and cosine distance $cos(\mathbf{w_k},\mathbf{w_l}) = 1 - \frac{\mathbf{w_l}^T \mathbf{w_k}}{\|\mathbf{w_k}\|_2^2 \| \mathbf{w_l} \|_2^2}$, and investigate their distribution. 
Here, too, varying the hyperparameters introduces higher amounts of diversity, while changing or fixing the seeds does not affect the weight diversity much. As these values are computed at the end of model training, repeated starting points due to fixed seeds appear not to reduce weight diversity significantly.
In a more hands-off approach, we compute 2d reductions of the weight over all epochs using UMAP ~\citep{mcinnesUMAPUniformManifold2018a}. 
In the 2d reductions (see Figure \ref{fig:model_zoos:umap_visualizations}), the zoos varying in seed only show little to no structure. Zoos with hyperparameter changes and random seeds are similarly unstructured. Zoos with varying hyperparameters and fixed seeds show clear clusters with models of the same initialization method and activation function. These findings are further supported by the predictability of the initialization method and activation function (Table \ref{tab:model_zoos:benchmark_prediction}). The structures are unsurprising considering that the activation function is very influential in shaping the loss surface, while the initialization method and the seed determine the starting point on it. Depending on the downstream task, this property can be desirable or should be avoided, which is why we provide both configurations.

\paragraph*{Model Property Prediction}
As a set of benchmark results on the proposed model zoos and to further evaluate the zoos, we use linear models to predict hyperparameters or performance values of the individual models. 
As features, we use the model weights $\mathbf{w}$ or per-layer quintiles of the weights $s(\mathbf{w})$ as in~\citep{unterthinerPredictingNeuralNetwork2020}.
Linear models are used to evaluate the properties of the dataset and the quality of the features.
We report these results in Table \ref{tab:model_zoos:benchmark_prediction}.
The layer-wise weight statistics ($s(\mathbf{w})$) have generally higher predictive performance than the raw weights $\mathbf{w}$. 
In particular, $s(\mathbf{w})$ are not affected by using fixed or random seeds and thus generalize well to unseen seeds.
For the ResNet-18 zoos, $\mathbf{w}$ becomes too large to be used as a feature and is therefore omitted.
Across all zoos, the accuracy as well as the hyperparameters can be predicted very accurately. 
The generalization gap and epoch appear to be more difficult to predict. 
These findings hold for all zoos, regardless of the different architectures, model sizes, task complexity, and performance range.
$\mathbf{w}$ can be used to predict the initialization method and activation function to very high accuracy, if the seeds are fixed. The performance drops drastically if seeds are varied. These results confirm our expectation of diversity in weight space induced by fixing or varying seed.
These results show i) that the model weights of our zoos contain rich information on their properties; ii) confirm the notions of diversity that were design goals for the zoos; and iii) leave room for improvements on the more difficult properties to predict, in particular the generalization gap.\looseness-1

\begin{table}[ht!]
{
\scriptsize
\caption[Model Zoo Benchmark Property Prediction Results]{Benchmark results for predicting model properties from the weights 
($\mathbf{w}$) and layer-wise weight statistics ($s(\mathbf{w})$) using linear models. 
We report the prediction $R^2$ for accuracy, generalization gap (GGap), epoch, learning rate (LR) and dropout (Drop), and prediction accuracy for initialization method (Init) and activation function (Act). 
Values reported in \%, higher values are better.
}
\label{tab:model_zoos:benchmark_prediction}
}
\begin{minipage}{\linewidth}
{
\scriptsize

\setlength{\tabcolsep}{6pt}
\begin{tabular}{@{}lllcccccccccc@{}}

\toprule
                               & &               & \multicolumn{2}{c}{Accuracy} & \multicolumn{2}{c}{GGap} & \multicolumn{2}{c}{Epoch} & \multicolumn{2}{c}{Init} & \multicolumn{2}{c}{Act} \\ 
\cmidrule(rl){4-5} \cmidrule(rl){6-7} \cmidrule(rl){8-9} \cmidrule(rl){10-11} \cmidrule(rl){12-13} 
Dataset                       & Architecture        & Config & $\mathbf{w}$               & $s(\mathbf{w})$         & $\mathbf{w}$             & $s(\mathbf{w})$       & $\mathbf{w}$             & $s(\mathbf{w})$        & $\mathbf{w}$            & $s(\mathbf{w})$        & $\mathbf{w}$           & $s(\mathbf{w})$        \\
\cmidrule(rl){1-1} \cmidrule(rl){2-2} \cmidrule(rl){3-3} \cmidrule(rl){4-5} \cmidrule(rl){6-7} \cmidrule(rl){8-9} \cmidrule(rl){10-11} \cmidrule(rl){12-13} 
\multirow{3}{*}{MNIST}         & CNN (s) & Seed          & 92.3           & 98.7        & 2.1          & 68.8      & 87.2         & 97.8       & n/a            & n/a           &     n/a       &  n/a          \\
                               & CNN (s) & Hyp-10-r   & -11.2          & 69.4        & -49.8        & 13.7      & -95.5        & 14.3       & 42.6        & 77.6       & 45.5       & 78.5       \\
                               & CNN (s) & Hyp-10-f    & 66.5           & 70.1        & 5.4          & 12.5      & -4.8         & 14.5       & 94.3        & 79.8       & 81.2       & 76.8       \\
\cmidrule(rl){1-1} \cmidrule(rl){2-2} \cmidrule(rl){3-3} \cmidrule(rl){4-5} \cmidrule(rl){6-7} \cmidrule(rl){8-9} \cmidrule(rl){10-11} \cmidrule(rl){12-13} 
\multirow{3}{*}{F-MNIST}       & CNN (s) & Seed          & 87.5           & 97.2        & 20.9         & 60.5      & 89.1         & 97.1       &      n/a       &      n/a      &      n/a      &   n/a        \\
                              & CNN (s) & Hyp-10-r   & 8.7            & 76.9        & -47.5        & 13.7      & -70.1        & 18.9       & 48.4        & 81.5       & 47.9       & 79.6      \\
                              & CNN (s) & Hyp-10-f    & 62.4           & 75.6        & 3.9          & 12.6      & -2.0        & 17.0         & 95.4        & 81.6       & 84.6       & 77.7      \\
\cmidrule(rl){1-1} \cmidrule(rl){2-2} \cmidrule(rl){3-3} \cmidrule(rl){4-5} \cmidrule(rl){6-7} \cmidrule(rl){8-9} \cmidrule(rl){10-11} \cmidrule(rl){12-13} 
\multirow{3}{*}{SVHN}          & CNN (s) & Seed          & 91.0             & 98.6        & -42.8        & 65.9      & 66.9         & 92.5       &   n/a          &      n/a      &     n/a       &   n/a        \\
                              & CNN (s) & Hyp-10-r   & -8.6           & 90.3        & -55.3        & 27.6      & -30.5        & 11.1       & 38.2        & 58.5       & 55.7       & 72.3       \\
                              & CNN (s) & Hyp-10-f    & 64.2           & 89.9        & 17.5         & 27.4      & -0.1         & 11.1       & 67.3        & 58.2       & 76.1       & 73.6      \\
\cmidrule(rl){1-1} \cmidrule(rl){2-2} \cmidrule(rl){3-3} \cmidrule(rl){4-5} \cmidrule(rl){6-7} \cmidrule(rl){8-9} \cmidrule(rl){10-11} \cmidrule(rl){12-13} 
\multirow{3}{*}{USPS}          & CNN (s) & Seed          & 92.5           & 98.7        & 44.3         & 71.8      & 86.0         & 98.4       &     n/a        &    n/a        &       n/a     &    n/a       \\
                              & CNN (s) & Hyp-10-r   & -11.5          & 70.3        & -35.2        & 13.6      & -75.7        & 21.3       & 49.2        & 88.8       & 43.7       & 66.2      \\
                              & CNN (s) & Hyp-10-f    & 73.2           & 70.8        & 10.8         & 14.7      & 18.9         & 23.0         & 96.3        & 88.1       & 74.5       & 72.7       \\
\cmidrule(rl){1-1} \cmidrule(rl){2-2} \cmidrule(rl){3-3} \cmidrule(rl){4-5} \cmidrule(rl){6-7} \cmidrule(rl){8-9} \cmidrule(rl){10-11} \cmidrule(rl){12-13} 
\multirow{3}{*}{CIFAR10} & CNN (s) & Seed          & 75.3           & 96.0          & 27.0           & 90.2      & 68.6         & 91.1       &     n/a        &        n/a    &    n/a        &   n/a        \\
                              & CNN (s) & Hyp-10-r   & 50.1           & 88.0          & -4.3         & 40.5      & -2.7         & 34.2       & 34.0          & 50.5       & 71.5       & 80.9       \\
                              & CNN (s) & Hyp-10-f    & 67.0             & 87.9        & 38.2         & 42.9      & 27.0           & 31.8       & 72.0          & 52.2       & 75.6       & 80.0        \\
\cmidrule(rl){1-1} \cmidrule(rl){2-2} \cmidrule(rl){3-3} \cmidrule(rl){4-5} \cmidrule(rl){6-7} \cmidrule(rl){8-9} \cmidrule(rl){10-11} \cmidrule(rl){12-13} 

\multirow{3}{*}{CIFAR10} & CNN (l) & Seed          & 83.6           & 98.2        & 33.4         & 92.9      & 86.5         & 95.7       &     n/a        &        n/a    &  n/a          &       n/a    \\
                              & CNN (l) & Hyp-10-r   & 32.6           & 90.5        & -0.9         & 47        & -10.5        & 35.5       & 41.6        & 51.6       & 69.1       & 83.1     \\
                              & CNN (l) & Hyp-10-f    & 64.5           & 91.4        & 30.4         & 40.7      & 29.8         & 35.3       & 74.5        & 54.9       & 77.7       & 86.0       \\
\cmidrule(rl){1-1} \cmidrule(rl){2-2} \cmidrule(rl){3-3} \cmidrule(rl){4-5} \cmidrule(rl){6-7} \cmidrule(rl){8-9} \cmidrule(rl){10-11} \cmidrule(rl){12-13} 
\multirow{3}{*}{STL}   & CNN (s) & Seed          & 17.8           & 91.2        & 2.0            & 30.2      & 45.3         & 95.0         &      n/a       &       n/a     &    n/a        &     n/a      \\
                              & CNN (s) & Hyp-10-r   & -8.7           & 77.1        & -44.0          & 9.3       & -68.8        & 19.1       & 41.3        & 93.9       & 46.3       & 66.8     \\
                              & CNN (s) & Hyp-10-f    & 76.1           & 76.5        & 6.7          & 10.7      & 21.2         & 22.4       & 98.1        & 91.3       & 78.1       & 62.6      \\
\cmidrule(rl){1-1} \cmidrule(rl){2-2} \cmidrule(rl){3-3} \cmidrule(rl){4-5} \cmidrule(rl){6-7} \cmidrule(rl){8-9} \cmidrule(rl){10-11} \cmidrule(rl){12-13} 
\multirow{3}{*}{STL}       & CNN (l) & Seed          & -112         & 94.2        & 2.8          & 37.3      & 5.6          & 98.7       &        n/a     &      n/a      &      n/a      &   n/a        \\
                              & CNN (l) & Hyp-10-r      & -79.6          & 74.1        & -118       & 10.7      & -106         & 18.8       & 43.8        & 90.4       & 49.4       & 68.3     \\
                              & CNN (l) & Hyp-10-f      & 84.1           & 77.7        & 10.4         & 11.7      & 14.6         & 19.1       & 97.8        & 92.8       & 78.8       & 68.0      \\
\cmidrule(rl){1-1} \cmidrule(rl){2-2} \cmidrule(rl){3-3} \cmidrule(rl){4-5} \cmidrule(rl){6-7} \cmidrule(rl){8-9} \cmidrule(rl){10-11} \cmidrule(rl){12-13} 
CIFAR10                       & ResNet-18 & Seed          & --.- & 96.8 & --.- & 76.7 & --.- & 99.6 & n/a & n/a & n/a & n/a\\
CIFAR100                      & ResNet-18 & Seed          & --.- & 97.4 & --.- & 95.4 & --.- & 99.9 & n/a & n/a & n/a & n/a\\
t-ImageNet                    & ResNet-18 & Seed          & --.- & 96.1 & --.- & 87.5 & --.- & 99.9 & n/a & n/a & n/a & n/a\\
\bottomrule
\end{tabular}
}
\end{minipage}
\end{table}

\section{Potential use cases \& Applications}
\label{sec:model_zoos:usecases}
While populations of NNs have been used in previous work, they still are relatively novel as a dataset. As use cases for such datasets may not be obvious, this section presents potential use cases and applications. For all use cases, we collect related work that uses model populations. Here, the zoos may be used as data or to evaluate the methods. For some of the use cases, the analysis above provides support. Lastly, we suggest ideas for future work that we hope can inspire the community to make use of the model zoos. 

\subsection{Model Analysis}
The analysis of trained models is an important and difficult step in the machine learning pipeline. 
Commonly, models are applied on hold-out test sets, which may contain difficult cases with specific properties~\citep{lecunDeepLearning2015}. 
Other approaches identify subsections of input data that are relevant for a specific output ~\citep{yosinskiUnderstandingNeuralNetworks2015,karpathyVisualizingUnderstandingRecurrent2015,zintgrafVisualizingDeepNeural2017}. 
A third group of methods compares the activations of models, e.g. the cka method used in Sec. \ref{sec:model_zoos:analysis} to measure diversity~\citep{kornblithSimilarityNeuralNetwork2019}.

Populations of models have been used to identify commonalities in model weights, activations, or graph structure which are predictive for model properties. 
Some methods use the weights, weight statistics or eigenvalues of the weight matrices as features to predict a model's accuracy or hyper-parameters ~\citep{unterthinerPredictingNeuralNetwork2020,eilertsenClassifyingClassifierDissecting2020,martinTraditionalHeavyTailedSelf2019}. Recently, ~\citep{schurholtSelfSupervisedRepresentationLearning2021} have learned self-supervised representation of the weights and demonstrate their usefulness for predicting model properties. Other publications use activations to approximate intermediate margins ~\citep{yakTaskArchitectureIndependentGeneralization2018, jiangPredictingGeneralizationGap2019} or graph connectivity features ~\citep{corneanuComputingTestingError2020} to predict the generalization gap or test accuracy. 
Standardized, diverse model zoos may facilitate the development of new methods, or be used as evaluation datasets for existing model analysis, interpretability, or comparison methods.
 
Previous work as well as the experiment results in Sec \ref{sec:model_zoos:analysis} indicate that even more complex model properties might be predicted from the weights. By studying populations of models, in-depth diagnostics of models, such as whether a model learned a specific bias, may be based on the weights or topology of models. 
Lastly, model properties as well as the weights may be used to derive a model 'identity' along the training trajectory, to allow for NN versioning. \looseness-1

\subsection{Learning Dynamics}
Analyzing and utilizing the learning dynamics of models has been a useful practice. 
For example, early stopping~\citep{FinnoffImprovingModelSelection1993}, which determines when to end training at minimal generalization error based on a cross-validation set has become standard in machine learning practice. 

More recently, methods have exploited zoos of models. Population-based training~\citep{jaderbergPopulationBasedTraining2017} evaluates the performance of model candidates in a population, and decides which of the candidates to pursue further and which to give up. HyperBand evaluates performance metrics for groups of models to optimize hyperparameters \citep{liHyperbandNovelBanditBased2018, liSystemMassivelyParallel2020}.
Research in Neural Architecture Search was greatly simplified by the NASBench dataset family \citep{yingNASBench101ReproducibleNeural2019}, which contains performance metrics for varying hyperparameter choices. Our model zoos extend these datasets by adding models including their weights at states throughout training, which may open new doors for new approaches.

The accuracy distribution of our model zoos becomes relatively broad if hyperparameters are varied (Figure \ref{fig:model_zoos:boxplots}).
For early stopping or population-based methods, identifying a good range of hyperparameters to try, and then identifying those candidates that will perform best towards the end of training, is a challenging and relevant task. 
Our model zoos may be used to develop and evaluate methods to that end.
Beyond that, diverse model zoos offer the opportunity to make further steps of understanding and exploiting the learning dynamics of models, i.e., by studying the regularities of generalizing and overfitting models. 
The shape and curvature of training trajectories may contain rich information on the state of model training.
Such information could be used to monitor model training or adjust hyperparameters to achieve better results.
The sparsified model zoos add several potential use cases. They may be used to study the sparsification performance on a population level, study emerging patterns of populations of sparse models, or the relation of full models and their sparse counterparts. 
\looseness-1

\subsection{Representation Learning}
NN models have grown in recent years and with them the dimensionality of their parameter space. 
Empirically, it is more effective to train large models to high performance and distill them in a second step, than to directly train the small models~\citep{hoeflerSparsityDeepLearning2021,liuWeActuallyNeed2021}.
This and other related problems raise interesting questions. What are useful regularities in NN weights? How can the weight space be navigated in a more efficient way?

Recent work has attempted to learn lower dimensional representations of the weights of NNs~\citep{haHyperNetworks2017,ratzlaffHyperGANGenerativeModel2019,zhangGraphHyperNetworksNeural2019,knyazevParameterPredictionUnseen2021,schurholtSelfSupervisedRepresentationLearning2021,schurholtHyperRepresentationsGenerativeModels2022,schurholtHyperRepresentationsPreTrainingTransfer2022}. 
Such representations can reveal the latent structure of NN weights.
Other approaches identify subspaces in the weight space that relate to high performance or generalization \citep{wortsmanLearningNeuralNetwork2021,lucasMonotonicLinearInterpolation2021,bentonLossSurfaceSimplexes2021}.
In~\citep{schurholtSelfSupervisedRepresentationLearning2021}, representations learned on model zoos achieve higher performance in predicting model properties than weights or weight statistics. 
~\citep{knyazevParameterPredictionUnseen2021} proposes a method to learn from a population of diverse neural architectures to generate weights for unseen architectures in a single forward pass.

Our model zoos can be either a dataset to train representations on as in~\citep{schurholtSelfSupervisedRepresentationLearning2021} or \citep{bentonLossSurfaceSimplexes2021}, or as a common dataset to validate such methods.
Learned representations may bring a better understanding of the weight space and thus help to reduce the computational cost and improve the performance of NNs.\looseness-1

\subsection{Generating New Models}
In conventional machine learning, models are randomly initialized and then trained on data. 
As that procedure may require large amounts of data and computational resources, fine-tuning and transfer learning are more efficient training approaches that re-use already trained models for a different task or dataset~\citep{yosinskiHowTransferableAre2014,fengTransferredDiscrepancyQuantifying2020}.
%
Other publications have extended the concept of transfer learning from a one-to-one setup to many-to-one setups~\citep{liuKnowledgeFlowImprove2019,shuZooTuningAdaptiveTransfer2021}. 
Both approaches attempt to combine learned knowledge from several source models into a single target model. 
Most recently, ~\citep{schurholtHyperRepresentationsGenerativeModels2022,schurholtHyperRepresentationsPreTrainingTransfer2022} have generated unseen NN models with desirable properties from representations learned on model zoos. The generated models were able to outperform random initialization and pretraining in transfer-learning regimes. In \citep{peeblesLearningLearnGenerative2022}, a transformer is trained on a population of models with diffusion to generate model weights.\looseness-1

All these approaches require suitable and diverse models to be available. Further, the exact properties of models suitable for generative use, transfer learning, or ensembles are still in discussion~\citep{fengTransferredDiscrepancyQuantifying2020}.
Population-based transfer learning methods such as zoo-tuning \citep{shuZooTuningAdaptiveTransfer2021}, knowledge flow \citep{liuKnowledgeFlowImprove2019} or model-zoo \citep{rameshModelZooGrowing2022} have been demonstrated on populations with only a few models. Populations for these methods ideally are as diverse as possible, so that they provide different features. 
Investigating the models in the proposed zoos may help identify models that lend themselves to transfer learning or ensembling.\looseness-1

\section{Conclusion}
To enable the investigation of populations of neural network models, we release a novel dataset of model zoos with this work. 
These model zoos contain systematically generated and diverse populations of 50'360 neural network models comprised of 3'844'360 collective model states. 
The released model zoos come with a comprehensive analysis and initial benchmarks for multiple downstream tasks and invite further work in the direction of the following use cases: (i) model analysis, (ii) learning dynamics, (iii) representation learning and (iv) model generation.

\begin{subappendices}
%
\section{Model Zoo Generation Details}
\label{app:zoo_generation}
In our model zoos, we use three architectures. Two of them rely on a general CNN architecture, the third is a common ResNet-18\citep{heDeepResidualLearning2016}. 
For the first two architectures, use the general CNN architecture in two sizes, detailed in Table \ref{tab:model_zoos:model_zoo_architecture}. 
By varying different generating factors listed in Table \ref{tab:model_zoos:generating_factors}, we create a grid of configurations, where each node represents a model. 
Each node is instantiated as a model and trained with the exact same training protocol. 
We chose the hyperparameters with diversity in mind.
The ranges for each of the generating factors are chosen such that they can lead to functioning models with a corresponding set of other generating factors.
Nonetheless, that leads to some nodes with uncommon and less-than-promising configurations.

The code to generate the models can be found on \href{www.modelzoos.cc}{www.modelzoos.cc}. With that code, the model zoos can be replicated, changed, or extended. 
We trained our model zoos on CPU nodes with up to 64 CPUs. Training a zoo takes between 3h (small models, small configuration, and small dataset) and 3 days (large models, large configuration, and large dataset). 
Overall, the generation of the zoos took around 30'000 CPU hours.


\begin{table}[ht!]
    \centering
    \begin{minipage}{0.8\textwidth}
    \centering
    {\small
    \caption[CNN Architecture Details of the Model Zoos.]{CNN architecture details for the models in model zoos. }
    \label{tab:model_zoos:model_zoo_architecture}
    \begin{tabular}{@{}llcc@{}}
        \toprule
        \textbf{Layer}          & \textbf{Component} & \textbf{CNN small} & \textbf{CNN large} \\
        \cmidrule(r){1-1} \cmidrule(rl){2-2}  \cmidrule(rl){3-3}  \cmidrule(rl){4-4}
        \multirow{5}{*}{Conv 1} & input channels     & 1 or 3          & 3       \\
                                & output channels    & 8            & 16        \\
                                & kernel size        & 5            & 3        \\
                                & stride             & 1            & 1        \\
                                & padding            & 0            & 0        \\
        \cmidrule(r){1-1} \cmidrule(rl){2-2}  \cmidrule(rl){3-3}  \cmidrule(rl){4-4}
        Max Pooling             & kernel size        & 2            & 2        \\
        \cmidrule(r){1-1} \cmidrule(rl){2-2}  \cmidrule(rl){3-3}  \cmidrule(rl){4-4}
        Activation              &                    &                      \\
        \cmidrule(r){1-1} \cmidrule(rl){2-2}  \cmidrule(rl){3-3}  \cmidrule(rl){4-4}
        \multirow{5}{*}{Conv 2} & input channels     & 8            & 16        \\
                                & output channels    & 6            & 32        \\
                                & kernel size        & 5            & 3         \\
                                & stride             & 1            & 1        \\
                                & padding            & 0            & 0        \\
        \cmidrule(r){1-1} \cmidrule(rl){2-2}  \cmidrule(rl){3-3}  \cmidrule(rl){4-4}
        Max Pooling             & kernel size        & 2            & 2        \\
        \cmidrule(r){1-1} \cmidrule(rl){2-2}  \cmidrule(rl){3-3}  \cmidrule(rl){4-4}
        Activation              &          &             &          \\
        \cmidrule(r){1-1} \cmidrule(rl){2-2}  \cmidrule(rl){3-3}  \cmidrule(rl){4-4}
        \multirow{5}{*}{Conv 3} & input channels     & 6            & 32        \\
                                & output channels    & 4            & 15        \\
                                & kernel size        & 2            & 3        \\
                                & stride             & 1            & 1        \\
                                & padding            & 0            & 0       \\
        \cmidrule(r){1-1} \cmidrule(rl){2-2}  \cmidrule(rl){3-3}  \cmidrule(rl){4-4}
        Activation              &          &                      \\
        \cmidrule(r){1-1} \cmidrule(rl){2-2}  \cmidrule(rl){3-3}  \cmidrule(rl){4-4}
        \multirow{2}{*}{Linear 1} & input channels     & 36         & 60          \\
                                & output channels    & 20           & 20        \\
        \cmidrule(r){1-1} \cmidrule(rl){2-2}  \cmidrule(rl){3-3}  \cmidrule(rl){4-4}
        Activation              &          &                      \\
        \cmidrule(r){1-1} \cmidrule(rl){2-2}  \cmidrule(rl){3-3}  \cmidrule(rl){4-4}
        \multirow{2}{*}{Linear 2} & input channels     & 20         & 20          \\
                                & output channels    & 10           & 10        \\
        \midrule
        Total Parameters        &                   & 2464 or 2864     & 10853 \\
        \bottomrule
    \end{tabular}
    }
    \end{minipage}

\end{table}

\section{Data Management and Accessibility of Model Zoos}
\label{app:data_management}

\paragraph*{Data Management and Documentation:}
To ensure that every zoo is reproducible, expandable, and understandable, we document each zoo. 
For each zoo, a Readme file is generated, displaying basic information about the zoo.
The exact search pattern and the training protocol used to train the zoo are saved in a machine-readable JSON file. 
To make the zoos expandable, the dataset used to train the zoo and a file describing the model architecture are included. 
The model class definition in pytorch is included with the zoo.
Each model is saved along with a JSON file containing its exact hyperparameter combination. A second JSON file contains the the performance metrics during training. Model checkpoints are saved for every epoch.
To enable further training of the models in the zoo, a checkpoint recording the optimizer state is saved for the final epoch of each model. 
All data can be found on the model zoo website as well as directly from Zenodo. \looseness-1

\paragraph*{Accessibility:}
We ensure the technical accessibility of the data by hosting it on Zenodo, where the data will be hosted for at least 20 years.
Further, we take steps to reduce access barriers by providing code for data loading and preprocessing. 
With that, we reduce the friction associated with analyzing the raw zoo files. 
Further, it improves consistency by reducing errors associated with extracting information from the zoo.
To that end, we provide a PyTorch dataset class encapsulating all model zoos for easy and quick access within the PyTorch framework. A Tensorflow counterpart will follow.
All code can be found on the model zoo website as well as a code repository on github.
To ensure conceptional accessibility, we include detailed insights, visualizations, and the analysis of the model zoo (Sec. \ref{sec:model_zoos:analysis}) with each zoo.
Mode details can be found on the dataset website \href{www.modelzoos.cc}{www.modelzoos.cc}.

\section{Dataset Documentation and Intended Uses}
The main dataset documentation can be found at \href{www.modelzoos.cc}{www.modelzoos.cc} and is detailed in the paper in Section \ref{sec:model_zoos:data_management}. 
There, we provide links to the zoos, which are hosted on Zenodo as well as analysis of the zoos. In the future, the analysis will be systematically extended. 
The documentation includes code to reproduce, adapt, or extend the zoos, code to reproduce the benchmark results, as well as code to load and preprocess the datasets.
Dataset Metadata and DOIs are automatically provided by Zenodo, which also guarantees the long-term availability of the data. 
Files are stored as \texttt{zip}, \texttt{json} and \texttt{pt} (pytorch) files. All libraries to read and use the files are common and open source. We provide the code necessary to read and interpret the data.\looseness-1

The datasets are synthetic and intended to investigate populations of neural network models, i.e., to develop or evaluate model analysis methods, progress the understanding of learning dynamics, serve as datasets for representation learning on neural network models, or as a basis for new model generation methods. More information regarding the usage is given in the paper.


\section{Author Statement}
The dataset is publicly available under \href{www.modelzoos.cc}{www.modelzoos.cc} and licensed under the Creative Commons Attribution 4.0 International license (CC-BY 4.0).
The authors state that they bear responsibility under the CC-BY 4.0 license.

\section{Hosting, Licensing, and Maintenance Plan}
The dataset is publicly available under \href{www.modelzoos.cc}{www.modelzoos.cc} and licensed under the Creative Commons Attribution 4.0 International license (CC-BY 4.0).
The landing page contains documentation, code, and references to the datasets, as detailed in the paper in Section \ref{sec:model_zoos:data_management}.
The datasets are hosted on Zenodo, to ensure (i) long-term availability (at least 20 years), (ii) automatic searchable dataset metadata, (iii) DOIs for the dataset, and (iv) dataset versioning. 
The authors will maintain the datasets, but invite the community to engage. Code to recreate, correct, adapt, or extend the datasets is provided, and s.t. maintenance can be taken over by the community at need. The github repository allows the community to discuss, interact, add, or change code. \looseness-1

\end{subappendices}
\FloatBarrier
\newpage
\clearpage

\newpage
\thispagestyle{empty}
\hbox{}
\afterpage{
    \thispagestyle{emptychaptertransition}
    \vspace*{\fill}
    \newpage
    \thispagestyle{plain}
}

\chapter[Hyper-Representations: Learning Representations on Neural Network Weights]{Hyper-Representations: Self-Supervised Representation Learning on Neural Network Weights for Model Characteristic Prediction
}
\label{chap::hyper_reps}

\blfootnote{This work was accepted for publication at NeurIPS 2021 \citep{schurholtSelfSupervisedRepresentationLearning2021} }

\section*{Abstract}
Self-Supervised Learning (SSL) has been shown to learn useful and information-preserving representations. 
Neural Networks (NNs) are widely applied, yet their weight space is still not fully understood.
Therefore, we propose to use SSL to learn \textit{hyper-representations} of the weights of populations of NNs.
To that end, we introduce domain-specific data augmentations and an adapted attention architecture.  Our empirical evaluation demonstrates that self-supervised representation learning in this domain is able to recover diverse NN model characteristics. 
Further, we show that the proposed learned representations outperform prior work for predicting hyper-parameters, test accuracy, and generalization gap as well as transfer to out-of-distribution settings. Code and datasets are publicly available\footnote{\url{https://github.com/HSG-AIML/NeurIPS_2021-Weight_Space_Learning}}.

\section{Introduction}
\label{sec:hyper_reps:introduction}
%
This work investigates populations of Neural Network (NN) models and aims to learn representations of them using Self-Supervised Learning.
Within NN populations, not all model training is successful, i.e., some overfit and others generalize.
This may be due to the non-convexity of the loss surface during optimization \citep{goodfellowQualitativelyCharacterizingNeural2015}, the high dimensionality of the optimization space, or the sensitivity to hyperparameters \citep{haninHowStartTraining2018}, which causes models to converge to different regions in weight space. 
What is still not yet fully understood, is how different regions in weight space are related to model characteristics. 

Previous work has made progress investigating characteristics of NN models, e.g. by visualizing learned features \citep{zeilerVisualizingUnderstandingConvolutional2014,karpathyVisualizingUnderstandingRecurrent2015}. 
Another line of work compares the activations of pairs of NN models \citep{raghuSVCCASingularVector2017, morcosInsightsRepresentationalSimilarity2018, kornblithSimilarityNeuralNetwork2019}. 
Both approaches rely on the expressiveness of the data, and are, in the latter case, limited to two models at a time. 
Other approaches predict model properties, such as accuracy, generalization gap, or hyperparameters from the margin distribution \citep{yakTaskArchitectureIndependentGeneralization2018, jiangPredictingGeneralizationGap2019}, graph topology features \citep{corneanuComputingTestingError2020} or eigenvalue decomposition of the weight matrices \citep{martinTraditionalHeavyTailedSelf2019}. In a similar direction, other publications propose to investigate populations of models and to predict properties directly from their weights or weight statistics in a supervised way \citep{unterthinerPredictingNeuralNetwork2020,eilertsenClassifyingClassifierDissecting2020}.
However, these manually designed features may not fully capture the latent model characteristics embedded in the weight space.

Therefore, our goal is to learn task-agnostic representations from populations of NN models able to reveal such characteristics.
Self-Supervised Learning (SSL) is able to reveal latent structure in complex data without the need of labels, e.g., by compressing and reconstructing data \citep{goodfellowDeepLearning2016,kingmaAutoEncodingVariationalBayes2013}. 
Recently, a specific approach to SSL called contrastive learning has gained popularity \citep{misraSelfSupervisedLearningPretextInvariant2020,chenSimpleFrameworkContrastive2020,chenImprovedBaselinesMomentum2020,grillBootstrapYourOwn2020}. 
Contrastive learning leverages inherent symmetries and equivariances in the data, allows to encode inductive biases and thus structure the learned representations. 

\begin{figure*}[t]
\begin{center}
\begin{minipage}[b]{1.0\linewidth}    
{\small
\begin{tabular}{ccccc}
\text{ $ $} I. Model Zoo Generation & \text{ $ $} \text{ $ $} II. Representation Learning Approach & \text{ $ $} III. Down. Tasks
\end{tabular}
\vspace{-8pt}
}
\end{minipage}
\centerline{\includegraphics[trim=2 2 2 2, clip, 
width=\linewidth]{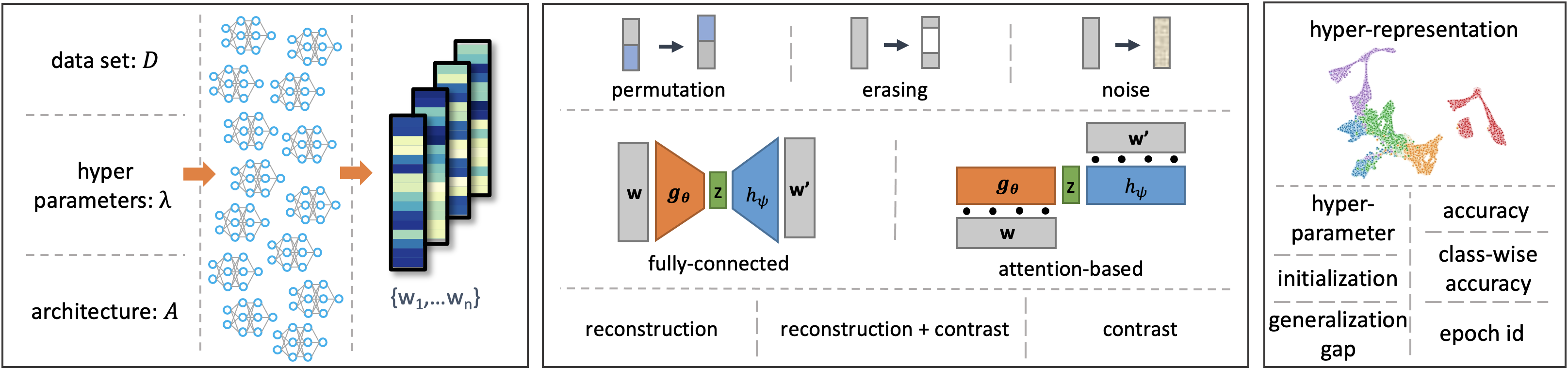}}
\caption[Schematic Overview of the Hyper-Representation Method]{An overview of the proposed self-supervised representation learning approach.
        \textbf{I.} Populations of trained NNs form model zoos; each model is transformed in a vectorized form of its weights. 
        \textbf{II.} Hyper-representations are learned from the model zoos using different augmentations, architectures, and Self-Supervised Learning tasks. 
        \textbf{III.} Hyper-representations are evaluated on downstream tasks which predict model characteristics. 
        }
\label{figure.intro}
\end{center}
\end{figure*}

In this paper, we propose a novel approach to apply SSL to learn representations of the weights of NN populations. We learn representations using reconstruction, contrast, and a combination of both. To that end, we adapt a transformer architecture to NN weights.
Further, we propose three novel data augmentations for the domain of NN weights.
We introduce structure preserving permutations of NN weights as augmentations, which make use of the structural symmetries within the NN weights that we find necessary for learning generalizing representations. We also adapt erasing \citep{zhong2020random} and noise \citep{goodfellowDeepLearning2016} as augmentations for NN weights. 
We evaluate the learned representations by linear-probing for the generating factors and characteristics of the models.
An overview of our learning approach is given in Figure \ref{figure.intro}.

\newpage
To validate our approach, we perform extensive numerical experiments over different populations of trained NN models. 
We find that \textit{hyper-representations}\footnote{By \textit{hyper-representation}, we refer to a learned representation from a population of NNs , i.e., a model zoo, in analogy to HyperNetworks \citep{haHyperNetworks2017}, which are trained to generate weights for larger NN models.} can be learned and reveal the characteristics of the model zoo.
We show that hyper-representations have high utility on tasks such as the prediction of hyper-parameters, test accuracy, and generalization gap. 
This empirically confirms our hypothesis on meaningful structures formed by NN populations. 
Furthermore, we demonstrate improved performance compared to the state-of-the-art for model characteristic prediction and outlay the advantages in out-of-distribution predictions.
For our experiments, we use publicly available NN model zoos and introduce new model zoos. 
In contrast to \citep{unterthinerPredictingNeuralNetwork2020, eilertsenClassifyingClassifierDissecting2020}, our zoos contain models with different initialization points and diverse configurations and include densely sampled model versions during training. 
Our ablation study confirms that the various factors for generating a population of trained NNs play a vital role in how and which properties are recoverable for trained NNs. 
In addition, the relation between generating factors and model zoo diversity reveals that seed variation for the trained NNs in the zoos is beneficial and adds another perspective when recovering NNs' properties.

\section{Model Zoos and Augmentations}
\paragraph*{Model Zoo.} We denote as $\mathcal{D}$ a data set that contains data samples with their corresponding labels. We denote as $\lambda$ the set of hyper-parameters used for training, (\textit{e.g.}, loss function, optimizer, learning rate, weight initialization, batch-size, epochs). We define as $A$ the specific NN architecture. Training under different prescribed configurations $\{ \mathcal{D}, \lambda, A\}$ results in a population of NNs which we refer to as \textit{model zoo}. We convert the weights and biases of all NNs of each model into a vectorized form. In the resulting model zoo $\mathcal{W} = \{ {\bf w}_1, ...., {\bf w}_M \}$, ${\bf w}_i$ denotes the flattened vector of dimension $N$, representing the weights and biases for one trained NN model. 

\label{sec:hyper_reps:augmentations}
\paragraph*{Augmentations.} 
Data augmentation generally helps to learn robust models, both by increasing the number of training samples and preventing overfitting of strong features \citep{shorten_survey_2019}. 
In contrastive learning, augmentations can be used to exploit domain-specific inductive biases, e.g., know symmetries or equivariances \citep{chenIntriguingPropertiesContrastive2021}.
To the best of our knowledge, augmentations for NN weights do not yet exist. 
In order to enable self-supervised representation learning, we propose three methods to augment individual instances of our model zoos.

Neurons in dense layers can change position without changing the overall mapping of the network, if in-going and out-going connections are changed accordingly ~\citep{bishopPatternRecognitionMachine2006}. The relation between equivalent versions of the same network translates to permutations of incoming weights with matrix $\mathbf{P}$ and transposed permutation $\mathbf{P}^{\mathrm{T}}$ of the outgoing weights ($\mathbf{P}^{\mathrm{T}} \mathbf{P} = \mathbf{I}$).
Considering the output at layer $l+1$, with weight matrices $\mathbf{W}$, biases $\mathbf{b}$, activations $\mathbf{a}$ and activation function $\sigma$, we have 
\begin{align}
     \mathbf{z}^{l+1} &= \mathbf{W}^{l+1} \sigma(\mathbf{W}^{l} \mathbf{a}^{l-1} + \mathbf{b}^{l}) + \mathbf{b}^{l+1} = \mathbf{ \hat{W} }^{l+1} \sigma( \mathbf{ \hat{W} }^{l} \mathbf{a}^{l-1} + \mathbf{ \hat{b} }^{l}) + \mathbf{b}^{l+1},     
\end{align}
where $ \mathbf{ \hat{W} }^{l+1} = \mathbf{W}^{l+1} (\mathbf{P}^l)^{\mathrm{T}}$, $ \mathbf{ \hat{W} }^{l} = \mathbf{P}^l  \mathbf{W}^{l} $ and $ \mathbf{ \hat{b} }^{l} = \mathbf{P}^l \mathbf{b}^{l} $ are the permuted weight matrices and bias vector, respectively.
The equivalences hold not only for the forward pass but also for the backward pass and weight update.\footnote{Details, formal statements and proofs can be found in Appendix A.} 
The permutation can be extended to kernels of convolution layers. 
The \textit{permutation augmentation} differs significantly from existing augmentation techniques. As an analogy, flips along the axis of images are similar, but specific instances from the set of possible permutations in the image domain. 
Each permutable layer with dimension $N_l$, has $N_l !$ different permutation matrices, and in total there are $\prod_l N_l!$ distinct, but equivalent versions of the same NN. 
While new data can be created by training new models, the generation is computationally expensive. The permutation augmentation, however, allows to compute valid NN samples at almost no computational cost.
Empirically, we found the permutation augmentation crucial for our learning approach.

In computer vision and natural language processing, masking parts of the input has proven to be helpful for generalization  \citep{devlinBERTPretrainingDeep2018}.
We adapt the approach of \textit{random erasing} of sections in the vectorized forms of trained NN weights. 
As in \citep{zhong2020random}, we apply the erasing augmentation with a probability $p$ to an area that is randomly chosen with lower and upper bounds $b_{low}$ and $b_{up}$. 
In our experiments, we set $p=0.5$, $b_{low}=0.03$, $b_{up}=0.3$ and erase with zeros. Adding \textit{noise augmentation} is another way of altering the exact values of NN weights without overly affecting their mapping, and has long been used in other domains \citep{goodfellowDeepLearning2016}. 
%
\section{Hyper-Representation Learning}
With this work, we propose to learn representations of structures in weight space formed by populations of NNs. We evaluate the representations by predicting model characteristics.
A supervised learning approach has been demonstrated \citep{unterthinerPredictingNeuralNetwork2020,eilertsenClassifyingClassifierDissecting2020}. \citep{unterthinerPredictingNeuralNetwork2020} find that statistics of the weights (mean, var, and quintiles) are superior to the weights to predict test accuracy, which we empirically confirm in our results. 
However, we intend to learn representations of the weights, that contain rich information beyond statistics.
‘Labels’ for NN models can be obtained relatively simply, yet they can only describe predefined characteristics of a model instance (e.g., accuracy) and so supervised learning may overfit few features, as \citep{unterthinerPredictingNeuralNetwork2020} show. 
Self-supervised approaches, on the other hand, are designed to learn task-agnostic representations, that contain rich and diverse information and are exploitable for multiple downstream tasks \citep{lecunSelfsupervisedLearningDark2021}. 
Below, we present the used architectures and losses for the proposed self-supervised representation learning.

\label{sec:loss}
\paragraph*{Architectures and Self-Supervised Losses}. We apply variations of an encoder-decoder architecture. We denote the encoder as $g_{\theta}({\bf w}_i)$, its parameters as $\theta$, and the hyper-representation with dimension $L$ as ${\bf z}_i=g_{\theta}({\bf w}_i)$. We denote the decoder as $h_{\psi}({\bf z}_i)$, its parameters as ${\psi}$, and the reconstructed NN weights as $\hat{ {\bf w}}_i=h_{\psi}({\bf z}_i)=h_{\psi}(g_{\theta}({\bf w}_i))$. 
As is common in CL, we apply a projection head $p_{\gamma}({\bf z}_i)$, with parameters ${\gamma}$, and denote the projected embeddings as $\bar{ {\bf z}}_i=p_{\gamma}({\bf z}_i)=p_{\gamma}(g_{\theta}({\bf w}_i))$. 
In all of the architectures, we embed the hyper-representation ${\bf z}_i$ in a low dimensional space, $L<N$. 
We employ two common SSL strategies: reconstruction and contrastive learning. 
Autoencoders (AEs) with a reconstruction loss are commonly used to learn low-dimensional representations \citep{goodfellowDeepLearning2016}. As AEs aim to minimize the reconstruction error, the representations attempt to fully encode samples. 
Further, contrastive learning is an elegant method to leverage inductive biases of symmetries and inductive biases \citep{bronsteinGeometricDeepLearning2021}. 
\\
\paragraph*{Reconstruction (ED):} For reconstruction, we minimize the MSE $\mathcal{L}_{MSE}=\frac{1}{M}\sum_{i=1}^M \Vert {\bf w}_i - h_{\psi}({g_{\theta}({\bf w}_i)}) \Vert_2^2$.
We denote the encoder-decoder with a reconstruction loss as ED. 
\\
\paragraph*{Contrast (E$_c$):} For contrastive learning, we use the common NT\_Xent loss \citep{chenSimpleFrameworkContrastive2020} as $\mathcal{L}_{c}$. 
For a batch of $M_B$ model weights, each sample is randomly augmented twice to form the two \emph{views} $i$ and $j$. With the cosine similarity $\text{sim}(\bar{\mathbf{z}}_i,\bar{\mathbf{z}}_j) = \bar{\mathbf{z}}_i^T\bar{\mathbf{z}}_j / ||\bar{\mathbf{z}}_i|| ||\bar{\mathbf{z}}_j||$, the loss is given as 
\begin{align}
{\mathcal{L}_{c}} = \sum_{(i,j)} \, - \log \frac{ \exp( \text{sim}(\bar{\mathbf{z}}_i, \bar{\mathbf{z}}_j)/T}{\sum_{k=1}^{2M_B} \mathbb{I}_{k \neq i} \exp( \text{sim}(\bar{\mathbf{z}}_i, \bar{\mathbf{z}}_j)/ T}, 
\end{align}
where $\mathbb{I}_{k \neq i}$ is 1 if $k \neq i$ and 0 otherwise, and $T$ is the temperature parameter.
We denote the encoder with a contrastive loss as E$_c$. 
\\
\paragraph*{Reconstruction + Contrast (E$_c$D):} Further we combine reconstruction and contrast via $\mathcal{L} = \beta \mathcal{L}_{MSE}+(1-\beta)\mathcal{L}_{c}$
in order to achieve good-quality compression via reconstruction and well-structured representations via the contrast. We denote this architecture with its loss as E$_c$D.
\\
\paragraph*{Reconstruction + Positive Contrast (E${_{c\text{+}}}$D):} In contrastive learning, many methods prevented mode collapse by using negative samples. The combined loss contains a reconstruction term $\mathcal{L}_{MSE}$, which can be seen as a regularizer that prevents mode collapse. Therefore, we also experiment with replacing $\mathcal{L}_c$ in our loss with a modified contrastive term without negative samples: 
\begin{align}
{\mathcal{L}_{c\text{+}}} = \sum_{i} \, - \log \left( \exp( \text{sim}(\bar{\mathbf{z}}_i^j, \bar{\mathbf{z}}_i^k) )/T   \right) = \sum_{i} \, -  \text{sim}(\bar{\mathbf{z}}_i^j, \bar{\mathbf{z}}_i^k) + \log(T )  .
\end{align}
\citep{chenExploringSimpleSiamese2021,schwarzerDataEfficientReinforcementLearning2021} explore similar simplifications without reconstruction. We denote the encoder-decoder with the loss $\mathcal{L}=\beta \mathcal{L}_{MSE}+(1 - \beta) \mathcal{L}_{c\text{+}}$ as E${_{c\text{+}}}$D. 
\label{sec:architectures} 

\begin{figure*}[t]
\begin{minipage}[t]{0.48\textwidth}
\begin{center}
\includegraphics[trim=0in 0in 0in 0in, clip, width=1.00\linewidth]{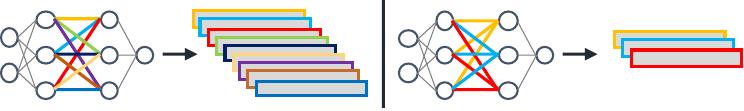}
\caption[Hyper-Rperesentation Tokenization Schemes]{Trained NN weights are the input sequence to a transformer. \textbf{Left}. Each element in the sequence represents the weight that connects two neurons at two different layers. \textbf{Right}. Each element in the sequence represents the set of weights related to one neuron.}
\label{fig:hyper_reps:transformer.input}    
\end{center}
\end{minipage}
\begin{minipage}[t]{0.02\textwidth}
\text{ }
\end{minipage}
\begin{minipage}[t]{0.48\textwidth}
\begin{center}
\includegraphics[trim=0 0 0 0, clip, width=.95\linewidth]{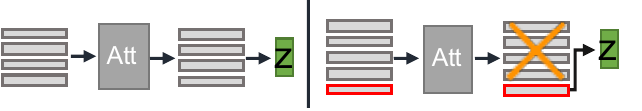}
\caption[Hyper-Representation Encoder Architecture]{
Multi-head attention-based encoder. 
\textbf{Left}. Regular sequence-to-sequence translation, each element of the output sequence is used.
\textbf{Right}. An additional \textit{compression token} is added to the sequence.
From the output sequence, only the compression token is taken.
}
\label{fig:hyper_reps:compression_token}    
\end{center}
\end{minipage}
\end{figure*}

\paragraph*{Attention Module.} Our encoder and decoder pairs are symmetrical and of the same type. As there is no intuition on good inductive biases in the weight space, we apply fully connected feed-forward networks (FFN) as baselines. Further, wee use multi-head self-attention modules (Att) \citep{vaswaniAttentionAllYou2017} as an architecture with very little inductive bias. 
In the multi-head self-attention module, we apply learned position encodings to preserve structural information \citep{dosovitskiyImageWorth16x162021}. The explicit combination of value and position makes attention modules ideal candidates to resolve the permutation symmetries of NN weight spaces.
We propose two methods to encode the weights into a sequence (Figure \ref{fig:hyper_reps:transformer.input}). In the first method, we encode each weight as a token in the input sequence. In the second method, we linearly transform the weights of one neuron or kernel and use it as a token. 
Further, we apply two variants to compress representations in the latent space (Figure \ref{fig:hyper_reps:compression_token}). In the first variant, we aggregate the output sequence of the transformer and linearly compress it to a hyper-representation ${\bf z}_i$. In the second variant, similarly to \citep{devlinBERTPretrainingDeep2018,zhong2020random}, we add a learned token to the input sequence that we dub \emph{compression token}. After passing the input sequence trough the transformer, only the compression token from the output sequence is linearly compressed to a hyper-representation ${\bf z}_i$. Without the compression token, the information is distributed across the output sequence. 
In contrast, the compression token is learned as an effective query to aggregate the relevant information from the other tokens, similar to \citep{jaeglePerceiverGeneralPerception2021}.
The capacity of the compression token can be an information bottleneck. Its dimensionality is directly tied to the dimension of the value tokens and so its capacity affects the overall memory consumption.

\label{proxy.about.representation.utility}
\paragraph*{Downstream Tasks.} We use linear probing \citep{grillBootstrapYourOwn2020} as a proxy to evaluate the utility of the learned hyper-representations. 
As downstream tasks we use accuracy prediction (Acc), generalization gap (GGap), epoch prediction (Eph) as proxy to model versioning, F1-Score prediction (F1$_C$), learning rate (LR), $\ell_2$-regularization ($\ell_2$-reg), dropout (Drop) and training data fraction (TF). Using such targets, we solve a regression problem and measure the $R^2$ score \citep{wright1921correlation}. We also evaluate for hyper-parameters prediction tasks, like the activation function (Act), optimizer (Opt), and initialization method (Init). Here, we train a linear perceptron by minimizing a cross-entropy loss \citep{goodfellowDeepLearning2016} and measure the prediction accuracy.  

\section{Empirical Evaluation}

\subsection{Model Zoos}
\label{sec:zoos_description}

\begin{table}[t]
\begin{minipage}[t]{0.48\textwidth}
{\small
\setlength{\tabcolsep}{1.5pt}
\begin{tabularx}{\linewidth}{c|ccccccc}
\toprule
\multicolumn{8}{c}{\texttt{TETRIS-SEED}}  \\
 & Rec. & Eph & Acc & F1$_{C0}$ & F1$_{C1}$ & F1$_{C2}$ & F1$_{C3}$ \\
\midrule
E${_{c}}$ & \text{  $ $ $ $ -} & 96.7 & \textbf{90.8} & 67.7 & 72.0 & \textbf{74.4} & 85.8 \\ 
ED & \textbf{96.1} & 88.3 & 68.9 & 47.8 & 57.2 & 33.0 & 58.1 \\
E${_{c}}$D & 84.1 & \textbf{97.0} & 90.2 & \textbf{70.7} & \textbf{75.9} & 69.4 & \textbf{86.6} \\
E${_{c\text{+}}}$D & 95.9 & 94.0 & 69.9 & 48.9 & 58.1 & 32.5 & 58.8 \\
\multicolumn{6}{c}{ } \\
\bottomrule
\end{tabularx}
}
\caption[Ablation Study over Hyper-Representation Training Losses]{Ablation results over self-supervised learning losses. All models are implemented with attention-based reference architecture. All values are given in \%.}
\label{table.ablation:task}
\end{minipage}
\begin{minipage}[t]{0.015\textwidth}
\text{ }
\end{minipage}
\begin{minipage}[t]{0.485\textwidth}
{\small
\setlength{\tabcolsep}{1.5pt}
\begin{tabularx}{\linewidth}{c|ccccccc}
\toprule
\multicolumn{8}{c}{\texttt{TETRIS-SEED}}  \\
 & Rec. & Eph & Acc & F1$_{C0}$ & F1$_{C1}$ & F1$_{C2}$ & F1$_{C3}$ \\
\midrule
FF          & 0.0  & 80.0 & 85.3 & 57.3 & 53.9 & 64.5 & 80.1 \\
Att$_W$     & 6.8  & 95.3 & 71.1 & 47.9 & 69.0 & 49.9 & 61.3 \\
Att$_{W+t}$ & 74.1 & 95.4 & 88.6 & 65.6 & 69.8 & \textbf{69.9} & \textbf{86.8} \\
Att$_{N}$   & \textbf{89.4} & \textbf{97}.1 & 88.4 & \textbf{71.4} & \textbf{80.8} & 69.1 & 82.3 \\
Att$_{N+t}$ & 84.1 & 97.0 & \textbf{90.2} & 70.7 & 75.9 & 69.4 & 86.6 \\
\bottomrule
\end{tabularx}
}
\caption[Ablation Study over Hyper-Representation Architectures]{Ablation results in \% under E$_c$D setup. 
We use feed-forward FF and attention-based variants with weight and neuron encoding Att$_W$ and Att$_N$ each with $+t$ and without compress. token.}
\label{table.ablation:architecture}
\end{minipage}
\vspace{-.2in}
\end{table}

\paragraph*{Publicly Available Model Zoos.} \citep{unterthinerPredictingNeuralNetwork2020} introduced model zoos of CNNs with 4970 parameters trained on MNIST \citep{lecunGradientbasedLearningApplied1998}, Fashion-MNIST \citep{xiaoFashionMNISTNovelImage2017}, CIFAR10 \citep{krizhevskyLearningMultipleLayers2009} and SVHN \citep{netzerReadingDigitsNatural2011}, and made them available under CC BY 4.0. We refer to these as \texttt{MNIST-HYP}, \texttt{FASHION-HYP}, \texttt{CIFAR10-HYP} and \texttt{SVHN-HYP}. We categorize these zoos as \emph{large} due to their number of parameters. In their model zoo creation, the CNN architecture and seed were fixed, while the activation function, initialization method, optimizer, learning rate, $\ell_2$ regularization, dropout, and the train data fraction were varied between the models.

\begin{wraptable}[11]{l}{0pt}
{\small
\setlength{\tabcolsep}{2.8pt}
\begin{tabularx}{0.55\linewidth}{c|c|ccc|ccc|c}
\toprule
\multicolumn{9}{c}{\texttt{TETRIS-SEED}}  \\
&\text{  $ $ $ $ -}& P & E & N & P,E & P,N & E,N & P,E,N  \\
\midrule
ED & 88.1 & \textbf{95.5} & 89.8 & 88.8 & \textbf{96.1} & 95.6 & 89.7 & \textbf{96.1}\\
E$_c$D & 49.2 & \textbf{72.9} & 67.3 & 59.4 & \textbf{84.1} & 81.9 & 65.4 & \textbf{84.1} \\
E$_{c\text{+}}$D  & 86.3 & \textbf{95.6} & 88.5 & 87.6 & \textbf{95.9} & 95.8 & 89.0 & \textbf{96.0} \\
\bottomrule
\end{tabularx}
}
\caption[Ablation Study over Hyper-Representation Data Augmentations]{Ablation results for different augmentations and representation learning tasks. We use: permutation (P), erasing (E) and noise (N) augmentation and report test-split reconstruction $R^2$ scores in \%.
}
\label{ablation_augmentation}
\end{wraptable}
\paragraph*{Our Model Zoos.} We hypothesize that using only one fixed seed may limit the variation in characteristics of a zoo. To address that, we train zoos where we also vary the seed and perform an ablation study below. As a toy example to test architectures, SSL tasks, and augmentations, we first create a 4x4 grey-scaled image data set that we call \emph{tetris} by using four tetris shapes. 

We introduce two zoos, which we call \texttt{TETRIS-SEED} and \texttt{TETRIS-HYP}, which we group under \emph{small}. Both zoos contain FFN with two layers and have a total number of 100 learnable parameters. In the \texttt{TETRIS-SEED} zoo, we fix all hyper-parameters and vary only the seed to cover a broad range of the weight space. The \texttt{TETRIS-SEED} zoo contains 1000 models that are trained for 75 epochs. To enrich the diversity of the models, the \texttt{TETRIS-HYP} zoo contains FFNs, which vary in activation function [\texttt{tanh}, \texttt{relu}], the initialization method [\texttt{uniform}, \texttt{normal}, \texttt{kaiming normal}, \texttt{kaiming uniform}, \texttt{xavier normal}, \texttt{xavier uniform}] and the learning rate [$1e$-$3$, $1e$-$4$, $1e$-$5$]. In addition, each combination is trained with seeds 1-100.
Out of the 3600 models in total, we have successfully trained 2900 for 75 epochs - the remainders crashed and are disregarded. Similarly to \texttt{TETRIS-SEED}, we further create zoos of CNN models with 2464 parameters, each using the MNIST and Fashion-MNIST data sets, called \texttt{MNIST-SEED} and \texttt{FASHION-SEED} and grouped them as \emph{medium}. To maximize the coverage of the weight space, we again initialize models with seeds 1-1000.
\footnote{\label{fn:hyper_reps:details_appendix}Full details on the generation of the zoos can be found in the Appendix Section C}

\paragraph*{Model Zoo Generating Factors.}  
Prior work discusses the impact of random seeds on the properties of model zoos. 
While \citep{yakTaskArchitectureIndependentGeneralization2018} use multiple random seeds for the same hyper-parameter configuration,  \citep{unterthinerPredictingNeuralNetwork2020} explicitly argue against that to prevent information leakage between samples. To disentangle the generating factors (seeds and hyper-parameters) and model properties, we have created five zoos with approximately the same number of CNN models trained on MNIST 
\textsuperscript{\ref{fn:hyper_reps:details_appendix}}.
\texttt{MNIST-SEED} varies only the random seed (1-1000), \texttt{MNIST-HYP-1-FIX-SEED} varies the hyper-parameters with one fixed seed per configuration (similarly to \citep{unterthinerPredictingNeuralNetwork2020}). 
To decouple the hyper-parameter configuration from one specific seed, \texttt{MNIST-HYP-1-RAND-SEED} draws 1 random seed for each hyper-parameter configuration. To investigate the influence of repeated configurations, in  \texttt{MNIST-HYP-5-FIX-SEED} and \texttt{MNIST-HYP-5-RAND-SEED} for each hyper-parameter configurations we add 5 models with different seeds, either 5 fixed seeds or randomly drawn seeds.\looseness-1
%
\begin{figure*}[t!]
\vskip -0.0in
\begin{center}

\vskip -0.05in
\begin{minipage}[b]{1.0\linewidth}    
    \includegraphics[trim=0 0 0 0,clip, width=\linewidth]{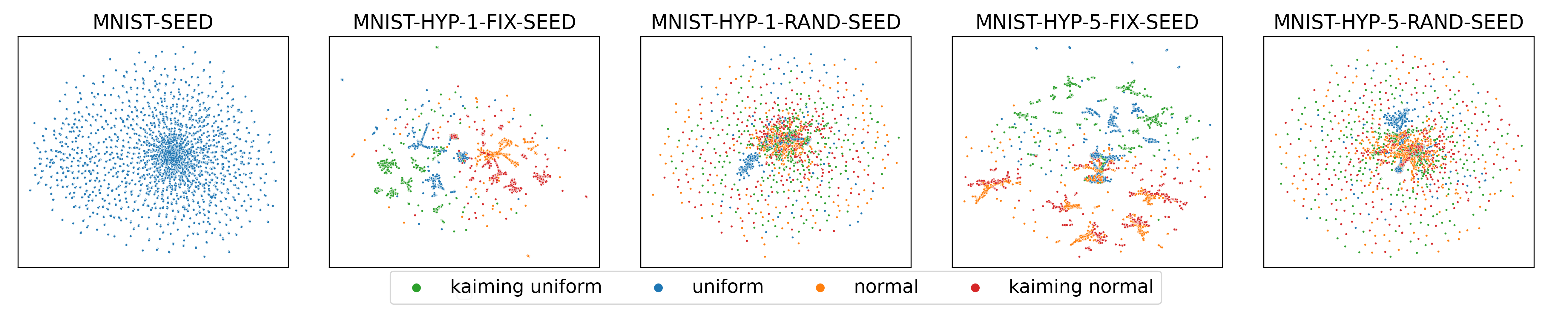}
\end{minipage}        
\vskip -0.1in
    \caption[UMAP Weight Visualization of Model Zoos in Comparison]{UMAP dimensionality reduction for NN model zoos created with different generating factors. Colors represent initialization methods. Compare numerical results in Table \ref{tab:hyper_reps:zoo_comparison}. 
    }
    \label{fig:hyper_reps:dataset_comparison_embeddings}    
\end{center}
\vskip -0.2in
\end{figure*}
\begin{table*}[t!]
\centering
\vspace{.05in}
\begin{minipage}[b]{.99\linewidth}  
\begin{center}
\begin{small}
\begin{sc}
\hspace{-.1in}
{\small
\setlength{\tabcolsep}{1.1pt}
\begin{tabularx}{\linewidth}{c|ccc|ccc|ccc|ccc|ccc}
\toprule
& \multicolumn{3}{c}{\texttt{}} & \multicolumn{3}{c}{\texttt{MNIST-HYP-} }& \multicolumn{3}{c}{\texttt{MNIST-HYP-} }&
\multicolumn{3}{c}{\texttt{MNIST-HYP-} } & \multicolumn{3}{c}{\texttt{MNIST-HYP-} } \\
& \multicolumn{3}{c}{\texttt{MNIST-SEED}} & \multicolumn{3}{c}{\texttt{1-FIX-SEED}}
& \multicolumn{3}{c}{\texttt{1-RAND-SEED}}
&
\multicolumn{3}{c}{\texttt{5-FIX-SEED}}
& \multicolumn{3}{c}{\texttt{5-RAND-SEED}}\\
\midrule
var & \multicolumn{3}{c}{.234} & \multicolumn{3}{c}{.155} & \multicolumn{3}{c}{.152} & \multicolumn{3}{c}{.091} & \multicolumn{3}{c}{.092} \\
var$_c$ & \multicolumn{3}{c}{.234} & \multicolumn{3}{c}{.101} & \multicolumn{3}{c}{.164} & \multicolumn{3}{c}{.094} & \multicolumn{3}{c}{.100}\\
\midrule
$R_{tr}$ & \multicolumn{3}{c}{73.4} & \multicolumn{3}{c}{91.3} & \multicolumn{3}{c}{80.0} & \multicolumn{3}{c}{81.3} & \multicolumn{3}{c}{65.6} \\
$R_{tes}$ & \multicolumn{3}{c}{59.0} & \multicolumn{3}{c}{72.9} & \multicolumn{3}{c}{37.2} & \multicolumn{3}{c}{70.0} & \multicolumn{3}{c}{38.6} \\
\toprule
& W & s(W) & E${_{c\text{+}}}$D & W & s(W) & E${_{c}}$D & W & s(W) & E${_{c\text{+}}}$D & W & s(W) & E${_{c}}$D  & W & s(W) & E${_{c}}$D\\
\midrule
Eph & 84.5 & \textbf{97.7}& 97.3 & 07.2 & 06.2 & \textbf{11.6} & \text{-}34 & 00.5 & \textbf{04.1} & \text{-}2.7 & 7.4 & \textbf{10.2} & 
\text{-}14 & \textbf{09.6} & 07.2 \\
Acc & 91.3 & 98.7 & \textbf{98.9} & 73.6 & 72.9 & \textbf{89.5} & \text{-}06 & 60.1 & \textbf{85.7} & 57.7 & 75.1 & \textbf{79.4} & 
\text{-}13 & 74.5 & \textbf{82.5}\\
GGap & 56.9 & 66.2 & \textbf{66.7} & 55.3  & 42.1 & \textbf{61.9} & \text{-}36 & 31.1 & \textbf{54.9} & 37.8 & 36.7 & \textbf{57.7} &
\text{-}3.2 & 46.1 & \textbf{60.5} \\
Init &  -- & -- & -- & \textbf{87.7} & 63.3 & 75.4 & 38.3 & \textbf{56.1} & 48.2 & \textbf{80.6} & 54.5 & 51.4 & 
35.5 & \textbf{47.2} & 40.4 \\
\bottomrule
\end{tabularx}
}
\end{sc}
\end{small}
\end{center}
\end{minipage}
\caption[Model Property Prediction Results Comparing Model Zoo Types]{$R^2$ score in \%. Results on the impact of the generating factors for the model zoos. \textbf{Top:} VAR is the variance of the weights. VAR$_c$ denotes the mean of the variances for groups of samples with shared initialization method and activation function. $R_{tr}$ and $R_{tes}$ are the reconstruction $R^2$ of train and test split on a reference E$_{c\text{+}}$D architecture trained for 500 epochs. \textbf{Bottom:} Downstream task performance compared to baselines predicting epochs, accuracy, generalization gap, and initialization.}
\label{tab:hyper_reps:zoo_comparison}
\end{table*}

\subsection{Training and Testing Setup}
\paragraph*{Architectures.} We evaluate our approach with different types of architectures, including E${_c}$, ED, E${_c}$D and E${_{c\text{+}}}$D as detailed in Section \ref{sec:loss}. 
The encoder E and decoders D in the FFN baseline are symmetrical 10 [FC-ReLU]-layers each and linearly reduce dimensionality for the input to the latent space $\mathbf{z}$. Considering the attention-based encoder and decoder, on the \texttt{TETRIS-SEED} and \texttt{TETRIS-HYP} zoos, we used 2 attention blocks with 1 attention head each, token dimensions of 128 and FC layers in the attention module of dimension 512. On the larger zoos, we use up to 4 attention heads in 4 attention blocks, token dimensions of up to size 800 and FC layers in the attention module of dimension 1000. For the combined losses, we evaluate different $\beta \in [0.05,0.85]$.

\paragraph*{Hyper-representation Learning and Downstream Tasks.} We apply the proposed data augmentation methods for representation learning (see Section \ref{sec:hyper_reps:augmentations}). We run our representation learning algorithms for up to 2500 epochs, using the adam optimizer \citep{Kingma:2014:Adam}, a learning rate of 1e-4, weight decay of 1e-9, dropout of 0.1 percent and batch-sizes of 500. In all of our experiments, we use 70\% of the model zoos for training, 15\% for validation and 15\% for testing. We use checkpoints of all epochs, but ensure that samples from the same models are either in the train, validation or the test split of the zoo. As quality metric for the hyper-representation learning, we track the reconstruction $R^2$ on the validation split of the zoo and report the reconstruction $R^2$ on the test split. 
As a proxy for how much useful information is contained in the hyper-representation, we evaluate on downstream tasks as described in Section \ref{proxy.about.representation.utility}. To ensure numerical stability of the solution to the linear probing, we apply Tikhonov regularization \citep{tikhonov1977solutions} 
with regularization parameter ${\alpha}$ in the range [$1e$-$5$, $1e3$] (we choose ${\alpha}$ by cross-validating over the $R^2$ score of the validation split of the zoo) and report the $R^2$ score of the test split of the zoo. To minimize the cross entropy loss for the categorical hyper-parameter prediction, we use the Adam optimizer with learning rate of $1e$-$4$ and weight-decay of $1e$-$6$. The linear probing is applied to the same train-validation-test splits as it is in our representation learning setup.  

\paragraph*{Out-of-Distribution Experiments.} We follow a setup for out-of-distribution experiments similar to \citep{unterthinerPredictingNeuralNetwork2020}. We investigate how well the linear probing estimator computed over hyper-representations generalizes to yet unseen data. 
Therefore, we use the zoos \texttt{MNIST-HYP}, \texttt{FASHION-HYP}, \texttt{CIFAR10-HYP} and \texttt{SVHN-HYP}. 
On each zoo, we apply our self-supervised approach to learn their corresponding hyper-representations and fit a linear probing estimator to each of them (in-distribution).
We then apply both the hyper-representation mapper and the linear probing estimator of one zoo 
on the weights of the other zoos (out-of-distribution). 
The target ranges and distributions vary between the zoos. Linear probe prediction may preserve the relation between predictions but include a bias. Therefore, we use Kendall’s $\tau$ coefficient as performance metric, which is a measure of rank correlation. 
It measures the ordinal association between two measured quantities
\citep{kendall:1938:measure:Biometrika}. 

\begin{table}[]
\centering
\begin{center}
{ \small
\begin{minipage}[b]{.49\linewidth}
\setlength{\tabcolsep}{1.3pt}
\begin{tabularx}{\linewidth}{l|ccccccc}
\toprule
& \multicolumn{7}{c|}{\texttt{TETRIS-SEED}}  \\
 & W & PCA$_l$ & PCA$_c$ & PCA$_{\text{r}}$ & U$_{m}$ & s(W) & E${_{c}}$D \\
\midrule
Eph & 55.4 & 46.9 & 77.8 & 93.5 & 0.00 & 96.4 & \textbf{97.0} \\
Acc & 49.1 & 23.9 & 74.7 & 74.9 & 0.01 & 86.9 & \textbf{90.2}  \\
Ggap & 43.9 & 28.3 & 72.9 & 75.7 & 0.01 & 81.0 & \textbf{81.9}  \\
\midrule
$F_{C0}$ & 26.6 & 0.07 & 52.8 & 49.1 & 0.02 & 62.5 & \textbf{70.7}  \\
$F_{C1}$ & 41.4 & 30.2 & 48.6 & 58.6 & 0.01 & 63.1 &  \textbf{75.9} \\
$F_{C2}$  & 41.5 & 0.06 & 38.3 & 38.5 & 0.01 & 60.9 & \textbf{69.4}  \\
$F_{C3}$  & 53.9 & 44.6 & 73.8 & 72.2 & 0.00 & 75.2 & \textbf{86.6}  \\
\bottomrule
\end{tabularx}
\end{minipage}
\begin{minipage}[b]{.48\linewidth}
\centering
\setlength{\tabcolsep}{1.3pt}
\begin{tabularx}{\linewidth}{l|ccccccc}
\toprule
& \multicolumn{7}{c}{\texttt{TETRIS-HYP}}  \\
& W & PCA$_l$ & PCA$_c$ & PCA$_{\text{r}}$ & U$_{m}$ & s(W) & E${_{c}}$D \\
\midrule
Eph & 02.8 & 02.7 &  0.04 &  13.1 & 0.03 & 16.3 & \textbf{16.7}  \\
Acc & 11.0 & 14.0 & 0.81 & 73.8 & 0.7 & 80.2 & \textbf{83.7}    \\
GGap & 12.9 & 12.9 & 0.31 & 76.0 & 0.3 & 79.2 & \textbf{81.6}    \\
\midrule
LR & \text{-}4.7 & \text{-}3.2 & \text{-}1.3 & \text{-}0.4 &  \text{-}1.4 & \textbf{53.4} & 51.1  \\
Act & 74.7 & 72.7 & 45.1 &  74.2 & 45.9 &  73.6 & \textbf{86.9}   \\
Init & 38.5 & 36.4 & 32.7 & 35.4 & 33.6 & 46.6 & \textbf{47.1} \\
\multicolumn{8}{c}{ } \\
\bottomrule
\end{tabularx}
\end{minipage}
}
\end{center}
\caption[Model Property Prediction Results Tetris Model Zoo]{$R^2$ given in \%. \textbf{Left}.  Reconstruction, epoch, accuracy, generalization gap and class-wise F-scores prediction. \textbf{Right}. Epoch, accuracy, generalization gap, learning rate, seed, activation function and initialization prediction. 
}
\label{table.tetris-seed.and.tetris-hyp}
\end{table}

\paragraph*{Baselines, Computing Infrastructure and Run Time.} 
As baseline, we use the model weights (W). In addition, we also consider PCA with linear (PCA$_l$), cosine (PCA$_c$), and radial basis kernel (PCA$_{\text{r}}$), as well as UMAP (U$_{\text{m}}$) \citep{mcinnesUMAPUniformManifold2018a}. We further compare to layer-wise weight-statistics (mean, var, quintiles) s(W) as in \citep{unterthinerPredictingNeuralNetwork2020,eilertsenClassifyingClassifierDissecting2020}. 
As computing hardware, we use half of the available resources from NVIDIA DGX2 station with 3.3GHz CPU and 1.5TB RAM memory, which has a total of 16 1.75GHz GPUs, each with 32GB memory. To create one small and medium zoo, it takes 1 to 2 days and 10 to 12 days, respectively. For one experiment over the small zoo, it takes around 3 hours to learn the hyper-representation on a single GPU and evaluate on the downstream tasks. It takes approximately 1 day for the medium zoos and 2 to 3 days for the large scale zoos for the same experiment. 
The representation learning model size ranges from ~225k on the small zoos to 65M parameters on the largest zoos.
We use ray.tune \citep{liawTuneResearchPlatform2018} for hyperparameter optimization and track experiments with W\&B \citep{wandb}.
\begin{table*}[t]
\vspace{-.1in}
\centering
\vspace{.05in}
\begin{minipage}[b]{.99\linewidth}  
\begin{center}
\begin{small}
\begin{sc}
\hspace{-.1in}
{\small
\setlength{\tabcolsep}{4pt}
\begin{tabularx}{\linewidth}{l|ccc|ccc|ccc|ccc}
\toprule
& \multicolumn{3}{c}{\texttt{MNIST-HYP}} & \multicolumn{3}{c}{\texttt{FASHION-HYP}} & \multicolumn{3}{c}{\texttt{CIFAR10-HYP}} & \multicolumn{3}{c}{\texttt{SVHN-HYP}} \\
& W & s(W) & E${_{c}}$D & W & s(W) & E${_{c}}$D & W & s(W) & E${_{c}}$D & W & s(W) & E${_{c}}$D   \\

\midrule
Eph & 25.8 & 33.2 & \textbf{50.0} & 26.6 & 34.6 & \textbf{51.3} & 25.7 & 30.3 & \textbf{53.3} & 
22.8 & 37.8 & \textbf{52.6} \\
Acc & 74.7 & 81.5 & \textbf{94.9} & 70.9 & 78.5 & \textbf{96.2} & 76.4 & 82.9 & \textbf{92.7} & 
80.5 & 82.1 & \textbf{91.1} \\
\midrule
GGap & 23.4 & 24.4 & \textbf{27.4} & 48.1 & 41.1 & \textbf{49.0} & 37.7 & 37.4 & \textbf{40.4} & 
38.7 & 42.2 & \textbf{44.2} \\
\midrule
LR & 29.3 & 34.3 & \textbf{37.1} & 33.5 & 35.6 & \textbf{42.4} & 27.4 & 32.3 &\textbf{44.7}  & 
24.5 & 33.4 & \textbf{49.1} \\
$\ell_2$-reg & 12.5 & 16.5 & \textbf{20.1} & 11.9 & 16.3 & \textbf{25.0} & 08.7 & 13.8 & \textbf{28.0}  &
09.0 & 13.6 & \textbf{28.0} \\
Drop & 28.5 & 19.2 & \textbf{35.8} & 26.7 & 21.3 & \textbf{38.3} & 16.7 & 16.5 & \textbf{33.8}  & 
09.0 & 14.6 & \textbf{23.3} \\
TF & 03.8 & 07.8 & \textbf{15.9} & 08.1 & 08.2 & \textbf{22.1} & 08.4 & 06.9 & \textbf{35.4}  &  
03.2 & 08.8 &  \textbf{21.4} \\
\midrule
Act & 88.6 & 81.1 & \textbf{88.7} & 89.8 & 82.4 & \textbf{90.1} & 88.3 & 80.3 & \textbf{90.0} & 
86.9 & 78.8 & \textbf{87.2} \\
Init & \textbf{94.6} & 72.0 & 80.6 & \textbf{95.7} & 76.5 & 86.7 & \textbf{93.5} & 73.3 & 82.6  & 
\textbf{91.0} & 73.0 & 82.8 \\
Opt & \textbf{76.7} & 65.4 & 66.4 & \textbf{79.9} & 67.4 & 73.0 & \textbf{74.0} & 65.5 & 71.0 & 
\textbf{72.5} & 68.2 & 72.3 \\

\end{tabularx}
}
\end{sc}
\end{small}
\end{center}
\end{minipage}
\caption[Model Property Prediction Results Unterthiner Model Zoo]{\textbf{Top 7 Rows}. $R^2$ score in $\%$ for Eph, Acc, GGap, LR $\ell_2$-reg, Drop and TF prediction. \textbf{Bottom 3 Rows}. Accuracy score for Act, Init, and Opt prediction. 
}
\label{comparisament.ref}
\vspace{-.15in}
\end{table*}
\subsection{Results}

\paragraph*{Augmentation Ablation.} 
To evaluate the impact of the proposed augmentations (Section \ref{sec:hyper_reps:augmentations}) for our representation learning method, we present an ablation analysis on the \texttt{TETRIS-SEED} zoo, in which we measure $R^2$ for ED, E$_c$D and E$_{c\text{+}}$D\footnote{We leave out E$_c$ as it does not use a reconstruction loss}. We use 120 permutations, a probability of $0.5$ for erasing the weights, and zero-mean noise with standard deviation $0.05$ (see Table \ref{ablation_augmentation}). We find the permutation augmentation to be necessary for generalization - particularly under higher compression ratios. The additional samples generated with the permutation appear to effectively prevent overfitting of the training set. Without the permutation augmentation, the test performance diverges after a few training epochs.  
Erasing further improves test performance and allows for extended training without overfitting. The addition of noise yields inconsistent results and is difficult to tune, so, we omit it in our further experiments.

\paragraph*{Architecture Ablation.} 
The different architectures are compared in Table \ref{table.ablation:task}. The results show that within the set of used NN architectures for hyper-representation learning, the attention-based architectures learn considerably faster, 
yield lower reconstruction error, and have the highest performance on the downstream tasks compared to the FFN-based architectures. We attribute this to the attention modules, which are able to reliably capture long-range relations on complex data due to the global field of visibility in each layer.  
While tokenizing each weight individually (Att$_{W}$) is able to learn, the computational load is significant, even for a small zoo, due to the large number of tokens in the sequences. The memory load prevents the application of that encoding on larger zoos. 
We find Att$_{N+t}$ embedding all weights of one neuron (or convolutional kernel) to one token in combination with compression tokens shows the overall best performance and scales to larger architectures. Compression tokens achieve higher performance, which we confirm on larger and more complex datasets. The dedicated token gathers information from all other tokens of the sequence in several attention layers. This appears to enable the hyper-representation to grasp more relevant information than linearly compressing the entire sequence. On the other hand, compression tokens are only an advantage, if their capacity is high enough, in particular higher than the bottleneck.

\paragraph*{Self-Supervised Learning Ablation.} In Table \ref{table.ablation:architecture}, we evaluate the usefulness of the self-supervised learning tasks (Section \ref{sec:loss}). The application of purely contrastive loss in E$_c$, learns very useful representations for the downstream tasks. However, E$_c$ heavily depends on expressive projection heads and by design cannot reconstruct samples to further investigate the representation. 
Pure reconstruction ED results in embeddings with a low reconstruction loss, but comparably low performance on downstream tasks. Among our losses, the combination of reconstruction with contrastive loss as in E$_c$D, provides hyper-representations ${\bf z}$ that have the best overall performance. The variation E$_{c+}$D, is closer to ED in general performance but still outperforms it on the downstream tasks.  The addition of a contrastive loss with a projection head helps to pronounce distinctiveness so that the hyper-representations are good at reconstructing the NN weights, and in revealing properties of the NNs through the encoder. 
Empirically, we found that on some of the larger zoos, E$_{c+}$D performed better than E$_{c}$D. On these zoos, it appears that E$_{c+}$D is most suitable for homogeneous zoos without distinct clusters, while E$_{c}$D is suitable for zoos with more sub-structures, compare Figure \ref{fig:hyper_reps:dataset_comparison_embeddings} and Table \ref{tab:hyper_reps:zoo_comparison}. We therefore applied both learning strategies on all zoos and report the more performant one.

\paragraph*{Zoo Generating Factors Ablation.} 
Figure \ref{fig:hyper_reps:dataset_comparison_embeddings} visualizes the weights of the zoos, which contain models of all epochs\footnote{Further visualizations can be found in Appendix Section C}. In Table \ref{tab:hyper_reps:zoo_comparison}, we report numerical properties. 
Only changing the seeds appears to result in homogeneous development with very high correlation between $s(W)$ and the properties of the samples in the zoo, as previous work already indicated \citep{schurholtInvestigationWeightSpace2021}.
Varying the hyper-parameters reduces the correlation. With fixed seeds, we observe clusters of models with shared initialization method and activation function. Quantitatively we obtain lower VAR$_c$ and high predictive value of $W$ for the initialization method. 
That seems a plausible outcome, given that the architecture and activation function determine the shape of the loss surface, while the seed and initialization method decide the starting point.  
Such clustering appears to facilitate the prediction of categorical characteristics from the weights.
We observe similar properties in the zoos of \citep{unterthinerPredictingNeuralNetwork2020}, see Appendix Section C.
%
Initializing models with random seeds disperses the clusters, compare Figure \ref{fig:hyper_reps:dataset_comparison_embeddings} and Table \ref{tab:hyper_reps:zoo_comparison}. 
While VAR between 1 fixed and 1 random seed is comparable, VAR$_c$ is considerably smaller with fixed seed, the predictive value of $W$ for the initialization methods drops significantly. 
Random seeds also appear to make both the reconstruction as well as NN property prediction more difficult. 
The repetition of configurations with five seeds hinders shortcuts in the weight space make reconstruction and the prediction of characteristics harder. 
Thus, we conclude that changing only the seeds results in models with very similar evolution during learning. In these zoos, statistics of the weights perform best. In contrast, using one seed shared across models might create shortcuts in the weight space. Zoos that vary both appear to be most diverse and hardest for learning and NN property prediction. Across all zoos with hyperparameter changes, learned hyper-representations significantly outperform the baselines.
\begin{table*}[t!]
\vspace{-.1in}
\centering
\vspace{.05in}
\begin{minipage}[b]{.99\linewidth}  
\begin{center}
\begin{small}
\begin{sc}
\hspace{-.1in}
{\small
\setlength{\tabcolsep}{2.6pt}
\begin{tabularx}{\linewidth}{l|ccc|ccc|ccc|ccc}
\toprule
& \multicolumn{3}{c}{\texttt{MNIST-HYP}} & \multicolumn{3}{c}{\texttt{FASHION-HYP}} &  \multicolumn{3}{c}{\texttt{SVHN-HYP}}  & \multicolumn{3}{c}{\texttt{CIFAR10-HYP}}\\
& W & s(W) & E${_{c\text{+}}}$D & W & s(W) & E${_{c\text{+}}}$D & W & s(W) & E${_{c\text{+}}}$D & W & s(W) & E${_{c\text{+}}}$D   \\
\midrule
\texttt{MNIST-HYP} & \cellcolor{darkgray!10} \textbf{.36} & \cellcolor{darkgray!10} .29 & \cellcolor{darkgray!10} \textbf{.36} & .21 & .14 & \textbf{.27} & \textbf{.26} & .12 &   .23 & \text{-}.01 & \text{-}.04 & \textbf{.02} \\
\midrule
\texttt{FASHION-HYP} & \text{-}.02 & \textbf{.08} & .02 & \cellcolor{darkgray!10} .54 & \cellcolor{darkgray!10} .48 & \cellcolor{darkgray!10} \textbf{.56} & .06 &  \textbf{.14} & .01 &  .07 & .10 & \textbf{.27} \\
\midrule
\texttt{SVHN-HYP} & .05 & \textbf{.15} & \text{-}.04 & \text{-}.02 & \textbf{.27} & .10 & \cellcolor{darkgray!10} .44 & \cellcolor{darkgray!10} .34 & \cellcolor{darkgray!10} \textbf{.45} & \text{-}.02 & .08 & \textbf{.10} \\
\midrule
\texttt{CIFAR10-HYP} & \textbf{.11} & .09 & .06 & .38 & .36 & \textbf{.39} & .14 & .14 & \textbf{.15} & \cellcolor{darkgray!10} \textbf{.41} & \cellcolor{darkgray!10} .28 & \cellcolor{darkgray!10} .35 \\
\bottomrule
\end{tabularx}
}
\end{sc}
\end{small}
\end{center}
\end{minipage}
\caption[Model Property Prediction Results for Cross-Dataset Experiments]{Kendall's $\tau$ score for the generalization gap (GGap) prediction. 
We train estimators for each zoo (rows) and evaluate on all zoos (columns). The block diagonal elements contain the in-distribution prediction values. The remaining values are for out-of-distribution prediction.}
\label{table.out-of-distribution.predictioin}
\vspace{-.15in}
\end{table*}

\paragraph*{Downstream Tasks.} 
We learn and evaluate our hyper-representations on 11 different zoos: \texttt{TETRIS-SEED}, \texttt{TETRIS-HYP}, 5 variants of \texttt{MNIST}, 
\texttt{MNIST-HYP},  \texttt{FASHION-HYP},  \texttt{CIFAR10-HYP} and \texttt{SVHN-HYP}. We compare to multiple baselines and s(W). The results are shown in Tables \ref{tab:hyper_reps:zoo_comparison}, \ref{table.tetris-seed.and.tetris-hyp} and \ref{comparisament.ref}. On all model zoos, hyper-representations learn useful features for the downstream tasks, which outperform the actual NN weights and biases, all of the baseline dimensionality reduction methods as well as s(W) \citep{unterthinerPredictingNeuralNetwork2020}.
On the \texttt{TETRIS-SEED},  and \texttt{MNIST-SEED} model zoos, s(W) achieves high $R^2$ scores on all downstream tasks (see Table \ref{table.tetris-seed.and.tetris-hyp} left). 
As discussed above, these zoos contain a strong correlation between $s(W)$ and sample properties.
Nonetheless, learned hyper-representations achieve higher $R^2$ scores on all characteristics, with the exception of \texttt{MNIST-SEED} Eph, where E${_{c\text{+}}}$D is competitive to s(W). 
On \texttt{TETRIS-HYP}, the overall performance of all methods is lower compared to \texttt{TETRIS-SEED} (see Table \ref{table.tetris-seed.and.tetris-hyp} right), making it the more challenging zoo. 
Here, too, hyper-representations have the highest $R^2$ score on all downstream tasks, except for the LR prediction. 
On \texttt{MNIST-HYP},  \texttt{FASHION-HYP},  \texttt{CIFAR10-HYP} and \texttt{SVHN-HYP}, hyper-representations outperform s(W) on all downstream tasks and achieve higher $R^2$ score compared to W on the prediction of continuous hyper-parameters and activation prediction. 
On the remaining categorical hyper-parameters initialization method and optimizer, the weight space achieves the highest $R^2$ scores (see Table \ref{comparisament.ref}). We explain this with the small amount of variation in the model zoo (see Section \ref{sec:zoos_description}), which allows to separate these properties in weight space. 

\paragraph*{Out-of-Distribution Prediction}. Table \ref{table.out-of-distribution.predictioin} shows the out-of-distribution results for generalization gap prediction\footnote{In Appendix Section D, we also give results for other tasks, including epoch id and test accuracy prediction}, which is a very challenging task. The task is rendered more difficult by the fact that many of the samples which have to be ordered by their performance are very close in performance. 
As the results show, hyper-representations are able to recover the order of samples by performance to a significant degree. Further, hyper-representations outperform the baselines in Kendall's $\tau$ measure in the majority of the results. 
This verifies that our approach indeed preserves the distinctive information about trained NNs while compactly relating to their common properties, including the characteristics of the training data. Presumably, learning representations on zoos with multiple different datasets would improve the out-of-distribution capabilities even further.

\section{Related Work}

There is ample research evaluating the structures of NNs by visualizing activations, \textit{e.g.}, ~\citep{zeilerVisualizingUnderstandingConvolutional2014,karpathyVisualizingUnderstandingRecurrent2015,yosinskiUnderstandingNeuralNetworks2015}, which allow some insights in the patterns of, \textit{e.g.}, the kernels of CNNs. 
Other research evaluated networks by computing a degree of similarity between networks. \citep{laaksoContentClusterAnalysis2000} compared the activations of NNs by a measure of "sameness". \citep{liConvergentLearningDifferent2015} computed correlations between the activations of different nets.~\citep{wangUnderstandingLearningRepresentations2018} tried to match the subspaces of the activation spaces in different networks~\citep{johnsonSubspaceMatchProbably2019}, which showed to be unreliable. \citep{raghuSVCCASingularVector2017, morcosInsightsRepresentationalSimilarity2018, kornblithSimilarityNeuralNetwork2019} applied correlation metrics to NN activations in order to study the learning dynamics and compare NNs. 
\citep{dinhSharpMinimaCan2017a} link model properties to characteristics of the loss surface around the minimum solution. 
In contrast, comparing models with similarity metrics, other efforts map models to an absolute representation.
~\citep{jiaGeometricStructureActivation2019} approximated the space of DNN activations with a convex hull.~\citep{jiangPredictingGeneralizationGap2019} also used activations to approximate the margin distribution and predict the generalization gap. \citep{corneanuComputingTestingError2020} proposed persistent homology by using connectivity patterns in the NN activation, and compute topological summaries. \\

While previously mentioned related work studied measures defined on the activations for insights about the NN characteristics, the methods applied on populations of NN weights have not received much attention for the same purpose.
\citep{martinTraditionalHeavyTailedSelf2019} relate the empirical spectral density of the weight matrices to accuracy, indicating that the weight space alone contains relevant information about the model. 
\citep{eilertsenClassifyingClassifierDissecting2020} evalutaed a classifier for hyper-parameter prediction directly from the weights. In contrast, we learn a general purpose hyper-representations in self-supervised fashion.~\citep{unterthinerPredictingNeuralNetwork2020} proposed layer-wise statistics derived from the NN models to predict the test accuracy of NN model. 
In \citep{schurholtInvestigationWeightSpace2021}, subspaces of the weights space of populations of NNs are investigated for model uniqueness and epoch order. \\

In the proposed work, we model associations between the different weights in NN using an attention-based module. This helps us to learn representations that compactly extract the relevant and meaningful information while considering the correlations between the weights in an NN. 
\newpage
\section{Conclusions}
In this work, we present a novel approach to learn hyper-representations from the weights of neural networks. 
To that end, we proposed novel augmentations, self-supervised learning losses and adapted multi-head attention-based architectures with suitable weight encoding for this domain. Further, we introduced several new model zoos and investigated their properties. We showed that not only learned neural networks but also their hyper-representations contain the latent footprint of their training data. 
We demonstrated high performance on downstream tasks, exceeding existing methods in hyper-parameters, test accuracy, and generalization gap prediction and showing the potential in an out-of-distribution setting. 



\begin{subappendices}
\newcommand{\icol}[1]{
  \begin{smallmatrix}#1\end{smallmatrix}%
}

\section{Permutation Augmentation}
In this appendix section, we give the full derivation about the permutation equivalence in the proposed permutation augmentation (Section \ref{sec:hyper_reps:augmentations} in the paper).
\label{ap:augmentation}
In the following appendix subsections, we show the equivalence in the \textit{forward} and \textit{backward} pass through the
neural network with original learnable parameters and the permutated neural network.

\subsection{Neural Networks and Back-propagation}
Consider a common, fully-connected feed-forward neural network (FFN). It maps inputs $\mathbf{x} \in \mathbb{R}^{N_0}$ to outputs $\mathbf{y} \in \mathbb{R}^{N_L}$. For a FFN with $L$ layers, the forward pass reads as
\begin{equation}
\begin{aligned}  
  \mathbf{a}^0 &= \mathbf{x}, \\
  \mathbf{n}^l &= \mathbf{W}^{l} \mathbf{a}^{l-1} + \mathbf{b}^{l}, \qquad l \in \{1,\cdots,L\}, \\
  \mathbf{a}^l &= \sigma(\mathbf{n}^l), \qquad l \in \{ 1,\cdots,L \}. 
\end{aligned}  
\end{equation}
Here, $\mathbf{W}^l \in \mathbb{R}^{N_{l} \times N_{l-1}}$ is the weight matrix of layer $l$, $\mathbf{b}^l$ the corresponding bias vector. Where $ N_{l}$ denotes the dimension of the layer $l$. 
The activation function is denoted by $ \sigma $, it processes the layer's weighted sum $\mathbf{n}^l$ to the layer's output $\mathbf{a}^l$. 

Training of neural networks is defined as an optimization against a objective function on a given dataset, \textit{i.e.} their weights and biases are chosen to minimize a cost function, usually called \emph{loss}, denoted by $\mathcal{L}$. The training is commonly done using a gradient based rule. Therefore, the update relies on the gradient of $\mathcal{L}$ with respect to weight ${\bf W}^l$ and the bias ${\bf b}^l$, that is it relies on $ \nabla_{\mathbf{W}} \mathcal{L} $ and $ \nabla_{\mathbf{b}} \mathcal{L} $, respectively.
Back-propagation facilitates the computation of these gradients and makes use of the chain rule to back-propagate the prediction error through the network 
~\citep{rumelhartLearningRepresentationsBackpropagating1986}. We express the error vector at layer $l$ as
\begin{equation}
  \mathbf{\delta}^l = \nabla_{\mathbf{n}^l} \mathcal{L},
\end{equation}
and further use it to express the gradients as
\begin{equation}
\begin{aligned}  
  \nabla_{\mathbf{W}^l} \mathcal{L} &= \mathbf{\delta}^l (\mathbf{a}^{l-1})^{\mathrm{T}}, \\
  \nabla_{\mathbf{b}^l} \mathcal{L} &= \mathbf{\delta}^l.
\end{aligned}  
\end{equation}

The output layer's error is simply given by
\begin{equation}
  \mathbf{\delta}^L = \nabla_{\mathbf{a}^L} \mathcal{L} \odot \sigma'(\mathbf{n}^L),
\end{equation}
where $\odot$ denotes the Hadamard element-wise product and $\sigma'$ is the activation's derivative with respect to its argument. Subsequent earlier layer's error are computed with
\begin{equation}
\begin{aligned}  
  \mathbf{\delta}^l = (\mathbf{W}^{l+1})^{\mathrm{T}} \mathbf{\delta}^{l+1} \odot \sigma'(\mathbf{n}^l), 
  \qquad l \in  \{1,\cdots,L-1 \}.
\end{aligned}    
\end{equation}
A usual parameter update takes on the form
\begin{equation}
  (\mathbf{W}^l)_{\text{new}} = \mathbf{W}^l - \beta \nabla_{\mathbf{W}^l} \mathcal{L}, \label{ap:eq:hyper_reps:update}
\end{equation}
where $\beta$ is a positive learning rate.

\subsection{Proof: Permutation Equivalence}

In the following appendix subsection, we show the permutation equivalence for feed-forward and convolutional layers.

\label{sec:hyper_reps:headings}
\paragraph*{Permutation Equivalence for Feed-forward Layers.} Consider the permutation matrix $\mathbf{P}^l \in \mathbb{N}^{N_i \times N_l}$, such that $(\mathbf{P}^l)^{\mathrm{T}} \mathbf{P}^l = \mathbf{I}$, where $ \mathbf{I}$ is the identity matrix. 
We can write the weighted sum for layer $l$ as
\begin{equation}
\begin{aligned}  
  \mathbf{n}^{l+1} &= \mathbf{W}^{l+1} \; \mathbf{a}^{l} + \mathbf{b}^{l+1} \\
               &= \mathbf{W}^{l+1} \;\sigma(\mathbf{n}^{l}) +
               \mathbf{b}^{l+1} \\
               &= \mathbf{W}^{l+1} \; \sigma(\mathbf{W}^{l} \mathbf{a}^{l-1} + \mathbf{b}^{l}) + \mathbf{b}^{l+1}.
               \label{data.feed.forward}
\end{aligned}  
\end{equation}
As $\mathbf{P}^l$ is a permutation matrix and since we use the element-wise nonlinearity $\sigma(.)$, it holds that 
\begin{align}
 \mathbf{P}^l \sigma(\mathbf{n}^l) = \sigma(\mathbf{P}^l \mathbf{n}^l),
\end{align}
which implies that we can write
\begin{equation}
\begin{aligned}  
    \mathbf{n}^{l+1}  &= \mathbf{W}^{l+1} \; \mathbf{I} \; \sigma(\mathbf{W}^{l} \; \mathbf{a}^{l-1} + \mathbf{b}^{l}) + \mathbf{b}^{l+1} \\
                &= \mathbf{W}^{l+1} \;(\mathbf{P}^l)^{\mathrm{T}} \; \mathbf{P}^l \; \sigma(\mathbf{W}^{l} \; \mathbf{a}^{l-1} + \mathbf{b}^{l}) + \mathbf{b}^{l+1} \\
                &= \mathbf{W}^{l+1} \; (\mathbf{P}^l)^{\mathrm{T}} \; \sigma( \mathbf{P}^l \; \mathbf{W}^{l} \; \mathbf{a}^{l-1} + \mathbf{P}^l \; \mathbf{b}^{l}) + \mathbf{b}^{l+1} \\
                &= \mathbf{ \hat{W} }^{l+1} \; \sigma( \mathbf{ \hat{W} }^{l} \; \mathbf{a}^{l-1} + \mathbf{ \hat{b} }^{l}) + \mathbf{b}^{l+1}, \label{ap:eq:hyper_reps:forward_perm}
\end{aligned}  
\end{equation}
where $ \mathbf{ \hat{W} }^{l+1} = \mathbf{W}^{l+1} (\mathbf{P}^l)^{\mathrm{T}}$, $ \mathbf{ \hat{W} }^{l} = \mathbf{P}^l  \mathbf{W}^{l} $ and $ \mathbf{ \hat{b} }^{l} = \mathbf{P}^l \mathbf{b}^{l} $ are the permuted weight matrices and bias vector.

Note that rows of weight matrix and bias vector of layer $l$ are exchanged together with columns of the weight matrix of layer $l+1$. In turn,  $\forall l \in \{ 1,\cdots,L-1 \}$,  \eqref{ap:eq:hyper_reps:forward_perm} holds true. At any layer $l$, there exist $N_l \!$ different permutation matrices $\mathbf{P}^l$. Therefore, in total there are $ \prod_{l=1}^{L-1} \, N_l!$ equivalent networks.

Additionally, we can write
\begin{equation}
\begin{aligned}
(\mathbf{P}^l \mathbf{W}^l)_{\text{new}} =& \mathbf{P}^l\mathbf{W}^l\! - \alpha \mathbf{P}^l \nabla_{\mathbf{W}^l} \mathcal{L} \\ 
\!\!\!\!\!=&  \mathbf{P}^l\mathbf{W}^l\! - \alpha  \mathbf{P}^l \mathbf{\delta}^l (\mathbf{a}^{l-1})^{\mathrm{T}}  \\
\!\!\!\!\!=& \mathbf{P}^l\mathbf{W}^l \!- \alpha  \mathbf{P}^l \left[ (\mathbf{W}^{l+1})^{\mathrm{T}} \mathbf{\delta}^{l+1} \odot \sigma'(\mathbf{n}^l) \right] \, (\mathbf{a}^{l-1})^{\mathrm{T}} \\
\!\!\!\!\!=& \mathbf{P}^l\mathbf{W}^l \!- \alpha  \left[ (\mathbf{W}^{l+1} \mathbf{P}^{\mathrm{T}})^{\mathrm{T}} \mathbf{\delta}^{l+1} \odot \sigma'(\mathbf{P}^l \mathbf{n}^l) \right] \, (\mathbf{a}^{l-1})^{\mathrm{T}} \\
\!\!\!\!\!=& \mathbf{P}^l\mathbf{W}^l \!- \alpha [ (\mathbf{W}^{l+1} (\mathbf{P}^l)^{\mathrm{T}})^{\mathrm{T}} \mathbf{\delta}^{l+1}  \odot  \sigma'( \mathbf{P}^l \mathbf{W}^{l} \mathbf{a}^{l-1}  +  \mathbf{P}^l \mathbf{b}^{l} ) ] \! (\mathbf{a}^{l-1})^{\mathrm{T}}.
\end{aligned}
\end{equation}

If we apply a permutation $\mathbf{P}^l$ to our update rule (equation \ref{ap:eq:hyper_reps:update}) at any layer except the last, then using the above, we can express the gradient-based update as 
\begin{align}
    &(\mathbf{\hat{W}}^l)_{\text{new}} = \mathbf{\hat{W}}^l - \alpha \left[ (\mathbf{\hat{W}}^{l+1})^{\mathrm{T}} \mathbf{\delta}^{l+1} \odot \sigma'(\mathbf{\hat{W}}^{l} \mathbf{a}^{l-1} +  \mathbf{\hat{b}}^{l} ) \right] \, (\mathbf{a}^{l-1})^{\mathrm{T}} \square \label{ap:eq:hyper_reps:update_perm}
\end{align}
The above implies that the permutations not only preserve the structural flow of information in the forward pass, 
but also preserve the structural flow of information during the update with the backward pass. That is we preserve the structural flow of information about the gradients with respect to the parameters during the backward pass through the feed-forward layers.

\newpage
\paragraph*{Permutation Equivalence for Convolutional Layers.} We can easily extend the permutation equivalence in the feed-forward layers to convolution layers. 
Consider the 2D-convolution with input channel $\mathbf{x}$ 
and $O$ output channels $\mathbf{a}_{s_1} = [\icol{ \mathbf{a}_1 \\ . \\ \mathbf{a}_O }]$. 
We express a single convolution operation for the input channel ${\bf x}$ with a convolutional kernel ${\bf K}_o, o \in \{1,..., O\}$ as
\begin{equation}
\begin{aligned} 
    \mathbf{a}_o = & \mathbf{b}_o +  \mathbf{K}_o \star \mathbf{x},   o \in \{ 1,\cdots, O \}, \label{permutations.on.kernel} 
\end{aligned} 
\end{equation}
where $\star$ denotes the discrete convolution operation. 

Note that in contrast to the whole set of permutation matrices ${\bf P}^l$ (which were introduced earlier) now we consider only a subset that affects the order of the input channels (if we have multiple) and the order of the concatenation of the output channels. 

We now show that changing the order of input channels does not affect the output if the order of kernels is changed accordingly.

The proof is similar to the permutation equivalence for the feed-forward layer. The difference here is that we take into account only the change in the order of channels and kernels. In order to prove permutation equivalence here it suffices to show that we can represent the convolution of multiple input channels by multiple convolution kernels in an alternative form, that is as matrix-vector operation.

To do so we first show that we can express the convolution of one input channel with one kernel to its equivalent matrix-vector product form. Formally, we have that
\begin{equation}
\begin{aligned} 
    \mathbf{a}_o = & \mathbf{b}_o +  \mathbf{K}_o \star \mathbf{x}  =    & \mathbf{b}_o +  \mathbf{R}_o \mathbf{x}, \label{permutations.on.kernel.alternative} 
\end{aligned} 
\end{equation}
where $\mathbf{R}_o$ is the convolution matrix. We build the matrix $\mathbf{R}_o$ in this alternative form \eqref{permutations.on.kernel.alternative} for the convolution operation from the convolutional kernel ${\bf K}_o$. $\mathbf{R}_o$ has a special structure (if we have 1D convolution then it is known as a circulant convolution matrix), while the input channel ${\bf x}$ remains the same. The number of columns in $\mathbf{R}_o$ equals the dimension of the input channel, while the number of rows in $\mathbf{R}_o$ equals the dimension of the output channel. In each row of ${\bf R}_o$, we store the elements of the convolution kernel. That is we sparsely distribute the kernel elements such that the multiplication of one row ${\bf R}_{o,j}$ of ${\bf R}_o$ with the input channel ${\bf x}$ results in the convolution output ${a}_{o,j}$ for the corresponding position $j$ at the output channel ${\bf a}_{o}$. 

The convolution of one input channel by multiple kernels can be expressed as a matrix-vector operation. In that case, the matrix in the equivalent form for the convolution with multiple kernels over one input represents a block concatenated matrix, where each of the block matrices has the previously described special structure, \textit{i.e.},
\begin{equation}
\begin{aligned} 
    \mathbf{a}_{s_1} = \left[ \icol{ \mathbf{a}_1 \\ . \\ \mathbf{a}_O } \right] = & \left[ \icol{ \mathbf{b}_{1} \\ . \\ \mathbf{b}_{O} }  \right] +  
    \left[ \begin{matrix}
    \mathbf{R}_1 \\
    . \\
    \mathbf{R}_O
    \end{matrix} \right]  
    \mathbf{x} = 
     \mathbf{b}_{f} +  {\bf R}_{f} \mathbf{x},
    \label{permutations.on.kernel.alternative.block.structure} 
\end{aligned} 
\end{equation}
where $\mathbf{b}_{f} = \left[ \icol{ \mathbf{b}_{1} \\ . \\ \mathbf{b}_{O} }  \right]$ and ${\bf R}_f = \left[ \begin{matrix}
    \mathbf{R}_1 \\
    . \\
    \mathbf{R}_O
    \end{matrix} \right]$.

In the same way the convolution of multiple input channels $\mathbf{x}_{1}, ..., \mathbf{x}_{S}$ by multiple kernels ${\bf K}_1, ..., {\bf K}_O$ can be expressed as a matrix vector operation. In that case, the matrix in the equivalent form for the convolution with multiple kernels over multiple inputs represents a block diagonal matrix, where each of the blocks in the block diagonal matrix has the previously described special structure, \textit{i.e.},
\begin{equation}
\begin{aligned} 
    \left[ \icol{ \mathbf{a}_{s_1} \\ . \\ \mathbf{a}_{s_s}}  \right] = & \left[ \icol{ \mathbf{b}_{f} \\ . \\ \mathbf{b}_{f}}  \right] +  
    \left[ \begin{matrix}
    \mathbf{R}_{f} & {\bf 0} & ... &  {\bf 0} \\
    . &. &. & . \\
    {\bf 0} & ... &  {\bf 0} & \mathbf{R}_{f} \\
    \end{matrix} \right]  
    \left[ \icol{ \mathbf{x}_{1} \\. \\ \mathbf{x}_{S} } \right]=  \left[ \icol{ \mathbf{b}_{f} \\ . \\ \mathbf{b}_{f}}  \right] +  
    {\bf R}  \left[ \icol{ \mathbf{x}_{1} \\. \\ \mathbf{x}_{S} } \right],
    \label{permutations.on.kernel.alternative.block.structure.0.5} 
\end{aligned} 
\end{equation}
where ${\bf R}=\left[ \begin{matrix}
    \mathbf{R}_{f} & {\bf 0} & ... &  {\bf 0} \\
    . &. &. & . \\
    {\bf 0} & ... &  {\bf 0} & \mathbf{R}_{f} \\
    \end{matrix} \right]$, which we can also express as
\begin{equation}
\begin{aligned} 
    \mathbf{a} =  \mathbf{b} +  {\bf R}\left[ \icol{\mathbf{x}_1 \\ . \\ \mathbf{x}_S} \right],
    \label{permutations.on.kernel.alternative.block.structure.1.0} 
\end{aligned} 
\end{equation}
where ${\bf a}=\left[ \icol{ \mathbf{a}_{s_1} \\ . \\ \mathbf{a}_{s_s}}  \right]$ and ${\bf b}=\left[ \icol{ \mathbf{b}_{f} \\ . \\ \mathbf{b}_{f}}  \right]$.

Note that the above equation has an equivalent form with equation \eqref{data.feed.forward}, therefore, the previous proof is valid for the update with respect to hole matrix ${\bf R}$. However, ${\bf R}$ has a special structure, therefore for the update of of each element in ${\bf R}$, we can use the chain rule, which results in 
\begin{equation}
\begin{aligned} 
    \frac{\partial f({\bf R})}{\partial R_{ij}}=&\sum_{k}\sum_{l}\frac{\partial f({\bf R})}{\partial R_{kl}}\frac{ \partial R_{kl}}{\partial R_{ij}}= Tr\left[ \left[ \frac{\partial f({\bf R})}{\partial {\bf R}} \right]^T  \frac{\partial {\bf R}}{\partial R_{ij}} \right].
    \label{permutations.on.kernel.alternative.block.structure.element.wise} 
\end{aligned} 
\end{equation}
Replacing $f()$ by $\mathcal{L}$ in the above and using the update rule equation \eqref{ap:eq:hyper_reps:update} gives us the update equation for ${R}_{ij}$. Using similar argumentation and derivation that leads to equation \eqref{ap:eq:hyper_reps:update_perm} concludes the proof for permutation equivalence for a convolutional layer $\square$

\begin{figure}[t!]
\begin{center}
\begin{minipage}[b]{.49\linewidth}    
    \includegraphics[trim=0in 0in 0in 0in, clip, width=1\linewidth]{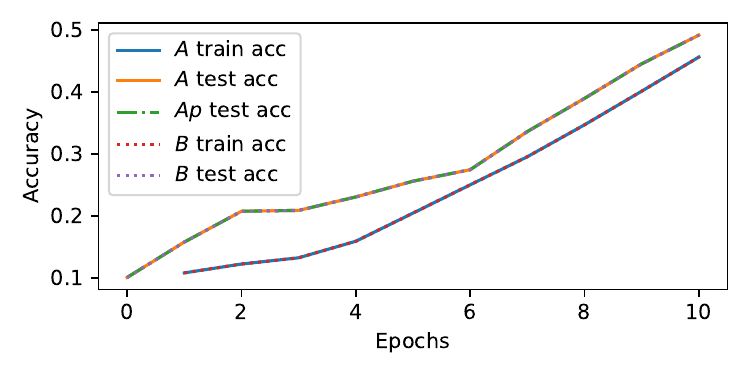}
\end{minipage}        
\begin{minipage}[b]{.49\linewidth}    
    \includegraphics[trim=0in 0in 0in 0in, clip, width=1\linewidth]{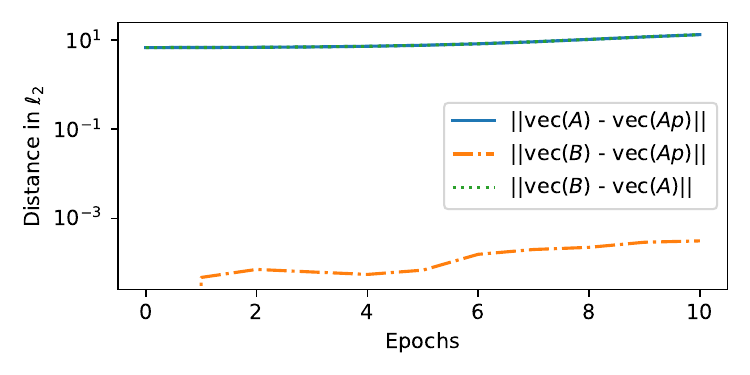}
\end{minipage}        
\caption[Empirical Evaluation of Permutation Equivalence]{Empirical Evaluation of Permutation Equivalence. \textbf{Left:} Accuracies of orignal model $A$, permuted model $Ap$ and model $B$ trained from $Ap$'s initialization. All models are indistinguishable in their accuracies. \textbf{Right:} pairwise distances of vectorized weights over epochs of models $A$, $Ap$ and $B$. The distance between $A$ and $Ap$ as well as $A$ and $B$ is equally large and does not change much over the epochs. The distance between $Ap$ and $B$, which start from the same point in weight space, is small and remains small over the epochs. \textbf{both figures:} permuted versions of models are indistinguishable in their mapping, but far apart in weight space.}
\label{fig:hyper_reps:perm_empirical_eval}    
\end{center}
\end{figure}

\paragraph*{Empirical Evaluation.} We empirically confirm the permutation equivalence (see Figure \ref{fig:hyper_reps:perm_empirical_eval}). We begin by comparing the accuracies of permuted models (Figure \ref{fig:hyper_reps:perm_empirical_eval} left). To that end, we randomly initialize model $A$ and train it for 10 epochs. We pick one random permutation and permute all epochs of model $A$. For the permuted version $Ap$ we compute the test accuracy for all epochs. The test accuracy of model $A$ and $Ap$ lie on top of each other, so the permutation equivalence holds for the forward pass.
To test the backward pass, we create model $B$ as a copy of $Ap$ at initialization and train for 10 epochs. Again, train and test epochs of $A$ and $B$ lie on top of each other, which indicates that the equivalence empirically holds for the backward pass, too.

To track how models develop in weight space, we compute the mutual $\ell_2$ distances between the vectorized weights (Figure \ref{fig:hyper_reps:perm_empirical_eval} right). The distance between $A$ and $Ap$, as well as between $A$ and $B$ is high and identical. Therefore, model $A$ is far away from models $Ap$ and $B$. Further, the distance between $Ap$ and $B$ is small, confirming the backward pass equivalence. We attribute the small difference to numerical errors.

\paragraph*{Further Weight Space Symmetries.}
It is important to note that besides the symmetry used above, other symmetries exist in the model weight space, which changes the representation of a NN model, but not its mapping, \textit{e.g.}, scaling of subsequent layers with piece-wise linear activation functions ~\citep{dinhSharpMinimaCan2017a}. While some of these symmetries may be used as augmentation, these particular mappings only create equivalent networks in the forward pass, but different gradients and updates in the backward pass when back propagating. Therefore, we did not consider them in our work.

\clearpage
\newpage

\clearpage
\newpage

\section{Downstream Tasks Additional Details}

In this appendix section, we provide additional details about the downstream tasks that we use to evaluate the utility of the hyper-representations obtained by our self-supervised learning approach.

\subsection{Downstream Tasks Problem Formulation}
\label{proxy.about.representation.utility2}
We use linear probing as a proxy to evaluate the utility of the learned hyper-representations, similar to \cite{grillBootstrapYourOwn2020}. 

We denote the training and testing hyper-representations as ${\bf Z}_{train}$ and ${\bf Z}_{test}$. We assume that training ${\bf t}_{train}$ and testing ${\bf t}_{test}$ target vectors are given. We compute the closed form solution $\hat{\bf r}$ to the regression problem 
\begin{equation*}
    (Q1):\hat{\bf r} = \arg \min_{ {\bf r} } \Vert {\bf Z}_{train}{\bf r}- {\bf t}_{train} \Vert_2^2,
\end{equation*}
and evaluate the utility of ${\bf Z}_{test}$ by measuring the $R^2$ score \cite{wright1921correlation} as discrepancy between the predicted ${\bf Z}_{test}\hat{\bf r}$ and the true test targets ${\bf t}_{test}$. 

Note that by using different targets ${\bf t}_{train}$ in $(Q1)$, we can estimate different ${\bf r}$ coefficients. This enables us to evaluate on different downstream tasks, including, accuracy prediction (Acc), epoch prediction (Eph) as proxy to model versioning, F-Score \cite{goodfellowDeepLearning2016} prediction (F$_c$), learning rate (LR), $\ell_2$-regularization ($\ell_2$-reg), dropout (Drop) and training data fraction (TF). For these target values, we solve $(Q1)$, but for categorical hyper-parameters prediction, like the activation function (Act), optimizer (Opt), and initialization method (Init), we train a linear perceptron by minimizing a cross-entropy loss \cite{goodfellowDeepLearning2016}. Here, instead of $R^2$ score, we measure the prediction accuracy.

\subsection{Downstream Tasks  Targets}

In this appendix subsection, we give the details about how we build the target vectors in the respective problem formulations for all of the downstream tasks.

\paragraph*{Accuracy Prediction (Acc).} In the accuracy prediction problem, we assume that for each trained NN model on a particular data set, we have its accuracy. Regarding the task of accuracy prediction, the value $a_{train,i}$ for the training NN models represents the training target value $t_{train,i}=a_{train,i}$, while the value $a_{test,j}$ for the testing NN models represents the true testing target value $t_{test,j}=a_{test,j}$.

\paragraph*{Generalization Gap Prediction (GGap).} In the generalization gap prediction problem, we assume that for each trained NN model on a particular data set, we have its train and test accuracy. The generalization gap represents the target value $g_{i}=a_{train,i}-a_{test,i}$ for the training NN models, while $g_{j}=a_{train,j}-a_{test,j}$ is the true target $t_{test,j}=g_{j}$ for the testing NN models.

\paragraph*{Epoch Prediction (Eph).} We consider the simplest setup as a proxy to model versioning, where we try to distinguish between NN weights and biases recorded at different epoch numbers during the training of the NNs in the model zoo. To that end, we assume that we construct the model zoo such that during the training of a NN model, we record its different evolving versions, \textit{i.e.}, the zoo includes versions of one NN model at different epoch numbers $e_{i}$.  
Similarly to the previous task, our targets for the task of epoch prediction are the actual epoch numbers, $t_{train,i}=e_{train, i}$ and $t_{test,j}=e_{test, j}$, respectively.

\paragraph*{F Score Prediction} (F$_c$). To identify more fine-grained model properties, we therefore consider the class-wise F score. We define the F score prediction task similarly as in the previous downstream task. We assume that for each NN model in the training and testing subset of the model zoo, we have computed F score \cite{goodfellowDeepLearning2016} for the corresponding class with label $c$ that we denote as F$_{train, c,i}$ and  F$_{test, c, j}$, respectively. Then we use  F$_{train,c,i}$ and F$_{test,c,i}$ as a target value $t_{train,c,i}=$F$_{train,c,i}$ in the regression problem and set  $t_{test,c,j}=$F$_{test,c,j}$ during the test evaluation.

\paragraph*{Hyper-parameters Prediction.} We define the hyper-parameter prediction task identically as the previous downstream tasks. Where for continuous hyper-parameters, like learning rate (LR), $\ell_2$-regularization ($\ell_2$-reg), dropout (Drop), and nonlinear thresholding function (TF), we solve the linear regression problem $(Q1)$. Similarly to the previous task, our targets for the task of hyperparameters prediction are the actual hyperparameters values. 

In particular, for learning rate $t_{train,i}=learning\text{ }rate$, for $\ell_2$-regularization $t_{train,i}=\ell_2\text{-}regularization\text{ }type$, for dropout (Drop)  $t_{train,i}=dropout\text{ } value$ and for nonlinear thresholding function (TF) $t_{train,i}=nonlinear\text{ } thresholding\text{ }function$. In a similar fashion, we also define the test targets ${\bf t}_{test}$.

For categorical hyper-parameters, like activation function (Act), optimizer (Opt), and initialization method (Init), instead of regression loss $(Q1)$, we train a linear perception by minimizing a cross-entropy loss \cite{goodfellowDeepLearning2016}. Here, we also define the targets as detailed above. The only difference here is that the targets have discrete categorical values.

\clearpage
\newpage

\section{Model Zoos Details}

\begin{table}[ht!]
 {
 \centering
 \scriptsize
 \sc
 \setlength{\tabcolsep}{2pt}
 \begin{tabularx}{\linewidth}{lcccccc}
 \toprule
Our Zoos & Data & NN Type & No. Param. &  Varying Prop. & No. Eph & No. NNs  \\
 \cmidrule(r){1-1} \cmidrule(l){2-7}
 \texttt{TETRIS-SEED} & TETRIS & MLP & 100 & seed (1-1000) & 75  & 1000*75\\
 \texttt{TETRIS-HYP} & TETRIS & MLP & 100 & seed (1-100), act, init, lr & 75 & 2900*75  \\
 \cmidrule(r){1-1} \cmidrule(l){2-7}
 \texttt{MNIST-SEED} & MNIST & CNN & 2464 & seed (1-1000) & 25  & 1000*25\\
  \texttt{FASHION-SEED} & F-MNIST & CNN & 2464 & seed (1-1000) & 25  & 1000*25\\
 \cmidrule(r){1-1} \cmidrule(l){2-7}
 \texttt{MNIST-HYP-1-FIX-SEED} & MNIST & CNN & 2464 & fixed seed, act, int, lr & 25  & $\sim$ 1152*25\\
 \texttt{MNIST-HYP-1-RAND-SEED} & MNIST & CNN & 2464 & random seed, act, int, lr & 25  & $\sim$ 1152*25\\
 \cmidrule(r){1-1} \cmidrule(l){2-7}
 \texttt{MNIST-HYP-5-FIX-SEED} & MNIST & CNN & 2464 & 5 fixed seeds, act, int, lr & 25  & $\sim$ 1280*25 \\
 \texttt{MNIST-HYP-5-RAND-SEED} & MNIST & CNN & 2464 & 5 random seeds, act, int, lr & 25  & $\sim$ 1280*25 \\
 \multicolumn{7}{c}{\text{ }} \\
\multicolumn{7}{c}{\text{ }} \\ 
\toprule
 Existing Zoos  & Data & NN Type & No. Param. & Varying Prop. & No. Eph & No. NNs  \\
 \cmidrule(r){1-1} \cmidrule(l){2-7}
 \texttt{MNIST-HYP} & MNIST & CNN & 4970 & act, init, opt, lr, & 9 & $\sim$ 30000*9  \\
 & & &  & $\ell_2$-reg, drop, tf & &  \\
 \texttt{FASHION-HYP} & F-MNIST & CNN & 4970 & act, init, opt, lr, & 9 & $\sim$ 30000*9  \\
 & & &  & $\ell_2$-reg, drop, tf & &  \\
 \texttt{CIFAR10-HYP} & CIFAR10 & CNN & 4970 & act, init, opt, lr,  & 9 & $\sim$ 30000*9  \\
 & & &  & $\ell_2$-reg, drop, tf & &  \\
 \texttt{SVHN-HYP} & SVHN & CNN & 4970 & act, init, opt, lr, & 9 & $\sim$ 30000*9  \\
 & & &  & $\ell_2$-reg, drop, tf & &  \\
 \bottomrule
 \end{tabularx}
 }
\end{table}
\begin{table}[ht!]
{
\centering
 \sc
 \setlength{\tabcolsep}{6pt}
\begin{tabularx}{\linewidth}{lccccc}
\toprule
&$1$ Init & $M$ Init & No data leakage & Dense checkpoints \\ 
 \cmidrule(r){1-1} \cmidrule(l){2-6}
\cite{unterthinerPredictingNeuralNetwork2020}
& ~~~ $\surd$    & $\surd$ & $\times$ & $\times$ \\ 
\cite{eilertsenClassifyingClassifierDissecting2020}
& ~~~$\surd$    & $\times$& $\surd$ & $\times$ \\ 
 \cmidrule(r){1-1} \cmidrule(l){2-6}
proposed zoos & ~~~ $\surd$ & $\surd$ & $\surd$ & $\surd$ \\
\bottomrule
\end{tabularx}
}
\caption[Overview of Model Zoo Characteristics]{Overview of the characteristics for the model zoos proposed and used (existing) in this work.}
 \label{table.intro.characteristics}
\end{table}

In Table \ref{table.intro.characteristics}, we give an overview of the characteristics for the used model zoos in this paper. This includes
\begin{itemize}
\item The data sets used for zoo creation.
\item The type of the NN models in the zoo.
\item Number of learnable parameters for each of the NNs.
\item Used number of model versions that are taken at the corresponding epochs during training.
\item Total number of NN models contained in the zoo.
\end{itemize}

In Table \ref{table.intro.characteristics} we also compare the existing and the introduced model zoos in prior and this work in terms of properties like initialization, data leakage, and presence of dense model versions obtained by recording the NN model during training evolution.

In Table \ref{tab:hyper_reps:zoo_arch_details} we provide the architecture configurations and exact modes of variation of our model zoos. 
\begin{table}[t!]
{
 \centering
 \tiny
 \setlength{\tabcolsep}{4.5pt}
 \begin{tabularx}{\linewidth}{cccccccc}
 \toprule
Our Zoos & INIT & SEED & OPT &  ACT & LR & DROP & $\ell_2$-Reg  \\
 \cmidrule(r){1-1} \cmidrule(l){2-8}
 \texttt{TETRIS-SEED} & Uniform  & 1-1000  & Adam & tanh & 3e-5 & 0.0 & 0.0 \\
 \cmidrule(r){1-1} \cmidrule(l){2-8}
                    & uniform, normal,     & 1-100  & Adam & tanh, relu & 1e-3, 1e-4, & 0.0 & 0.0 \\
 \texttt{TETRIS-HYP} &   kaiming-no, kaiming-un & & & & 1e-5\\
                    &   xavier-no, xavier-un & \\
  \cmidrule(r){1-1} \cmidrule(l){2-8}
 \texttt{MNIST-SEED}  & Uniform  & 1-1000  & Adam & tanh & 3e-4 & 0.0 & 0.0 \\
 \cmidrule(r){1-1} \cmidrule(l){2-8}
 \texttt{MNIST-HYP- 1-FIX-SEED}  & uniform, normal  & 42  & Adam, SGD & tanh, relu, & 3e-3, 1e-3,  & 0.0, 0.3, & 0, 1e-3, 1e-1 \\
                    &   kaiming-un, kaiming-no &    &   & sigmoid, gelu & 3e-4, 1e-4 & 0.5\\
 
  \cmidrule(r){1-1} \cmidrule(l){2-8}
  \texttt{MNIST-HYP- 1-RAND-SEED}   & uniform, normal  & 1$\in [1e0, 1e6]$  & Adam, SGD & tanh, relu & 3e-3, 1e-3 & 0.0, 0.3, & 0, 1e-3, 1e-1 \\
                    &   kaiming-un, kaiming-no &    &   & sigmoid, gelu & 3e-4, 1e-4 & 0.5\\
  
  \cmidrule(r){1-1} \cmidrule(l){2-8}
 \texttt{MNIST-HYP- 5-FIX-SEED}  & uniform, normal & 1,2,3,4,5  & Adam, SGD & tanh, relu& 1e-3, 1e-4 & 0.0, 0.5 & 1e-3, 1e-1 \\
                    &   kaiming-un, kaiming-no &    &   & sigmoid, gelu\\
 
  \cmidrule(r){1-1} \cmidrule(l){2-8}
 \texttt{MNIST-HYP- 5-RAND-SEED}  & uniform, normal  & 5$\in [1e0, 1e6]$  & Adam, SGD & tanh, relu & 1e-3, 1e-4 & 0.0, 0.5 & 1e-3, 1e-1 \\
                    &   kaiming-un, kaiming-no &    &   & sigmoid, gelu\\
 
  \cmidrule(r){1-1} \cmidrule(l){2-8}
  \texttt{FASHION-SEED}  & Uniform  & 1-1000  & Adam & tanh & 3e-4 & 0.0 & 0.0 \\
\bottomrule
 \end{tabularx}
}
\caption{Architecture configurations and modes of variation of our model zoos.}
\label{tab:hyper_reps:zoo_arch_details}
\end{table}

\begin{figure}[t!]
\begin{center}
\begin{minipage}[b]{.24\linewidth}    
    \includegraphics[trim=0in 0in 0in 0in, clip, width=1\linewidth]{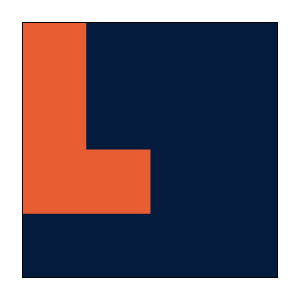}
\end{minipage}        
\begin{minipage}[b]{.24\linewidth}    
    \includegraphics[trim=0in 0in 0in 0in, clip, width=1\linewidth]{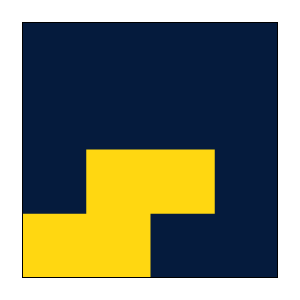}
\end{minipage}        
\begin{minipage}[b]{.24\linewidth}    
    \centering
    \includegraphics[trim=0in 0in 0in 0in, clip, width=1\linewidth]{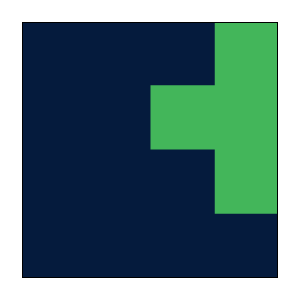}
\end{minipage}        
\begin{minipage}[b]{.24\linewidth}    
    \centering
    \includegraphics[trim=0in 0in 0in 0in, clip, width=1\linewidth]{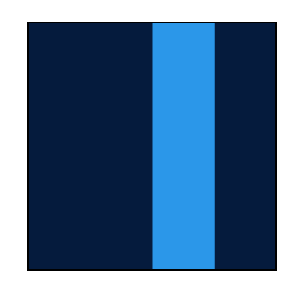}
\end{minipage}
\vskip -0.1in
    \caption[Tetris Dataset Visualization]{Visualization of samples representing the four basic shapes in our Tetris data set.}
    \label{figure.tetris.shapes}    
\end{center}
\end{figure}

\subsection{Zoos Generation Using Tetris Data}
\begin{figure}[t]
\begin{center}
\begin{minipage}[b]{1\linewidth}
\centerline{\includegraphics[trim=0.9in 3.in 5.5in .1in,clip, width=.45\linewidth]{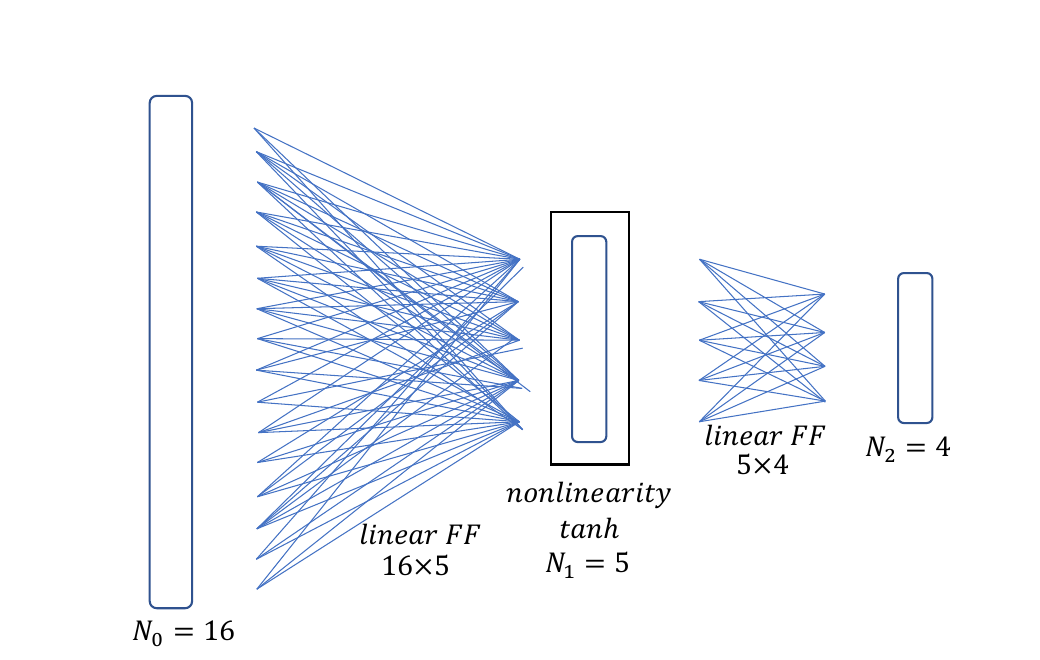}}
\end{minipage}
\caption[MLP Model Zoo Architecture Diagram]{A diagram for the feed-forward architecture of the NNs in the  \texttt{TETRIS-SEED} and \texttt{TETRIS-HYP} zoos.}
\label{figure.FNN}
\end{center}
\end{figure}

As a toy example, we first create 4x4 grey-scaled image data set that we call \emph{Tetris} by using four tetris shapes. In Figure \ref{figure.tetris.shapes} we illustrate the basic shapes of the tetris data set. We introduce two zoos, which we call \texttt{TETRIS-SEED} and \texttt{TETRIS-HYP}, which we group under \emph{small}. Both zoos contain FFN with two layers. In particular, the FFN has an input dimension of $16$ a latent dimension of $5$, and output dimension of $4$. In total the FFN has $16 \times 5 + 5\times 4 =100$ learnable parameters (see Table \ref{tab-ffn-arch}). We give an illustration of the used FFN architecture in Figure \ref{figure.FNN}.

In the \texttt{TETRIS-SEED} zoo, we fix all hyper-parameters and vary only the seed to cover a broad range of the weight space. The \texttt{TETRIS-SEED} zoo contains 1000 models that are trained for 75 epochs. In total, this zoo contains $1000 \times 75 = 75 000$ trained NN weights and biases.

To enrich the diversity of the models, the \texttt{TETRIS-HYP} zoo contains FFNs, which vary in activation function [\texttt{tanh}, \texttt{relu}], the initialization method [\texttt{uniform}, \texttt{normal}, \texttt{kaiming normal}, \texttt{kaiming uniform}, \texttt{xavier normal}, \texttt{xavier uniform}] and the learning rate [$1e$-$3$, $1e$-$4$, $1e$-$5$]. In addition, each combination is trained with 100 different seeds.
Out of the 3600 models in total, we have successfully trained 2900 for 75 epochs - the remainder crashed and are disregarded. So in total, this zoo contains $2900 \times 75 = 217 500$ trained NN weights and biases.

\subsection{Zoos Generation Using MNIST Data.}
\begin{figure}[t!]
\begin{center}
\begin{minipage}[b]{1\linewidth}
\centerline{\includegraphics[trim=.0in .0in .0in .0in,clip, width=.9\linewidth]{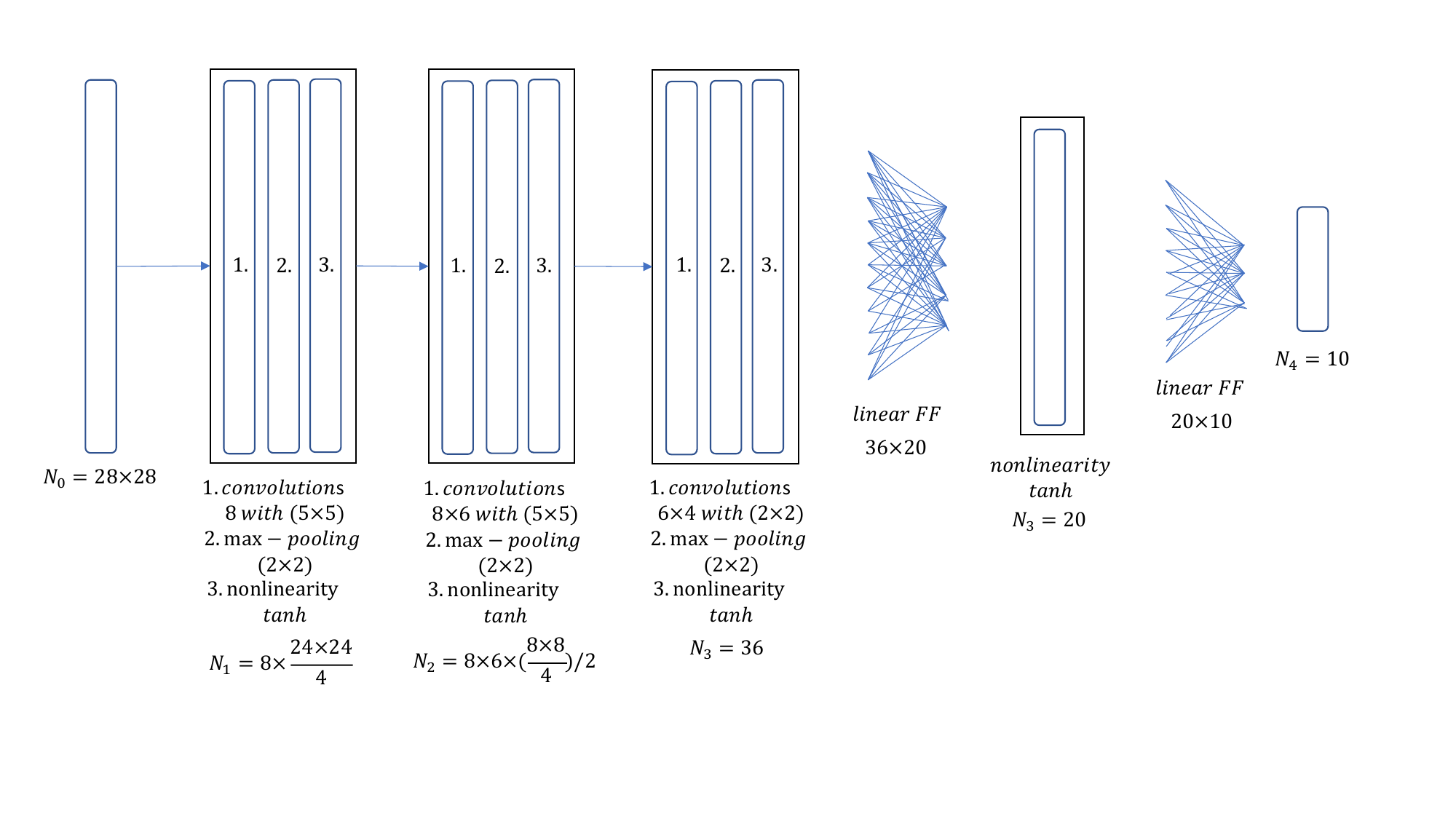}}
\end{minipage}
\caption[CNN Model Zoo Architecture Diagram]{A diagram for the CNN architecture of the NNs in the  \texttt{MNIST} zoos.}
\label{figure.CNN}
\end{center}
\end{figure}

Similarly to \texttt{TETRIS-SEED}, we further create medium-sized zoos of CNN models. 
In total the CNN has 2464 learnable parameters, distributed over 3 convolutional and 2 fully connected layers. The full architecture is detailed in Table \ref{tab-cnn-arch}. We give an illustration of the used CNN architecture in Figure \ref{figure.CNN}. Using the MNIST data set, we created five zoos with approximately the same number of CNN models.

In the \texttt{MNIST-SEED} zoo we vary only the random seed (1-1000), while using only one fixed hyper-parameter configuration. In particular, 

In \texttt{MNIST-HYP-1-FIX-SEED} we vary the hyper-parameters. We use only one fixed seed for all the hyper-parameter configurations (similarly to \citep{unterthinerPredictingNeuralNetwork2020}). The \texttt{MNIST-HYP-1-RAND-SEED} model zoo contains CNN models, where per each model we draw and use 1 random seed and different hyper-parameter configurations. 

We generate   \texttt{MNIST-HYP-5-FIX-SEED} ensuring that for each hyper-parameter configuration, we add 5 models that share 5 fixed seeds. We build \texttt{MNIST-HYP-5-RAND-SEED} such that for each hyper-parameter configuration we add 5 models that have different random seeds.

We 
grouped these model zoos as \emph{medium}. 
In total, each of these zoos approximately $1000 \times 25 = 25 000$ trained NN weights and biases.

\begin{figure}[t!]
\begin{center}
\begin{minipage}[b]{1\linewidth}
\centerline{\includegraphics[trim=0.5cm 0cm 0.1cm 0cm, clip, width=1\linewidth]{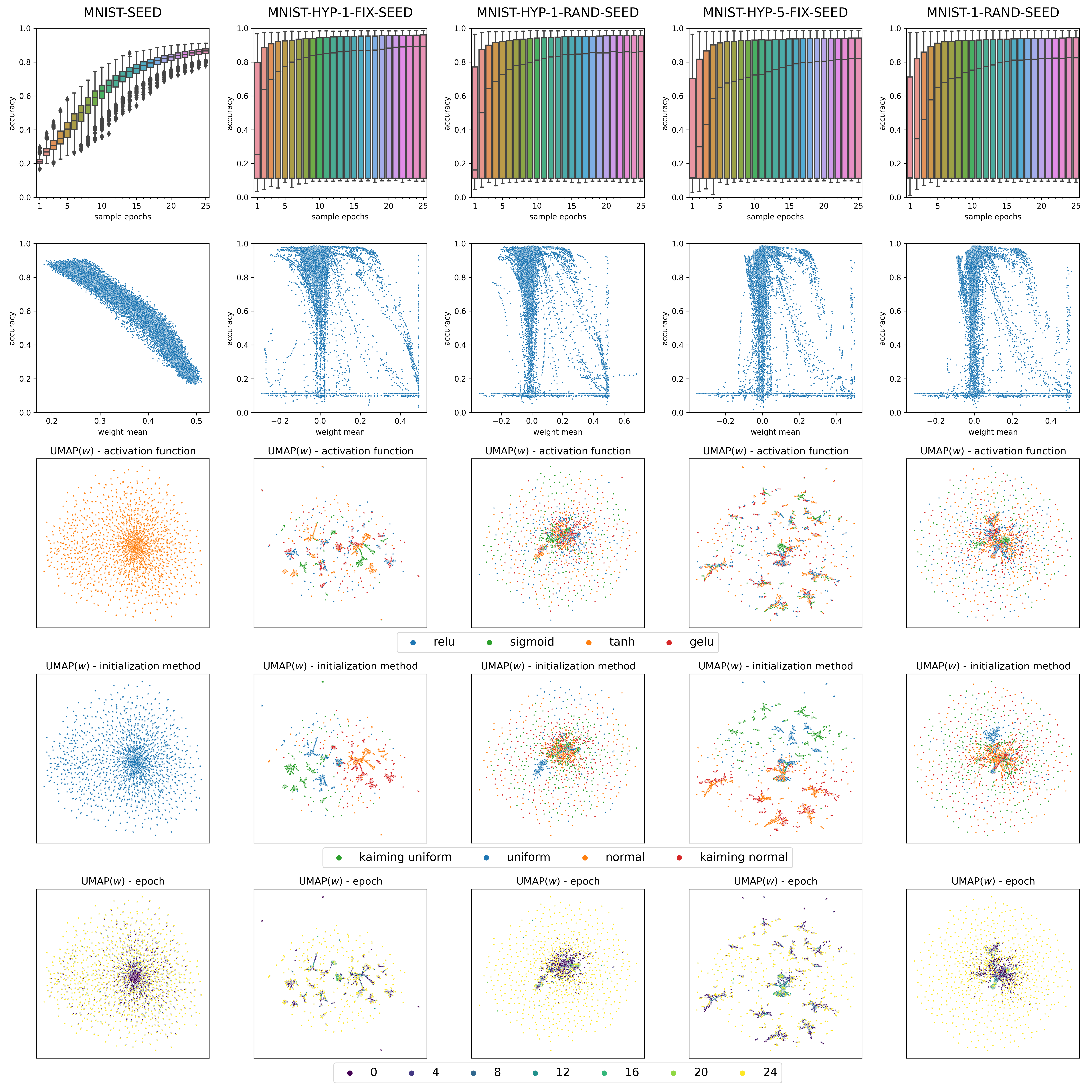}}
\end{minipage}
\caption[Visualization of MNIST Model Zoo Properties]{Visualization on the properties for the \texttt{MNIST-SEED}, \texttt{MNIST-HYP-1-FIX-SEED}, \texttt{MNIST-HYP-1-RANDOM-SEED}, \texttt{MNIST-HYP-5-FIX-SEED} and \texttt{MNIST-HYP-5-RANDOM-SEED} zoos. \textbf{Row One}. Boxplot of NNs accuracy over the epoch ids. \textbf{Row Two}. NNs accuracy plotted over the mean of the NNs weights of each sample. \texttt{MNIST-SEED} shows homogeneous development and a strong correlation between weight mean and accuracy, while varying the hyperparameters yields heterogeneous development without that correlation. \textbf{Rows Three to Five}. UMAP reductions of the weight space colored by activation function, initialization method, and sample epoch. Zoos with fixed seeds contain visible clusters of NNs that share the same initialization method or activation function. Zoos with varying hyperparameters and random seeds do not contain such clear clusters.} 
\label{figure.MNIST.zoo.props}
\end{center}
\end{figure}

In Figure \ref{figure.MNIST.zoo.props} we provide a visualization for different properties of the \texttt{MNIST-SEED}, \texttt{MNIST-HYP-1-FIX-SEED}, \texttt{MNIST-HYP-1-RANDOM-SEED}, \texttt{MNIST-HYP-5-FIX- SEED} and \texttt{MNIST- HYP-5-RANDOM-SEED} zoos. 
The visualization supports the empirical findings from the paper, that zoos that vary in seed only appear to contain a strong correlation between the mean of the weights and the accuracy. In contrast, the same correlation is considerably lower if the hyper-parameters are varied.
Further, we also observe clusters of models with shared initialization methods and activation functions for zoos with fixed seeds. Random seeds seem to disperse these clusters to some degree. This additionally confirms our hypothesis about the importance of the generating factors for the zoos. We find that zoos containing hyper-parameters variation and multiple (random) seeds have a rich set of properties, avoid 'shortcuts' between the weights (or their statistics) and properties, and therefore benefit hyper-representation learning.

\begin{figure}[ht!]
\begin{center}
\begin{minipage}[b]{1.00\linewidth}    
    \includegraphics[trim=0in 0in 0in 0in, clip, width=1\linewidth]{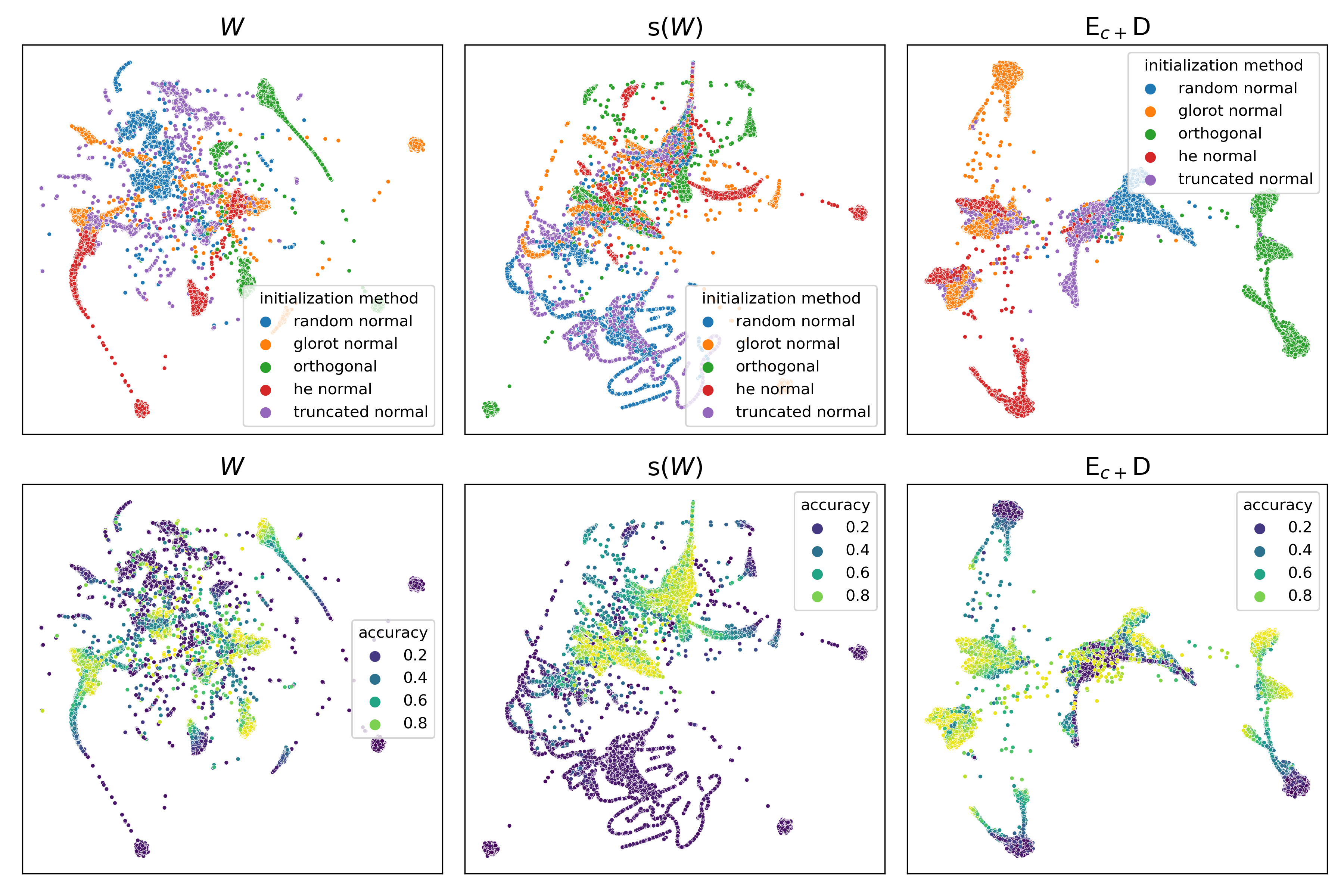}
\end{minipage}        
\vskip -0.1in
\begin{minipage}[b]{1\linewidth}    
\begin{center}
\begin{tabular}{p{3.9cm}p{3.9cm}p{3.9cm}p{3.9cm}}
\end{tabular}
\end{center}
\end{minipage}
\vskip -0.1in
    \caption[UMAP Weight Visualization of Hyper-Representations and Baselines]{UMAP dimensionality reduction of the weight space (left), weight statistics (middle) and learned hyper-representations (right) for the MNIST-HYP zoo \cite{unterthinerPredictingNeuralNetwork2020}. 
    The initialization methods for the trained NN weights 
    are already visually separable to a high degree 
    in weight space, which carries over to the learned embedding space, while the statistics introduce a mix between the initialization methods. For accuracy, in seems that the statistics filter out and contain more relevant information than the weight space. Learned embeddings appear to cluster the models according to their initialization methods and within the clusters help to preserve high accuracy.
    }
    \label{fig:hyper_reps:embedding_comparsion_mnist_hyper}    
\end{center}
\end{figure}

In Figure \ref{fig:hyper_reps:embedding_comparsion_mnist_hyper}, we show additional UMAP reductions of \texttt{MNIST-HYP}, which confirm our previous findings. Similarly to the UMAP for the \texttt{MNIST-HYP-1-FIX-SEED} zoo, the UMAP for the \texttt{MNIST-HYP} has distinctive and recognizable initialization points. The categorical hyper-parameters are visually separable in weight space. As we can see in the same figure, it seems that the UMAP for the \texttt{MNIST-HYP} zoo contains very few paths along which the evolution during learning of all the models can be tracked in weight space, facilitating both epoch and accuracy prediction.

\subsection{Zoo Generation Using F-MNIST Data}

We used the F-MNIST data set. 
As for the previous zoos for the MNIST data set, we have created one zoo with exactly the same number of CNN models as in \texttt{MNIST-SEED}.
In this zoo that we call \texttt{FASHION-SEED}, we vary only the random seed (1-1000), while using only one fixed hyper-parameter configuration.

\begin{table}[h!]
\centering
\begin{center}
\begin{small}
\begin{sc}
\begin{tabular}{l|ccc}
\toprule
 & type & details & Params \\
\midrule
1.1 & Linear    & ch-in=16, ch-out=5 & 80 \\
1.2 & Nonlin. & Tanh \\
\midrule
2.1 & Linear    & ch-in=5, ch-out=4 & 20\\
\end{tabular}
\end{sc}
\end{small}
\end{center}
\caption[MLP Model Zoo Architecture Details]{FFN Architecture Details. \textsc{ch-in} describes the number of input channels, \textsc{ch-out} the number of output channels.}
\label{tab-ffn-arch}
\end{table}

\begin{table}[h!]
\centering
\vskip 0.1in
\begin{center}
\begin{small}
\begin{sc}
\begin{tabular}{l|ccc}
\toprule
 & type & details & Params \\
\midrule
1.1 & Conv    & ch-in=1, ch-out=8, ks=5 & 208 \\
1.2 & MaxPool & ks= 2 \\
1.3 & Nonlin. & Tanh \\
\midrule
2.1 & Conv    & ch-in=8, ch-out=6, ks=5 & 1206 \\
2.2 & MaxPool & ks= 2 \\
2.3 & Nonlin. & Tanh \\
\midrule
3.1 & Conv    & ch-in=6, ch-out=4, ks=2 & 100 \\
3.2 & MaxPool & ks= 2 \\
3.3 & Nonlin. & Tanh \\
\midrule
4 & Flatten & \\
\midrule
5.1 & Linear    & ch-in=36, ch-out=20 & 740 \\
5.2 & Nonlin. & Tanh \\
\midrule
6.1 & Linear    & ch-in=20, ch-out=10 & 200
\end{tabular}
\end{sc}
\end{small}
\end{center}
\caption[CNN Model Zoo Architecture Details]{CNN Architecture Details. \textsc{ch-in} describes the number of input channels, \textsc{ch-out} the number of output channels. \textsc{ks} denotes the kernel size, kernels are always square.}
\label{tab-cnn-arch}
\end{table}

\clearpage
\newpage

\section{Additional Results}

In this appendix section, we provide additional results about the impact of the compression ratio $c=N/L$.

\subsection{Impact of the Compression Ratio N/L}

In this subsection, we first explain the experiment setup and then comment on the results about the impact of the compression ratio on the performance for downstream tasks.

\textbf{Experiment Setup.} To see the impact of the compression ratio $c=N/L$ on the performance over the downstream tasks, we use our hyper-representation learning approach under different types of architectures, including E${_c}$, ED, and E${_c}$D (see Section 3 in the paper). 
As encoders E and decoders D, we used the attention-base modules introduced in Section 3 in the paper. The attention-based encoder and decoder, on the \texttt{TETRIS-SEED} and \texttt{TETRIS-HYP} zoos, we used 2 attention blocks with 1 attention head each, token dimensions of 128 and FC layers in the attention module of dimension 512. 

We use our weight augmentation methods for representation learning (please see Section 3.1 in the paper). We run our representation learning algorithm for up to 2500 epochs, using the adam optimizer \cite{Kingma:2014:Adam}, a learning rate of 1e-4, weight decay of 1e-9, dropout of 0.1 percent, and batch-sizes of 500. In all of our experiments, we use 70\% of the model zoos for training and 15\% for validation and testing each. We use checkpoints of all epochs but ensure that samples from the same models are either in the train or in the test split of the zoo. As a quality metric for self-supervised learning, we track the reconstruction $R^2$ on the test split of the zoo.

\textbf{Results.} As Table \ref{ablation_latdim} shows, all NN architectures decrease in performance, as the compression ratio increases. 
The purely contrastive setup E$_c$ generally learns embeddings which are useful for the downstream tasks, which are very stable under compression. These results strongly depend on a projection head with enough complexity. The closer the contrastive loss comes to the bottleneck of the encoder, the stronger the downstream tasks suffer under compression.
Notably, the reconstruction of ED is very stable, even under high compression ratios. However, higher compression ratios appear to negatively impact the hyper-representations for the downstream tasks we consider here.
The combination of reconstruction and contrastive loss shows the best performance for $c=2$, but suffers under compression. Higher compression ratios perform comparably on the downstream tasks, but don't manage high reconstruction $R^2$. We interpret this as sign that the combination of losses requires high capacity bottlenecks. If the capacity is insufficient, the two objectives can't be both satisfied.

\begin{table}[t]
\begin{minipage}[b]{1\linewidth}
\centering
\text{\small Encoder with contrastive loss E$_c$}
\label{tab:compression_Ec}
\begin{center}
\begin{small}
\begin{sc}
\begin{tabular}{p{.6cm}|p{.5cm}c p{.6cm}p{.55cm}p{.55cm}p{.55cm}p{.6cm}p{.55cm}}
\toprule
$c$ & Rec & Eph &Acc & Ggap & F$_{C0}$ & F$_{C1}$ & F$_{C2}$ & F$_{C3}$  \\
\midrule
2 & -- & 96.7 & 90.8 & 82.5 & 67.7 & 72.0 & 74.4 & 85.8 \\
3 & -- & 96.6 & 89.4 & 81.5 & 68.4 & 69.4 & 71.1 & 85.1 \\
5 & -- & 96.4 & 89.5 & 81.8 & 67.1 & 68.7 & 69.7 & 84.0 \\
\end{tabular}
\end{sc}
\end{small}
\end{center}
\end{minipage}
\vskip 0.1in
\begin{minipage}[b]{1\linewidth}
\centering
\text{\small Encoder and decoder with reconstruction loss ED }
\label{tab:compression_ED}
\begin{center}
\begin{small}
\begin{sc}
\begin{tabular}{p{.6cm}|p{.5cm}c p{.6cm}p{.55cm}p{.55cm}p{.55cm}p{.6cm}p{.55cm}}
\toprule
$c$ & Rec & Eph &Acc & Ggap & F$_{C0}$ & F$_{C1}$ & F$_{C2}$ & F$_{C3}$  \\
\midrule
2 & 96.1 & 88.3 & 68.9 & 69.9 & 47.8 & 57.2 & 33.0 & 58.1 \\
3 & 93.0 & 74.6 & 69.4 & 66.9 & 53.5 & 46.5 & 38.9 & 48.3 \\
5 & 87.7 & 80.5 & 60.0 & 63.3 & 37.9 & 48.8 & 24.4 & 52.6 \\
\end{tabular}
\end{sc}
\end{small}
\end{center}
\end{minipage}
\vskip 0.1in
\begin{minipage}[b]{1\linewidth}
\centering
\text{\small Encoder and decoder with}
\text{\small reconstruction and contrastive loss E$_{c}$D }
\label{tab:compression_EcD}
\begin{center}
\begin{small}
\begin{sc}
\begin{tabular}{p{.6cm}|p{.5cm}c p{.6cm}p{.55cm}p{.55cm}p{.55cm}p{.6cm}p{.55cm}}
\toprule
$c$ & Rec  & Eph & Acc & Ggap & F$_{C0}$ & F$_{C1}$ & F$_{C2}$ & F$_{C3}$  \\
\midrule
2 & 84.1 & 97.0 & 90.2 & 81.9 & 70.7 & 75.9 & 69.4 & 86.6 \\
3 & 75.6 & 96.3 & 88.3 & 80.7 & 66.9 & 70.8 & 66.1 & 83.2  \\
5 & 64.5 & 96.3 & 85.2 & 80.0 & 63.5 & 68.0 & 61.3 & 73.6 \\
\end{tabular}
\end{sc}
\end{small}
\end{center}
\end{minipage}
\vskip 0.2in
\caption[Hyper-Representation Compression Rationn Ablation Study]{The impact of the compression ratio $c=N/L$ in the different NN architectures of our approach for learning hyper-representations over the \texttt{Tetris-Seed} Model Zoo. All values are $R^2$ scores and given in \%.}
\label{ablation_latdim}
\end{table}

\subsection{NN Model Characteristics Prediction on FASHION-SEED}

Due to space limitations, here in Figure \ref{result.mnist-seed-fashion-seed}, we present the results on the \texttt{FASHION-SEED} together with the results on \texttt{MNIST-SEED}. The experimental setup is the same as for the \texttt{MNIST-SEED} zoo, which is explained in the paper. Here, we add a complementary result to our ablation study about the seed variation, that we presented in section 4.3 in the paper. Similarly to the discussion in the paper, random seed variation in the \texttt{FASHION-SEED} again appears to make the prediction more challenging. The results show that the proposed approach is on par with the comparing $s(W)$  for this type of model zoo.

\begin{table}[ht!]
\begin{minipage}[b]{1\linewidth}
\begin{center}
\begin{small}
\begin{sc}
\begin{tabular}{l|ccc|ccc}
\toprule
& \multicolumn{3}{c}{\texttt{MNIST-SEED}}& \multicolumn{3}{c}{\texttt{FASHION-SEED} } \\
 & W & s(W) & E${_{c+}}$D & W & s(W) & E${_{c+}}$D  \\
\midrule
Eph &  84.5 & \textbf{97.7} & 97.3 & 87.5 & \textbf{97.0} & 95.8 \\
\midrule
Acc   & 91.3 & 98.7 & \textbf{98.9} & 88.5 & 97.9 & \textbf{98.0} \\
\midrule
GGap   & 56.9 & 66.2 & \textbf{66.7}  & 70.4 & 81.4 & \textbf{83.2} \\

\end{tabular}
\end{sc}
\end{small}
\end{center}
\end{minipage}
\caption[Model Property Prediction Results for MNIST and Fashion-MNIST Zoos]{$R^2$ score in \% for epoch, accuracy and generalization gap. 
}
\label{result.mnist-seed-fashion-seed}
\end{table}

\subsection{In-distribution and Out-of-distribution Prediction}

\begin{figure}[ht!]
\begin{center}
\begin{minipage}[b]{1\linewidth}
\centerline{\includegraphics[trim=0.7in 0in 0.7in 0in, clip, width=1\linewidth]{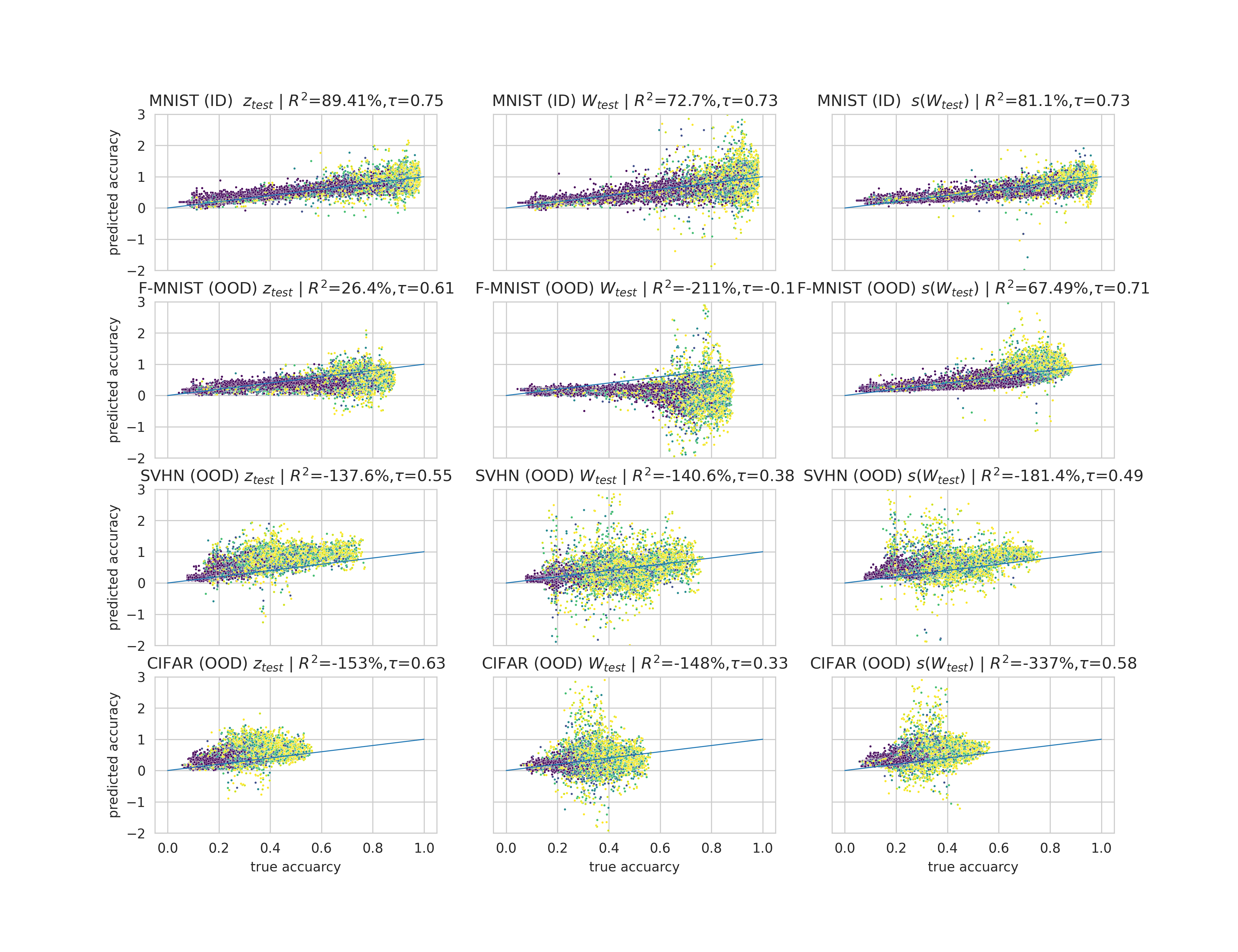}}
\end{minipage}
\begin{minipage}[b]{.99\linewidth}  
\begin{center}
\begin{small}
\begin{sc}
{\small
\setlength{\tabcolsep}{2pt}
\begin{tabularx}{\linewidth}{l|ccc|ccc|ccc|ccc}
\toprule
& \multicolumn{3}{c}{\texttt{MNIST-HYP}} & \multicolumn{3}{c}{\texttt{FASHION-HYP}} &  \multicolumn{3}{c}{\texttt{SVHN-HYP}}  & \multicolumn{3}{c}{\texttt{CIFAR10-HYP}}\\
& W & s(W) & E${_{c\text{+}}}$D & W & s(W) & E${_{c\text{+}}}$D & W & s(W) & E${_{c\text{+}}}$D & W & s(W) & E${_{c\text{+}}}$D   \\
\midrule
\texttt{MNIST-HYP} ($\tau$) & \cellcolor{Gray!10} .73 & \cellcolor{Gray!10} .73 & \cellcolor{Gray!10} \textbf{.75} & \text{-}.08 & \textbf{.71} & .61 & .38 & .49 & \textbf{.55} & .33 & .58 & \textbf{.63} \\
\texttt{MNIST-HYP} ($R^2$) & \cellcolor{Gray!10} 72.7 & \cellcolor{Gray!10} 81.1 & \cellcolor{Gray!10} \textbf{89.4} & \text{-}211 & \textbf{67} & 26 & \text{-}140 & \text{-}180 & \text{-}\textbf{137} 
& \text{-}\textbf{148} & \text{-}337 & \text{-}153 \\
\end{tabularx}
}
\end{sc}
\end{small}
\end{center}
\end{minipage}
\caption[Accuracy Prediction for Hyperparameter Variation Zoos]{In-distribution and out-of-distribution results for test accuracy prediction. Representation learning model and linear probes are trained on \texttt{MNIST-HYP}, and evaluated on \texttt{MNIST-HYP}, \texttt{FASHION-HYP}, \texttt{SVHN-HYP} and \texttt{CIFAR-HYP}.}
\label{figure.ood.MNIST.accuracy}
\end{center}
\end{figure}

\begin{figure}[ht!]
\begin{center}
\begin{minipage}[b]{1\linewidth}
\centerline{\includegraphics[trim=0.7in 0in 0.7in 0in, clip, width=1\linewidth]{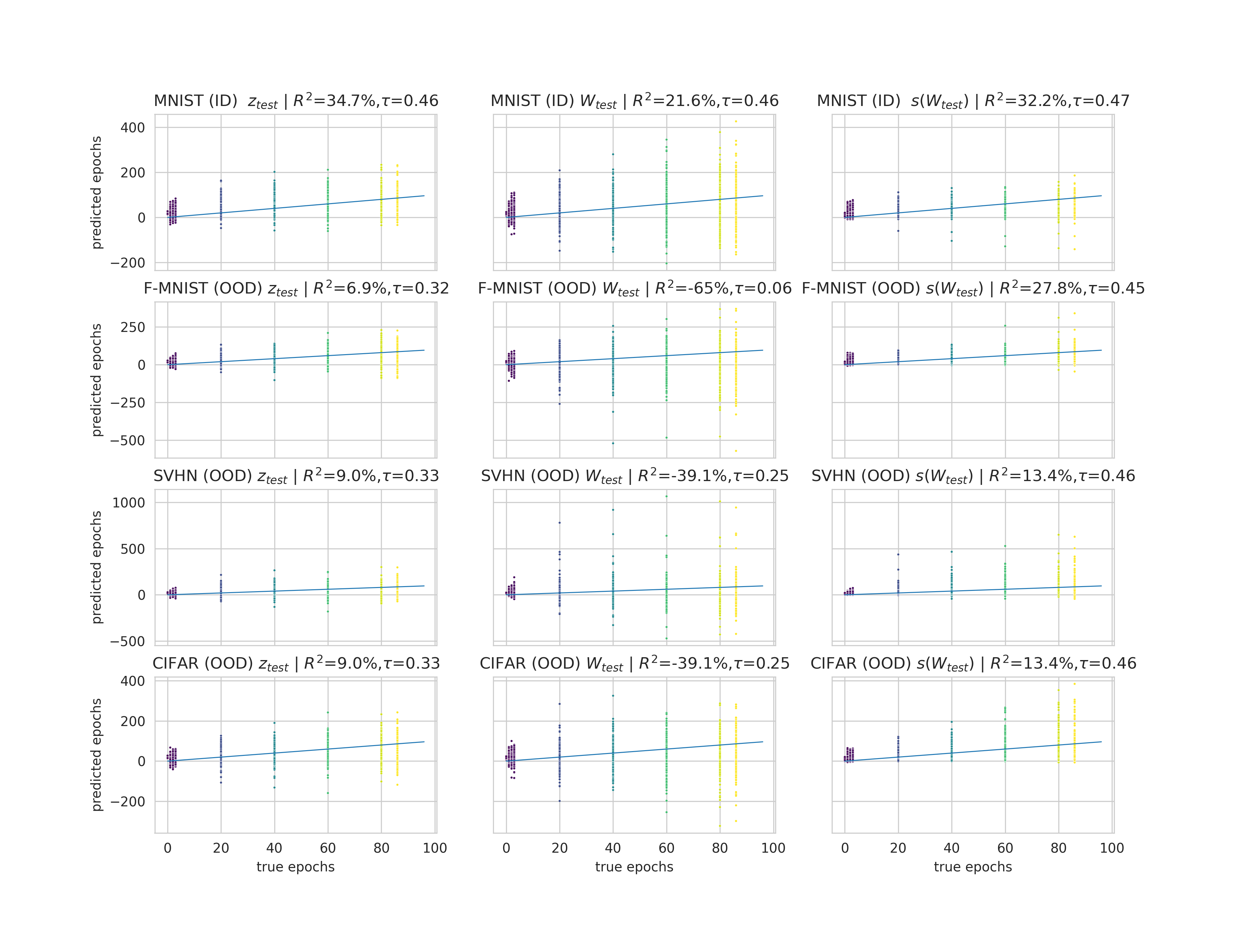}}
\end{minipage}
\vspace{.05in}
\begin{minipage}[b]{.99\linewidth}  
\begin{center}
\begin{small}
\begin{sc}
{\small
\setlength{\tabcolsep}{2pt}
\begin{tabularx}{\linewidth}{l|ccc|ccc|ccc|ccc}
\toprule
& \multicolumn{3}{c}{\texttt{MNIST-HYP}} & \multicolumn{3}{c}{\texttt{FASHION-HYP}} &  \multicolumn{3}{c}{\texttt{SVHN-HYP}}  & \multicolumn{3}{c}{\texttt{CIFAR10-HYP}}\\
& W & s(W) & E${_{c\text{+}}}$D & W & s(W) & E${_{c\text{+}}}$D & W & s(W) & E${_{c\text{+}}}$D & W & s(W) & E${_{c\text{+}}}$D   \\
\midrule
\texttt{MNIST-HYP} ($\tau$) & \cellcolor{Gray!10} .46 & \cellcolor{Gray!10} \textbf{.47} & \cellcolor{Gray!10} .46 & .06 & \textbf{.45} & .32 & .25 & \textbf{.46} & .33 & .18 & \textbf{.41} & .16 \\
\texttt{MNIST-HYP} ($R^2$) & \cellcolor{Gray!10} 21.6 & \cellcolor{Gray!10} 32.2 & \cellcolor{Gray!10} \textbf{34.7} & 
\text{-}64.9 & \textbf{27.8} & 6.9 & 
\text{-}39.1 & \textbf{13.4} & 9. & 
\text{-}21.9 & \textbf{19.2} & \text{-}13.\\
\end{tabularx}
}
\end{sc}
\end{small}
\end{center}
\end{minipage}
\caption[Epoch Prediction for Hyperparameter Variation Zoos]{In-distribution and out-of-distribution results for the epoch id predictions. Representation learning model and linear probes are trained on \texttt{MNIST-HYP}, and evaluated on \texttt{MNIST-HYP}, \texttt{FASHION-HYP}, \texttt{SVHN-HYP} and \texttt{CIFAR-HYP}.}
\label{figure.ood.MNIST.epoch}
\end{center}
\end{figure}

\begin{figure}[ht!]
\begin{center}
\begin{minipage}[b]{1\linewidth}
\centerline{\includegraphics[trim=0.7in .1in 0.7in 0in, clip, width=1\linewidth]{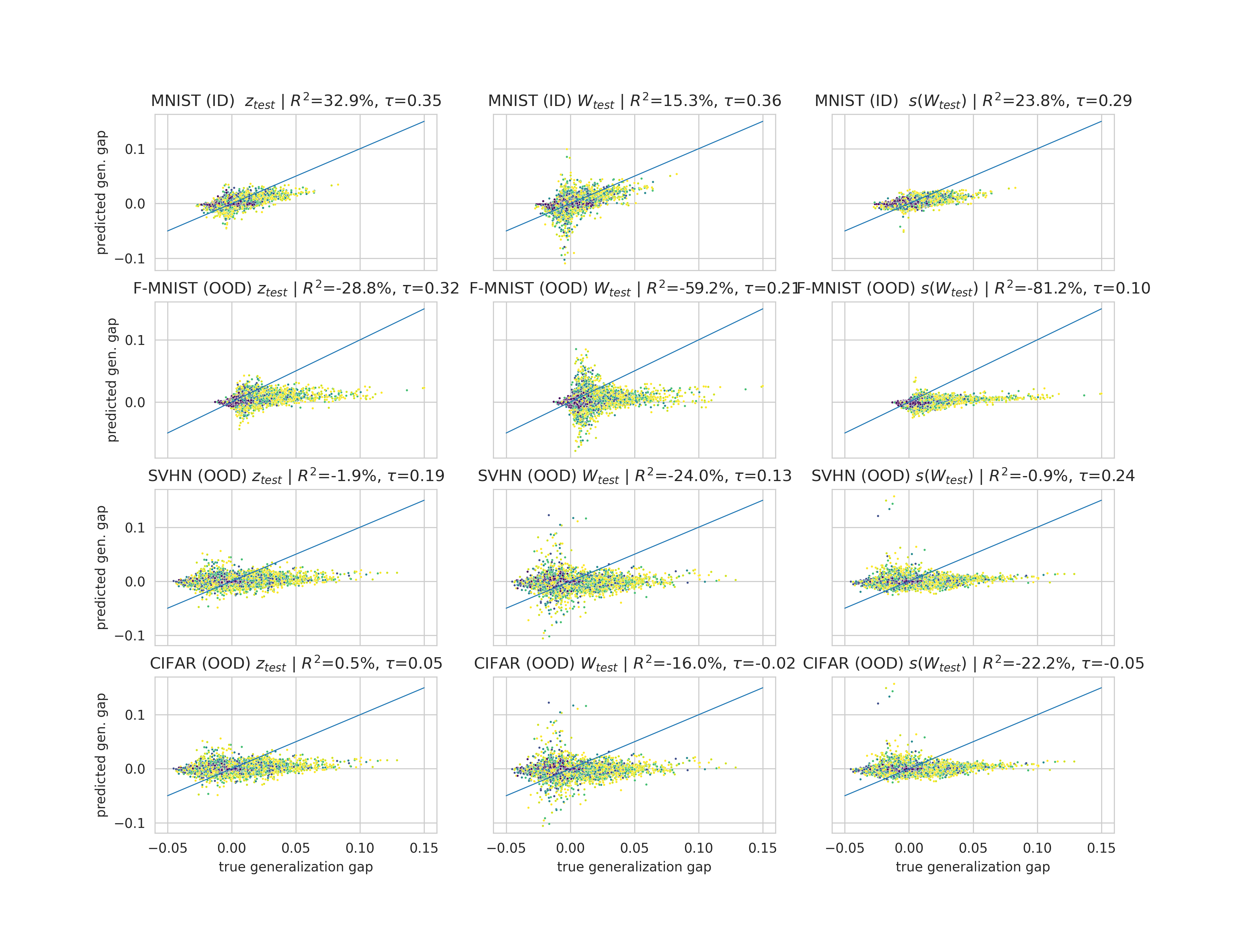}}
\end{minipage}
\begin{minipage}[b]{.99\linewidth}  
\begin{center}
\begin{small}
\begin{sc}
{\small
\setlength{\tabcolsep}{2pt}
\begin{tabularx}{\linewidth}{l|ccc|ccc|ccc|ccc}
\toprule
& \multicolumn{3}{c}{\texttt{MNIST-HYP}} & \multicolumn{3}{c}{\texttt{FASHION-HYP}} &  \multicolumn{3}{c}{\texttt{SVHN-HYP}}  & \multicolumn{3}{c}{\texttt{CIFAR10-HYP}}\\
& W & s(W) & E${_{c\text{+}}}$D & W & s(W) & E${_{c\text{+}}}$D & W & s(W) & E${_{c\text{+}}}$D & W & s(W) & E${_{c\text{+}}}$D   \\
\midrule
\texttt{MNIST-HYP} ($\tau$)& \cellcolor{Gray!10} \textbf{.36} & \cellcolor{Gray!10} .29 & \cellcolor{Gray!10} .35 & 
.20 & .10 & \textbf{.32} &
.13 & \textbf{.24} & .19 &
\text{-}.05 & \text{-}.02 & \textbf{.05} \\
\texttt{MNIST-HYP} ($R^2$)& \cellcolor{Gray!10} 15.3 & \cellcolor{Gray!10} 24.8 & \cellcolor{Gray!10} \textbf{32.9} & 
\text{-}56.2 & \text{-}81.8 & \text{-}\textbf{27.8}  
& \text{-}24. & \text{-}\textbf{.9} &
\text{-}1.9 
& \text{-}16. & \text{-}22.2 & \textbf{.5}
\end{tabularx}
}
\end{sc}
\end{small}
\end{center}
\end{minipage}
\caption[Generalization Gap Prediction for Hyperparameter Variation Zoos]{In-distribution and out-of-distribution results for the generalization gap predictions. Representation learning model and linear probes are trained on \texttt{MNIST-HYP}, and evaluated on \texttt{MNIST-HYP}, \texttt{FASHION-HYP}, \texttt{SVHN-HYP} and \texttt{CIFAR-HYP}.}
\label{figure.ood.MNIST.ggap}
\end{center}
\end{figure}

In Figures \ref{figure.ood.MNIST.accuracy},  
\ref{figure.ood.MNIST.epoch} and
\ref{figure.ood.MNIST.ggap}  
we show in-distribution and out-of-distribution comparative results for test accuracy, epoch id, and generalization gap prediction using the \texttt{MNIST-HYP} zoo. 

In the majority of the results for accuracy and generalization gap prediction, our learned representations have higher $R^2$ and Kendall's $\tau$ score. Also, in the baseline methods the distribution of predicted target values is more dispersed compared to the true target values. On the epoch id prediction we have comparable results but with lower score, we attribute this to the fact that the zoos contain sparse checkpoints and we suspect that there are not enough so that our learning model could capture the present variability. Overall in the in-distribution and out-of-distribution results for test accuracy, epoch id, and generalization gap prediction, the proposed approach has a slight advantage.

\begin{figure}[ht!]
\begin{center}
\begin{minipage}[b]{1\linewidth}
\centerline{\includegraphics[trim=0.7in 0in 0.7in 0in, clip, width=1\linewidth]{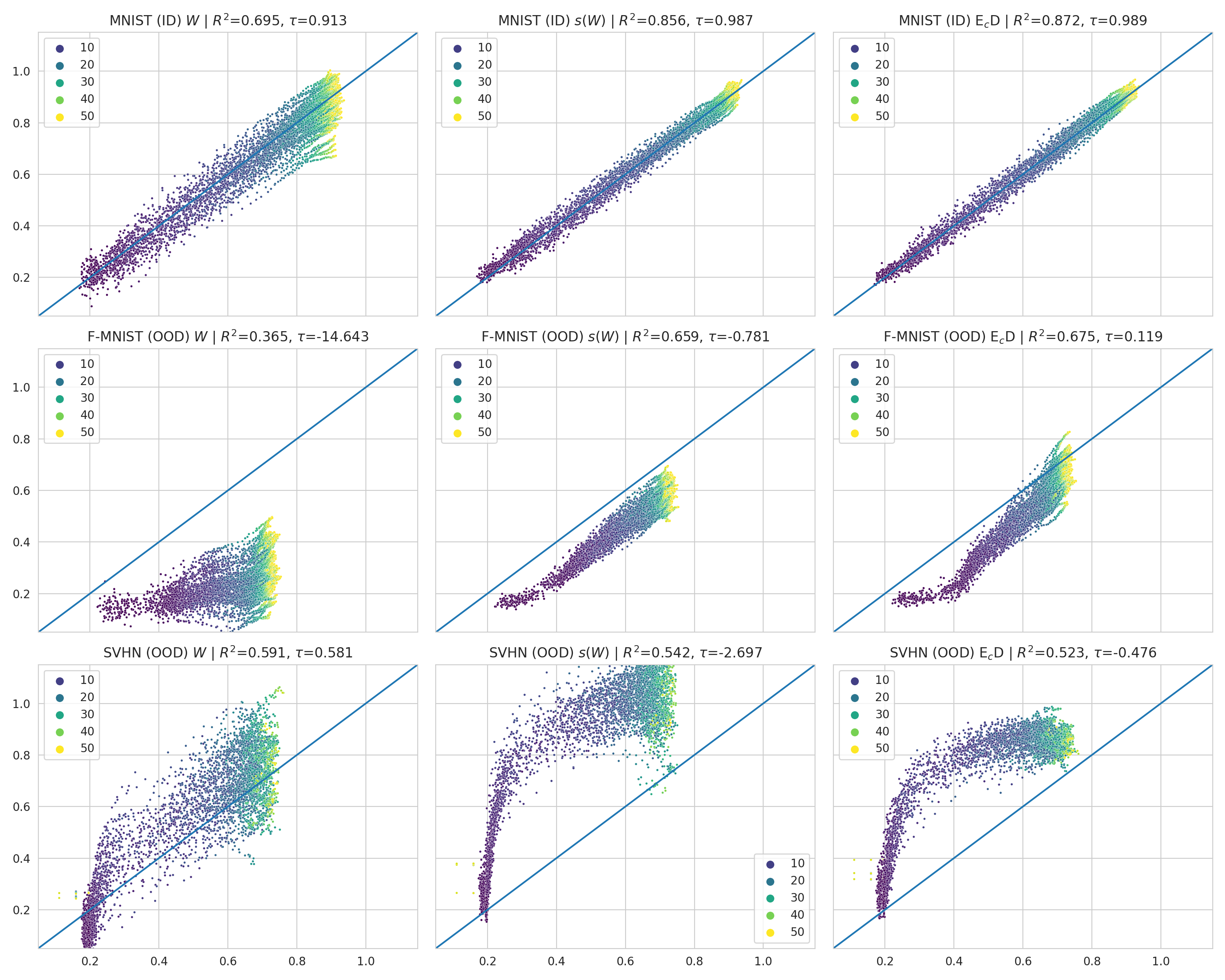}}
\end{minipage}
\begin{minipage}[b]{.99\linewidth}  
\begin{center}
\begin{small}
\begin{sc}
{\small
\setlength{\tabcolsep}{4pt}
\begin{tabularx}{\linewidth}{l|ccc|ccc|ccc}
\toprule
& \multicolumn{3}{c}{\texttt{MNIST-SEED}} & \multicolumn{3}{c}{\texttt{FASHION-SEED}} &  \multicolumn{3}{c}{\texttt{SVHN-SEED}}\\
& W & s(W) & E${_{c\text{+}}}$D & W & s(W) & E${_{c\text{+}}}$D & W & s(W) & E${_{c\text{+}}}$D   \\
\midrule
\texttt{MNIST-SEED} ($\tau$) & 
\cellcolor{Gray!10} .913 & \cellcolor{Gray!10}.987 & \cellcolor{Gray!10} \textbf{.989} 
& -14 & -.781 & \textbf{.12} 
& \textbf{.581} & -2.7 & -.48 \\
\texttt{MNIST-SEED} ($R^2$) & 
\cellcolor{Gray!10} 69.5 & \cellcolor{Gray!10} 85.6 & \cellcolor{Gray!10} \textbf{87.2} 
& 36.5 & 65.9 & \textbf{67.5}
& \textbf{.591} & .542 & .523 \\
\\
\end{tabularx}
}
\end{sc}
\end{small}
\end{center}
\end{minipage}
\caption[Accuracy Prediction for the MNIST-Seed Zoo]{In-distribution and out-of-distribution results for test accuracy prediction. Representation learning model and linear probes are trained on \texttt{MNIST-SEED}, and evaluated on \texttt{MNIST-SEED}, \texttt{FASHION-SEED} and \texttt{SVHN-SEED}.}
\label{figure.ood.MNIST-seed.accuracy}
\end{center}
\end{figure}

Due to space limitations, for the \texttt{MNIST-SEED}, \texttt{FASHION-SEED} zoos and an additional \texttt{SVHN-SEED} zoo we only include out-of-distribution results for accuracy prediction in Figure \ref{figure.ood.MNIST-seed.accuracy}. Here, too, our learned representations have higher scores in both Kendall's $\tau$ as well as $R^2$. Further, the accuracy prediction for \texttt{SVHN-SEED} clearly preserves the order, but has a noticeable bias. We attribute that effect to the different accuracy distributions of \texttt{MNIST-SEED} (ID, accuracy: [0.2,0.95]) and \texttt{SVHN-SEED} (OOD, accuracy: [0.2,0.75]). Due to the higher accuracy in \texttt{MNIST-SEED}, we suspect that the accuracy in \texttt{SVHN-SEED} is overestimated. 

\end{subappendices}
\FloatBarrier
\newpage
\clearpage
\cleardoublepage

\newpage
\thispagestyle{empty}
\hbox{}
\afterpage{
    \thispagestyle{emptychaptertransition}
    \vspace*{\fill}
    \newpage
    \thispagestyle{plain}
}

\addtocontents{toc}{\newpage}
\chapter{Hyper-Representations as Generative Models: Sampling Unseen Neural Network Weights}
\label{chap::generative_hyper_reps}

\blfootnote{This work was accepted for publication at NeurIPS 2022 \citep{schurholtHyperRepresentationsGenerativeModels2022} }

\section*{Abstract}
Learning representations of neural network weights given a model zoo is an emerging and challenging area with many potential applications from model inspection, to neural architecture search or knowledge distillation.
Recently, an autoencoder trained on a model zoo was able to learn a \textit{hyper-representation}, which captures intrinsic and extrinsic properties of the models in the zoo.
In this work, we extend hyper-representations for generative use to sample new model weights.
We propose layer-wise loss normalization which we demonstrate is key to generate high-performing models and several sampling methods based on the topology of hyper-representations.
The models generated using our methods are diverse, performant, and capable of outperforming strong baselines as evaluated on several downstream tasks: initialization, ensemble sampling, and transfer learning. 
Our results indicate the potential of knowledge aggregation from model zoos to new models via hyper-representations thereby paving the avenue for novel research directions.\looseness-1

\section{Introduction}
Over the last decade, countless neural network models have been trained and uploaded to different model hubs.
Many factors such as random initialization and no global optimum ensure that the trained models are different from one another.
What could we learn from such a population of neural network models?
Since the parameter space of neural networks is complex and high-dimensional, representation learning from such populations (often referred to as model zoos) has become an emerging and challenging area.

Recent work along that direction has demonstrated the ability of such learned representations to capture intrinsic and extrinsic properties of the models in a zoo~\citep{unterthinerPredictingNeuralNetwork2020,
schurholtSelfSupervisedRepresentationLearning2021,martinPredictingTrendsQuality2021}.
According to ~\citep{schurholtSelfSupervisedRepresentationLearning2021}, NNs populate a low dimensional manifold, which can be learned with an autoencoder via self-supervised learning directly from the model parameters (weights and biases)
without access to the original image data and labels. This so-called \textit{hyper-representation} has been demonstrated to be useful for predicting several model properties such as accuracy, hyperparameters, or architecture configurations.\looseness-1

However,  ~\citep{schurholtSelfSupervisedRepresentationLearning2021} focused on discriminative downstream tasks by exploiting the encoder only. We take one step further and extend their work towards the generative downstream tasks by sampling model weights directly from the task-agnostic hyper-representation. 
To that end, we introduce a layer-wise normalization that improves the quality of decoded neural network weights significantly. Based on a careful analysis of the geometry, smoothness, and robustness of this space, we also propose several sampling methods to generate weights in a single forward pass from the hyper-representation.
We evaluate our approach on four image datasets and three generative downstream tasks of (i) model initialization, (ii) ensemble sampling, and (iii) transfer learning. Our results demonstrate its capability to outperform previous hyper-representation learning and conventional baselines.\looseness-1

Previous work on generating model weights proposed (Graph) HyperNetworks~\citep{haHyperNetworks2017,zhangGraphHyperNetworksNeural2019,knyazevParameterPredictionUnseen2021}, Bayesian HyperNetworks~\citep{deutschGeneratingNeuralNetworks2018}, HyperGANs~\citep{ratzlaffHyperGANGenerativeModel2019} and HyperTransformers~\citep{zhmoginovHyperTransformerModelGeneration2022} for neural architecture search, model compression, ensembling, transfer- or meta-learning.
These methods learn representations by using images and labels of the target domain. In contrast, our approach only uses model weights and does not need access to underlying data samples and labels -- 
an emergent use case, e.g. of deep learning monitoring services or model hubs. 
In addition to the ability to generate novel and diverse model weights, compared to previous works our approach (a) can generate novel weights conditionally on model zoos from unseen tasks and (b) can be conditioned on the latent factors of the underlying hyper-representation. Notably, both (a) and (b) can be done without the need to retrain hyper-representations.\looseness-1

\begin{figure*}[t]
\begin{minipage}[t]{0.98\textwidth}
\begin{center}
\includegraphics[trim=0in 0in 0in 0in, clip, width=1.0\linewidth]{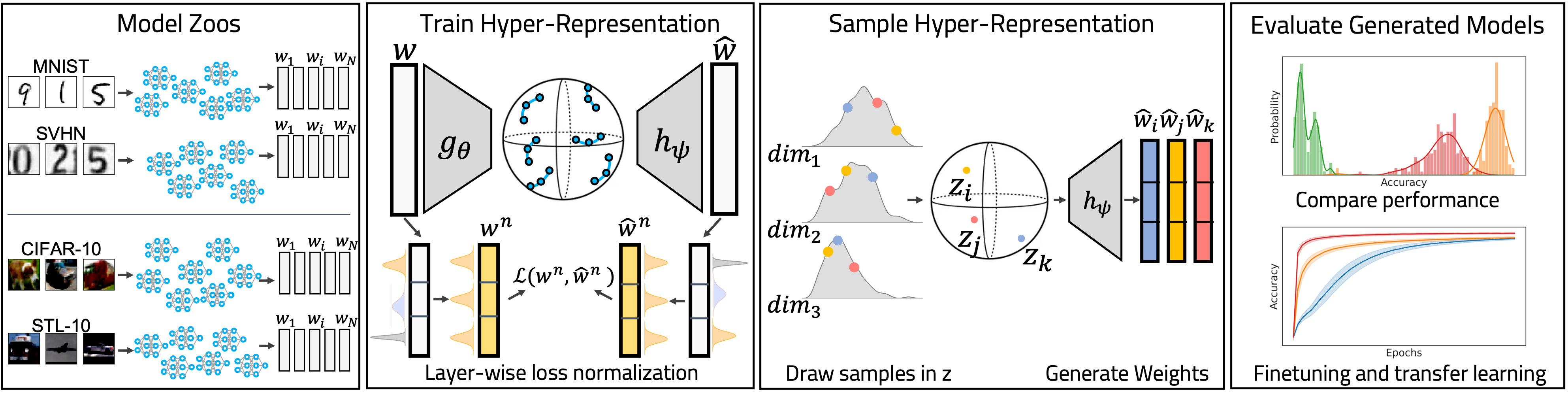}
\caption[Generative Hyper-Representations Schematic Overview]{\small Outline of our approach: Model zoos are trained on image classification tasks. Hyper-representations are trained with self-supervised learning on the weights of the model zoos using layer-wise loss normalization in the reconstruction loss. 
We sample new embeddings in hyper-representation space and decode to weights. 
Generated models perform significantly better than random initialization or models sampled from baseline hyper-representations. 
Sampled models achieve high performance fine-tuned and transfer learned on new datasets.\looseness-1
}
\label{fig:gen_hyper_reps:scheme}    
\end{center}
\end{minipage}
\end{figure*}

\newpage
The results suggest our approach (Figure \ref{fig:gen_hyper_reps:scheme}) to be a promising step towards a general-purpose hyper-representation encapsulating knowledge of model zoos to advance different downstream tasks.
The hyper-representations and code to reproduce our results are available at 
\url{https://github.com/HSG-AIML/NeurIPS_2022-Generative_Hyper_Representations}.
\looseness-1

\section{Background: Training Hyper-Representations}
\label{sec:gen_hyper_reps:training}

We summarize the first stage of our method that corresponds to learning a hyper-representation of a population of neural networks, called a \textit{model zoo}~\citep{schurholtSelfSupervisedRepresentationLearning2021}. In~\citep{schurholtSelfSupervisedRepresentationLearning2021} and this paper, a model zoo consists of models trained on the same task such as CIFAR-10 image classification~\citep{krizhevskyLearningMultipleLayers2009}.
Specifically, a hyper-representation is learned using an autoencoder $\hat{\mathbf{w}}_i = h(g(\mathbf{w}_i))$ on a zoo of $M$ models $\{ {\bf w}_i\}_1^M$, 
where ${\bf w}_i$ is the flattened vector of dimension $N$ of all the weights of the $i$-th model. 
The encoder $g$ compresses vector $\mathbf{w}_i$ to fixed-size hyper-representation $\mathbf{z}_i=g(\mathbf{w}_i)$ of lower dimension. The decoder $h$ decompresses the hyper-representation to the reconstructed vector $\hat{\mathbf{w}}_i$. Both encoder and decoder are built on a self-attention block~\citep{vaswaniAttentionAllYou2017}. The samples from model zoos are understood as sequences of convolutional or fully connected neurons. Each of the neurons is encoded as a token embedding and concatenated to form a sequence. 
The sequence is passed through several layers of multi-head self-attention. Afterward, a special compression token summarizing the entire sequence is linearly compressed to the bottleneck. The output is fed through a tanh-activation to achieve a bounded latent space $\mathbf{z}_i$ for the hyper-representation. The decoder is symmetric to the encoder, the embeddings are linearly decompressed from hyper-representations $\mathbf{z}_i$, and position encodings are added.\looseness-1

Training is done in a multi-objective fashion, minimizing the composite loss $\mathcal{L} = \beta \mathcal{L}_{MSE}+(1-\beta)\mathcal{L}_{c}$, where
$\mathcal{L}_{c}$ is a contrastive loss and $\mathcal{L}_{MSE}$ is a weight reconstruction loss (see details in~\citep{schurholtSelfSupervisedRepresentationLearning2021}). We can write the latter in a layer-wise way to facilitate our discussion in \S~\ref{sec:gen_hyper_reps:lwln}:\looseness-1
\begin{equation}
    \label{eq:gen_hyper_reps:baseline_loss}
    \mathcal{L}_{MSE} = \frac{1}{MN}\sum\nolimits_{i=1}^M \sum\nolimits_{l=1}^L || \hat{\mathbf{w}}^{(l)}_i - {\mathbf{w}}^{(l)}_i ||^2_2,
\end{equation}
\noindent where $\hat{\mathbf{w}}^{(l)}_i$, ${\mathbf{w}}^{(l)}_i$ are reconstructed and original weights for the $l$-{th} layer of the $i$-{th} model in the zoo. 
The contrastive loss $\mathcal{L}_{c}$ leverages two types of data augmentation at train time to impose structure on the latent space: permutation exploiting inherent symmetries of the weight space and random erasing.

\section{Methods}
\label{sec:gen_hyper_reps:methods}
In the following, we present (i) layer-wise loss normalization to ensure that decoded models are performant, and (ii) sampling methods to generate diverse populations of models.\looseness-1
\subsection{Layer-Wise Loss Normalization}\label{sec:gen_hyper_reps:lwln}
We observed that hyper-representations as proposed by~\citep{schurholtSelfSupervisedRepresentationLearning2021} decode to dysfunctional models, with performance around random guessing. 
To alleviate that, we propose a novel layer-wise loss normalization (LWLN), which we motivate and detail in the following.
\begin{figure}[h!]
\begin{minipage}[t]{0.98\textwidth}
\begin{center}
\includegraphics[trim=0mm 0mm 0mm 0mm, clip, width=0.9\linewidth]{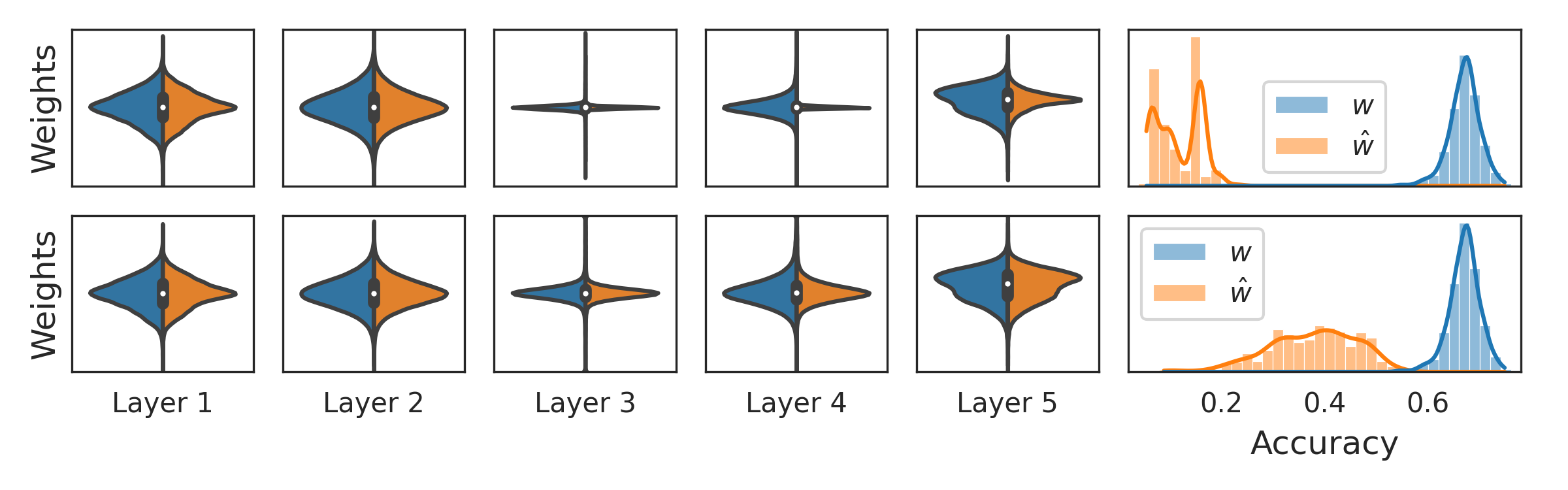}
\caption[Loss Normalization Effect on Weight Distribution]{
\small Comparison of the distributions of SVHN zoo weights $\mathbf{w}$ (blue) and reconstructed weights $\hat{\mathbf{w}}$ (orange) as well as their test accuracy on the SVHN test set. 
\textbf{Top:} Baseline hyper-representation as proposed by~\citep{schurholtSelfSupervisedRepresentationLearning2021}, the weights of layers 3, 4 collapse to the mean. These layers form a weak link in reconstructed models. The accuracy of reconstructed models drops to random guessing.
\textbf{Bottom:} Hyper-representation trained with layer-wise loss normalization (LWLN). The normalized distributions are balanced, all layers are evenly reconstructed, and the accuracy of reconstructed models is significantly improved.
}
\label{fig:gen_hyper_reps:layer_norm_eval}    
\end{center}
\end{minipage}
\end{figure}

\newpage
Due to the MSE training loss in~\eqref{eq:gen_hyper_reps:baseline_loss}, the reconstruction error can generally be expected to be uniformly distributed over all weights and layers of the weight vector $\mathbf{w}$.
However, the weight magnitudes of many of our zoos are unevenly distributed across different layers. 
In these zoos, the even distribution of reconstruction errors leads to undesired effects.
Layers with broader distributions and large-magnitude weights are reconstructed well, while layers with narrow distributions and small-magnitude weights are disregarded. The latter layers can become a weak link in the reconstructed models, causing performance to drop significantly down to random guessing. 
The top row of Figure \ref{fig:gen_hyper_reps:layer_norm_eval} shows an example of a baseline hyper-representation learned on the zoo of SVHN models~\citep{netzerReadingDigitsNatural2011}.
Common initialization schemes~\citep{heDelvingDeepRectifiers2015,glorotUnderstandingDifﬁcultyTraining2010} produce distributions with different scaling factors per layer, so the issue is not an artifact of the zoos, but can exist in real-world model populations. Similarly, recent work on generating models normalizes weights to boost performance~\citep{knyazevParameterPredictionUnseen2021}.
In order to achieve equally accurate reconstruction across the layers, we introduce a layer-wise loss normalization (LWLN) with the mean $\mu_l$ and standard deviation $\sigma_l$ of all weights in layer $l$ estimated over the train split of the zoo:\looseness-1
\begin{equation}
    \label{eq:gen_hyper_reps:lwln_loss}
    \mathcal{L}_{\bar{MSE}} = \frac{1}{MN} \sum_{i=1}^{M} \sum_{l=1}^{L} \left\| \frac{\hat{\mathbf{w}}_i^{(l)}-\mu_l}{\sigma_l}-\frac{\mathbf{w}_i^{(l)}-\mu_l}{\sigma_l} \right\|_2^2 = \frac{1}{MN}\sum_{i=1}^M \sum_{l =1}^{L}\frac{\|\hat{\mathbf{w}}_i^{(l)}-\mathbf{w}_i^{(l)} \|_2^2}{\sigma_l^2}. 
\end{equation}

\subsection{Sampling from Hyper-Representations}
\label{sec:gen_hyper_reps:sampling}
We introduce methods to draw diverse and high-quality samples $\mathbf{z}^* \sim p(\mathbf{z})$ from the learned hyper-representation space to generate model weights $\mathbf{w}^* = h(\mathbf{z}^*)$. 
Such sampling is facilitated if there is knowledge on the topology of the space spanned by $\mathbf{z}$. One way to achieve that is to train a variational autoencoder (VAE) with a predefined prior~\citep{kingmaAutoEncodingVariationalBayes2013} instead of the autoencoder of ~\citep{schurholtSelfSupervisedRepresentationLearning2021}. 
While training VAEs on common domains such as images has become well-understood and feasible, in our relatively novel weight domain, we found it problematic (see details in Appendix \ref{app:sampling}).
Other generative methods avoid a predefined prior of VAEs, either by analyzing the topology of the space learned by the autoencoder or fitting a separate density estimation model on top of the learned representation~\citep{liu2019acceleration,guo2019auto}. These methods assume the representation space to have strong regularities.
The hyper-representation space learned by the autoencoder of~\citep{schurholtSelfSupervisedRepresentationLearning2021} is already regularized by dropout regularization applied to the encoder and decoder as in~\citep{ghoshVariationalDeterministicAutoencoders2020}. The contrastive loss component requiring similar models to be embedded close to each other may also improve the regularity of the representation space.
Empirically, we found our layer-wise loss normalization (LWLN) to further regularize the representation space by ensuring robustness and smoothness (see Figure \ref{fig:gen_hyper_reps:rob_smooth} in \S~\ref{sec:gen_hyper_reps:experiments}). 

Given the smoothness and robustness of the learned hyper-representation space, we follow ~\citep{liu2019acceleration,guo2019auto,ghoshVariationalDeterministicAutoencoders2020} in estimating the density and topology to draw samples from a regularized autoencoder. To that end, we introduce three strategies to sample from that space: $S_{\text{KDE}}, S_{\text{Neigh}}, S_{\text{GAN}}$.
To model the density and topology in representation space, we use the embeddings of the train set as anchor samples $\{\mathbf{z}_i\}$. 
We observe that many anchor samples from $\{\mathbf{z}_i\}$ correspond to the models with relatively poor accuracy (Figure \ref{fig:gen_hyper_reps:layer_norm_eval}), so to improve the quality of sampled weights, we consider the variants of these methods using only those embeddings of training samples corresponding to the top 30\% performing models. We denote these sampling methods as $S_{\text{KDE30}}, S_{\text{Neigh30}}, S_{\text{GAN30}}$ respectively.
These methods can potentially decrease sample diversity, however, we found that the generated weights are still diverse enough (e.g. to construct high-performant ensembles, Figure~\ref{fig:gen_hyper_reps:ensembles}).
Finally, as baseline and sanity check we explore sampling uniformly in representation space $S_U$ and sampling in low-probability regions $S_C$.\looseness-1

\subsubsection*{Uniform $S_U$}
\vspace{-4pt}
%
%
As a naive baseline, we draw samples uniformly in hyper-representation space (bounded by tanh, \S~\ref{sec:gen_hyper_reps:training}) and denote it as $S_U$. This is naive, because we found that the embeddings $\mathbf{z}$ populate only sections of a shell of a high-dimensional sphere (see Figures~\ref{fig:gen_hyper_reps:hyper_sphere_z} and \ref{fig:gen_hyper_reps:hyper_sphere_cos} in Appendix \ref{app:analysis}). So most of the uniform samples lie in the low-probability regions of the space and are not expected to be decoded to useful models.\looseness-1

%
\subsubsection*{Density estimation $S_{\text{KDE}}$ and counterfactual sampling $S_C$}
\vspace{-4pt}
The dimensionality $D$ of hyper-representations $\mathbf{z}$ in~\citep{schurholtSelfSupervisedRepresentationLearning2021}, as well as in our work, is relatively high due to the challenge of compressing weights $\mathbf{w}$. 
Fitting a probability density model to such a high-dimensional distribution is feasible by making a conditional independence assumption: $p(\mathbf{z}^{(j)}|\mathbf{z}^{(k)}, \mathbf{w}) = p(\mathbf{z}^{(j)}| \mathbf{w})$, where $\mathbf{z}^{(j)}$ is the $j$-th dimensionality of the embedding $\mathbf{z}$.
To model the distribution of each $j$-th dimensionality, we choose kernel density estimation (KDE), as it is a powerful yet simple, non-parametric, and deterministic method with a single hyperparameter.
We fit a KDE to the $M$ anchor samples $\{\mathbf{z}_{i}^{(j)}\}_{i=1}^M$ of each dimension $j$, and draw samples $z^{(j)}$ from that distribution: $ z^{(j)} \sim p(\mathbf{z}^{(j)}) = \frac{1}{M h} \sum\nolimits_{i=1}^{M} K(\frac{\mathbf{z}^{(j)} - \mathbf{z}^{(j)}_i}{h})
$,
\noindent where $K(x)=(2\pi)^{-1/2} \exp{(-\frac{x^2}{2})}  $
is the Gaussian kernel and $h$ is a bandwidth hyperparameter.
The samples of each dimension $z^{(j)}$ are concatenated to form samples $\mathbf{z}^* = [z^{(1)}, z^{(2)}, \cdots, z^{(D)}]$. This method is denoted as $S_{\text{KDE}}$.

As a sanity check, we invert the $S_{\text{KDE}}$ method and explicitly draw samples from regions not populated by anchor samples, i.e. with a low probability according to the KDE. This method, denoted as $S_C$, essentially samples counterfactual embeddings and similarly to $S_U$ is expected to perform poorly.\looseness-1
%
\subsubsection*{Neighbor sampling $S_{\text{Neigh}}$}

Sampling neighbors of anchor samples $\{\mathbf{z}_i\}$ could be a simple and effective sampling strategy, but due to the high sparsity of the hyper-representation space, this strategy results in poor-quality samples. We therefore propose to use a neighborhood-based dimensionality reduction function $k: \mathbb{R}^D \to \mathbb{R}^{d}$ that maps $\mathbf{z}_i$ to low-dimensional embeddings $\mathbf{n}_i \in \mathbb{R}^{d}$ where sampling is facilitated. 
The assumption is that due to the low dimensionality of $\mathbb{R}^{d}$ (we choose $d=3$) there will be fewer low-probability regions so that uniform sampling in $\mathbb{R}^{d}$ can be effective.
Specifically, given low-dimensional embeddings $\mathbf{n}_i = k(\mathbf{z}_i)$, we sample $\mathbf{n}^*$ uniformly from the cube: $\mathbf{n}^* \sim U({min(\mathbf{n}),max(\mathbf{n})})$.
Samples $\mathbf{n}^*$ are then mapped back to hyper-representations $ \mathbf{z}^* = k^{-1}(\mathbf{n}^*)$. 
To preserve the neighborhood topology of $\mathbb{R}^D$ in $\mathbb{R}^d$ and enable mapping back to $\mathbb{R}^D$, we choose $k$ to be an approximate inverse neighborhood-based dimensionality reduction function based on UMAP~\citep{mcinnesUMAPUniformManifold2018a}.\looseness-1
\\
%
\subsubsection*{Latent space GAN $S_{\text{GAN}}$}\label{sec:gen_hyper_reps:gan}

A common choice for generative representation learning is generative adversarial networks (GANs)~\citep{goodfellowGenerativeAdversarialNets2014}. While training a GAN directly to generate weights is a promising yet challenging avenue for future research~\citep{ratzlaffHyperGANGenerativeModel2019}, we found the GAN framework to work reasonably well when trained on the hyper-representations. This idea follows \citep{liu2019acceleration,guo2019auto} that showed improved training stability and efficiency compared to training GANs on inputs directly.
We train a generator $G: \mathbb{R}^d \to \mathbb{R}^D$ with $\mathbf{z}^* = G(\mathbf{n}^*)$ to generate samples in hyper-representation space from the Gaussian noise $\mathbf{n}^*$.  We choose $d=16$ as a compromise between size and capacity.
See a detailed architecture of our GAN in Appendix~\ref{app:sampling}.\looseness-1
%

%
\section{Experiments}
\label{sec:gen_hyper_reps:experiments}
%
\vspace{-0.15cm}
\subsection{Experimental Setup}

We train and evaluate our approaches on four image classification datasets: MNIST~\citep{lecunGradientbasedLearningApplied1998}, SVHN~\citep{netzerReadingDigitsNatural2011}, 
CIFAR-10~\citep{krizhevskyLearningMultipleLayers2009}, STL-10~\citep{coatesAnalysisSingleLayerNetworks2011}. For each dataset, there is a model zoo that we use to train an autoencoder following~\citep{schurholtSelfSupervisedRepresentationLearning2021}.\looseness-1

\paragraph*{Model zoos:} 
In practice, there are already many available model zoos, e.g., on Hugging Face or GitHub, that can be used for hyper-representation learning and sampling. 
Unfortunately, these zoos are not systematically constructed and require further effort to mine and evaluate.
Therefore, in order to control the experiment design, ensure feasibility and reproducibility, we generate novel or use the model zoos of~\citep{schurholtSelfSupervisedRepresentationLearning2021,schurholtModelZoosDataset2022} created in a systematic way.
With controlled experiments, we aim to develop and evaluate inductive biases and methods to train and utilize hyper-representation, which can be scaled up efficiently to large-scale and non-systematically constructed zoos later. 
For each image dataset, a zoo contains $M=1000$ convolutional networks of the same architecture with three convolutional layers and two fully-connected layers. 
Varying only in the random seeds, all models of the zoo are trained for 50 epochs with the same hyperparameters following~\citep{schurholtSelfSupervisedRepresentationLearning2021}. 
To integrate higher diversity in the zoo, initial weights are uniformly sampled from a wider range of values rather than using well-tuned initializations of~\citep{glorotUnderstandingDifﬁcultyTraining2010,heDelvingDeepRectifiers2015}.
Each zoo is split into the train (70\%), validation (15\%), and test (15\%) splits. 
To incorporate the learning dynamics, we train autoencoders on the models trained for
21-25 epochs following~\citep{schurholtSelfSupervisedRepresentationLearning2021}. Here the models have already achieved high performance, but have not fully converged. The development in the remaining epochs of each model is treated as hold-out data to compare against. 
We use the MNIST and SVHN zoos from~\citep{schurholtSelfSupervisedRepresentationLearning2021} and based on them create the CIFAR-10 and STL-10 zoos. Details on the zoos can be found in Appendix~\ref{app:zoos}.\looseness-1

\paragraph*{Experimental details:} We train separate hyper-representations on each of the model zoos. Images and labels are not used to train the hyper-representations (see \S~\ref{sec:gen_hyper_reps:training}).
Using the proposed sampling methods (\S~\ref{sec:gen_hyper_reps:sampling}), we generate new embeddings and decode them to weights. We evaluate sampled populations as initializations (epoch 0) and by fine-tuning for up to 25 epochs.
We distinguish between in-dataset and transfer-learning. 
For in-dataset, the same image dataset is used for training and evaluating our hyper-representations and baselines.
For transfer learning, hyper-representations (and pre-trained models in baselines) are trained on a source dataset, then all populations are evaluated and fine-tuned on a different target dataset. 
Full details on training, including infrastructure and compute are detailed in the Appendix  \ref{app:training}.\looseness-1
%
%

%
\paragraph*{Baselines:}
As the first baseline, we consider the autoencoder of~\citep{schurholtSelfSupervisedRepresentationLearning2021}, which is the same as ours but without the proposed layer-wise loss-normalization (LWLN, \S~\ref{sec:gen_hyper_reps:lwln}). We combine this autoencoder with the $S_\text{KDE30}$ sampling method and, hence, denote it as $B_{\text{KDE}30}$.
We consider two other baselines based on training models with stochastic gradient descent (SGD): training from scratch on the target classification task $B_T$, and training on a source followed by fine-tuning on the target task $B_F$. The latter remains one of the strongest transfer learning baselines~\citep{chen2019closer,dhillon2019baseline,kolesnikov2020big}.
%

\paragraph*{Reproducibility, reliability and comparability:}
We compare populations of at least 50 models to evaluate each method reliably.
We report standard deviation in Tables~\ref{tab:gen_hyper_reps:initialization}-\ref{tab:gen_hyper_reps:transfer} and statistical significance, effect size and 95\% confidence interval in Appendix \ref{app:results}. 
To ensure fairness and comparability, all methods share training hyperparameters. 
Fine-tuning uses the hyperparameters of the target domain. 

%
\subsection{Results}
\label{sec:gen_hyper_reps:results}
In the following, we first analyze the learned hyper-representations further justifying our sampling methods and assumptions made in \S~\ref{sec:gen_hyper_reps:sampling}. 
We then confirm the effectiveness of our approach for model initialization without and with fine-tuning in the in-dataset and transfer learning settings.
%
\subsubsection*{Hyper-Representations are Robust and Smooth}
\label{sec:gen_hyper_reps:analysis}

\begin{figure}[htpb]
\begin{center}
\begin{tabular}{cccc}
     \multicolumn{2}{c}{Baseline hyper-representation} & \multicolumn{2}{c}{Our hyper-representation} \\
     \multicolumn{2}{c}{\includegraphics[trim=0in 0in 0in 3mm, clip, width=0.48\linewidth]{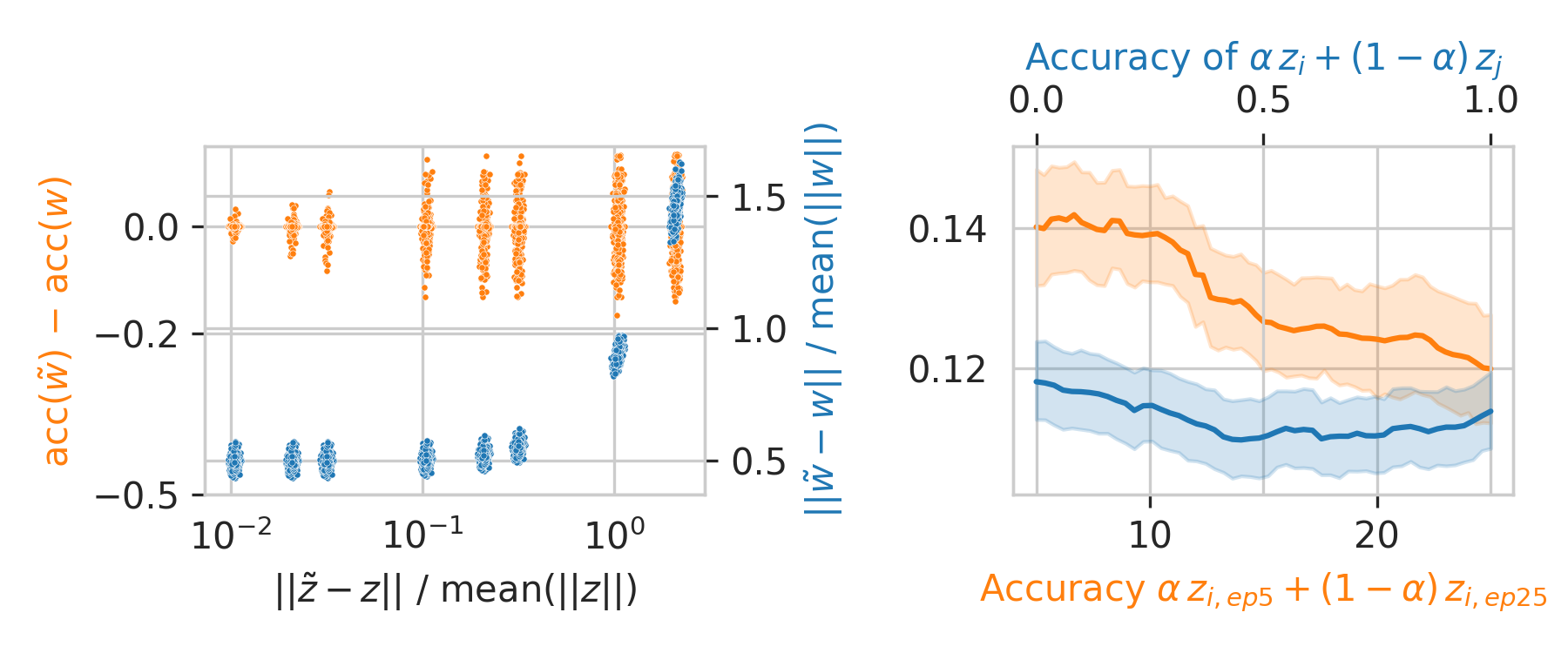}} 
     & 
     \multicolumn{2}{c}{\includegraphics[trim=0in 0in 0in 3mm, clip, width=0.48\linewidth]{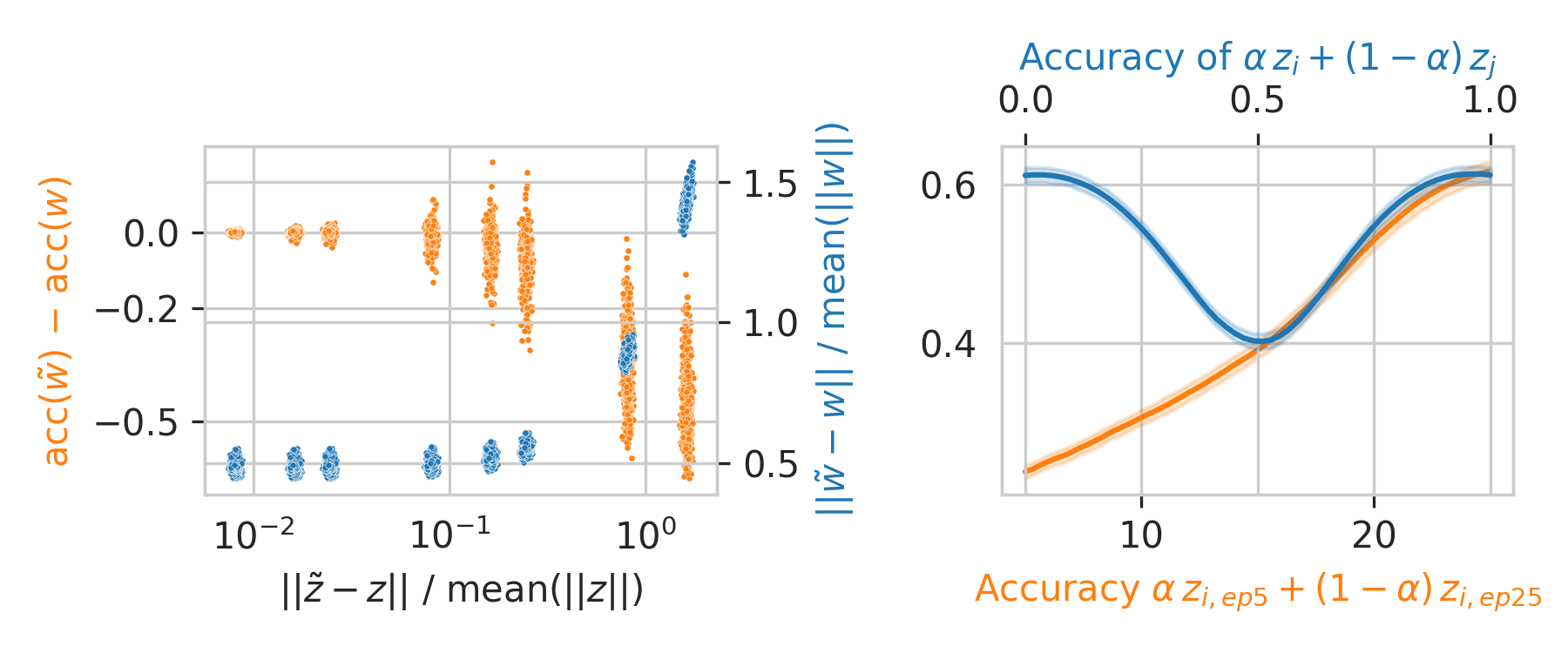}}  \\
     \multicolumn{1}{c}{\hspace{50pt}(a)} & \multicolumn{1}{c}{\hspace{50pt}(b)} & \multicolumn{1}{c}{\hspace{50pt}(c)} & \multicolumn{1}{c}{\hspace{50pt}(d)} \\
\end{tabular}

\end{center}
\vspace{-4mm}
\small
\captionof{figure}[Robustness and Smoothness of Hyper-Representations]{
\textbf{(a,c):} Robustness of hyper-representations.
For both baseline and our hyper-representation, relatively large levels of relative noise >10\% are necessary to degrade the test accuracy ({\color{orange}orange}) or reconstruction ({\color{cyan}blue}); see the text for further discussion.
\textbf{(b,d):} 
Interpolations along model trajectories ({\color{orange}orange}) and 
between $\mathbf{z}$ of different models ({\color{cyan}blue}) show the smoothness of our hyper-representation.
}
\vspace{-1mm}
\label{fig:gen_hyper_reps:rob_smooth}    
\end{figure}

We evaluate the robustness and smoothness of the hyper-representation space with two experiments on the SVHN zoo. First, to evaluate robustness, we add different levels of noise to the embeddings of the test set to create $\tilde{\mathbf{z}}$, decode them to model weights $\tilde{\mathbf{w}}$ and compute models' accuracies on the SVHN classification task. 
We found that both the baseline as well as our hyper-representations are robust to noise as large levels of relative noise >10\% are required to affect performance (Figure \ref{fig:gen_hyper_reps:rob_smooth}, a,c). 
Second, to probe for smoothness, we linearly interpolate between the test set embeddings (i) along the trajectory of the same model at different epochs ($\mathbf{z}_{i, ep5}$ and $\mathbf{z}_{i, ep25}$) and (ii) between 250 random pairs of embeddings on the trajectories of different models ($\mathbf{z}_{i}$ and $\mathbf{z}_{j}$).
We decode the interpolated embeddings and compute the models' accuracies on the classification task. 
For our model, we found remarkably smooth development of accuracy along the interpolation in both schemes (Figure \ref{fig:gen_hyper_reps:rob_smooth}, d). The lack of fluctuations along and between trajectories supports both local and global notions of smoothness in hyper-representation space.

For the baseline autoencoder (without LWLN) decoded models all perform close to 10\% accuracy, so these representations do not support similar notions of smoothness (Figure \ref{fig:gen_hyper_reps:rob_smooth}, b), while robustness can be misleading since the accuracy even without adding noise is already low (Figure \ref{fig:gen_hyper_reps:rob_smooth}, a).
Therefore, LWLN together with regularizations added to the autoencoder allows for learning robust and smooth hyper-representation. This property makes sampling from that representation more meaningful as we show next.\looseness-1

%
%
\subsubsection*{Sampling for In-dataset Initialization}
\paragraph*{Comparison between sampling methods:} 
\begin{wrapfigure}{r}{0.62\linewidth}
\vspace{-9mm}
\includegraphics[width=1.0\linewidth]{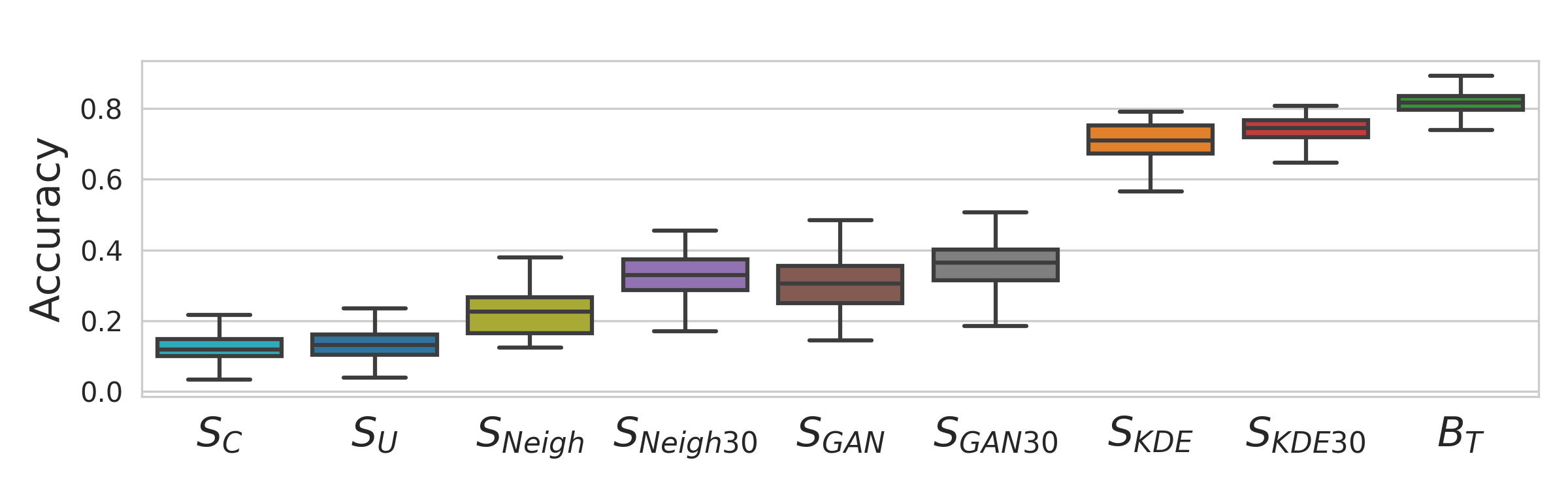}
\vspace{-8mm}
\captionof{figure}[Performance Comparison of Sampled Populations]{
MNIST results of sampled weights (no fine-tuning) compared to training from scratch with SGD ($B_T$).\looseness-1
}
\vspace{-6mm}
\label{fig:gen_hyper_reps:accuracy_distributions}    
\end{wrapfigure} 
We evaluate the performance of different sampled populations (obtained with LWLN) \textit{without fine-tuning} generated weights.
On MNIST, all sampled models except those obtained using $S_U$ and $S_C$ perform better than random initialization (10\% accuracy), but worse than models trained from scratch $B_T$ for 25 epochs (Figure \ref{fig:gen_hyper_reps:accuracy_distributions}). 
Distribution-based samples ($S_{\text{KDE}}$ and $S_{\text{GAN}}$) perform better than neighborhood based samples ($S_{\text{Neigh}}$).
The populations based on the top 30\% perform better than their 100\% counterparts with $S_{\text{KDE30}}$ as the strongest sampling method overall. This demonstrates that the learned hyper-representation and sampling methods are able to capture complex subtleties in weight space differentiating high and low performing models.\looseness-1


\paragraph*{Comparison to the baseline hyper-representations:}  We also compare $S_{\text{KDE30}}$ that is based on our autoencoder with layer-wise loss normalization (LWLN) to the baseline autoencoder using the same sampling method ($B_{\text{KDE}30}$) without fine-tuning. 
On all datasets except for MNIST, $S_{\text{KDE30}}$ considerably outperform $B_{\text{KDE}30}$ with the latter performing just above 10\% (random guessing), see Table~\ref{tab:gen_hyper_reps:initialization} (rows with epoch 0).
We attribute the success of LWLN to two main factors.
First, LWLN prevents the collapse of reconstruction to the mean (compare Figure \ref{fig:gen_hyper_reps:layer_norm_eval} top to bottom). Second, by fixing the weak links, the reconstructed models perform significantly better (see Appendix \ref{app:loss} for more results).\looseness-1


\paragraph*{In-dataset fine-tuning:} 
When fine-tuning, our $S_{\text{KDE}30}$ and baseline $B_{\text{KDE}30}$ appear to gradually converge to similar performance (Table \ref{tab:gen_hyper_reps:initialization}). While unfortunate, this result aligns well with previous findings that longer training and enough data make initialization less important~\citep{mishkinAllYouNeed2016,he2019rethinking,rasmus2015semi}.

\begin{wraptable}{r}{0.65\linewidth}
\vspace{-6mm}
\captionof{table}[Fine-tuning Results of Sampled Populations]{
Mean and std of test accuracy (\%) of sampled populations with LWLN ($S_{\text{KDE}30}$) and without ($B_{\text{KDE}30}$) compared to models trained from scratch $B_T$. Best results for each epoch and dataset are bolded.
}\label{tab:gen_hyper_reps:initialization}
\vspace{2mm}
\small
\setlength{\tabcolsep}{2pt}
\begin{tabularx}{\linewidth}{rccccc}
\toprule
\multicolumn{1}{r}{\textbf{Method}} & \textbf{Ep.} & \multicolumn{1}{c}{\textbf{MNIST}} & \multicolumn{1}{c}{\textbf{SVHN}} &
\multicolumn{1}{c}{\textbf{CIFAR-10}} & \multicolumn{1}{c}{\textbf{STL-10}} \\
\midrule
$B_{T}$                & 0 &  \multicolumn{4}{c}{$\approx$10\% (random guessing)}\\
$B_{\text{KDE}30}$   & 0     & {63.2±7.2}   & 10.1±3.2           & 
15.5±3.4           & 12.7±3.4  \\
$S_{\text{KDE}30}$   & 0     & \textbf{68.6±6.7}   & \textbf{51.5±5.9}  & 
\textbf{26.9±4.9}  & \textbf{19.7±2.1}  \\
\midrule
$B_{T}$                & 1      & 20.6±1.6           & 19.4±0.6           & 
27.5±2.1           & 15.4±1.8  \\
$B_{\text{KDE}30}$   & 1     & {83.2±1.2}   & 67.4±2.0 	        & 
39.7±0.6           &	\textbf{26.4±1.6} \\
$S_{\text{KDE}30}$   & 1     & \textbf{83.7±1.3}   & \textbf{69.9±1.6}  &
\textbf{44.0±0.5}  &	{25.9±1.6}  \\
\midrule
$B_{T}$                & 25     & 83.3±2.6           & 66.7±8.5           & 
46.1±1.3           & 35.0±1.3  \\
$B_{\text{KDE}30}$   & 25    & \textbf{93.2±0.6}   & \textbf{75.4±0.9}  & 
{48.1±0.6}  & \textbf{38.4±0.9}  \\
$S_{\text{KDE}30}$   & 25    & {93.0±0.7}    & {74.2±1.4} &
\textbf{48.6±0.5}  & {38.1±1.1}  \\
\midrule
$B_{T}$                & 50     & 91.1±2.6           & 70.7±8.8           & 
48.7±1.4           & 39.0±1.0  \\

\bottomrule 
\end{tabularx}
\vspace{-4mm}
\end{wraptable} 
We also compare $S_{\text{KDE}30}$ and $B_{\text{KDE}30}$ to training models from scratch ($B_T$).
On all four datasets, both ours and the baseline hyper-representations outperform $B_T$ when generated weights are fine-tuned for the same number of epochs as $B_T$.
Notably, on MNIST and SVHN generated weights fine-tuned for 25 epochs are even better than $B_T$ run for 50 epochs. Comparison to 50 epochs is more fair though, since the hyper-representations were trained on model weights trained for up to 25 epochs.
These findings show that the models initialized with generated weights learn faster achieving better results in 25 epochs than $B_T$ in 50 epochs.


\begin{wrapfigure}{r}{0.40\linewidth}
\vspace{-2mm}
\centering
\includegraphics[trim=4mm 4.5mm 3.5mm 3.8mm, clip, width=1.0\linewidth]{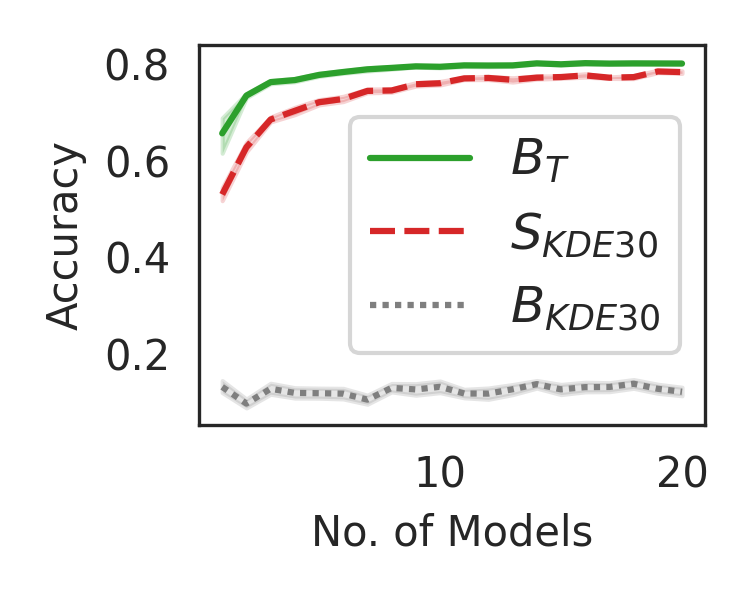}
\captionof{figure}[Performance of Sampled Ensembles]{
Generated ensembles evaluated on SVHN. 
Test accuracy is averaged over 15 ensembles of randomly chosen models.
}
\label{fig:gen_hyper_reps:ensembles}    
\end{wrapfigure}
\paragraph*{Sampling ensembles:} We found that a potentially useful by-product of learning hyper-representations is the ability to generate high-performant ensembles at almost no extra computational cost since both sampling and generation are computationally cheap. 
To demonstrate this effect, we compare ensembles formed using the baseline autoencoder ($B_{\text{KDE}30}$) and ours ($S_{\text{KDE}30}$) to the ensembles composed of models trained from scratch for 25 epochs ($B_T$) on SVHN.
Ensembles generated using the baseline $B_{\text{KDE}30}$ stagnate below 20\% (Figure \ref{fig:gen_hyper_reps:ensembles}).
In contrast, ensembles generated using our $S_{\text{KDE}30}$ gracefully improve with the ensemble size outperforming single $B_T$ models and almost matching $B_T$ ensembles with enough models in the ensembles.
Remarkably, the average test accuracy of generated ensembles of 15 models is 77.6\%, which is considerably higher than 70.7\% of models trained on SVHN for 50 epochs.
We conclude that hyper-representations learned with LWLN generate models that are not only performant but also diverse.
Although generating ensembles requires learning hyper-representation and model zoo first, we assume that in the future such a hyper-representation can be trained once and reused in unseen scenarios as we tentatively explore below (see results in Table~\ref{tab:gen_hyper_reps:cross_ds_recon} and the discussion therein).\looseness-1


\paragraph*{Do reconstructed models become similar to the original during fine-tuning?}
\begin{wrapfigure}{r}{0.68\linewidth}
\includegraphics[trim=4mm 2mm 3.8mm 3.8mm, clip, width=1.0\linewidth]{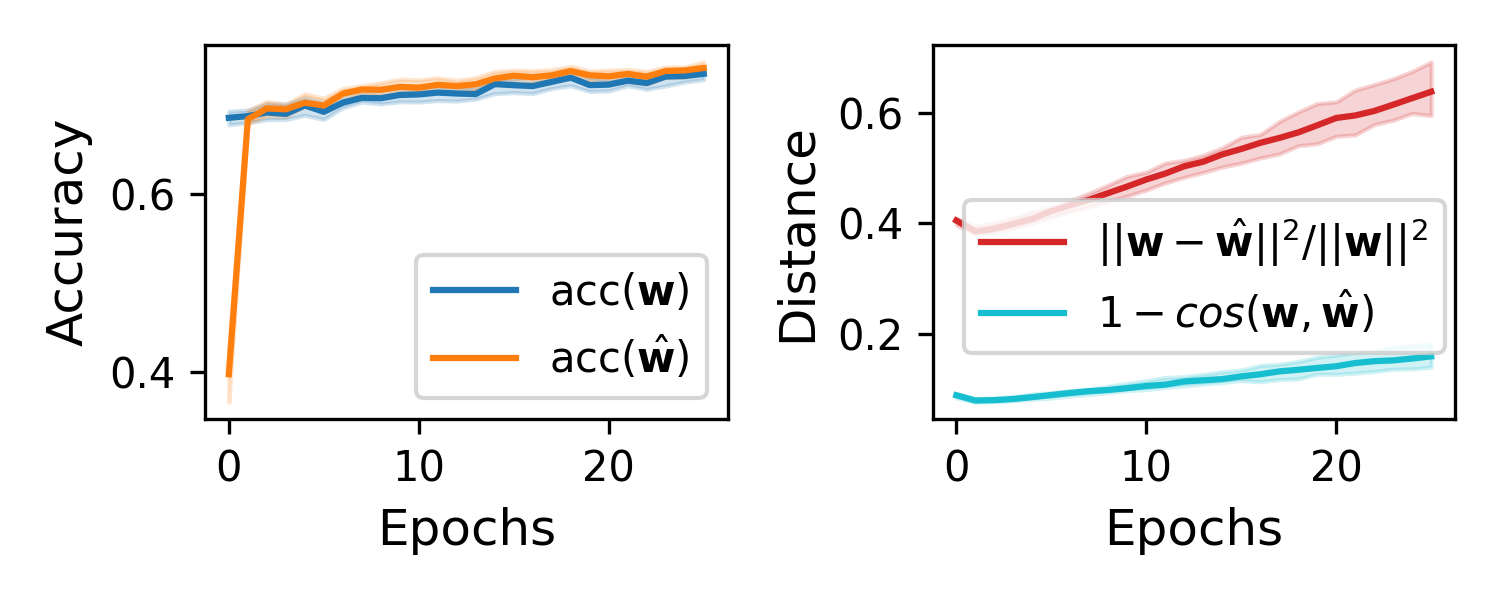}
\captionof{figure}[Comparison of Trained and Reconstructed Models]{
Progression of test accuracy (left) and distance (right) between weights during fine-tuning on SVHN; $\mathbf{w}$ -- initialization with the weights trained using SGD for 25 epochs; $\hat{\mathbf{w}}$ -- initialization with reconstructed weights.}
\label{fig:gen_hyper_reps:new_weight_solutions}    
\end{wrapfigure}
Sampled hyper - representations often learn faster and to a higher performance than the population of models they were trained on (Table~\ref{tab:gen_hyper_reps:initialization}).
We therefore explore the question, if reconstructed models develop in weight space in the same direction as their original, or find a different solution.
On SVHN, we found that the reconstructed models ($\hat{\mathbf{w}}$) after one epoch of fine-tuning perform similar to their originals ($\mathbf{w}$) and slightly outperform from there on (Figure \ref{fig:gen_hyper_reps:new_weight_solutions}, left).
At the same time, pairs of original and reconstructed models move further apart and become less aligned in weight space (Figure \ref{fig:gen_hyper_reps:new_weight_solutions}, right).
It appears that reconstructed models perform better and explore different solutions in weight space to do so. 
This confirms the intuition that hyper-representations impress useful structure on decoded weights. 
A pass through encoder and decoder thus results not just in a noisy reconstruction of the original sample. 
Instead, it maps to a different region on the loss surface, which leads to faster learning and better solutions.
Combining this with the ensembling results in Figure~\ref{fig:gen_hyper_reps:ensembles}, hyper-representations do not collapse to a single solution but decode to diverse and useful weights.\looseness-1

\subsubsection*{Sampling Initializations for Transfer Learning}
\label{sec:gen_hyper_reps:cross_dataset}
\paragraph*{Setup:}
We investigate the effectiveness of our method in a transfer-learning setup across image datasets. In particular, we report transfer learning results from SVHN to MNIST and from STL-10 to CIFAR-10 as two representative scenarios. Results on the other pairs of datasets can be found in Appendix \ref{app:results}. 
In these experiments, pre-trained models $B_F$ and the hyper-representation model are trained on a source domain. Subsequently, the pre-trained models $B_F$ and the samples $S_{\text{KDE}}$, $S_{\text{Neigh}}$ and $S_{\text{GAN}}$ are fine-tuned on the target domain. The baseline approach ($B_T$) is based on training models from scratch on the target domain. 
\begin{table}[ht!]
\centering
\caption[Performance of Sampled Models in Transfer Learning]{Transfer-learning results (mean and standard deviation of the test accuracy in \%). Note that for STL-10 to CIFAR-10 the performance of all methods saturate quickly due to the limited capacity of models in the zoo making further improvements challenging as we discuss in \S~\ref{sec:gen_hyper_reps:limitations}.
}
\label{tab:gen_hyper_reps:transfer}
{\small
\setlength{\tabcolsep}{6pt}
\centering
\begin{tabular}{lcccccc}
\toprule

\textbf{Method} & \multicolumn{3}{c}{\textbf{SVHN to MNIST}} & \multicolumn{3}{c}{\textbf{STL-10 to CIFAR-10}} \\
\cmidrule(r){1-1} \cmidrule(rl){2-4} \cmidrule(l){5-7} 
                  & \textbf{Ep. 0}          & \textbf{Ep. 1}        & \textbf{Ep. 50}       & \textbf{Ep. 0}        & \textbf{Ep. 1} & \textbf{Ep. 50}        \\
\cmidrule(r){1-1} \cmidrule(rl){2-2} \cmidrule(rl){3-3} \cmidrule(rl){4-4} \cmidrule(rl){5-5} \cmidrule(rl){6-6} \cmidrule(rl){7-7} 
$B_T$  & 10.0±0.6         & 20.6±1.6 & 91.1±1.0         & 10.1±1.3  & 27.5±2.1 & {48.7±1.4} \\ 
$B_F$    & \textbf{33.4±5.4} & 84.4±7.4 & 95.0±0.8 & \textbf{15.3±2.3} & {29.4±1.9} & \textbf{49.2±0.7} \\

\cmidrule(r){1-1} \cmidrule(rl){2-2} \cmidrule(rl){3-3} \cmidrule(rl){4-4} \cmidrule(rl){5-5} \cmidrule(rl){6-6} \cmidrule(rl){7-7} 
$S_{\text{KDE}30}$   & 31.8±5.6          & \textbf{86.9±1.4} & \textbf{95.5±0.4} & {14.5±1.9} & \textbf{29.6±2.0} & {48.8±0.9} \\
$S_{\text{Neigh}30}$ & 10.7±2.7          & 79.2±3.3  & \textbf{95.5±0.7}          & 10.1±2.1          & {29.2±1.9} & {48.9±0.7}  \\
$S_{\text{GAN}30}$   & 10.4±2.4          & 75.0±6.3 & 94.9±0.7          & 10.2±2.5          & {28.6±1.8} & {48.8±0.8} \\
\bottomrule

\end{tabular}
}
\end{table}

\begin{figure}
\begin{minipage}{0.98\textwidth}
\begin{center}
\includegraphics[trim=0in 0in 0in 0in, clip, width=0.95\linewidth]{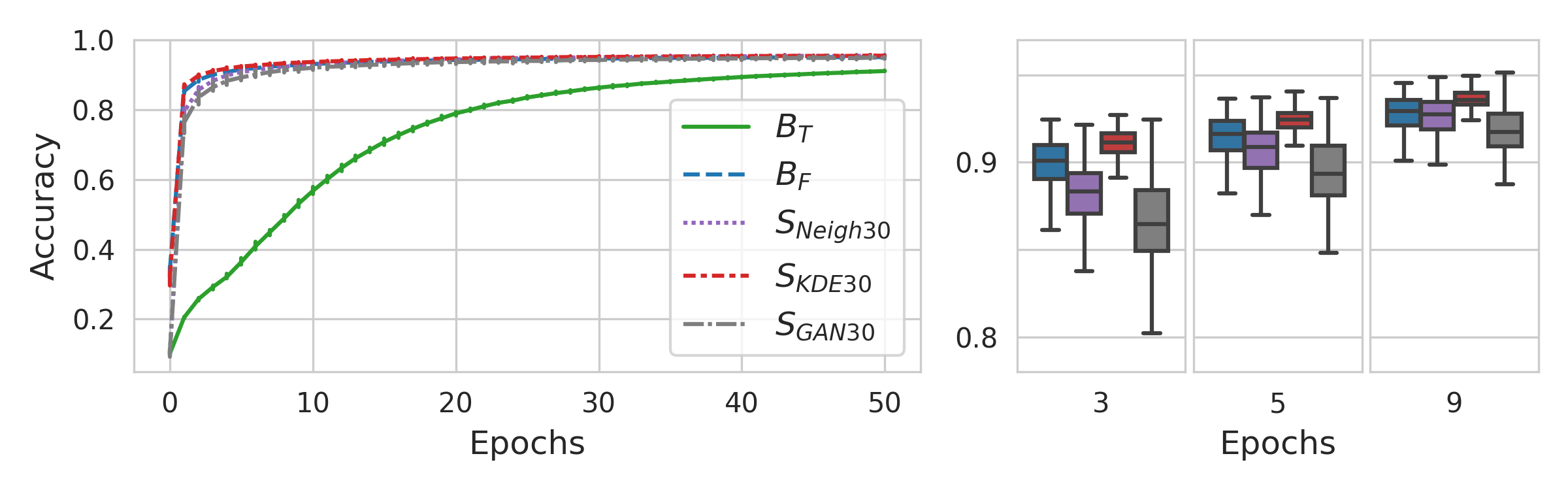}
\vspace{-4mm}
\caption[Performance of Sampled models in Transfer Learning Experiments]{SVHN to MNIST transfer learning experiment: test accuracy over epochs. Our sampling methods outperform the baselines after the first epoch. \textbf{Left:} epochs from 0 to 50. \textbf{Right}: epochs from 3 to 9, where $B_T$ is significantly lower than 80\% and thus is not visible.\looseness-1}
\vspace{-4mm}

\label{fig:gen_hyper_reps:sampling_accuracy_ood}    
\end{center}
\end{minipage}
\end{figure}

\paragraph*{Results:}
When transfer learning is performed from SVHN to MNIST, the sampled populations on average learn faster and achieve significantly higher performance than the $B_T$ baseline and generally compare favorably to $B_F$ (Figure~\ref{fig:gen_hyper_reps:sampling_accuracy_ood}, Table~\ref{tab:gen_hyper_reps:transfer}). 
In the STL-10 to CIFAR-10 experiment, all populations appear to saturate with only small differences in their performances (Table \ref{tab:gen_hyper_reps:transfer}).
Different sampling methods perform differently at the beginning versus the end of transfer learning. Generally, $S_{\text{KDE30}}$ performs better in the first epochs, while all methods perform comparably at the end of transfer learning. 
These discrepancies underline the difficulty of developing a single strong sampling method, which is an interesting area of future research.
We further found that all datasets are useful sources for all targets (see Appendix \ref{app:results}). Interestingly and other than in related work~\citep{mensinkFactorsInfluenceTransfer2021}, even transfer from the simpler to harder datasets (e.g., MNIST to SVHN) improves performance.
This might be explained by the ability of hyper-representations to capture a generic inductive prior useful across different domains, which we further investigate next.

\begin{wraptable}{r}{0.6\linewidth}
\vspace{-2mm}
\captionof{table}[Performance of Sampled Models Conditioned on Unseen Zoos]{
Test accuracy (\%) of models generated conditioned on the models of unseen zoos.}
\label{tab:gen_hyper_reps:cross_ds_recon}
\vspace{2mm}
\small
\setlength{\tabcolsep}{3pt}
\begin{tabularx}{\linewidth}{llcc}
\toprule
\textbf{ Training} & \textbf{ Conditioning} & \multicolumn{2}{c}{\textbf{ Mean / max accuracy}}  \\
\textbf{ zoo} & \textbf{unseen} & \textbf{ One model} & \textbf{ Ensemble} \\
\midrule
MNIST & SVHN & 12.7 / \textbf{19.8} & 13.4 / \textbf{18.7} \\
SVHN & MNIST & 16.2 / \textbf{26.0} & 22.1 / \textbf{29.8} \\
CIFAR-10 & STL-10 & 18.0 /  \textbf{24.4} & 23.8 / \textbf{26.7} \\
STL-10 & CIFAR-10 & 16.3 / \textbf{21.2} & 20.0 / \textbf{23.0} \\
\bottomrule
\end{tabularx}
\vspace{-2mm}
\end{wraptable} 

\paragraph*{Conditioning on unseen zoos:}
We explore if the hyper-representation trained on the models of one zoo (e.g. MNIST) can reconstruct the weights of another unseen zoo (e.g. SVHN).
This can be useful to enable the generation of weights for novel tasks without the need to retrain a hyper-representation. This is analogous to instance-conditioned GANs that recently were able to generate images from unseen domains without retraining GANs~\citep{casanova2021instance}.
Our results in Table \ref{tab:gen_hyper_reps:cross_ds_recon} show that while the performance on the unseen zoos is reduced, it is still well above random guessing (10\%), especially when multiple model weights are sampled and ensembled. This is promising, as the hyper-representations were trained on single-dataset zoos. 

\subsubsection*{Sampling Initializations for Unseen Architectures}
Generalization to unseen large architectures with complex connectivity (ResNet, MobileNet, and EfficientNet) is a very interesting and ambitious research problem. As a step towards that goal, we perform experiments in which we attempted to use our hyper-representation beyond the same simple architecture. Surprisingly, our results indicate the promise of leveraging the hyper-representation for more diverse architectures and settings. 
Further experiments investigating the cross-architecture generalization capabilities of hyper-representations can be found in Appendix \ref{app:analysis}. 

\paragraph*{Setup:} With this experiment, we aim to verify if it is possible to adapt our approach to architectures not seen during training, e.g., with skip connections and/or with more layers. We follow the transfer-learning setup of \S~\ref{sec:gen_hyper_reps:cross_dataset} and use an existing MNIST hyper-representation to sample weights as initialization for training on SVHN. However, we now also vary the architecture. 
While the decoder outputs a fixed-sized vector of weights, we can assign these weights to new architectures by either making sure that the new architecture still has the same number of parameters or by initializing randomly the extra parameters introduced. 
Specifically, we create three cases: (1) we add ResNet-style skip connections~\citep{heDeepResidualLearning2016} (1x1 conv) to the convolutional layers (3-conv + res-skip), (2) re-distribute the weights to smaller four convolutional layers (4-conv), (3) re-distribute to smaller four convolutional layers and add identity skip connections (4-conv + id.-skip).

\begin{wraptable}{r}{0.71\linewidth}
\vspace{-6mm}
\captionof{table}[Performance of Weight Sampled for Unseen Architectures]{
Test accuracy (\%) on SVHN of populations with generated weights compared to models trained from scratch $B_T$. Best results for each epoch and dataset are bolded.
\texttt{r. i.} indicates random initialization, \texttt{gen.} denotes weights generated with our ($S_{\text{KDE}30}$).
}
\label{tab:gen_hyper_reps:cross_architecture}
\vspace{2mm}
\small
\setlength{\tabcolsep}{2pt}
\begin{tabularx}{\linewidth}{rccc}
\toprule
\multicolumn{1}{c}{\textbf{Initialization}} & \multicolumn{1}{c}{\textbf{Epoch 1}} & \multicolumn{1}{c}{\textbf{Epoch 5}} & \multicolumn{1}{c}{\textbf{Epoch 50}} \\ 
\cmidrule(r){1-1} \cmidrule(rl){2-2} \cmidrule(rl){3-3} \cmidrule(rl){4-4}
3-conv (r. i.) + res-skip (r.i.)   & 18.9±1.6                         & 31.4±17                        & 50.6±28                         \\
3-conv (gen.) + res-skip (r.i.)     & \textbf{34.5±14 }              & \textbf{60.5±21 }              & \textbf{68.0±21}                \\
\cmidrule(r){1-1} \cmidrule(rl){2-2} \cmidrule(rl){3-3} \cmidrule(rl){4-4}
4-conv (r.i.)                        & 19.2±1.0                         & 19.2±0.9                         & 55.2±11                         \\
4-conv (gen.)                          & \textbf{44.0±4.5 }               & \textbf{57.8±3.5 }               & \textbf{67.6±1.9 }                \\
\cmidrule(r){1-1} \cmidrule(rl){2-2} \cmidrule(rl){3-3} \cmidrule(rl){4-4}
4-conv + id.-skip (r.i.)        & 18.9±1.0                         & 19.6±1.7                         & 56.4±7.9                          \\
4-conv + id.-skip (gen.)          & \textbf{48.0±4.0 }               & \textbf{59.9±2.5 }               & \textbf{66.4±1.7 }                \\ 
\bottomrule
\end{tabularx}
\vspace{-4mm}
\end{wraptable} 
\paragraph*{Results:} Surprisingly, despite training our hyper- representation on the models of the same architecture, generated weights for all three cases outperform random initialization and converge significantly faster across all the variations (Table \ref{tab:gen_hyper_reps:cross_architecture}). In all the variations even just after 5 epochs, the models with generated weights are better than training the baseline for 50 epochs. 
In the 3-conv + res-skip experiments, some models in both populations did not learn, which leads to high standard deviation. 
Further analysis is required to explain the gains of our approach in this challenging setup. 
To extend and scale up our method further, future work could combine it with the methods of growing networks \citep{chenNet2NetAcceleratingLearning2016,wangRecurrentParameterGenerators2021}, so that some layers are generated while some are initialized in a sophisticated way to preserve the functional form of the network.\looseness-1
\subsection{Limitations of Zoos with Small Models}\label{sec:gen_hyper_reps:limitations}
To thoroughly investigate different methods and make experiments feasible, we chose to use the model zoos of the same small scale as in~\citep{schurholtSelfSupervisedRepresentationLearning2021}. 
While on MNIST and SVHN, the architectures of such model zoos allowed us to achieve high performance, on CIFAR-10 and STL-10, the performance of all populations is limited by the low capacity of the model zoo's architecture. The models saturate at around 50\% and 40\% accuracy, respectively. 
The sampled populations reach the saturation point and fluctuate, but cannot outperform the baselines, see Appendix \ref{app:results} for details. 
We hypothesize that due to the high remaining loss, the weight updates are correspondingly large without converging or improving performance. 
This may cause the weights to contain relatively little signal and high noise. 
Larger model architectures might mitigate this behavior. Corresponding model zoos have recently been made available in \citep{schurholtModelZoosDataset2022} to tackle this issue\footnote{\href{www.modelzoos.cc}{www.modelzoos.cc}}.\looseness-1

\section{Related Work} 

\paragraph*{HyperNetworks:}
Recently, representation learning on neural networks is typically based on HyperNetworks that
learn low-dimensional structure of model weights to generate weights in a deterministic fashion~\citep{haHyperNetworks2017,bertinetto2016learning,knyazevParameterPredictionUnseen2021,zhangGraphHyperNetworksNeural2019}. HyperNetworks have also been extended to meta-learning by conditioning weight generation on data~\citep{zhmoginovHyperTransformerModelGeneration2022,requeima2019fast}. Closely related to our work, HyperGANs~\citep{ratzlaffHyperGANGenerativeModel2019} can sample model weights by combining the hypernetworks and the GAN framework. Similarly, \citep{deutschGeneratingNeuralNetworks2018} allow for sampling model weights by conditioning the hypernetwork on a noise vector. However, training hypernetwork-based methods require input data (e.g. images) to feed to the neural networks. 
In practice, there may already be large collections of trained models, while their training data may not always be accessible.
Learning representations of model weights without data, called hyper-representations, has been recently introduced in~\citep{schurholtSelfSupervisedRepresentationLearning2021}. Our methods build on that work to allow for better reconstruction and sampling.
\citep{denilPredictingParametersDeep2013} showed that given a few parameters of a network, the remaining values of a single model can be accurately reconstructed. However, in our work, we leverage the autoencoder to train a representation of the entire model zoo. 
Very recently, \citep{peeblesLearningLearnGenerative2022} used diffusion on a population of models to generate model weights for the original task via prompting.\looseness-1

\paragraph*{Transfer Learning:}
Transfer learning via fine-tuning aims at re-using models and their learned knowledge from a source to a target task~\citep{yosinskiHowTransferableAre2014,chen2019closer,dhillon2019baseline,mensinkFactorsInfluenceTransfer2021,kolesnikov2020big}. 
Transfer learning models make training less expensive, boost performance, or allow training on datasets with very few samples and have been applied on a wide range of domains~\citep{zhuangComprehensiveSurveyTransfer2020}. 
The common transfer learning methods however only consider transferring from a single model, and so disregard the large variety of pre-trained models and the potential benefit of combining them.

\paragraph*{Knowledge distillation:} Our work is related to~\citep{wang2018adversarial,liuKnowledgeFlowImprove2019,shuZooTuningAdaptiveTransfer2021} that allows to distill knowledge from a model zoo into a single network. Knowledge distillation overcomes the inherent limitation of transfer learning by transferring the knowledge from many large teacher models to a relatively small student model~\citep{liuKnowledgeFlowImprove2019,shuZooTuningAdaptiveTransfer2021}. 
Knowledge distillation however requires the source models at training as in~\citep{liuKnowledgeFlowImprove2019} and at inference as in\citep{shuZooTuningAdaptiveTransfer2021} thus increasing memory cost. Further, the learned knowledge cannot be shared between different target models.
\textbf{Learnable initialization} \citep{dauphinMetaInitInitializingLearning2019,zhu2021gradinit} provide methods to improve initialization by leveraging the meta-learning and gradient-flow ideas.
In contrast to knowledge distillation and learnable initialization, we train a hyper-representation of a model zoo in a latent space, which is a more general and powerful approach that can enable sampling an ensemble, property estimation, improved initialization, and implicit knowledge distillation across datasets.


\section{Conclusion}
\label{sec:gen_hyper_reps:conclusion}

In this paper, we propose a new method to sample from hyper-representations to generate neural network weights in one forward pass.
We extend the training objective of hyper-representations by a novel layer-wise loss normalization which is key to the capability of generating functional models.
Our method allows us to generate diverse populations of model weights, which show high performance as ensembles.
We evaluate sampled models both in-dataset as well as in transfer learning and find them capable of outperforming both models trained from scratch, as well as pre-trained and fine-tuned models.
Populations of sampled models, even for some unseen architectures, generally learn faster and achieve statistically significantly higher performance. 
This demonstrates that such hyper-representation can be used as a generative model for neural network weights and therefore might serve as a building block for transfer learning from different domains, meta-learning, or continual learning.\looseness-1

\begin{subappendices}
\section{Model Zoo Details}
\label{app:zoos}
\begin{wraptable}{r}{0.6\linewidth}
\vspace{-6mm}
\small
\captionof{table}[Overview of the model zoos]{
Model zoo overview.}\label{tab:gen_hyper_reps:zoo_overview}\vspace{2mm}
\setlength{\tabcolsep}{2pt}
{\small
\begin{tabularx}{\linewidth}{lccc}
\toprule
\textbf{Zoo} & \textbf{Channels} & \textbf{Parameters} & \textbf{Population Size} \\ 
\midrule
MNIST        & 1                       & 2464                & 1000                     \\
SVHN         & 1                       & 2464                & 1000                     \\
CIFAR-10     & 3                       & 2864                & 1000                     \\
STL-10       & 3                       & 2864                & 1000                     \\ 
\bottomrule
\end{tabularx}
}
\vspace{-4mm}
\end{wraptable} 
The model zoos are generated following the method of ~\citep{schurholtSelfSupervisedRepresentationLearning2021,schurholtModelZoosDataset2022}
An overview of the model zoos is given in Table \ref{tab:gen_hyper_reps:zoo_overview}.
All model zoos share one general CNN architecture, outlined in Table \ref{tab:gen_hyper_reps:model_zoo_architecture}. 
The hyperparameter choices for each of the populations are listed in Table \ref{tab:gen_hyper_reps:model_zoo_hyperparameters}. 
The hyperparameters are chosen to generate zoos with smooth, continuous development and spread in performance.

\begin{table}[ht!]
    \begin{center}
    {\small
    \caption[CNN Architecture Details of the Model Zoos.]{CNN architecture details for the models in model zoos. }
    \setlength{\tabcolsep}{8pt}
    \begin{tabularx}{0.5\linewidth}{llc}
        \toprule
        \textbf{Layer}          & \textbf{Component} & \textbf{Value} \\
        \cmidrule(r){1-1} \cmidrule(rl){2-2}  \cmidrule(rl){3-3}
        \multirow{5}{*}{Conv 1} & input channels     & 1/3                 \\
                                & output channels    & 8                    \\
                                & kernel size        & 5                    \\
                                & stride             & 1                    \\
                                & padding            & 0                    \\
        \cmidrule(r){1-1} \cmidrule(rl){2-2}  \cmidrule(rl){3-3}
        Max Pooling             & kernel size        & 2                    \\
        \cmidrule(r){1-1} \cmidrule(rl){2-2}  \cmidrule(rl){3-3}
        Activation              & tanh / gelu         &                      \\
        \cmidrule(r){1-1} \cmidrule(rl){2-2}  \cmidrule(rl){3-3}
        \multirow{5}{*}{Conv 2} & input channels     & 8                    \\
                                & output channels    & 6                    \\
                                & kernel size        & 5                    \\
                                & stride             & 1                    \\
                                & padding            & 0                    \\
        \cmidrule(r){1-1} \cmidrule(rl){2-2}  \cmidrule(rl){3-3}
        Max Pooling             & kernel size        & 2                    \\
        \cmidrule(r){1-1} \cmidrule(rl){2-2}  \cmidrule(rl){3-3}
        Activation              & tanh / gelu         &                      \\
        \cmidrule(r){1-1} \cmidrule(rl){2-2}  \cmidrule(rl){3-3}
        \multirow{5}{*}{Conv 3} & input channels     & 6                    \\
                                & output channels    & 4                    \\
                                & kernel size        & 2                    \\
                                & stride             & 1                    \\
                                & padding            & 0                    \\
        \cmidrule(r){1-1} \cmidrule(rl){2-2}  \cmidrule(rl){3-3}
        Activation              & tanh / gelu         &                      \\
        \cmidrule(r){1-1} \cmidrule(rl){2-2}  \cmidrule(rl){3-3}
        \multirow{2}{*}{Linear 1} & input channels     & 36                   \\
                                & output channels    & 20                   \\
        \cmidrule(r){1-1} \cmidrule(rl){2-2}  \cmidrule(rl){3-3}
        Activation              & tanh / gelu         &                      \\
        \cmidrule(r){1-1} \cmidrule(rl){2-2}  \cmidrule(rl){3-3}
        \multirow{2}{*}{Linear 2} & input channels     & 20                   \\
                                & output channels    & 10                   \\
        \bottomrule
    \end{tabularx}
    \label{tab:gen_hyper_reps:model_zoo_architecture}
    }
    \end{center}
\end{table}
\begin{table}[ht]
    \setlength{\tabcolsep}{8pt}
    \centering
    {\small
    \caption[Hyperparameter Details for the Model Zoos. ]{Hyperparameter choices for the model zoos. }
    \begin{tabularx}{0.63\linewidth}{lll}
        \toprule
        \textbf{Model Zoo}     & \textbf{Hyperparameter} & \textbf{Value}   \\
        \cmidrule(r){1-1} \cmidrule(rl){2-2}  \cmidrule(rl){3-3}
        \multirow{5}{*}{MNIST} & input channels          & 1                \\
                               & activation              & tanh             \\
                               & weight decay                      & 0                \\
                               & learning rate                      & 3e-4             \\
                               & initialization          & uniform          \\
                               & optimizer               & Adam             \\
                               & seed                    & [1-1000]         \\
        \cmidrule(r){1-1} \cmidrule(rl){2-2}  \cmidrule(rl){3-3}
        \multirow{5}{*}{SVHN}  & input channels          & 1                \\
                               & activation              & tanh             \\
                               & weight decay                      & 0                \\
                               & learning rate                      & 3e-3             \\
                               & initialization          & uniform          \\
                               & optimizer               & adam             \\
                               & seed                    & [1-1000]         \\
        \cmidrule(r){1-1} \cmidrule(rl){2-2}  \cmidrule(rl){3-3}
        \multirow{5}{*}{CIFAR-10} & input channels          & 3                \\
                               & activation              & gelu             \\
                               & weight decay                      & 1e-2             \\
                               & learning rate                      & 1e-4             \\
                               & initialization          & kaiming-uniform  \\
                               & optimizer               & adam             \\
                               & seed                    & [1-1000]         \\
        \cmidrule(r){1-1} \cmidrule(rl){2-2}  \cmidrule(rl){3-3}
        \multirow{5}{*}{STL-10}& input channels          & 3                \\
                               & activation              & tanh             \\
                               & weight decay                      & 1e-3             \\
                               & learning rate                      & 1e-4             \\
                               & initialization  a        & kaiming-uniform  \\
                               & optimizer               & adam             \\
                               & seed                    & [1-1000]         \\
        \bottomrule
    \end{tabularx}
    \label{tab:gen_hyper_reps:model_zoo_hyperparameters}

    }
\end{table}
\newpage
\section{Hyper-Representation Architecture and Training Details}
\label{app:training}
\begin{figure}[ht!]
\begin{minipage}[t]{1.0\textwidth}
\begin{center}
\includegraphics[trim=0in 0in 0in 0in, width=0.99\linewidth]{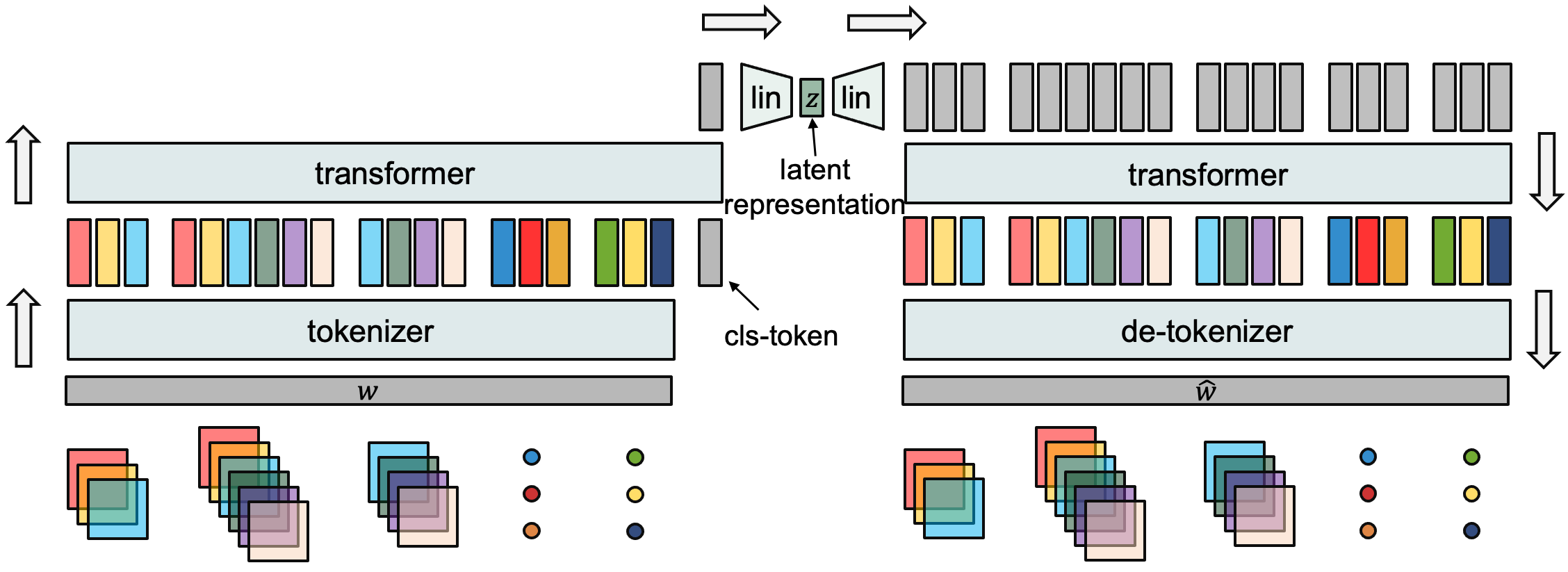}
\caption[Schematic of the Hyper-Representations Auto-Encoder Architecture.]{
Schematic of the auto-encoder architecture to learn hyper-representations. 
}
\label{fig:gen_hyper_reps:autoencoder_architecture}    
\end{center}
\end{minipage}
\end{figure}
Hyper-representations are learned with an autoencoder based on multi-head self-attention. The architecture is outlined in Figure \ref{fig:gen_hyper_reps:autoencoder_architecture}. 
Convolutional and fully connected neurons are embedded to token embeddings of dimension $d_{token}$. 
Learned position encodings are added to provide relational information.
A learned compression token (CLS) is appended to the sequence of token embeddings. 
The sequence of token embeddings is passed to $N_{layers}$ layers of multi-head self-attention with $N_{heads}$ heads with hidden embedding dimension $d_{hidden}$. 
The CLS token is compressed to the bottleneck of dimension $d_{z}$ with an MLP or a linear layer.
For the decoder, an MLP or a linear layer maps the bottleneck to a sequence of token embeddings.
The sequence is passed through another stack of multi-head self-attention, which is symmetric to the encoder.
Debedders map the token embeddings back to convolutional and fully connected neurons.
The reconstruction and contrastive loss are balanced with a parameter $\beta$. The contrastive loss is computed on the embeddings $\mathbf{z}$ mapped through a projection head $\mathbf{\bar{z}}=p(\mathbf{z}$, where $p$ is a learned MLP with four layers with 400 neurons each and $\mathbf{\bar{z}}$ has 50 dimensions.
In Table \ref{tab:gen_hyper_reps:hyper_rep_training}, the exact hyper-parameters for each of the hyper-representations are listed to reproduce our results.

\begin{table}[ht!]
    \centering
    \begin{minipage}{0.7\linewidth}
    \centering
    {\small
    \caption[Hyper-Representation Architecture and Training Details.]{Hyper-representation architecture and training details. }
    \begin{tabularx}{\linewidth}{lcccc}
    \toprule
                        & \texttt{MNIST}    & \texttt{SVHN}     & \texttt{CIFAR-10} & \texttt{STL-10}   \\
    \cmidrule(r){2-2} \cmidrule(rl){3-3}  \cmidrule(rl){4-4} \cmidrule(rl){5-5}
                        & \multicolumn{4}{c}{\textit{Architecture}} \\
    \cmidrule(r){2-2} \cmidrule(rl){3-3}  \cmidrule(rl){4-4} \cmidrule(rl){5-5}
    $d_{inpot}$         & 2464     & 2464     & 2864     & 2864     \\
    $d_{token}$         & 972      & 1680     & 1488     & 1632     \\
    $d_{hidden}$        & 1140     & 1800     & 1164     & 1680     \\
    $N_{layers}$        & 2        & 4        & 2        & 4        \\
    $N_{heads}$         & 12       & 12       & 12       & 24       \\
    $d_{z}$             & 700      & 1000     & 700      & 700      \\
    Compression         & linear   & linear   & linear   & linear   \\
    \cmidrule(r){2-2} \cmidrule(rl){3-3}  \cmidrule(rl){4-4} \cmidrule(rl){5-5}
                        & \multicolumn{4}{c}{\textit{Training}} \\
    \cmidrule(r){2-2} \cmidrule(rl){3-3}  \cmidrule(rl){4-4} \cmidrule(rl){5-5}
    Optimizer           & Adam     & Adam     & Adam     & Adam     \\
    Learning rate       & 0.0001   & 0.0001   & 0.0001   & 0.0001   \\
    Dropout             & 0.1      & 0.1      & 0.1      & 0.1      \\
    Weight Decay        & 1e-09    & 1e-09    & 1e-09    & 1e-09    \\
    $\beta$             & 0.977    & 0.920    & 0.950    & 0.950    \\
    training epochs     & 1750     & 1750     & 500      & 2000     \\
    batch size          & 500      & 250      & 200      & 200     \\
    \bottomrule
    \end{tabularx}
    \label{tab:gen_hyper_reps:hyper_rep_training}
    }
    \end{minipage}

\end{table}

\newpage
\section{Evaluation of Layer-Wise Loss Normalization }
\label{app:loss}
To evaluate layer-wise loss normalization, we compare two hyper-representations with comparable reconstruction. Both have a $R^2=1-\frac{mse(\hat{\mathbf{w}},\mathbf{w})}{mse(\mathbf{w_{mean},\mathbf{w}}}$ as a measure of the explained variance of around 70\%. One is trained with the baseline hyper-representation MSE, the other with layer-wise-normalization. 
\begin{figure}[ht!]
\begin{minipage}[t]{1.0\textwidth}
\begin{center}
\includegraphics[trim=0in 0in 0in 0in, width=0.8\linewidth]{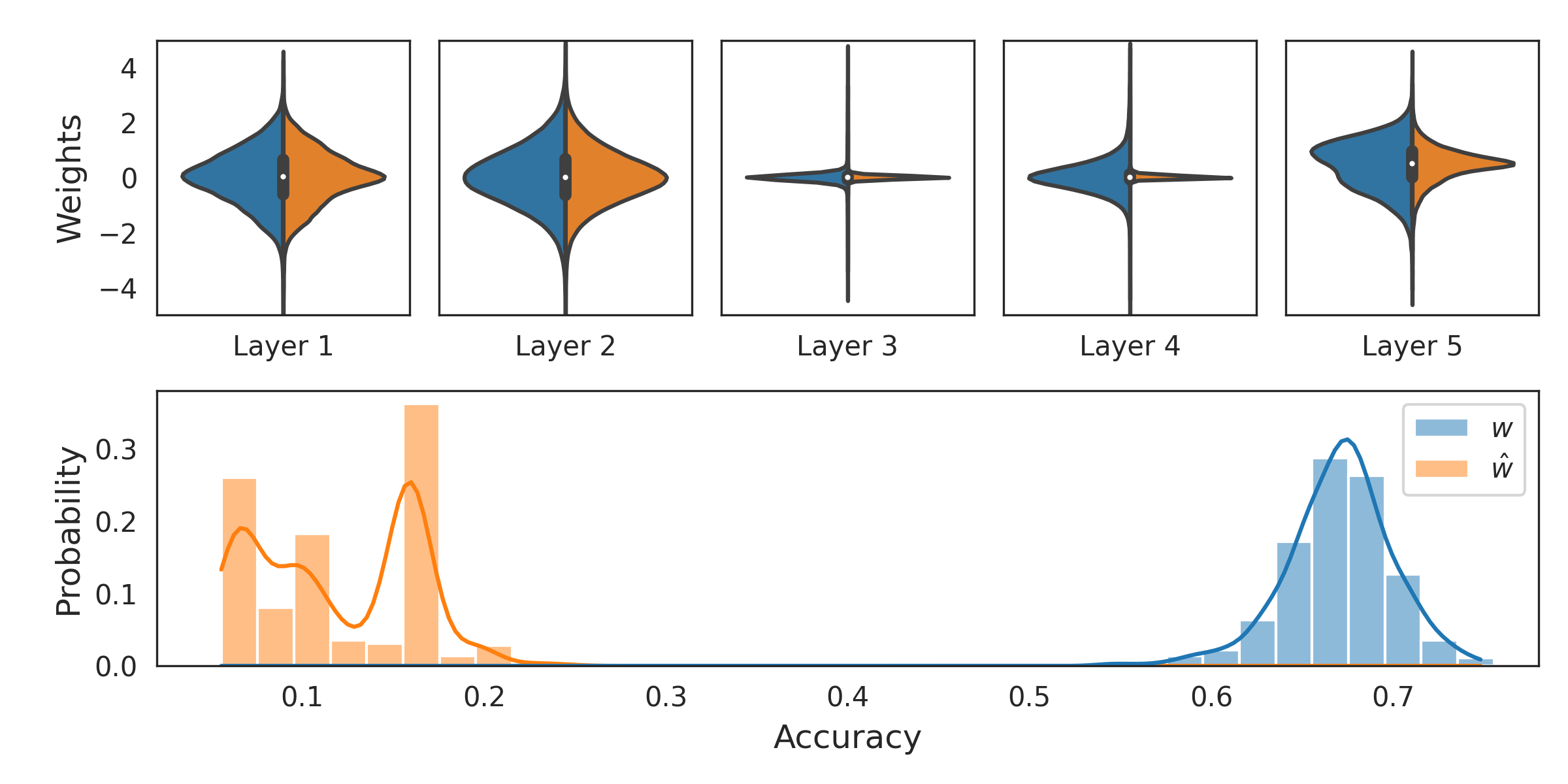}
\vskip -0.1in
\caption[Weight Distribution of the SVHN Model Zoo]{
\textbf{Top:} Weight distribution per layer (1-5) of the SVHN test set before $w$ and after reconstruction $\hat{w}$ with the baseline hyper-representation training loss. 
Layers 3 and 4 have small weight distributions, therefore adding little penalty to the MSE, and are consequently poorly reconstructed. 
\textbf{Bottom:} Accuracy distribution of the same population before and after reconstruction. 
The badly reconstructed layers (top) cause the reconstructed models to perform around random guessing.
}
\label{fig:gen_hyper_reps:weight_distro_raw}    
\end{center}
\end{minipage}
\vspace{2mm}
\begin{minipage}[t]{1.0\textwidth}
\begin{center}
\includegraphics[trim=0in 0in 0in 0in, width=0.8\linewidth]{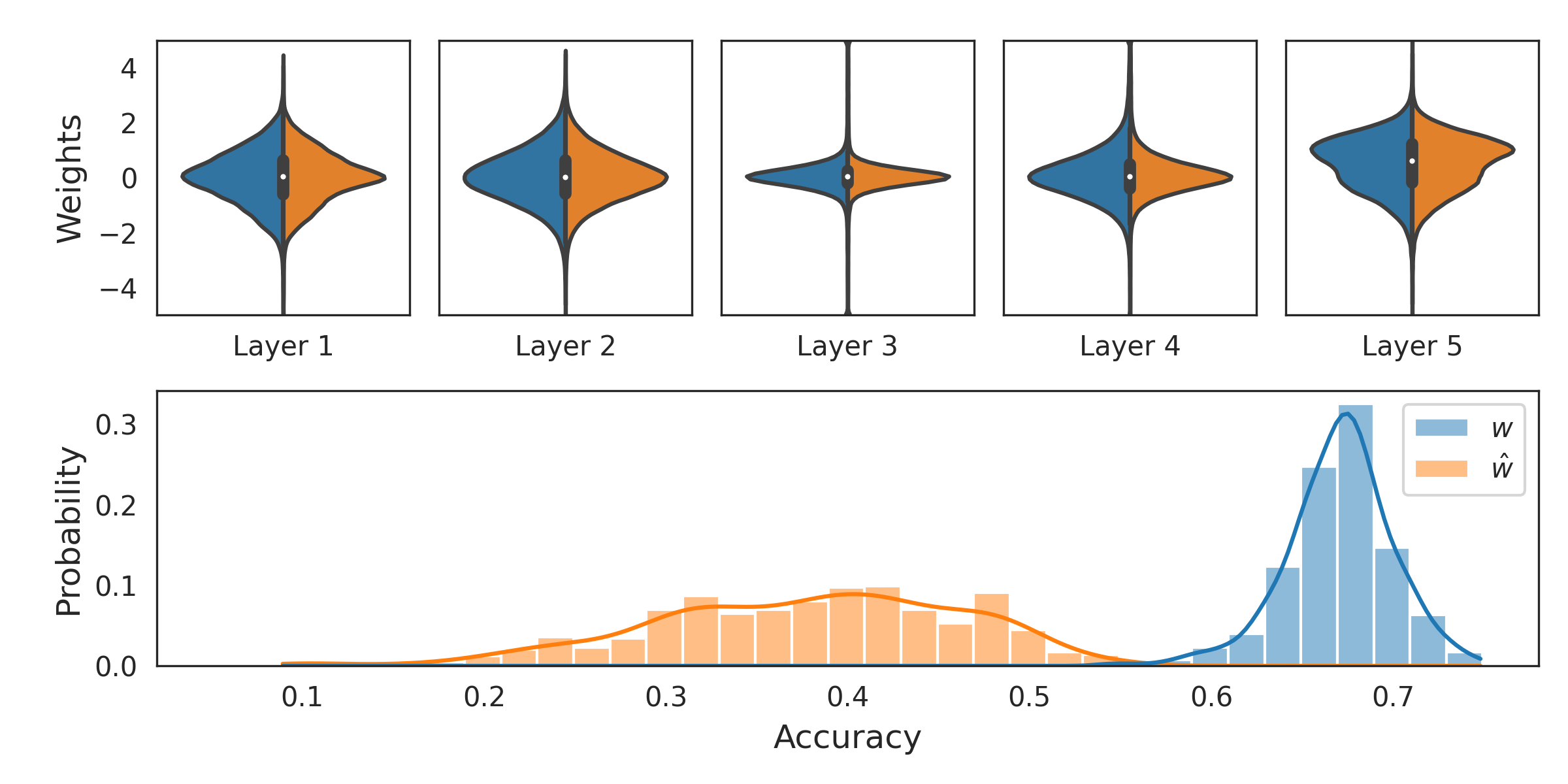}
\vskip -0.1in
\caption[Normalized Weight Distribution of the SVHN Model Zoo]{
\textbf{Top:} Weight distribution per layer (1-5) of the SVHN test set before $w$ and after reconstruction $\hat{w}$ with layer-wise loss normalization.
The distributions of all layers are more similar, the reconstruction is equally distributed across the layers.
\textbf{Bottom:} Accuracy distribution of the same population before and after reconstruction. 
The normalization fixes the catastrophic failure of the models. The remaining loss in accuracy can be explained by the remaining reconstruction error. 
}
\label{fig:gen_hyper_reps:weight_distro_norm}    
\end{center}
\end{minipage}
\end{figure}
Figures \ref{fig:gen_hyper_reps:weight_distro_raw} and \ref{fig:gen_hyper_reps:weight_distro_norm} show the distribution of weights per layer before and after reconstruction, as well as the accuracy distribution of both populations on the SVHN image test set. With the baseline learning scheme in Figure \ref{fig:gen_hyper_reps:weight_distro_raw}, the distributions in layers 3 and 4 do not match. In these layers, the original weight distribution is smaller, and so there is only a small error even if the reconstructions predict the mean. These layers become a weak link of the reconstructed models and cause performance around random guessing.
With layer-wise loss normalization in Figure \ref{fig:gen_hyper_reps:weight_distro_norm}, the weight distribution between the layers becomes more similar. As a consequence, the reconstruction error is more evenly distributed across the layers, there are no single layers that aren't reconstructed at all.  
This appears to allow information to flow forward through the model and significantly improves the performance of reconstructed models. 
We find layer-wise-normalization necessary to reconstruct or sample functional models across all populations, where the weights are unevenly distributed. 

\section{Hyper-Representation Analysis}
\label{app:analysis}
In this section, we detail the analysis of hyper-representations. We begin with their geometry, followed by the distributions of individual dimensions of hyper-representations, and finally investigate robustness and smoothness.

\begin{figure}[ht!]
\begin{minipage}[t]{0.49\textwidth}
\begin{center}
\includegraphics[trim=0in 0in 0in 0in, clip, width=0.9\linewidth]{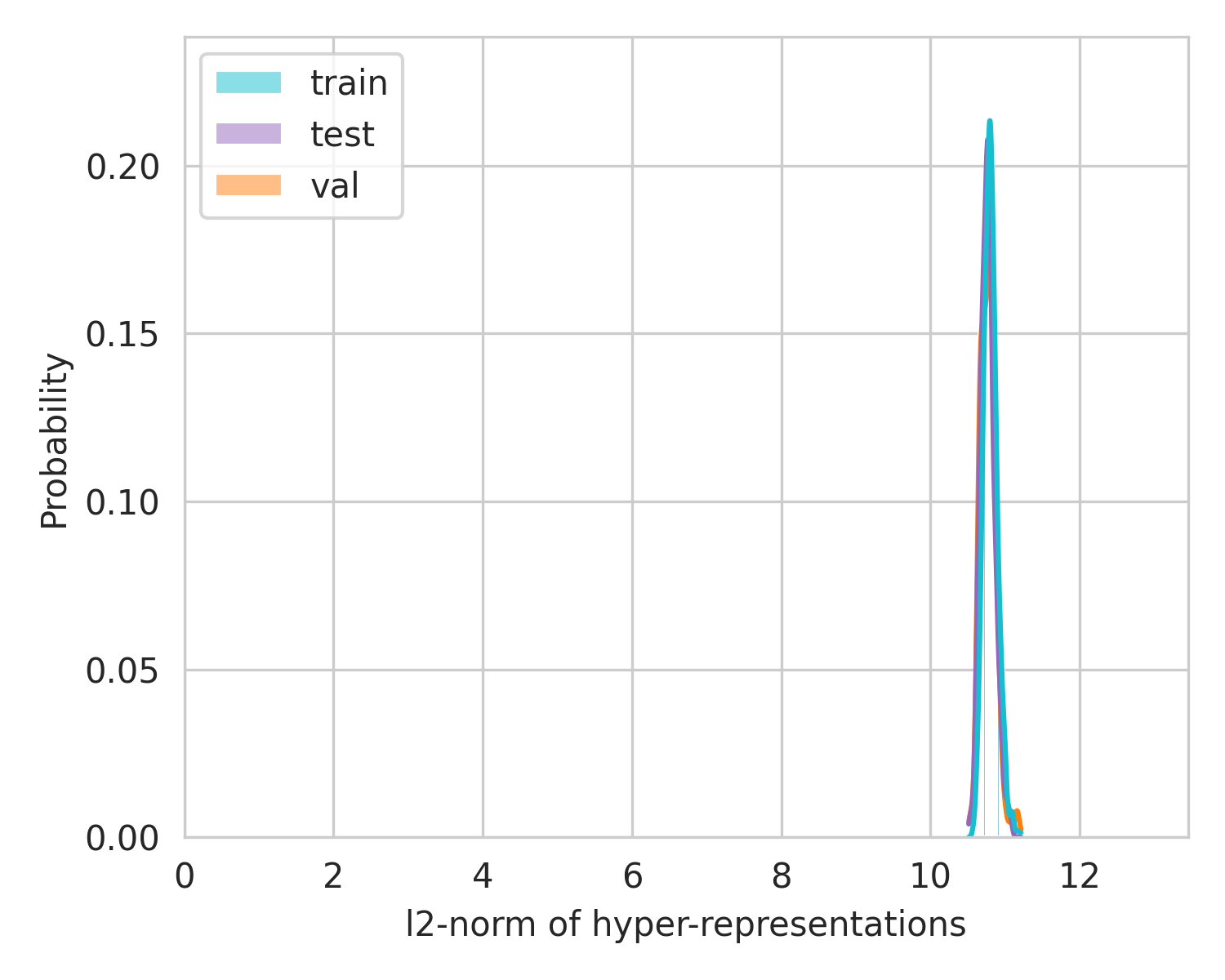}
\vskip -0.1in
\caption[Distribution of Euclidean Distance in Hyper-representations]{Distributions of $\ell_2$ norm of hyper-representations $\mathbf{z}$ of the MNIST zoo.}
\label{fig:gen_hyper_reps:hyper_sphere_z}    
\end{center}
\end{minipage}
\hskip 0.2cm
\begin{minipage}[t]{0.49\textwidth}
\begin{center}
\includegraphics[trim=0in 0in 0in 0in, clip, width=0.9\linewidth]{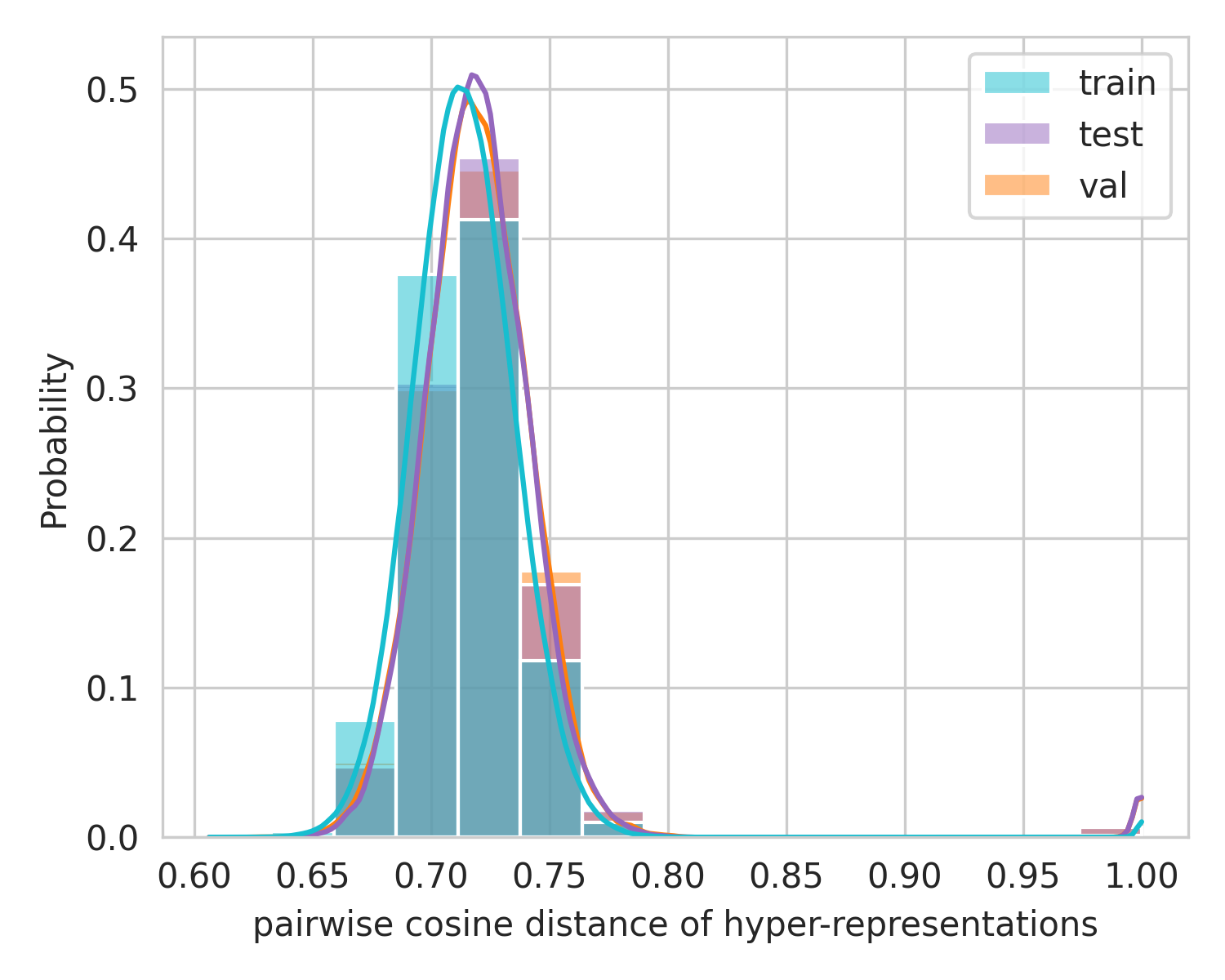}
\vskip -0.1in
\caption[Distribution of Cosine Distance in Hyper-representations]{Distributions of pairwise cosine distance of hyper-representations $\mathbf{z}$ of the MNIST zoo.}
\label{fig:gen_hyper_reps:hyper_sphere_cos}    
\end{center}
\end{minipage}
\end{figure}

\paragraph*{Embeddings in Hyper-Representation Space Populate a Hyper-Sphere}
We analyze the geometry of hyper-representations $\mathbf{z}$. 
The space of hyper-representations is bounded to a high dimensional box by a tanh activation. Surprisingly, hyper-representations do not populate the entire space, but sections on a shell of a high-dimensional sphere.
Figure \ref{fig:gen_hyper_reps:hyper_sphere_z} shows the distribution of the norm of the embeddings of the MNIST zoo. 
All embeddings are distributed on a small band between lengths 10 and 12, therefore they must populate the shell of a hyper-sphere.
In Figure \ref{fig:gen_hyper_reps:hyper_sphere_cos} we investigate pairwise cosine distances between the embeddings of the MNIST zoo. The majority of the embeddings populate the region between 0.6 and 0.8. 
The outliers around 1.0 are embeddings of the same model at different epochs. 
This indicates that models are not entirely orthogonal, but mutually equally far apart, populating a section of the shell of the hyper-sphere. 
While hyper-spheres are commonly found in embeddings of contrastive  learning~\citep{jingUnderstandingDimensionalCollapse2021}, in our experiments hyper-spheres form even without a contrastive loss. Properties of the models embedded on that hyper-sphere can be predicted from hyper-representations, therefore the topology on the sphere appears to encode model properties. \looseness-1

\begin{figure}[ht!]
\begin{minipage}[t]{0.99\textwidth}
\begin{center}
\includegraphics[trim=0mm 5.3mm 0mm 0mm, clip, width=1.0\linewidth]{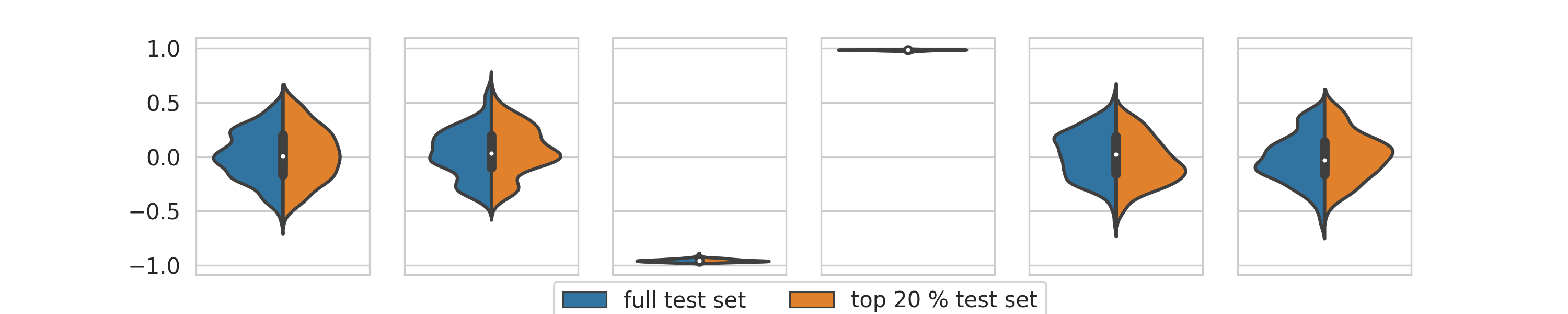}
\vskip 0mm
\caption[Distributions of Individual Dimension of Hyper-representations]{Distributions of individual dimensions of hyper-representations $\mathbf{z}$ of the MNIST zoo. In {\color{cyan}blue} is the distribution of all samples, in {\color{orange}orange} the subset of the 30 \% best samples.}
\label{fig:gen_hyper_reps:z_wise_distribution}    
\end{center}
\end{minipage}
\end{figure}

\paragraph*{Distributions of Dimensions of Embeddings in Hyper-Representation Encode Properties}
Previous work showed that linear probing from hyper-representations accurately predicts i.e. model accuracy. 
In these linear probes, the individual $z$ dimensions each linearly contribute to accuracy predictions. 
This allows us to investigate $z$ dimensions independently.
Figure \ref{fig:gen_hyper_reps:z_wise_distribution} shows examples for the distribution of selected individual dimensions of hyper-representations $\mathbf{z}$. On the left is the distribution of the entire population, and on the right of the top 30 \% performing models.
The individual dimensions show different types of distributions, with different modes. Most have a zero mean and span 3/4 of the available range, but some collapse to either $-1$ or $1$. 
Further, the distributions also differ in at least some dimension between the entire population, and the better-performing split of the population.

\paragraph*{Generalization Capabilities of Hyper-Representations to Diverse Model Zoos}
There are certain architectural changes such as adding/removing/changing pooling layers and nonlinearity that do not change the number of parameters (the dimensionality of the input/output required by our approach). These changes as well as changes in hyperparameters used to train models in a zoo may drastically alter the distribution of weights and pose a challenge to the proposed approach. Modern neural networks (ResNet, MobileNet, EfficientNet, etc.) are often trained with very different hyperparameters. With the experiment below, we investigate the generalization capabilities of hyper-representations to such changes, which might be important for modern large-scale settings as well.

\textbf{Setup:} We experimentally evaluate the generalizability of the proposed approach on models trained with a different choice of nonlinearity or other hyperparameters with two experiments (a and b). To that end, in addition to the original SVHN test zoo (zoo 1), we use two more diverse SVHN zoos (zoo 2 and zoo 3). In zoo 2, in addition to random seed, models differ in the activation (tanh, relu, gelu, sigmoid), l2-regularization (0, 0.001, 0.1), and dropout (0,0.3,0.5). In zoo 3 (extending zoo 2), we increase the diversity further by additionally varying the initialization method (uniform, normal, kaiming-uniform, kaiming-normal) and the learning rate (0.0001, 0.001, 0.01).

\begin{table}[ht!]
\setlength{\tabcolsep}{12pt}
\centering
{\small
\caption[Generalization of Hyper-Representations to Diverse Zoo]{Generalizability of hyper-representations towards more diverse model zoo configurations (measured as the reconstruction score, higher is better).}
\begin{tabularx}{\linewidth}{rccc}
\toprule
\textbf{Training zoo}    & \textbf{Test zoo 1}  & \textbf{Test zoo 2} & \textbf{Test zoo 3} \\
\textbf{}    & \textbf{original}  & \textbf{vary activation} & \textbf{ vary hyperparameters} \\
\midrule
Original        & 81.9\%               & 45.7\%                                              & 38.9\%                           \\
Diverse (zoo 3) & 25.8\%               & 89.1\%                                              & 75.6\%                          \\
\bottomrule
\end{tabularx}
}
\label{tab:gen_hyper_reps:hyper_rep_generalization}
\end{table}

\textbf{Experiment (a):} We first evaluate our original encoder-decoder trained on a model zoo varying in random seed only. For evaluation, we pass the test splits of zoo 2 and zoo 3 through the encoder-decoder. We measure the reconstruction $R^2$ score of the original encoder-decoder on the diverse test zoos. \\
\textbf{Results:} Our results (Table 5.D.1) indicate that our original encoder-decoder can still encode and decode weights even in such a challenging setting, although there is an expected drop in performance.

\textbf{Experiment (a):} We next evaluate if hyper-representations can be trained on diverse zoos. For this experiment, we train a hyper-representation on the train split of zoo 3. With this, we aim to show that training hyper-representations on diverse zoos improve generalization capabilities further. \\
\textbf{Results:} Our results show that training on diverse zoos is a much more difficult task to optimize, hence the reconstruction on the original zoo degrades. It nonetheless improves the reconstruction results on the test split of the diverse zoos 2 and 3. This indicates that varying seeds and hyperparameters may be different aspects of complexity that need to be considered.

\newpage
\section{Sampling Methods }
\label{app:sampling}
\subsection*{VAE}
A common extension of the autoencoder of \citep{schurholtSelfSupervisedRepresentationLearning2021} to enable sampling from its latent representation is to make the autoencoder variational~\citep{kingmaAutoEncodingVariationalBayes2013}.
In our experiments, VAEs could not be trained to reconstruct model weights without unweighting the KL-divergence to insignificance essentially making it deterministic as in~\citep{schurholtSelfSupervisedRepresentationLearning2021}. 
Empirically, embeddings in hyper-representations are mapped on the shell of a sphere (see Section \ref{app:analysis}) and leave the inside of the sphere entirely empty. 
On the other hand, a Gaussian prior allocates most of the probability mass near the center of the sphere. 
It therefore appears plausible that the two may be incompatible.
That issue of non-compatible priors is well known.
~\citep{ghoshVariationalDeterministicAutoencoders2020} find that regularizing embeddings and decoder yields equally smooth representation spaces as VAEs without restrictions to specific priors.
During training of hyper-representations, both encoder and decoder are regularized with a small $\ell_2$ penalty. Further, dropout is applied throughout the autoencoder, which serves as another regularizer and adds blurriness to the embeddings. The combination of dropout, the erasing augmentation and the contrastive loss further regularizes the hyper-representation space. 
In all our sampling methods, we draw samples from probability distributions, which effectively disconnects the drawn samples from training embeddings.
\looseness-1

\subsection*{Latent Space GAN Details}
The generator and discriminator of our GAN consist of four fully-connected layers interleaved with ReLU nonlinearities. The same architecture and training hyperparameters are used for all experiments.
The generator's input is a Gaussian noise $\mathbf{n}^*$ of dimensionality $d=16$, the hidden dimensionalities are 128, 256, and 512, and the output dimensionality is equal to the hyper-representation length $D$.
The discriminator's input is $D$-dimensional, the hidden dimensionalities are 1024, 512, and 256, and the output dimensionality is a scalar denoting either a real or fake sample.
The discriminator is regularized with Spectral Norm~\cite{miyatoSpectralNormalizationGenerative2018}. The discriminator and generator are trained for 1000 epochs and batch size 32 using Adam with a two-time-scale update rule~\cite{heusel2017gans}: learning rate is 1e-4 for the generator and 2e-4 for the discriminator.

\FloatBarrier
\newpage
\section{Full Experiment Results }
\label{app:results}

\begin{table}[ht!]
\centering
\setlength{\tabcolsep}{12pt}
{\small
\caption[Full Results on Sampling Experiments - Digits]{Accuracy of sampled models: median and 95\% confidence intervals. On the main diagonal are in-dataset experiments, otherwise transfer-learning from source to target. Bold numbers highlight the best source-to-target results. N/A denotes cases, in which the boot-strapped CI on the median could not be computed.}
\begin{tabularx}{0.8\linewidth}{lccc}
\toprule
Population           & Source                 & \multicolumn{2}{c}{Target}                    \\
\cmidrule(rl){1-1} \cmidrule(rl){2-2}  \cmidrule(rl){3-4} 
                     & \multicolumn{1}{l}{}   & MNIST                 & SVHN                  \\
 \cmidrule(rl){3-3} \cmidrule(rl){4-4}  
$B_T$                & \multirow{8}{*}{MNIST} & 91.1 {[}91.1, 91.2{]} & 72.3 {[}72.0, 72.4{]} \\
\cmidrule(rl){1-1} \cmidrule(rl){2-2}  \cmidrule(rl){3-4} 
$B_F$                &                        & 91.2 {[}91.0, 91.3{]} & 76.2 {[}75.8, 76.5{]} \\
$S_{\text{KDE}}$     &                        & 92.3 {[}92.1, 92.8{]} & 76.7 {[}76.2, 77.0{]} \\
$S_{\text{KDE}30}$   &                        & 93.1 {[}92.9, 93.4{]} & 77.2 {[}76.8, 77.6{]} \\
$S_{\text{Neigh}}$   &                        & 93.4 {[}93.2, 93.5{]} & 76.8 {[}76.4, 77.1{]} \\
$S_{\text{Neigh}30}$ &                        & \textbf{94.0 {[}93.8, 94.1{]}} & \textbf{77.0 {[}76.3, 77.4{]}} \\
$S_{\text{GAN}}$     &                        & 93.5 {[}93.3, 93.6{]} & 76.9 {[}76.6, 77.6{]} \\
$S_{\text{GAN}30}$   &                        & 93.9 {[}93.5, 93.9{]} & 76.5 {[}76.3, 76.8{]} \\
\cmidrule(rl){1-1} \cmidrule(rl){2-2}  \cmidrule(rl){3-4} 
$B_F$                & \multirow{7}{*}{SVHN}  & 95.1 {[}95.0, 95.3{]} & 73.2 {[}72.8, 73.4{]} \\
$S_{\text{KDE}}$     &                        & 95.1 N/A              & 73.0 {[}72.6, 73.3{]} \\
$S_{\text{KDE}30}$   &                        & 95.5 N/A              & 74.2 {[}73.9, 74.5{]} \\
$S_{\text{Neigh}}$   &                        & \textbf{97.2 {[}97.0, 97.3{]}} & \textbf{78.1 {[}77.9, 78.2{]}} \\
$S_{\text{Neigh}30}$ &                        & 95.5 {[}95.4, 95.7{]} & 76.5 {[}76.3, 76.7{]} \\
$S_{\text{GAN}}$     &                        & 94.3 {[}94.1, 94.6{]} & 74.5 {[}74.0, 74.9{]} \\
$S_{\text{GAN}30}$   &                        & 94.9 {[}94.8, 95.1{]} & 75.3 {[}75.0, 75.6   \\
\bottomrule
\end{tabularx}
}
\label{tab:gen_hyper_reps:accuracy_digits}
\vspace{-.2in}
\end{table}
\begin{table}[ht!]
\setlength{\tabcolsep}{12pt}
\centering
{\small
\caption[Statistical Significance of Sampling Experiments - Digits]{Mann-Whitney U test of Samples S vs Baselines B: p-value and CLES (Common Language Effect Size). p-values indicate the probability of the samples of two groups originating from the same distribution. CLES=0.5 indicates no effect, CLES=1.0 a strong positive, CLES=0.0 a strong negative effect.
As the results indicate, both proposed sampling methods are almost always statistically significantly better than the two baselines. Further, their effect is often very strong. 
}
\begin{tabularx}{0.85\linewidth}{@{}lccc@{}}
\toprule
Population Pairs                 & Source                  & \multicolumn{2}{c}{Target}                            \\
\cmidrule(rl){1-1} \cmidrule(rl){2-2}  \cmidrule(rl){3-4} 
                                 &                         & MNIST                     & SVHN                      \\
 \cmidrule(rl){3-3} \cmidrule(rl){4-4}
$S_{\text{KDE}}$ vs. $B_{T}$     & \multirow{12}{*}{MNIST} & \textbf{2.1e-18 | 0.8701} & \textbf{5.2e-27 | 0.9551} \\
$S_{\text{KDE}}$ vs. $B_{F}$     &                         & \textbf{0.0e+00 | 0.8639} & \textbf{1.1e-01 | 0.5920} \\
$S_{\text{KDE}30}$ vs. $B_{T}$   &                         & \textbf{7.0e-27 | 0.9539} & \textbf{2.5e-29 | 0.9754} \\
$S_{\text{KDE}30}$ vs. $B_{F}$   &                         & \textbf{6.9e-22 | 0.9545} & \textbf{1.7e-04 | 0.7180} \\
$S_{\text{Neigh}}$ vs. $B_{T}$   &                         & \textbf{1.5e-30 | 0.9857} & \textbf{6.6e-31 | 0.9888} \\
$S_{\text{Neigh}}$ vs. $B_{F}$   &                         & \textbf{4.5e-25 | 0.9889} & \textbf{5.2e-03 | 0.6622} \\
$S_{\text{Neigh}30}$ vs. $B_{T}$ &                         & \textbf{1.7e-35 | 0.9987} & \textbf{1.3e-29 | 0.9778} \\
$S_{\text{Neigh}30}$ vs. $B_{F}$ &                         & \textbf{3.1e-28 | 0.9994} & \textbf{1.4e-02 | 0.6426} \\
$S_{\text{GAN}}$ vs. $B_{T}$     &                         & \textbf{7.6e-31 | 0.9883} & \textbf{8.0e-25 | 0.9351} \\
$S_{\text{GAN}}$ vs. $B_{F}$     &                         & \textbf{3.0e-25 | 0.9907} & \textbf{7.8e-03 | 0.6546} \\
$S_{\text{GAN}30}$ vs. $B_{T}$   &                         & \textbf{1.1e-31 | 0.9953} & \textbf{2.1e-26 | 0.9496} \\
$S_{\text{GAN}30}$ vs. $B_{F}$   &                         & \textbf{6.8e-26 | 0.9973} & \textbf{4.9e-02 | 0.6144} \\
\cmidrule(rl){1-1} \cmidrule(rl){2-2}  \cmidrule(rl){3-4} 
$S_{\text{KDE}}$ vs. $B_{T}$     & \multirow{12}{*}{SVHN}  & \textbf{6.1e-79 | 0.9943} & \textbf{1.1e-04 | 0.6006} \\
$S_{\text{KDE}}$ vs. $B_{F}$     &                         & 7.8e-01 | 0.4904          & 3.8e-01 | 0.4704          \\
$S_{\text{KDE}30}$ vs. $B_{T}$   &                         & \textbf{1.7e-82 | 1.0000} & \textbf{1.6e-30 | 0.7985} \\
$S_{\text{KDE}30}$ vs. $B_{F}$   &                         & \textbf{0.0e+00 | 0.7292} & \textbf{3.0e-08 | 0.6850} \\
$S_{\text{Neigh}}$ vs. $B_{T}$   &                         & \textbf{2.9e-78 | 0.9867} & \textbf{8.6e-80 | 0.9916} \\
$S_{\text{Neigh}}$ vs. $B_{F}$   &                         & \textbf{2.8e-44 | 0.9661} & \textbf{1.8e-47 | 0.9833} \\
$S_{\text{Neigh}30}$ vs. $B_{T}$ &                         & \textbf{1.7e-82 | 1.0000} & \textbf{4.7e-76 | 0.9797} \\
$S_{\text{Neigh}30}$ vs. $B_{F}$ &                         & \textbf{8.2e-08 | 0.6791} & \textbf{1.7e-42 | 0.9563} \\
$S_{\text{GAN}}$ vs. $B_{T}$     &                         & \textbf{1.2e-31 | 0.9948} & \textbf{0.0e+00 | 0.8140} \\
$S_{\text{GAN}}$ vs. $B_{F}$     &                         & 1.5e-07 | 0.2517          & \textbf{7.5e-06 | 0.7118} \\
$S_{\text{GAN}30}$ vs. $B_{T}$   &                         & \textbf{4.2e-32 | 0.9987} & \textbf{6.7e-22 | 0.9067} \\
$S_{\text{GAN}30}$ vs. $B_{F}$   &                         & 3.6e-01 | 0.4565          & \textbf{0.0e+00 | 0.8335} \\
\bottomrule
\end{tabularx}
}
\label{tab:gen_hyper_reps:test_digits}
\vspace{-.2in}
\end{table}

\begin{table}[ht!]
\setlength{\tabcolsep}{12pt}
\centering
{\small
\caption[Full Results on Sampling Experiments - Natural Images]{Accuracy of sampled models: median and 95\% confidence intervals. On the main diagonal are in-dataset experiments, otherwise transfer-learning from source to target. Bold numbers highlight the best source-to-target results. N/A denotes cases, in which the boot-strapped CI on the median could not be computed.}
\begin{tabularx}{0.85\linewidth}{lcll}
\toprule
Population           & Source                    & \multicolumn{2}{c}{Target}                                      \\
\cmidrule(rl){1-1} \cmidrule(rl){2-2}  \cmidrule(rl){3-4} 
                     & \multicolumn{1}{l}{}      & \multicolumn{1}{c}{CIFAR-10}   & \multicolumn{1}{c}{STL-10}     \\
 \cmidrule(rl){3-3} \cmidrule(rl){4-4}  
$B_T$                & \multirow{8}{*}{CIFAR-10} & 49.0 {[}48.9, 49.0{]}          & 39.0 {[}38.9, 39.1{]}          \\
\cmidrule(rl){1-1} \cmidrule(rl){2-2}  \cmidrule(rl){3-4} 
$B_F$                &                           & 48.6 {[}48.3, 48.7{]}          & \textbf{42.8 {[}42.5, 42.9{]}} \\
$S_{\text{KDE}}$     &                           & 48.3 {[}48.1, 48.4{]}          & 40.7 {[}40.3, 40.9{]}          \\
$S_{\text{KDE}30}$   &                           & \textbf{48.7 {[}48.4, 48.8{]}} & 41.3 {[}40.9, 41.5{]}          \\
$S_{\text{Neigh}}$   &                           & 45.6 {[}44.9, 46.0{]}          & 36.7 {[}35.8, 37.4{]}          \\
$S_{\text{Neigh}30}$ &                           & 46.2 {[}45.8, 46.4{]}          & 37.9 {[}37.3, 38.2{]}          \\
$S_{\text{GAN}}$     &                           & 46.0 N/A                       & 38.6 {[}38.1, 39.0{]}          \\
$S_{\text{GAN}30}$   &                           & 47.0 {[}46.5, 47.2{]}          & 38.6 {[}38.2, 39.1{]}          \\
\cmidrule(rl){1-1} \cmidrule(rl){2-2}  \cmidrule(rl){3-4} 
$B_F$                & \multirow{7}{*}{STL-10}   & \textbf{49.3 {[}49.0, 49.4{]}} & \textbf{39.5 {[}38.9, 39.7{]}} \\
$S_{\text{KDE}}$     &                           & 48.6 {[}48.4, 48.9{]}          & 37.3 {[}37.0, 37.8{]}          \\
$S_{\text{KDE}30}$   &                           & 48.8 {[}48.4, 49.2{]}          & 38.3 {[}37.9, 38.4{]}          \\
$S_{\text{Neigh}}$   &                           & 10.0 N/A                       & 28.3 {[}26.8, 29.1{]}          \\
$S_{\text{Neigh}30}$ &                           & 49.0 {[}48.5, 49.1{]}          & 37.8 {[}37.6, 38.2{]}          \\
$S_{\text{GAN}}$     &                           & 49.0 {[}48.6, 49.4{]}          & 38.5 {[}37.9, 38.9{]}          \\
$S_{\text{GAN}30}$   &                           & 48.8 {[}48.5, 49.1{]}          & 37.9 N/A                      \\
\bottomrule
\end{tabularx}
}
\label{tab:gen_hyper_reps:accuracy_natural_images}
\vspace{-.2in}
\end{table}
\begin{table}[ht!]
\centering
\setlength{\tabcolsep}{12pt}
{\small
\caption[Statistical Significance of Sampling Experiments - Natural Images]{Mann-Whitney U test of Samples S vs Baselines B: p-value and CLES (Common Language Effect Size). p-values indicate the probability of the samples of two groups originating from the same distribution. CLES=0.5 indicates no effect, CLES=1.0 a strong positive, CLES=0.0 a strong negative effect.
}
\begin{tabularx}{0.85\linewidth}{@{}lccc@{}}
\toprule
Population Pairs                 & Source                     & \multicolumn{2}{c}{Target}                            \\
\cmidrule(rl){1-1} \cmidrule(rl){2-2}  \cmidrule(rl){3-4} 
                                 &                            & CIFAR-10                  & STL-10                    \\
 \cmidrule(rl){3-3} \cmidrule(rl){4-4}
$S_{\text{KDE}}$ vs. $B_{T}$     & \multirow{12}{*}{CIFAR-10} & 1.5e-06 | 0.2966          & \textbf{7.4e-19 | 0.8750} \\
$S_{\text{KDE}}$ vs. $B_{F}$     &                            & 3.7e-02 | 0.4014          & 1.7e-18 | 0.0849          \\
$S_{\text{KDE}30}$ vs. $B_{T}$   &                            & 3.6e-02 | 0.4114          & \textbf{4.8e-25 | 0.9371} \\
$S_{\text{KDE}30}$ vs. $B_{F}$   &                            & \textbf{2.9e-01 | 0.5498} & 0.0e+00 | 0.1266          \\
$S_{\text{Neigh}}$ vs. $B_{T}$   &                            & 5.7e-28 | 0.0364          & 7.4e-18 | 0.1359          \\
$S_{\text{Neigh}}$ vs. $B_{F}$   &                            & 3.1e-22 | 0.0413          & 7.1e-26 | 0.0024          \\
$S_{\text{Neigh}30}$ vs. $B_{T}$ &                            & 3.5e-25 | 0.0616          & 2.0e-07 | 0.2800          \\
$S_{\text{Neigh}30}$ vs. $B_{F}$ &                            & 2.2e-19 | 0.0741          & 3.0e-25 | 0.0089          \\
$S_{\text{GAN}}$ vs. $B_{T}$     &                            & 6.6e-25 | 0.0642          & 6.6e-02 | 0.4223          \\
$S_{\text{GAN}}$ vs. $B_{F}$     &                            & 2.8e-19 | 0.0754          & 1.0e-24 | 0.0145          \\
$S_{\text{GAN}30}$ vs. $B_{T}$   &                            & 2.1e-21 | 0.0983          & 1.1e-02 | 0.3928          \\
$S_{\text{GAN}30}$ vs. $B_{F}$   &                            & 8.8e-16 | 0.1195          & 2.7e-25 | 0.0084          \\
\cmidrule(rl){1-1} \cmidrule(rl){2-2}  \cmidrule(rl){3-4} 
$S_{\text{KDE}}$ vs. $B_{T}$     & \multirow{12}{*}{STL-10}   & 1.3e-01 | 0.4362          & 0.0e+00 | 0.1730          \\
$S_{\text{KDE}}$ vs. $B_{F}$     &                            & 6.9e-04 | 0.3028          & 6.0e-10 | 0.1404          \\
$S_{\text{KDE}30}$ vs. $B_{T}$   &                            & 6.1e-01 | 0.4783          & 1.2e-06 | 0.2948          \\
$S_{\text{KDE}30}$ vs. $B_{F}$   &                            & 1.1e-02 | 0.3528          & 9.1e-06 | 0.2424          \\
$S_{\text{Neigh}}$ vs. $B_{T}$   &                            & 2.9e-32 | 0.0000          & 3.0e-32 | 0.0000          \\
$S_{\text{Neigh}}$ vs. $B_{F}$   &                            & 3.3e-20 | 0.0000          & 7.1e-18 | 0.0000          \\
$S_{\text{Neigh}30}$ vs. $B_{T}$ &                            & 1.0e+00 | 0.5000          & 4.3e-09 | 0.2517          \\
$S_{\text{Neigh}30}$ vs. $B_{F}$ &                            & 2.1e-02 | 0.3654          & 5.4e-07 | 0.2090          \\
$S_{\text{GAN}}$ vs. $B_{T}$     &                            & 3.2e-01 | 0.5418          & 2.0e-04 | 0.3427          \\
$S_{\text{GAN}}$ vs. $B_{F}$     &                            & 2.7e-01 | 0.4360          & 2.4e-04 | 0.2864          \\
$S_{\text{GAN}30}$ vs. $B_{T}$   &                            & 6.2e-01 | 0.4788          & 5.4e-07 | 0.2880          \\
$S_{\text{GAN}30}$ vs. $B_{F}$   &                            & 1.2e-02 | 0.3532          & 4.6e-06 | 0.2340         \\
\bottomrule
\end{tabularx}
}
\label{tab:gen_hyper_reps:test_natural_images}
\vspace{-.2in}
\end{table}

\end{subappendices}
\FloatBarrier

\afterpage{
    \thispagestyle{emptychaptertransition}
    \vspace*{\fill}
    \newpage
    \thispagestyle{plain}
}

\chapter[Towards Scalable and Versatile Weight Space Learning]{Towards Scalable and \\ Versatile Weight Space Learning}
\label{chap::scalable_hyper_reps}
\newcommand{\ourmethodlong}{\texttt{Sequential Autoencoder for Neural Embeddings}\,}
\newcommand{\ourmethod}{\texttt{SANE}\,}

\blfootnote{This work was accepted for publication at ICML 2024 \citep{schurholtScalableWeightSpaceLearning2024} }

\section*{Abstract}
Learning representations of well-trained neural network models holds the promise to provide an understanding of the inner workings of those models.
However, previous work has either faced limitations when processing larger networks or was task-specific to either discriminative or generative tasks.
This paper introduces the \ourmethod approach to weight-space learning. 
\ourmethod overcomes previous limitations by learning task-agnostic representations of neural networks that are scalable to larger models of varying architectures and that show capabilities beyond a single task.
Our method extends the idea of \textit{hyper-representations} towards sequential processing of subsets of neural network weights, 
thus allowing one to embed larger neural networks as a set of tokens into the learned representation space. 
\ourmethod reveals global model information from layer-wise embeddings, and it can sequentially generate unseen neural network models, which was unattainable with previous \textit{hyper-representation} learning methods. 
Extensive empirical evaluation demonstrates that \ourmethod matches or exceeds state-of-the-art performance on several weight representation learning benchmarks, 
particularly in initialization for new tasks and larger ResNet architectures.\looseness-1

%
%
%
%
\section{Introduction}
The exploration of the ``weight space'' of neural network (NN) models, i.e., the high-dimensional space spanned by the model parameters of a population of trained NNs, allows us to gain insights into the inner workings of those models. 

\begin{figure}[t!]
\centering
\includegraphics[width=0.8\columnwidth]{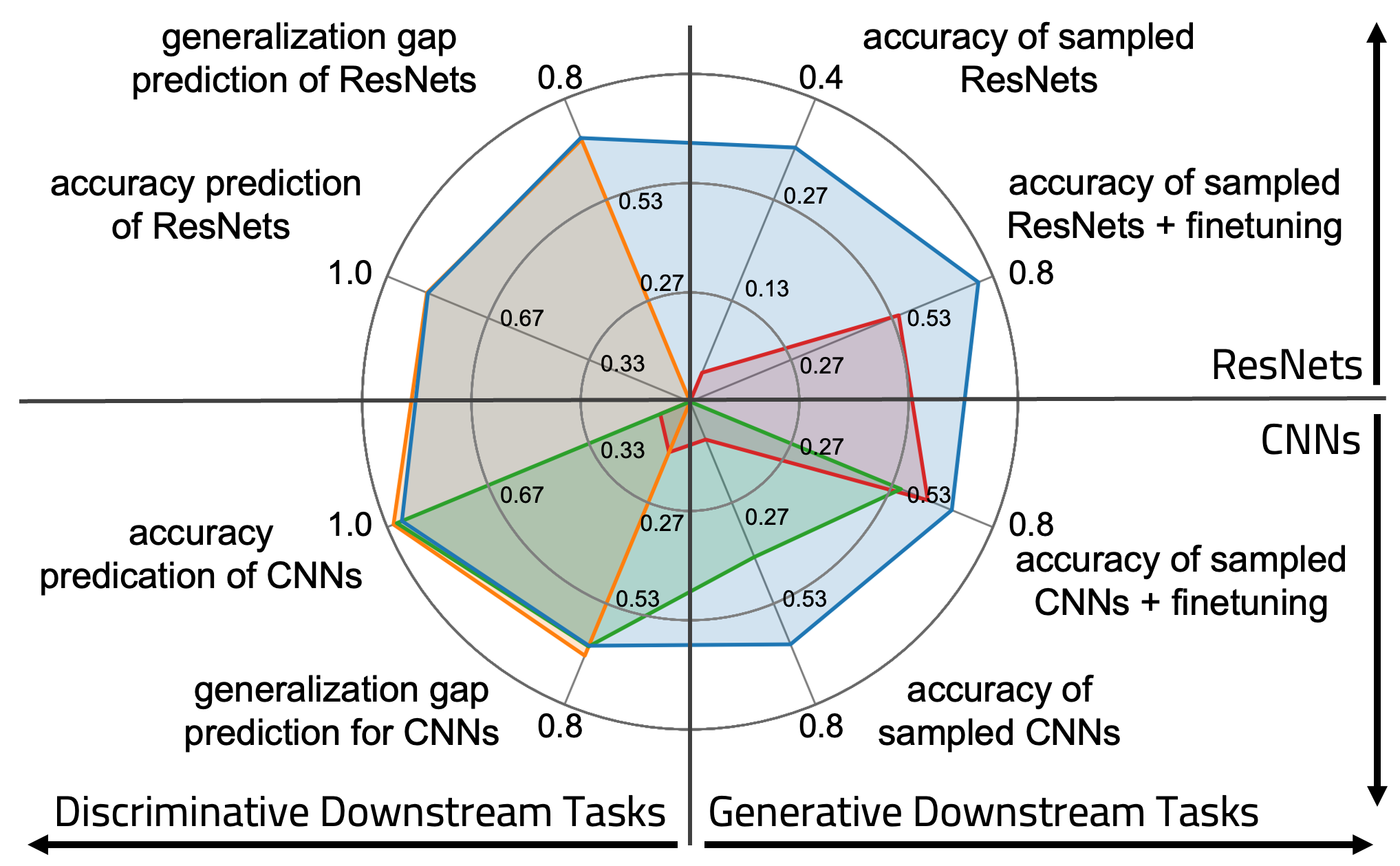}
\caption[Performance Overview on Generative and Discriminative Tasks]{Aggregated results of 56 experiments showing \textbf{(left:)} four discriminative downstream tasks in $R^2$, and (\textbf{right:}) four generative downstream tasks in accuracy, each evaluated on (\textbf{bottom:}) CNNs model zoos trained on 4 datasets (NMIST, SVHN, CIFAR-10, STL) and (\textbf{top:}) ResNet18 model zoos trained on three datasets (CIFAR-10, CIFAR-100, Tiny-ImageNet).  
The colors indicate performance of {\color{Red}\textbf{\texttt{Red:}}} raw NN weights, {\color{RedOrange}\textbf{\texttt{Orange:}}} weight statistics from \citet{unterthinerPredictingNeuralNetwork2020}, {\color{Green}\textbf{\texttt{Green:}}} trained \textit{hyper-representations} from \citet{schurholtSelfSupervisedRepresentationLearning2021, schurholtHyperRepresentationsGenerativeModels2022}, and {\color{Blue}\textbf{\texttt{Blue:}}} \ourmethod (ours).
While some methods perform well on specific tasks, or are restricted by the size of the underlying models, \ourmethod can deliver excellent performance on all tasks and model sizes.
}
\label{fig:scalable_hyper_reps:radar_plot_overview}
\end{figure}

%
%
In the discriminative context, previous works aim to link weight space properties to properties such as model quality, generalization gap, or hyperparameters, using either the margin distribution \citep{yakTaskArchitectureIndependentGeneralization2018,jiangPredictingGeneralizationGap2019}, graph topology features \citep{corneanuComputingTestingError2020}, or eigenvalue decompositions of weight matrices \citep{martinTraditionalHeavyTailedSelf2019,MM20_SDM,martinImplicitSelfregularizationDeep2021,martinPredictingTrendsQuality2021}.
%
%
Some works learn classifiers to map between statistics of weights and model properties \citep{eilertsenClassifyingClassifierDissecting2020, unterthinerPredictingNeuralNetwork2020},
or learn lower-dimensional manifolds to infer NN model properties \citep{schurholtSelfSupervisedRepresentationLearning2021}.

\begin{figure*}[t!]
\centering
\includegraphics[width=0.75\paperwidth]{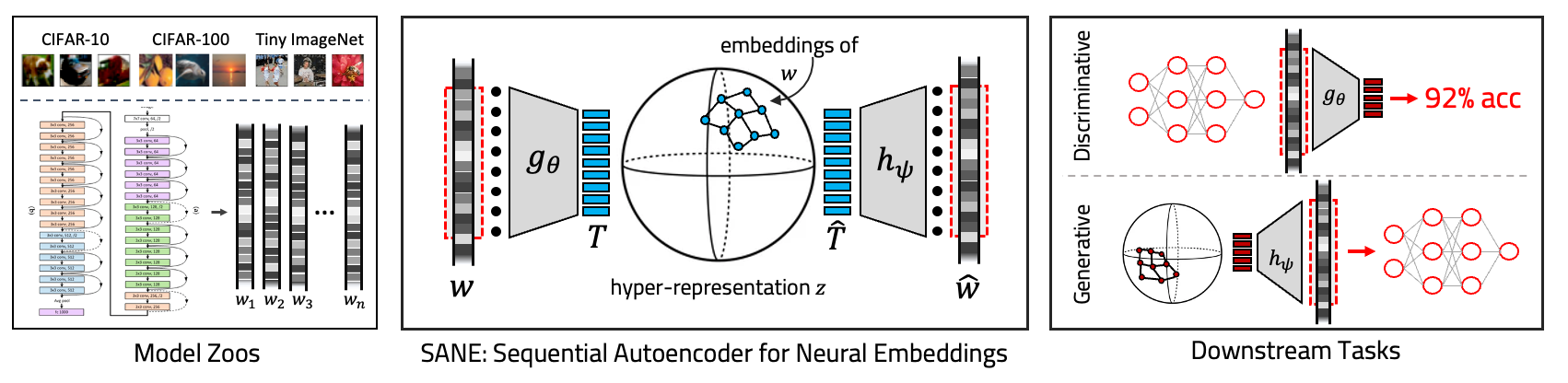}
\caption[Schematic Overview of \ourmethod]{
Given model zoos trained on different classification tasks, we extract and sequentialize the model weights. \ourmethod trains \textit{hyper-representations} on 
weights subsequences, i.e., individual layers.
\ourmethod can be used for multiple downstream tasks, either using the encoder for discriminative tasks such as the prediction of model accuracy, or the decoder for generative tasks such as sampling of new models.
}
\vspace{-4mm}
\label{fig:scalable_hyper_reps:approach_overview}
\end{figure*}

%
%
In the generative context, methods have been proposed to generate model weights using (Graph) HyperNetworks ~\citep{haHyperNetworks2017, zhangGraphHyperNetworksNeural2019, knyazevParameterPredictionUnseen2021}, Bayesian HyperNetworks \citep{deutschGeneratingNeuralNetworks2018}, HyperGANs \citep{ratzlaffHyperGANGenerativeModel2019}, and HyperTransformers \citep{zhmoginovHyperTransformerModelGeneration2022}.
These approaches have been used for tasks such as neural architecture search, model compression, ensembling, transfer learning, and meta-learning. They have in common that they derive their learning signal from the underlying (typically image) dataset. 
In contrast to these methods, 
so-called \textit{hyper-representations}~\citep{schurholtHyperRepresentationsGenerativeModels2022} learn a lower-dimensional representation directly from the weight space without the need to have access to data, e.g., the image dataset, to sample unseen NN models from that latent representation.

%
%
In this paper, we present \ourmethodlong (\ourmethod), an approach to learn task-agnostic representations of NN weight spaces capable of embedding individual NN models into a latent space to perform the above-mentioned discriminative or generative downstream tasks.
Our approach builds upon the idea of 
\textit{hyper-representations}~\citep{schurholtSelfSupervisedRepresentationLearning2021, schurholtHyperRepresentationsGenerativeModels2022}, which learn a lower-dimensional representation $z$ from a population of NN models. 
This is accomplished by auto-encoding their flattened weight vectors $w_i$ through a transformer architecture, with the bottleneck acting as a lower-dimensional embedding $z_i$ of each NN model. 
While the \textit{hyper-representation} method promises to be useful for discriminative and generative tasks, until now, separate \textit{hyper-representations} had to be trained specifically for either discriminative or generative tasks. 
Additionally, existing approaches have a major shortcoming: the underlying encoder-decoder model has to embed the entire flattened weight vectors $w_i$ at once into the learned lower-dimensional representation $z$. 
This drastically limits the size of NNs that can be embedded.
\ourmethod addresses these limitations by decomposing the entire weight vector $w_i$ into layers or smaller subsets, and then sequentially processes them. 
Instead of encoding the entire NN model by one embedding, \ourmethod encodes a potentially very large NN as multiple embeddings. 
%
%
The change from processing the entire flattened weight vector to subsets of weights is motivated by \citet{martinRethinkingGeneralizationRequires2019,martinImplicitSelfregularizationDeep2021}, who showed that global model information is preserved in the layer-wise components of NNs. 
An illustration of our approach can be found in Fig.~\ref{fig:scalable_hyper_reps:approach_overview}.\looseness-1

To evaluate \ourmethod, we analyze how NN embeddings encoded by \ourmethod behave in comparison to \citet{martinRethinkingGeneralizationRequires2019,martinImplicitSelfregularizationDeep2021} quality measures. 
We show that some of these weight matrix quality metrics show similar characteristics as the embeddings produced by \ourmethod. 
This holds not only for held-out NN models of the model zoo used for training \ourmethod but also for NN models of out-of-distribution model zoos with different architectures and training data.
Further, we demonstrate that \ourmethod can learn \textit{hyper-representations} of much larger NN models, and so it makes them applicable to real-world problems. 
%
%
In particular, the models in the ResNet model zoo used for training are three orders of magnitudes larger than all model zoos used for \textit{hyper-representation} learning in previous works. 
While previous \textit{hyper-representation} learning methods were structurally constrained to encode the entire NN model at once, \ourmethod is scalable by applying its sequential approach to encode layers or subsets of weights into hyper-representation embeddings.
While we demonstrate scaling up to ResNet-101 models, \ourmethod is not fundamentally limited to that size.
Finally, we evaluate \ourmethod on both discriminative and generative downstream tasks.
For discriminate tasks, we evaluate on six model zoos by linear-probing for properties of the underlying NN models. 
For generative tasks, we evaluate on seven model zoos by sampling targeted model weights for initialization and transfer learning. 

We provide an aggregated overview of our results in Fig.~\ref{fig:scalable_hyper_reps:radar_plot_overview}. 
On very small CNN models (evaluated on MNIST, SVHN, CIFAR-10, and STL, which we include for comparison with prior work), \ourmethod performs as well as previous state-of-the-art (SOTA) in discriminative tasks. In generative downstream tasks, \ourmethod outperforms SOTA by 25\% in accuracy for initialization on the same task and 17\% in accuracy for finetuning to new tasks.
On larger models such as ResNets (evaluated on CIFAR-10, CIFAR-100, Tiny-ImageNet, which were beyond the capabilities of prior work), we show results comparable to baselines for discriminative downstream tasks, and we report outperformance to baselines for generative downstream tasks by 31\% for initialization and 28\% for finetuning to new tasks. 
Additionally, we show that \ourmethod can sample targeted models by prompting with different architectures than it used for training. These sampled models can outperform models trained from scratch on the prompted architecture. Code is available at \href{https://github.com/HSG-AIML/SANE}{github.com/HSG-AIML/SANE}.\looseness-1
%

%
%
\newpage
\section{Methods}

\textit{Hyper-representations} learn an encoder-decoder model on the weights of NNs \citep{schurholtSelfSupervisedRepresentationLearning2021}: 
    \begin{align}
        \mathbf{z} = g_{\theta}(\mathbf{W}) \label{eq:scalable_hyper_reps:hrep_enc}\\
        \mathbf{\widehat{W}} = h_{\psi}( \mathbf{z} ),\label{eq:scalable_hyper_reps:hrep_dec}
    \end{align}
where $g_{\theta}$ is the encoder which maps the flattened weights $\mathbf{W}$ to embeddings $\mathbf{z}$, and $h_{\psi}$ decodes back to reconstructed weights $\mathbf{\widehat{W}}$.
Even though previous work realized both encoder and decoder with transformer backbones, the weight vector had to be of fixed size, and models are represented in a global embedding space \citep{schurholtSelfSupervisedRepresentationLearning2021, schurholtHyperRepresentationsGenerativeModels2022}. 
\textit{Hyper-representations} are trained with a reconstruction loss $\mathcal{L}_{rec}=\|\mathbf{W}-\mathbf{\widehat{W}} \|_2^2$ and contrastive guidance loss $\mathcal{L}_{c} = NTXent(p_{\phi}(\mathbf{z_i}),p_{\phi}(\mathbf{z_j}))$, where $p_{\phi}$ is a projection head. \citet{schurholtSelfSupervisedRepresentationLearning2021} proposed weight permutation, noise, and masking as augmentations to generate views $i,j$ of the same model.

Existing \textit{hyper-representation} methods have two major limitations: 
i) using the full weight vector to compute global model embeddings becomes infeasible for larger models; 
and ii) they can only embed models that share the architecture with the original model zoo.
Our \ourmethod method addresses both of these limitations. 
To make models more digestible for pretraining and inference, we propose to express models as sequences of token vectors.
To address i), \ourmethod learns per-token embeddings, which are trained on subsequences of the full base model sequence. 
This way, the memory and compute load are decoupled from the base model size. 
By decoupling the tokenization from the representation learning, we also address ii). 
The models in the model zoo set can have varying architectures, as long as they are expressed as a sequence with the same token-vector size. 
The transformer backbone and per-token embeddings also allow changes to the length of the sequence during or after training. 
Below, we provide technical details on \ourmethod.
We first provide details on pretraining \ourmethod, computing model embeddings, and sampling models; and we then introduce additional \textit{aligning}, \textit{haloing}, and \textit{bn-conditioning} methods to stabilize training and inference.

\paragraph*{\ourmethodlong}
\label{sec:scalable_hyper_reps:seq_hyper_reps}
To tokenize weights, we reshape the weights  $\mathbf{W}_{raw} \in \mathbb{R}^{c_{out}\times c_{1}\times\cdots \times c_{in}}$ to 2d matrices $\mathbf{W} \in \mathbb{R}^{c_{out}\times c_{r}}$, where $c_{out}$ are the outgoing channels, and where $c_{r}$ the remaining, flattened dimensions. 
We then slice the weights row-wise, along the outgoing channel. 
Using global token size $d_t$, we split the slices into multiple parts if $c_r>d_t$ and zero-pad to fill up to $d_t$. 
For weights $\mathbf{W}_l$ of layer $l$, this gives us tokens $\mathbf{T}_l \in \mathbb{R}^{n_l \times d_t}$, where $n_l = c_{out,l} \, \mbox{ceil}(\frac{c_r}{d_t})$. 
Since all tokens $\mathbf{T}_l$ share the same token size, the tokens of layer $l=1,...,L$ can be concatenated to get the model token sequence $\mathbf{T} \in \mathbb{R}^{N\times d_t}$.
To indicate the position of a token, we use a 3-dimensional position $\mathbf{P}_n = [n,l,k]$, where $n\in[1, N]$ indicates the global position in the sequence, $l \in [1, L]$ indicate the layer index, and $k \in [1, K(l)]$ is the position of the token within the layer.

Out of the full token sequence $\mathbf{T}$ and positions $\mathbf{P} \in \mathbb{N}^{N \times 3}$, we take a random consecutive sub-sequence $\mathbf{T}_{s,n}=\mathbf{T}_{n,...,n+ws}$ with positions $\mathbf{P}_{s,n}=\mathbf{P}_{n,...,n+ws}$ of length $ws$. 
We call these sub-sequences windows and the length of the sub-sequence the window size $ws$.

For \ourmethod on windows of tokens, we extend Eqs. \ref{eq:scalable_hyper_reps:hrep_enc} and \ref{eq:scalable_hyper_reps:hrep_dec} to encode and decode token windows as
    \begin{align}
        \mathbf{z}_{s,n} &= g_{\theta}(\mathbf{T}_{s,n},\mathbf{P}_{s,n}) \label{eq:scalable_hyper_reps:seqhrep_enc}\\
        \widehat{\mathbf{T}}_{s,n} &= h_{\psi}( \mathbf{z}_{s,n},\mathbf{P}_{s,n}),\label{eq:scalable_hyper_reps:seqhrep_dec}
    \end{align}
where $\mathbf{z}_{s,n} \in \mathbb{R}^{ws \times d_z}$ is the per-token latent representation of the window. 
In contrast to \textit{hyper-representations} Eqs. \ref{eq:scalable_hyper_reps:hrep_enc} and \ref{eq:scalable_hyper_reps:hrep_dec} which operate on the full flattened weights of a model, \ourmethod encodes sub-sequences of tokenized models.
For simplicity, we apply linear mapping to and from the bottleneck, to reduce tokens from $d_t$ to $d_z$. 

We adapt the composite training loss of \textit{hyper-representations}, $\mathcal{L} = (1-\gamma) \mathcal{L}_{rec} + \gamma \mathcal{L}_c$, for sequences as:
\begin{align}
    \mathcal{L}_{rec} &=\| \mathbf{M}_{s,n} \odot \left( \mathbf{T}_{s,n}-\widehat{\mathbf{T}}_{s,n} \right)\|_2^2 \label{eq:scalable_hyper_reps:seqhrep_recon_loss} \\
    \mathcal{L}_{c} &= NTXent(p_{\phi}(\mathbf{z_{s,n,i}}),p_{\phi}(\mathbf{z_{s,n,j}})). \label{eq:scalable_hyper_reps:seqhrep_con_loss}
\end{align}
Here, the mask $\mathbf{M}_{s,n}$ indicates signal with $1$ and padding with $0$, to ensure that the loss is only computed on actual weights. The contrastive guidance loss uses the augmented views $i,j$ and projection head $p_{\phi}$.

The pretraining procedure is detailed in Algorithm \ref{alg:scalable_hyper_reps:pretraining}. 
We preprocess model weights by standardizing weights per layer and aligning all models to a reference model; see \textit{Model Alignment} below.
As in previous work \citep{schurholtSelfSupervisedRepresentationLearning2021,peeblesLearningLearnGenerative2022}, 
the encoder and decoder are realized as transformer blocks. 
Training on the full sequence would memory-limit the base-model size by its sequence length.
\begin{algorithm}[]
   \caption{\ourmethod pretraining}
   \label{alg:scalable_hyper_reps:pretraining}
\begin{algorithmic}
    \STATE {\bfseries Input:} population of models
    \STATE \textbf{i:} standardize models weights
    \STATE \textbf{ii:} align models to one common reference model
    \STATE \textbf{iii:} tokenize models to tokens $\mathbf{T}$, positions $\mathbf{P}$, masks $\mathbf{M}$
    \STATE \textbf{iv:} draw \textit{k} windows per model: $\mathbf{T}_{s,n}$, $\mathbf{P}_{s,n}$,  $\mathbf{M}_{s,n}$
    \STATE \textbf{v:} train on $\mathcal{L}_{train}$ until convergence of $\mathcal{L}_{val}$
\end{algorithmic}
\end{algorithm}
Training the encoder and decoder on windows instead of the full model sequence decouples the memory requirement from the base model's full sequence length. 
The window size can be used to balance GPU memory load and the amount of context information.
Notably, since we disentangle the tokenization from the representation learning model, \ourmethod also allows us to embed sequences of models with varying architectures, as long as their token size is the same. 
To prevent potential overfitting to specific window positions, we propose to sample windows from each model sequence multiple times randomly.

\paragraph*{Computing \ourmethod Model Embeddings.}
\ourmethod can be used to analyse models in embedding space, e.g., by using embeddings as features to predict properties such as accuracy, or to identify other model quality metrics.
In contrast to \textit{hyper-representations}, \ourmethod can embed different model sizes and architectures in the same embedding space.
To embed any model, we begin by preprocessing weights by standardizing per layer and aligning models to a pre-defined reference model (see \textit{Model Alignment} below).
Subsequently, the preprocessed models are tokenized as described above.
For short model sequences, the embedding sequences can be directly computed as $\mathbf{z} = g_\theta(\mathbf{T},\mathbf{P})$. For larger models, the token sequences are too long to embed as one. We therefore employ \textit{haloing} (see below) to encode the entire sequence as coherent subsequences.
Algorithm \ref{alg:scalable_hyper_reps:embeddings} summarizes the embedding computation. 
\begin{algorithm}[]
\vspace{4pt}
   \caption{\ourmethod model embedding computation}
   \label{alg:scalable_hyper_reps:embeddings}
\begin{algorithmic}
    \STATE {\bfseries Input:} population of models
    \STATE \textbf{i:} preprocessing: standardize and align model weights
    \STATE \textbf{ii:} tokenize models: $\mathbf{T}$, positions $\mathbf{P}$, property $y$
    \STATE \textbf{iii:} split $\mathbf{T}$, $\mathbf{P}$ to consecutive chunks $\mathbf{T}_{hs,n}, \mathbf{P}_{hs,n}$ 
    \STATE \textbf{iv:} compute embeddings $\mathbf{z}_{hs,n} = g_{\theta}(\mathbf{T}_{hs,n},\mathbf{P}_{hs,n})$
    \STATE \textbf{v:} stitch model embeddings $\mathbf{z}$ together from chunks $\mathbf{z}_{hs,n}$
\end{algorithmic}
\vspace{4pt}
\end{algorithm}
To compare different models in embedding space, we aggregate the sequences of token embeddings. To that end, we understand the token sequence of one model to form a surface in embedding space and choose to represent that surface by its center of gravity. That is, we take the mean of all tokens along the embedding dimension as $\mathbf{\bar{z}} = \frac{1}{N} \sum_{n=1}^N(\mathbf{z}_n)$. That results in one vector in embedding space per model. Of course, one could use other aggregation methods with \ourmethod.

\textbf{Sampling Models with Few Prompt Examples.}
Sampling models from \ourmethod promises to transfer knowledge from existing populations to new models with different architectures.
Given pretrained encoders $g_{\theta}$ and decoder $h_\psi$, the challenge is to identify the distribution $\mathcal{P}$ in latent space which contains the targeted properties.
To approximate that distribution, previous work used a large number of well-trained models \citep{peeblesLearningLearnGenerative2022,schurholtHyperRepresentationsGenerativeModels2022}.
However, increasing the size of the sampled models makes generating a large number of high-performance models exceedingly expensive.
Instead of using expensive high-performance models to model $\mathcal{P}$ directly, we propose to find a rough estimate of $\mathcal{P}$, sample broadly, and refine $\mathcal{P}$ using the signal from the sampled models.
Using $E$ prompt examples $\mathbf{W}^e$ we compute the token sequence $\mathbf{T}^e$ and corresponding embedding sequence $\mathbf{z}^e = g_{\theta}(\mathbf{T}^e, \mathbf{P})$. Following previous work, we model the distribution $\mathcal{P}$ with a Kernel Density Estimation (KDE) per token as $\mathcal{P}_{e \in E}(\mathbf{z}_n^e)$ ~\citep{schurholtHyperRepresentationsGenerativeModels2022}.
We then draw $k$ new token samples as:
\begin{equation}
    \mathbf{z}_n^k \sim \mathcal{P}_{e \in E}(\mathbf{z}_n^e).
\end{equation}
We reconstruct the sampled embeddings to weight tokens $\mathbf{T}^k = h_{\psi}(\mathbf{z}^k,\mathbf{P})$ and then weights $\mathbf{W}^k$. 
Sampling tokens can be done cheaply, decoding and evaluating the weights using some performance metric involves only forward passes and is likewise cheap. 
Therefore, one can draw a large amount of samples and keep only the top $m$ models, according to the performance metric. 
We call this method \emph{subsampling}.
The process can be refined iteratively, by re-using the embeddings $\mathbf{z}^k$ of the best models as new prompt examples, to adjust the sampling distribution to best fit the needs of the performance metric. 
We call this sampling method \emph{bootstrapped}.
By only requiring a rough version of $\mathcal{P}$ and refining with the target signal, our sampling strategy reduces requirements on prompt examples such that only very few and slightly trained prompt examples are necessary. The overall sampling method is outlined in Algorithm~\ref{alg:scalable_hyper_reps:generating}. It makes use of \textit{model alignment}, \textit{haloing}, and \textit{batch-norm conditioning} which are detailed below.
In addition to the compute efficiency, these sampling methods learn the distribution of targeted models in embedding space. 
Further, they are not bound to the distribution of prompt examples, but instead they can find the distribution that best satisfies the target performance metric, independent of the prompt examples.
\begin{algorithm}[]
   \caption{Sampling models with \ourmethod }
   \label{alg:scalable_hyper_reps:generating}
\begin{algorithmic}
    \STATE {\bfseries Input:} model prompt examples $\mathbf{W}^e$
    \STATE \textbf{i:} tokenize prompt examples: tokens $\mathbf{T}^e$, positions $\mathbf{P}^e$
    \STATE \textbf{ii:} embed prompt examples $\mathbf{z}^e$ following Alg. \ref{alg:scalable_hyper_reps:embeddings}
    \FOR{$i_{boot} = 1$ {\bfseries to} \textit{bootstrap iterations}}
        \STATE \textbf{iii:} draw $k$ samples $\mathbf{z}_n^k \sim \mathcal{P}_{e \in E}(\mathbf{z}_n^e)$
        \STATE \textbf{iv:} decode to tokens $\mathbf{T^k} = h_{\psi}( \mathbf{z}^k )$
        \STATE \textbf{v:} apply batch-norm conditioning
        \STATE \textbf{vi:} compute target metric and keep best $m$ models
        \IF{\textit{bootstrap iterations}$\; > 1$}
            \STATE \textbf{vii:} $\mathbf{z}^e = \mathbf{z}^k for \; k \in m$ 
        \ENDIF
    \ENDFOR
\end{algorithmic}
\end{algorithm}
%
 
Growing sample model size poses several additional challenges, three of which we address with the following methods. We evaluate these methods in Appendix \ref{sec:scalable_hyper_reps:ablation}.\looseness-1

\newpage
\paragraph*{Model Alignment.}
Symmetries in the weight space of NN complicate representation learning of the weights. The number of symmetries grows fast with model size \citep{bishopPatternRecognitionMachine2006}. 
To make representation learning easier, we reduced all training models to a unique, canonical basis of a reference model.
With reference model $A$ we align model $B$ by finding the permutation $\pi = argmin_{\pi} \|\text{vec}(\Theta{(A)}) - \text{vec}(\Theta(B))\|^2$, where $\Theta(A)$ are the parameters of model $A$ \citep{ainsworthGitReBasinMerging2022}. 
We fix the same reference model across all dataset splits and use the last epoch of each model to determine the permutation for that model.

\paragraph*{Haloing.}
The sequential decomposition of \ourmethod decouples the pretraining sequence length from downstream task sequence lengths.
Since the memory load at inference is considerably lower, the sequences at inference can be longer. 
However, full model sequences may still not fit in memory and may have to be processed in slices.
To ensure consistency between the slices, we add context around the content windows. With added context halo before and after the content window, we get
$\mathbf{T}_{hs,n}=\mathbf{T}_{n-h,..,n,...,n+ws,n+ws+h}$. 
Similar to approaches in computer vision \citep{vaswaniScalingLocalSelfAttention2021}, this context halo is added for the pass through encoder and decoder, but disregarded after.

\paragraph*{Batch-Norm Conditioning.}
In most current NN models, some parameters like batch-norm weights are updated during forward passes instead of with gradients. 
Since that makes them structurally different, we exclude these parameters from representation learning and sampling with \ourmethod.
Nonetheless, these parameters need to be instantiated for sampled models to work well. 
For model sampling methods, we therefore propose to condition batch-norm parameters by performing a few forward passes with some target data. 
Importantly, this process does not update the learned weights of the model. 
It serves to align the batch norm statistics with the model's weights.

\newpage

%
%
\section{Training \ourmethod}
We pretrain \ourmethod following Alg. \ref{alg:scalable_hyper_reps:pretraining} on several populations of trained NN models, from the model zoo dataset \citep{schurholtModelZoosDataset2022}. 
We use zoos of small models to compare with previous work, as well as zoos containing larger ResNet-18 models. 
All zoos are split into training, validation, and test splits $70:15:15$.
\vspace{-10pt}
\begin{itemize}
\item \textbf{Smaller CNN zoos.}
The MNIST and SVHN zoos contain LeNet-style models with 3 convolution and 2 dense layers and only $\sim 2.5k$ parameters. 
The slightly larger CIFAR-10 and STL-10 zoos use the same architecture with wider layers and $\sim 12k$ parameters.
\vspace{-6pt}
\item \textbf{Larger ResNet zoos.}
We also use the CIFAR-10, CIFAR-100, and Tiny-Imagenet zoos containing ResNet-18 models \citep{schurholtModelZoosDataset2022} with $\sim 12M$ parameters to evaluate scalability to large models. 
\end{itemize}

\paragraph*{Pretraining.}
We train \ourmethod using Alg. \ref{alg:scalable_hyper_reps:pretraining}.
As augmentations, we use noise and permutation. 
The permutation is computed relative to the aligned model. For contrastive learning, the aligned model serves as one view, and a permuted version as the second view. \looseness-1

\paragraph*{Implementation Details.} 
To maintain diversity within each batch, we select only a single window from each model. Loading, preprocessing, and augmenting the entire sample only to use approximately 1\% of it is infeasible. To address this, we leverage FFCV \citep{leclercFFCVAcceleratingTraining2023} to compile datasets consisting of sliced and permuted windows of models. Each model is super-sampled for approximately full coverage within the training set, considering the ratio of window length to sequence length.
For the ResNet zoos, we include 140 models per zoo, a number that remains manageable in terms of memory and storage. 
We train for 50 epochs using a OneCycle learning rate scheduler \citep{smithSuperConvergenceVeryFast2018}. Seeds are recorded to ensure reproducibility.
We build \ourmethod in PyTorch \citep{paszkePyTorchImperativeStyle2019}, using automatic mixed precision and flash attention \citep{daoFlashAttentionFastMemoryEfficient2022} to enhance performance. We use ray.tune \citep{liawTuneResearchPlatform2018} for hyperparameter optimization. 

\newpage
\section{Embedding Analysis}
In this section, we analyze the embeddings of \ourmethod and compare to the weight-analysis methods \textit{WeightWatcher} (WW) \citep{martinPredictingTrendsQuality2021}. 
We focus on three aspects: 
i) global relation between accuracy and embeddings; 
ii) the trend of embeddings over layer index, as in  \citep{martinPredictingTrendsQuality2021}; and
iii) the identification of training phases as in \citep{martinTraditionalHeavyTailedSelf2019,martinImplicitSelfregularizationDeep2021}.

\begin{figure}[ht]
\centering
\includegraphics[trim=4mm 0mm 1.5mm 1.5mm, width=0.8\linewidth]{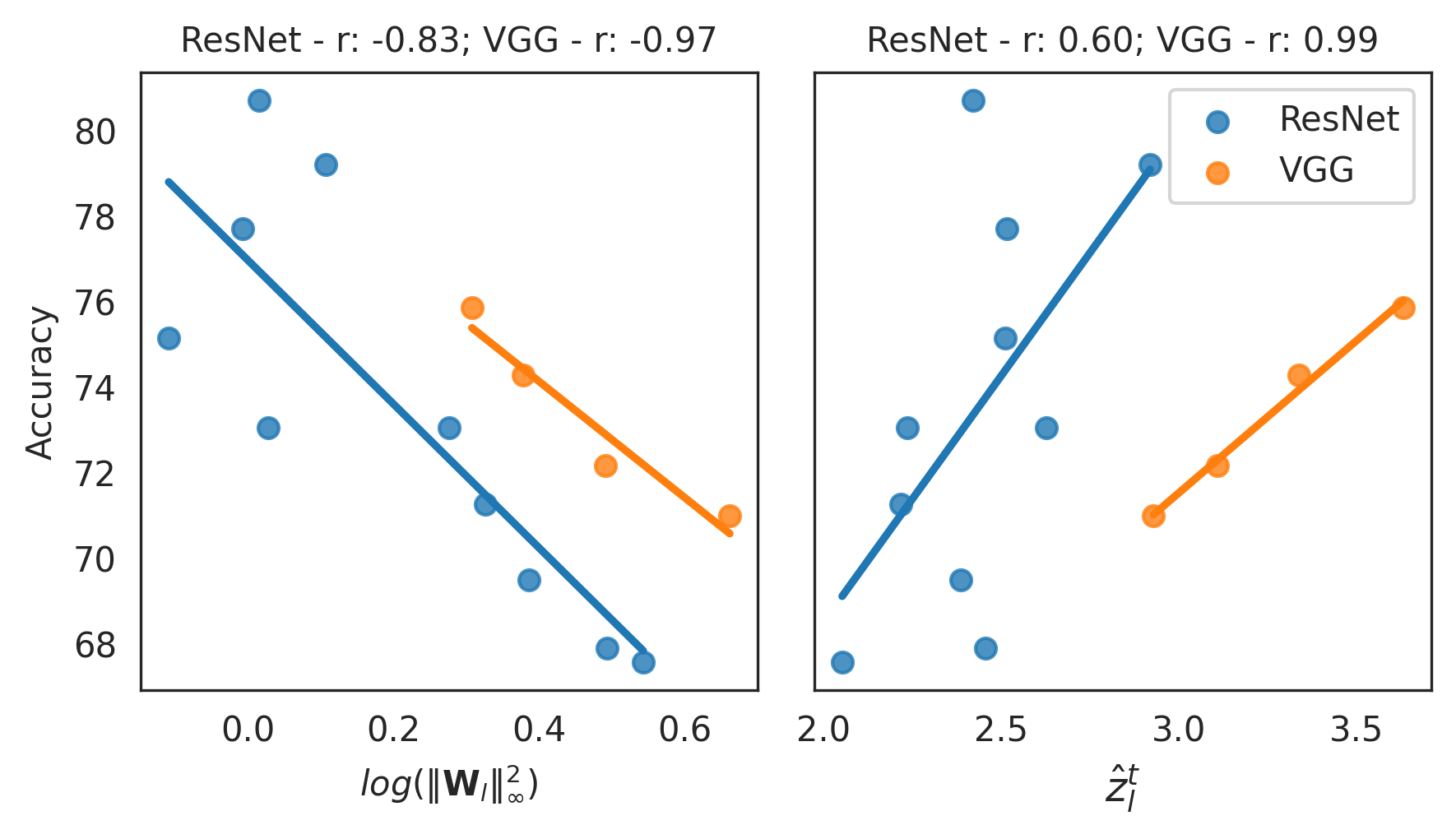}
\captionof{figure}[Accuracy over Weight Metrics for WeightWatcher and \ourmethod]{Comparison between WeightWatcher features (left) and \ourmethod (right). Accuracy over model features for ResNets and VGGs from pytorchcv of different sizes. Although \ourmethod is pretrained in a self-supervised fashion, it preserves the linear relation of a globally-aggregated embedding to model accuracy.
}
\label{fig:scalable_hyper_reps:ww_comparison_accuracy}    
\end{figure}

To analyze weights, we focus on two WW metrics which in previous work reveal model performance as well as internal model composition (\textit{correlation flow}); the log spectral norm $\log(\| \mathbf{W}\|^2_{\infty})$ and weighted $\alpha$, the coefficient of the power law fitted to the empirical spectral density \citep{martinPredictingTrendsQuality2021}. 
These two metrics describe different aspects of the eigenvalue distribution. 
To get a similar signal on the internal dependency of weight matrices, we compute per-layer scalars $\hat{z}_l$ as the spread of the tokens of one layer in hyper-representation space, i.e., their standard deviation: 
\begin{align}
    \hat{z}_l &=  std_t(\textbf{z}^t_m) \\
    \textbf{z}^t_m &= g(\textbf{W}^t_m),
\end{align}
where $g$ is the hyper-rep encoder, $\textbf{z}^t_m$ are the stacked tokens $t$ of layer $m$, and $\textbf{W}^t_m$ is the weight-slice $t$ of layer $m$.\\
%
\begin{figure}[ht]
\centering
\includegraphics[trim=4mm 0mm 1.5mm 1.5mm, width=0.8\linewidth]{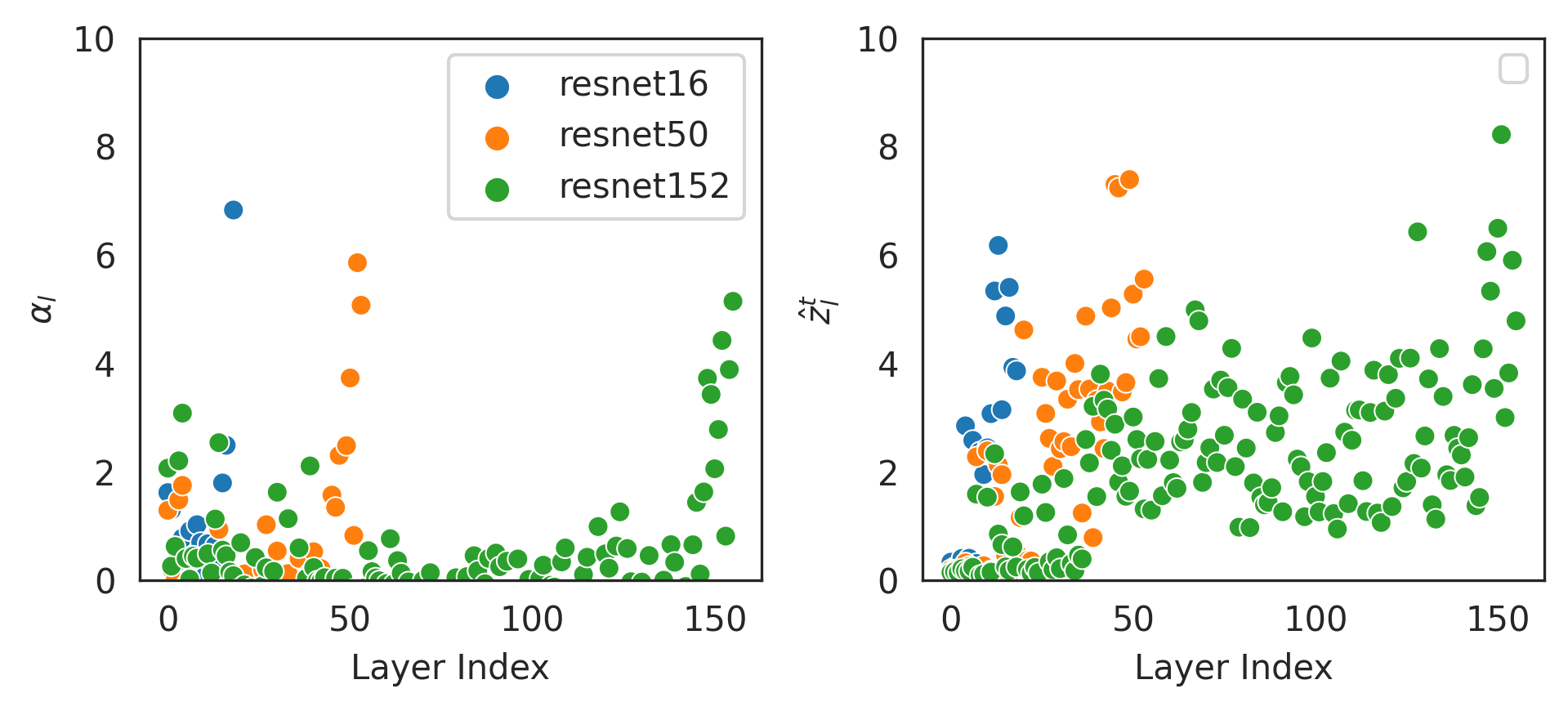}
\captionof{figure}[Weight Metrics over layers. Comparison betwen WeightWatcher and \ourmethod]{Comparison between WeightWatcher (WW) features (left) and \ourmethod (right). Features over layer index for ResNets from pytorchcv of different sizes.  
}
\label{fig:scalable_hyper_reps:ww_comparison_layers_resnet}
\end{figure}
To compare WW metrics to \ourmethod,
we pretrain \ourmethod on a Tiny-Imagenet ResNet-18 zoo and compute the two metrics on ResNets and VGGs of different sizes trained on ImageNet from pytorchcv \citep{semeryOsmrImgclsmob2024}.
On both ResNets in Figure \ref{fig:scalable_hyper_reps:ww_comparison_layers_resnet}, \ref{fig:scalable_hyper_reps:ww_comparison_layers_resnets_2x2} and VGGs in Figure \ref{fig:scalable_hyper_reps:ww_comparison_layers_vgg}, the WW metrics and our embeddings show similar global trends. 
On ResNets, our embeddings and WW have low values at early layers and a sharp increase at the end. 
However, our embeddings add an additional step for intermediate layers, which may indicate that \ourmethod is sensitive to a higher degree of variation in these layers which previous work found by comparing activations \citep{kornblithSimilarityNeuralNetwork2019}.
In a second experiment, we aggregate the layer-wise embeddings $\hat{z}_l$ to evaluate relations to model accuracy in Figures \ref{fig:scalable_hyper_reps:ww_comparison_accuracy}, \ref{fig:scalable_hyper_reps:ww_comparison_accuracy_2x2} and \ref{fig:scalable_hyper_reps:ww_comparison_accuracy_our_zoo}, similar to previous work \citep{martinPredictingTrendsQuality2021}. 
On models from pytorchcv and the Tiny-ImagNet model zoo from \citep{schurholtModelZoosDataset2022}, the WW features and \ourmethod embeddings both show strong correlations to model accuracy. 
However, while the WW metrics are negatively correlated to accuracy, our embeddings are positively correlated to accuracy. 
The reason for that may lie in the additional 'step' in Figure \ref{fig:scalable_hyper_reps:ww_comparison_layers_resnet}. 
That is, larger models with more layers generally have higher performance. As Figure \ref{fig:scalable_hyper_reps:ww_comparison_layers_resnet} shows, more layers add very small values reducing the global average for WW metrics. For our embeddings, deeper models have more layers with higher $\hat{z}_l$ values, due to the afore-mentioned step. This increases the global model average with growing model size.
Lastly, we compare the eigenvalue spectrum to embeddings. Previous work identified distinct shapes at different training phases or with varying training hyperparameters  \citep{martinTraditionalHeavyTailedSelf2019,martinImplicitSelfregularizationDeep2021}. While we can replicate the distributions of the eigenvalues, the distributions of our embeddings only show the change from early phases of training to the heavy-tailed distribution; see Figure \ref{fig:scalable_hyper_reps:ww_comparison_phases}. 
In summary, our embedding analysis indicates that \ourmethod represents several aspects of model quality (globally and on a layer level) that have been established previously.
%
%

\section{Empirical Performance}
In this section, we describe the general performance \ourmethod.
\subsection{Predicting Model Properties}
\label{sec:scalable_hyper_reps:property_prediction}
We evaluate \ourmethod for discriminative downstream tasks as a proxy for encoded model qualities. Specifically, we investigate whether \ourmethod matches the predictive performance of \textit{hyper-representations} on small CNN models (Table \ref{tab:scalable_hyper_reps:discr_small_zoos}) and whether similar performance can be achieved on ResNet-18 models (Table \ref{tab:scalable_hyper_reps:discr_resnet_zoos}). 
To that end, we compute model embeddings $\mathbf{\bar{z}}$ as outlined in Alg. \ref{alg:scalable_hyper_reps:embeddings}, and we compare against flattened weights $W$ and weight statistics $s(W)$.
Following the experimental setup of \citep{eilertsenClassifyingClassifierDissecting2020,unterthinerPredictingNeuralNetwork2020,schurholtSelfSupervisedRepresentationLearning2021}, we compute embeddings using the three methods and linear probe for test accuracy (\texttt{Acc}), epoch (\texttt{Ep}), and generalization gap (\texttt{Ggap}). We again use trained models from the modelzoo repository \citep{schurholtModelZoosDataset2022}, with the same train, test, val splits as above. 

\begin{table}[h]
\vspace{-4mm}
\caption[Property Prediction on small CNN Model Zoos]{Property prediction on populations of small CNNs used in previous work \citep{schurholtSelfSupervisedRepresentationLearning2021}. We report the regression $R^2$ on the test set prediction test accuracy \texttt{Acc.}, epoch \texttt{Ep.} and generalization gap \texttt{Ggap} for linear probing with model weights $W$, model weights statistics $s(W)$ or \ourmethod embeddings as inputs.\looseness-1}
\label{tab:scalable_hyper_reps:discr_small_zoos}
{
\centering
\sc
\small
\setlength{\tabcolsep}{6pt}
\begin{tabularx}{1.0\linewidth}{lccccccccccc}
\toprule
      & \multicolumn{3}{c}{MNIST}                                                      &  & \multicolumn{3}{c}{SVHN}                                                       &  & \multicolumn{3}{c}{CIFAR-10 (CNN)}                                             \\ 
      \cmidrule(r){2-4} \cmidrule(lr){6-8} \cmidrule(lr){10-12}
      & \multicolumn{1}{c}{W} & \multicolumn{1}{c}{s(W)} & \multicolumn{1}{c}{\ourmethod} &  & \multicolumn{1}{c}{W} & \multicolumn{1}{c}{s(W)} & \multicolumn{1}{c}{\ourmethod} &  & \multicolumn{1}{c}{W} & \multicolumn{1}{c}{s(W)} & \multicolumn{1}{c}{\ourmethod}  \\ 
      \cmidrule(r){1-1} \cmidrule(r){2-4} \cmidrule(lr){6-8} \cmidrule(lr){10-12}
Acc.   & 0.965                 & \textbf{0.987}           & 0.978                       &  & 0.910                 & 0.985                    & \textbf{0.991}              &  & -7.580                & \textbf{0.965}           & 0.885                        \\
Ep. & 0.953                 & \textbf{0.974}           & 0.958                       &  & 0.833                 & \textbf{0.953}           & 0.930                       &  & 0.636                 & \textbf{0.923}           & 0.771                       \\
Ggap  & 0.246                 & 0.393                    & \textbf{0.402}              &  & 0.479                 & 0.711                    & \textbf{0.760}              &  & 0.324                 & \textbf{0.909}           & 0.772                       \\ 
\bottomrule
\end{tabularx}
}
\vspace{-2mm}
\end{table} 

\paragraph*{\ourmethod matches baselines on small models.}
The results of linear probing on small CNNs in Tables \ref{tab:scalable_hyper_reps:discr_small_zoos} and \ref{tab:scalable_hyper_reps:discr_small_zoos_full} confirm the performance of $W$ (low) and $s(W)$ (very high) of previous work. 
\ourmethod embeddings show comparably high performance to the $s(W)$ and previous \textit{hyper-representations}. 
Additional experiments in Appendix \ref{app:scalable_hyper_reps:discr_comparison_previous_work} compare to previous work and confirm these findings. Sequential decomposition and representation learning as well as using the center of gravity does not significantly reduce the information contained in \ourmethod embeddings. 
%
\begin{table}[h]
\vspace{-2mm}
\caption[Property Prediction on ResNet-18 Model Zoos]{
Property prediction on ResNet-18 model zoos of \citep{schurholtModelZoosDataset2022}. We report the regression $R^2$ on the test set prediction test accuracy \texttt{Acc.}, epoch \texttt{Ep.} and generalization gap \texttt{Ggap} for linear probing with model weights statistics $s(W)$ or \ourmethod embeddings as inputs.\looseness-1
}
\label{tab:scalable_hyper_reps:discr_resnet_zoos}
\centering
\sc
\small
\setlength{\tabcolsep}{8pt}
\begin{tabularx}{0.8\linewidth}{lcccccccc}
\toprule
      & \multicolumn{2}{c}{CIFAR-10} &  & \multicolumn{2}{c}{CIFAR-100} &  & \multicolumn{2}{c}{Tiny-Imagenet} \\
      \cmidrule(){2-3} \cmidrule(){5-6} \cmidrule(){8-9} 
      &  s(W)     & \ourmethod    &  & s(W)     & \ourmethod    &      & s(W)      & \ourmethod      \\ 
\cmidrule(r){1-1} \cmidrule(){2-3} \cmidrule(){5-6} \cmidrule(){8-9} 
Acc.      & 0.880    & 0.879      &       & 0.923    & 0.922      &        &  0.802    &       0.795       \\
Ep.       & 0.999    & 0.999      &       & 0.999    & 0.992      &        &  0.999    &       0.980       \\
Ggap      & 0.490    & 0.512      &       & 0.882    & 0.879      &        &  0.704    &       0.699 
\\
\bottomrule
\end{tabularx}
\end{table}

\newpage
\paragraph*{\ourmethod performance prediction scales to ResNets.} 
Both $s(W)$ and \ourmethod embeddings show similarly high performance on populations of ResNet-18s; see Table \ref{tab:scalable_hyper_reps:discr_resnet_zoos}. 
On ResNet-18s, using the full weights $W$ for linear probing is infeasible due to the size of the flattened weights.
\ourmethod matches the high performance of $s(W)$. 
The results show that sequential \textit{hyper-representations} are capable of scaling to ResNet-18 models. 
Further, the aggregation even of long sequences (ca. 50k tokens) embedded in \ourmethod preserves meaningful information on model performance, which indicates the feasibility of applications like model diagnostics or targeted sampling.~\looseness-1
%
%
%
\subsection{Generating Models}
\label{sec:scalable_hyper_reps:generating_models}
\begin{figure}[t!]
\centering
\includegraphics[trim=0mm 0mm 0mm 0mm, width=1.0\linewidth]{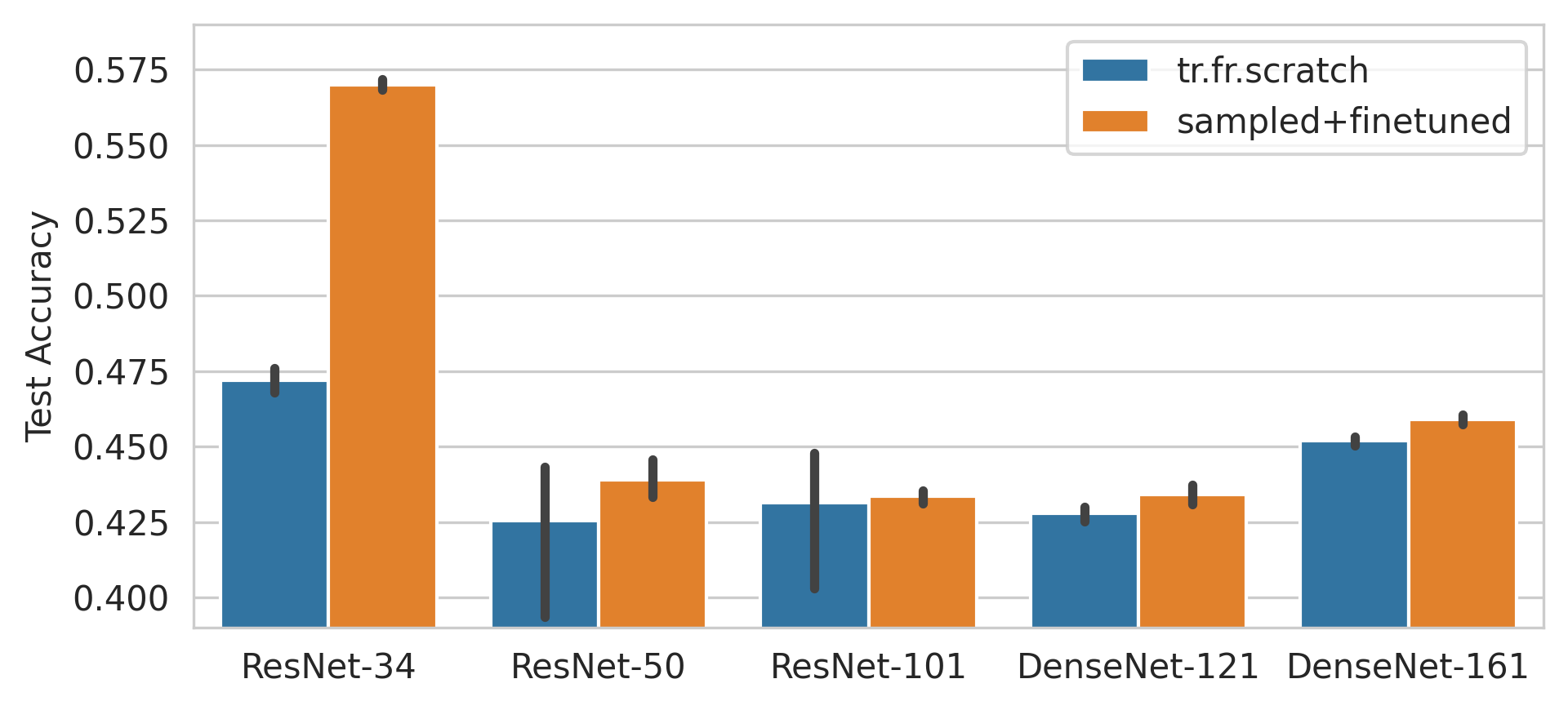}
\vspace{-2mm}
\captionof{figure}{Performance of Models Sampled with SANE for New Architectures}{Comparison between sampled models and random initialization trained for 5 epochs Tiny-Imagenet. Different architectures are sampled from \ourmethod pretrained on a ResNet-18 CIFAR-100 zoo. Although both models and tasks are changed, sampled models perform better.}
\label{fig:scalable_hyper_reps:sampling_new_arch_task}    
\end{figure}

We evaluate \ourmethod for the generative downstream tasks i.e., for sampling model weights. 
We generate weights following Alg. \ref{alg:scalable_hyper_reps:generating} and test them in fine-tuning, transfer learning, and how they generalize to new tasks and architectures. 
In the following paragraphs, we begin with experiments on small CNN models from the modelzoo repository to compare with previous work (Tables \ref{tab:scalable_hyper_reps:generative_cnn_zoos_in_domain}, \ref{tab:scalable_hyper_reps:generative_cnn_zoos_transfer}). 
Subsequently, we evaluate \ourmethod for sampling ResNet-18 models for finetuning and transfer learning (Tables \ref{tab:scalable_hyper_reps:generative_resnets_zoos_in_domain}, \ref{tab:scalable_hyper_reps:generative_resnets_zoos_transfer}). 
Lastly, we evaluate sampling for new tasks and new architectures using only few prompt examples (Figure \ref{fig:scalable_hyper_reps:sampling_new_arch_task} and Tables \ref{tab:scalable_hyper_reps:generative_resnets_zoos_fewshot_transfer}, \ref{tab:scalable_hyper_reps:generative_resnets_zoos_fewshot_resnet34}, \ref{tab:scalable_hyper_reps:generative_resnets_zoos_fewshot_ti_resnet34}, \ref{tab:scalable_hyper_reps:generative_resnets_zoos_fewshot_resnet34_one_prompt_example}).

We pretrain \ourmethod on models from the first half of the training epochs with Alg. \ref{alg:scalable_hyper_reps:pretraining}, and keep the remaining epochs (26-50) as holdout to compare against, following the experimental setup of \citep{schurholtHyperRepresentationsGenerativeModels2022}. 
We sample using Alg. \ref{alg:scalable_hyper_reps:generating} and use models from the last epoch in the pretraining set (epoch 25) as prompt examples. 
We denote subsampling with \ourmethod$_{SUB}$ and iteratively updating the distribution $\mathcal{P}$ as \ourmethod$_{BOOT}$. To evaluate the impact of the sampling method, we also combine \ourmethod with the $KDE30$ sampling approach that uses high-quality prompt examples \citep{schurholtHyperRepresentationsGenerativeModels2022}. We further evaluate sampling without prompt examples by bootstrapping off of a Gaussian prior $\mathcal{P}$, denoted as \ourmethod$_{GAUSS}$. We compare against training from scratch, as well as fine-tuning from the prompt examples.


\begin{table}[t]
\vspace{-2mm}
\caption[Performance of Sampled Models with SANE on Small CNN Model Zoos]{
Model generation on CNN model populations fine-tuned on the same task. We compare training from scratch with $S_{KDE30}$ from \citep{schurholtHyperRepresentationsGenerativeModels2022}, \ourmethod combined with the $KDE30$ sampling method, and our \ourmethod subsampled. Each of the sampled populations is fine-tuned over 25 epochs.
}\label{tab:scalable_hyper_reps:generative_cnn_zoos_in_domain}
\centering
\small
\setlength{\tabcolsep}{8pt}
\begin{tabularx}{0.85\linewidth}{clcccc}
\toprule
Ep.                  & \multicolumn{1}{c}{Method} & MNIST               & SVHN               & CIFAR-10           & STL                \\
\cmidrule(r){1-1} \cmidrule(lr){2-2} \cmidrule(lr){3-3} \cmidrule(lr){4-4} \cmidrule(lr){5-5} \cmidrule(l){6-6}
\multirow{5}{*}{0}     &  tr. fr. scratch                  & $\sim$10 /\%        & $\sim$10 /\%       & $\sim$10 /\%       & $\sim$10 /\%       \\
                       & $S_{KDE30}$                & 68.6$\pm$6.7           & 54.5$\pm$5.9          & \textit{n/a}                & \textit{n/a}                \\
                       & \ourmethod$_{KDE30}$                    & 84.8$\pm$0.8           & 70.7$\pm$1.4          & 56.3$\pm$0.5          & 39.2$\pm$0.8          \\
                       & \ourmethod$_{SUB}$                   & \textbf{86.7$\pm$0.8}  & \textbf{72.3$\pm$1.6} & \textbf{57.9$\pm$0.2} & \textbf{43.5$\pm$1.0} \\
                      & \ourmethod$_{GAUSS}$                 & 20.8$\pm$0.1                & 21.6$\pm$0.5                & 19.3$\pm$0.2               & 17.5$\pm$1.5               \\
\cmidrule(r){1-1} \cmidrule(lr){2-2} \cmidrule(lr){3-3} \cmidrule(lr){4-4} \cmidrule(lr){5-5} \cmidrule(l){6-6}
\multirow{5}{*}{1}     &  tr. fr. scratch                  & 20.6$\pm$1.6           & 19.4$\pm$0.6          & 37.2$\pm$1.4          & 21.3$\pm$1.6          \\
                       & $S_{KDE30}$                & 83.7$\pm$1.3           & 69.9$\pm$1.6          & \textit{n/a}                & \textit{n/a}                \\
                       & \ourmethod$_{KDE30}$                    & 85.5$\pm$0.8           & 71.3$\pm$1.4          & 58.2$\pm$0.2          & 43.5$\pm$0.7          \\
                       & \ourmethod$_{SUB}$                   & \textbf{87.5$\pm$0.6}  & \textbf{73.3$\pm$1.4} & \textbf{59.1$\pm$0.3} & \textbf{44.3$\pm$1.0} \\
                        & \ourmethod$_{GAUSS}$                 & 61.3$\pm$3.1                & 24.1$\pm$4.4                & 27.2$\pm$0.3               & 22.4$\pm$1.0               \\
\cmidrule(r){1-1} \cmidrule(lr){2-2} \cmidrule(lr){3-3} \cmidrule(lr){4-4} \cmidrule(lr){5-5} \cmidrule(l){6-6}
\multirow{5}{*}{5}     &  tr. fr. scratch                  & 36.7$\pm$5.2           & 23.5$\pm$4.7          & 48.5$\pm$1.0          & 31.6$\pm$4.2          \\
                       & $S_{KDE30}$                & \textbf{92.4$\pm$0.7} & 57.3$\pm$12.4         & \textit{n/a}                & \textit{n/a}                \\
                       & \ourmethod$_{KDE30}$                    & 87.5$\pm$0.7           & 72.2$\pm$1.2          & 58.8$\pm$0.4          & 45.2$\pm$0.6          \\
                       & \ourmethod$_{SUB}$                   & 89.0$\pm$0.4           & \textbf{73.6$\pm$1.5} & \textbf{59.6$\pm$0.3} & \textbf{45.3$\pm$0.9} \\
                       & \ourmethod$_{GAUSS}$                 & 83.4$\pm$0.8                & 35.6$\pm$8.9                & 43.3$\pm$0.3               & 34.2$\pm$0.7               \\
\cmidrule(r){1-1} \cmidrule(lr){2-2} \cmidrule(lr){3-3} \cmidrule(lr){4-4} \cmidrule(lr){5-5} \cmidrule(l){6-6}
\multirow{5}{*}{25}    &  tr. fr. scratch                  & 83.3$\pm$2.6           & 66.7$\pm$8.5          & 57.2$\pm$0.8          & 44.0$\pm$1.0          \\
                       & $S_{KDE30}$                & \textbf{93.0$\pm$0.7}           & 74.2$\pm$1.4          &       \textit{n/a}             &    \textit{n/a}                \\
                       & \ourmethod$_{KDE30}$                    & 92.0$\pm$0.3           & 74.7$\pm$0.8          & 60.2$\pm$0.6          & \textbf{48.4$\pm$0.5} \\
                       & \ourmethod$_{SUB}$                   & 92.3$\pm$0.4           & \textbf{75.1$\pm$1.0} & \textbf{61.2$\pm$0.1} & 48.0$\pm$0.4          \\
                       & \ourmethod$_{GAUSS}$                 & \textbf{94.2$\pm$0.4}       & 54.2$\pm$17.6               & 52.2$\pm$0.6               & 43.5$\pm$0.5               \\
\cmidrule(r){1-1} \cmidrule(lr){2-2} \cmidrule(lr){3-3} \cmidrule(lr){4-4} \cmidrule(lr){5-5} \cmidrule(l){6-6}
\multicolumn{1}{l}{50} &  tr. fr. scratch                  & 91.1$\pm$2.6           & 70.7$\pm$8.8          & 61.5$\pm$0.7          & 47.4$\pm$0.9      \\
\bottomrule
\end{tabularx}
\end{table} 

\paragraph*{Sampling High-Performing CNNs Zero-Shot.}
We begin with finetuning and transfer learning experiments on small CNNs from the modelzoo dataset to validate that the sequential decomposition for pretraining and sampling does not hurt performance.
The results of these experiments show dramatically improved performance zero-shot for fine-tuning and transfer learning over previous \textit{hyper-representations}; see Tables \ref{tab:scalable_hyper_reps:generative_cnn_zoos_in_domain} and \ref{tab:scalable_hyper_reps:generative_cnn_zoos_transfer}. 
At epoch 0, \ourmethod improves over previous \textit{hyper-representations} $S_{KDE30}$ by almost 20\%.
The effect becomes smaller during fine-tuning.
Nonetheless, \ourmethod consistently outperforms training from scratch with a higher epoch budget, often by several percentage points. 
This demonstrates on small CNNs that sequential pretraining and sampling of \ourmethod improves performance, particularly zero shot. 
This indicates the potential for scenarios with little labelled data.

\begin{table}[h]
\vspace{-2mm}
\caption[Performance of Sampled Models with SANE on ResNet-18 Model Zoos]{
Model generation on ResNet-18 model populations fine-tuned on the same task. We compare sampled models at different epochs with models trained from scratch.
}
\label{tab:scalable_hyper_reps:generative_resnets_zoos_in_domain}
\centering
\small
\setlength{\tabcolsep}{8pt}
\begin{tabularx}{0.8\linewidth}{clccc}
\toprule
Epoch               & \multicolumn{1}{c}{Method} & CIFAR-10           & CIFAR-100          & Tiny-Imagenet      \\
\cmidrule(r){1-1} \cmidrule(lr){2-2} \cmidrule(lr){3-3} \cmidrule(lr){4-4} \cmidrule(l){5-5} 
\multirow{2}{*}{0}  & tr. fr. scratch                  & $\sim$10 /\%       & $\sim$1 /\%        & $\sim$0.5 /\%      \\
                    & \ourmethod$_{KDE30}$                    & 64.8$\pm$2.0          & 19.8$\pm$2.5          & 8.4$\pm$0.9           \\
                    & \ourmethod$_{SUB}$                   & 68.1$\pm$0.7          & 19.8$\pm$1.3          & 11.1$\pm$0.5          \\
                    & \ourmethod$_{BOOT}$                  & \textbf{68.6$\pm$1.2} & \textbf{20.4$\pm$1.3} & \textbf{11.7$\pm$0.5} \\
\cmidrule(r){1-1} \cmidrule(lr){2-2} \cmidrule(lr){3-3} \cmidrule(lr){4-4} \cmidrule(l){5-5} 
\multirow{2}{*}{1}  & tr. fr. scratch                  & 43.7$\pm$1.3          & 17.5$\pm$0.7          & 13.8$\pm$0.8          \\
                    & \ourmethod$_{KDE30}$                    & 82.4$\pm$0.9          & 59.0$\pm$1.3          & 46.7$\pm$0.8          \\
                    & \ourmethod$_{SUB}$                   & \textbf{83.6$\pm$1.5} & \textbf{60.8$\pm$0.8} & \textbf{47.4$\pm$1.0}          \\
                    & \ourmethod$_{BOOT}$                  & 82.8$\pm$1.4          & 60.2$\pm$0.5          & 47.2$\pm$0.8 \\
\cmidrule(r){1-1} \cmidrule(lr){2-2} \cmidrule(lr){3-3} \cmidrule(lr){4-4} \cmidrule(l){5-5} 
\multirow{2}{*}{5}  & tr. fr. scratch                  & 64.4$\pm$2.9          & 36.5$\pm$2.0          & 31.1$\pm$1.6          \\
                    & \ourmethod$_{KDE30}$                    & \textbf{85.9$\pm$0.6} & 56.2$\pm$1.7          & 45.6$\pm$1.4          \\
                    & \ourmethod$_{SUB}$                   & 85.4$\pm$1.3          &\textbf{ 56.7$\pm$1.6  }        & 45.7$\pm$0.8          \\
                    & \ourmethod$_{BOOT}$                  & 85.4$\pm$0.7          & 56.4$\pm$1.2          & \textbf{49.1$\pm$1.7} \\
\cmidrule(r){1-1} \cmidrule(lr){2-2} \cmidrule(lr){3-3} \cmidrule(lr){4-4} \cmidrule(l){5-5} 
\multirow{2}{*}{10} & tr. fr. scratch                  &  76.5$\pm$2.7           & 49.0$\pm$2.0        &  39.9$\pm$2.2       \\
                    & \ourmethod$_{KDE30}$                    & 91.4$\pm$0.1          & \textbf{72.9$\pm$0.2} & \textbf{64.2$\pm$0.3} \\
                    & \ourmethod$_{SUB}$                   & \textbf{91.6$\pm$0.2}          & \textbf{72.9$\pm$0.1} & 64.0$\pm$0.2          \\
                    & \ourmethod$_{BOOT}$                  & 91.6$\pm$0.2          & 72.8$\pm$0.1          & 64.1$\pm$0.2          \\
\cmidrule(r){1-1} \cmidrule(lr){2-2} \cmidrule(lr){3-3} \cmidrule(lr){4-4} \cmidrule(l){5-5} 
25                  & tr. fr. scratch                  & 85.5$\pm$1.5          & 56.5$\pm$2.0          & 43.3$\pm$1.9          \\
50                  & tr. fr. scratch                  & 92.14$\pm$0.2         & 70.7$\pm$0.4          & 57.3$\pm$0.6          \\
60                  & tr. fr. scratch                  &   \textit{n/a}                 & 74.2$\pm$0.3          & 63.9$\pm$0.5         
\\
\bottomrule
\end{tabularx}
\vspace{-4mm}
\end{table}

\newpage
\paragraph*{\ourmethod Sequential Sampling Scales to ResNets.}
To evaluate how well sampling with \ourmethod scales to larger models, we continue with experiments on ResNet-18s. 
The results of these experiments Tables \ref{tab:scalable_hyper_reps:generative_resnets_zoos_in_domain} and \ref{tab:scalable_hyper_reps:generative_resnets_zoos_transfer} show that despite the long sequences, the sampled ResNet models perform well above random initialization. 
For example, sampled ResNet-18s achieve 68.1\% on CIFAR-10 without any fine-tuning (Table \ref{tab:scalable_hyper_reps:generative_resnets_zoos_in_domain}). 
These models are at least three orders of magnitude larger than previous models used for \textit{hyper-representation} learning \citep{schurholtHyperRepresentationsGenerativeModels2022}, rendering it computationally infeasible for the approach presented in \citep{schurholtHyperRepresentationsGenerativeModels2022} to be evaluated against.
As before, the performance difference to random initialization becomes smaller during fine-tuning. 
Similar to our experiments on CNNs, sampled ResNet-18s achieve competitive performance or even outperform training from scratch with a considerably smaller computational budget.\footnote{The base population is trained with a one-cycle learning rate scheduler. To avoid any bias, we adopt the same scheduler but train for only 10 epochs, which affects direct comparability.}
Transferred to a new task, sampled models outperform training from scratch and match fine-tuning from prompt examples (Table \ref{tab:scalable_hyper_reps:generative_resnets_zoos_transfer}). Interestingly, subsampling and bootstrapping appear to work well when there is a useful signal to start with, i.e., on easier tasks that are similar to the pretraining distribution. This suggests that the sampling distributions are not ideal, and may require a better fit, more samples, or iterative adjustment to fit new datasets zero-shot.
Nonetheless, even the relatively naive sampling methods can successfully sample competitive models, even at the scale of ResNet-sized architectures. This shows that our sequential sampling works even for long sequences of tokens.  

\paragraph*{Subsampling Improves Performance.}
Previous work requires high-quality prompt examples to target specific properties \citep{schurholtHyperRepresentationsGenerativeModels2022}. Our sampling methods drop these requirements and use prompt examples only to model a prior. We therefore compare \ourmethod with $S_{KDE30}$ from \citep{schurholtHyperRepresentationsGenerativeModels2022} to \ourmethod. Further, we compare the $KDE30$ sampling method with our subsampling approach on \ourmethod. 
On datasets where published results are available, using $KDE30$ with \ourmethod improves performance over previously published results with $S_{KDE30}$; see Table \ref{tab:scalable_hyper_reps:generative_cnn_zoos_in_domain} for MNIST and SVHN results, e.g., epoch 0. We credit this to the better reconstruction quality of pre-training with \ourmethod.
Further, our sampling methods improve performance over $S_{KDE30}$. We compare \ourmethod + $S_{KDE30}$ with \ourmethod + subsampling and \ourmethod + bootstrapping, e.g., in Table \ref{tab:scalable_hyper_reps:generative_resnets_zoos_in_domain} on CIFAR-10 at epoch 0 from 64.8\% to 68.1\%, or on Tiny Imagenet from 8.4\% to 11.1\%. 
Using bootstrapping to adjust $\mathcal{P}$ iteratively further improves the sampled models slightly. It even allows to replace prompt examples with a Gaussian prior $\mathcal{P}$. The results of \ourmethod$_{GAUSS}$ show high performance after fine-tuning, even the highest overall on MNIST. 
These results show that our sampling methods not only drop requirements for the prompt examples but even improve the performance of the sampled models.

\paragraph*{Few-Shot Model Sampling Transfers to New Tasks and Architectures.}
Lastly, we explore whether sampling models using \ourmethod generalizes beyond the original task and architecture with very few prompt examples. 
Such transfers are out of reach of previous \textit{hyper-representations}, which are bound to a fixed number of weights.
\ourmethod, on the other hand, represents models of different sizes or architectures simply as sequences of different lengths, which can vary between pretraining and sampling. 
Since we use the prompt examples only to roughly model the sampling distribution, we need only a few (1-5) prompt examples which are trained for only a few epochs (1-5). 
That way, sampling for new architectures and/or tasks can become very efficient.
We test that idea in three experiments: 
(i) \emph{changing the tasks} between pretraining and prompt-examples from CIFAR-100 to Tiny-Imagenet (Table \ref{tab:scalable_hyper_reps:generative_resnets_zoos_fewshot_transfer}); 
(ii) \emph{changing the architecture} between pretraining and prompt-examples from ResNet-18 to ResNet-34 (Table \ref{tab:scalable_hyper_reps:generative_resnets_zoos_fewshot_resnet34}); and 
(iii) \emph{changing both task and architecture} from ResNet-18 on CIFAR-100 to ResNet-34 on Tiny-Imagenet (Figure \ref{fig:scalable_hyper_reps:sampling_new_arch_task} and Table \ref{tab:scalable_hyper_reps:generative_resnets_zoos_fewshot_ti_resnet34}).

In all three experiments, using target prompt examples improves over random initialization as well as previous transfer experiments. This indicates that \ourmethod representations contain useful information even for new architectures or tasks. 
The sampled models outperform the prompt examples and training from scratch, considerably in earlier epochs, and preserve a performance advantage throughout fine-tining. 

\begin{wraptable}{r}{0.59\linewidth}
\vspace{-2mm}
\caption[Performance of Sampled ResNet-18s transferred to Tiny-Imagenet]{
Sampling ResNet-18 models for Tiny-Imagenet. \ourmethod was pre-trained on CIFAR-100, 15 samples are drawn using subsampling, and 5 prompt examples are taken from the Tiny-Imagenet ResNet-18 zoo at epoch 25 with a mean accuracy of 43\%.  
}
\label{tab:scalable_hyper_reps:generative_resnets_zoos_fewshot_transfer}
\small
\setlength{\tabcolsep}{3pt}
\begin{tabularx}{1.0\linewidth}{clc}
\toprule
\multicolumn{3}{c}{ResNet-18   CIFAR100 to TinyImagnet}                           \\
\midrule
Ep.               & Method                    & Acc TI \\
\cmidrule(r){1-1}  \cmidrule(l){2-2} \cmidrule(lr){3-3} 
\multirow{2}{*}{0}  & tr. fr. scratch           & 0.5$\pm$0.0           \\
                    & \ourmethod                & 0.6$\pm$0.0           \\
\cmidrule(r){1-1}  \cmidrule(l){2-2} \cmidrule(lr){3-3} 
\multirow{2}{*}{1}  & tr. fr. scratch           & 10.4$\pm$2.2          \\
                    & \ourmethod                & \textbf{39.4$\pm$1.5}  \\
\cmidrule(r){1-1}  \cmidrule(l){2-2} \cmidrule(lr){3-3} 
\multirow{2}{*}{2}  & tr. fr. scratch           & 28.5$\pm$0.9          \\
                    & \ourmethod                & \textbf{61.0$\pm$0.2}  \\
\cmidrule(r){1-1}  \cmidrule(l){2-2} \cmidrule(lr){3-3} 
2            & \ourmethod ensamble                          & 64.0     \\                          
\bottomrule
\end{tabularx}
\vspace{-2mm}
\end{wraptable} 
Sampling for a new task (Table~\ref{tab:scalable_hyper_reps:generative_resnets_zoos_fewshot_transfer}), the sampled models outperform the prompt examples after just two epochs of fine-tuning, which indicates that transfer-learning using \ourmethod is an efficient alternative. 
Sampling from ResNet-18 to ResNet-34 for the same task (Table \ref{tab:scalable_hyper_reps:generative_resnets_zoos_fewshot_resnet34}) shows likewise improved performance over training from scratch, which indicates that the learned representation generalizes to larger architectures as well.
Sampling for new tasks and different architecture (Figure \ref{fig:scalable_hyper_reps:sampling_new_arch_task} and Table \ref{tab:scalable_hyper_reps:generative_resnets_zoos_fewshot_ti_resnet34}) combines the previous experiments and confirms their results. Sampled models outperform training from scratch by a considerable margin. 
Figure \ref{fig:scalable_hyper_reps:sampling_new_arch_task} indicates that with increasing distance from the pretraining architecture for \ourmethod, the performance gain of sampled models decreases, e.g., with increasing ResNet size.
Additionally, since sampling models using \ourmethod is cheap and lends itself to ensembling, we investigate the diversity of sampled models in Appendix \ref{sec:scalable_hyper_reps:sampling_diversity}.
Taken together, the experiments show that \ourmethod learns representations that can generalize beyond the pretraining task and architecture, and can efficiently be sampled for both new tasks and architectures.

\subsection*{Limitations}
In this paper, we pretrain \ourmethod on homogeneous zoos with one architecture. This simplifies alignment for pre-training, but more importantly it simplifies evaluation for model generation. Since \ourmethod can train on varying architectures and model sizes, the model population requirement for pre-training is significantly relaxed. A sufficient number of models are available on public model hubs. 
Further, our sampling method requires access to prompt examples, to have an informed prior from which to sample. For small models, bootstrapping from a Gaussian finds the targeted distribution; see \ourmethod$_{GAUSS}$ in Table \ref{tab:scalable_hyper_reps:generative_cnn_zoos_in_domain}. For large models with correspondingly long sequences, that approach is too expensive, which is why we rely on prompt examples.  
Lastly, in this paper, we perform experiments only on computer vision tasks. This is a choice to simplify the experiment setup.

%
%

%
%
\section{Related Work}

Representation learning in the space of NN weights has become a growing field recently. 
Several methods with different approaches to deal with weight spaces have been proposed to predict model properties such as accuracy \citep{unterthinerPredictingNeuralNetwork2020,eilertsenClassifyingClassifierDissecting2020, andreisSetbasedNeuralNetwork2023, zhangNeuralNetworksAre2023} or to learn the encoded concepts \citep{ashkenaziNeRNLearningNeural2022, deluigiDeepLearningImplicit2023}. 
%
Other work investigates the structure of trained weights on a fundamental level, using their eigen or singular value decompositions to identify training phases or predict properties \citep{martinTraditionalHeavyTailedSelf2019,MM20_SDM,martinPredictingTrendsQuality2021,martinImplicitSelfregularizationDeep2021, YTHx22_TR, mellerSingularValueRepresentation2023}.
Taking an optimization perspective, other work has investigated the uniqueness of the basis of trained NNs \citep{ainsworthGitReBasinMerging2022, brownPrivilegedConvergentBases2023}. Other work identifies subspaces of weights that are relevant, which motivates our work \citep{bentonLossSurfaceSimplexes2021, lucasMonotonicLinearInterpolation2021, wortsmanLearningNeuralNetwork2021, fortLargeScaleStructure2019}. The mode connectivity of trained models has been investigated to improve understanding of how to train models \citep{draxlerEssentiallyNoBarriers2018, nguyenConnectedSublevelSets2019, frankleLinearModeConnectivity2019}. 

A different line of work trains models to generate weights for target models, such as HyperNetworks \citep{haHyperNetworks2017, nguyenHyperVAEMinimumDescription2019, zhangGraphHyperNetworksNeural2019, knyazevParameterPredictionUnseen2021,knyazevCanWeScale2023,kofinas2024graph}, with a recurrent backbone \citep{wangCompactOptimalDeep2023} as learned initialization \citep{dauphinMetaInitInitializingLearning2019} or for meta learning \citep{finnModelAgnosticMetaLearningFast2017,zhmoginovHyperTransformerModelGeneration2022,navaMetaLearningClassifierFree2022}.
While the last category uses data to get learning signals, another line of work learns representations of the weights directly. Hyper-Representations train an encoder-decoder architecture using reconstruction of the weights, with contrastive guidance, and has been proposed to predict model properties \citep{schurholtSelfSupervisedRepresentationLearning2021} or generate new models \citep{schurholtHyperRepresentationsPreTrainingTransfer2022, schurholtHyperRepresentationsGenerativeModels2022}. While previous work was limited to small models of fixed length, this paper proposes methods to decouple the representation learner size from the base model. Related approaches use convolutional auto-encoders \citep{berardiLearningSpaceDeep2022} or diffusion on the weights \citep{peeblesLearningLearnGenerative2022}.

%
%
\section{Conclusion}

In this work, we propose \ourmethod, a method to learn task-agnostic representations of Neural Network models. \ourmethod decouples model tokenization from \textit{hyper-representation} learning and can scale to much larger neural network models and generalize to models of different architectures. 
We analyze \ourmethod embeddings and find they reveal model quality metrics. Empirical evaluations show that 
i) \ourmethod embeddings contain information on model quality both globally and on a layer level,
ii) \ourmethod embeddings are predictive of model performance, and 
iii) sampling models with \ourmethod achieves higher performance and generalizes to larger models and new architectures. Further, we propose sampling methods that reduce quality and quantity requirements for prompt examples and allow targeting new model distributions.\looseness-1

\begin{subappendices}
\renewcommand{\ourmethodlong}{\texttt{Sequential Autoencoder for Neural Embeddings}\,}
\renewcommand{\ourmethod}{\texttt{SANE}\,}

\section{Ablation Studies}
\label{sec:scalable_hyper_reps:ablation}
In this section, we perform ablation studies to assess the effectiveness of the methods proposed above: 
model alignment to simplify the learning task; 
inference window size to improve inference quality; 
haloing and batch-norm conditioning to increase sample quality.
\paragraph*{Impact of Model Alignment.}
\begin{wraptable}{r}{0.48\linewidth}
\vspace{-8mm}
\captionof{table}[Ablation Study on Alignment and Permutation]{
Impact of alignment ablation and permutation on reconstruction loss.
}
\label{tab:scalable_hyper_reps:ablation_alignment}
\small
\setlength{\tabcolsep}{6pt}
\begin{tabularx}{1.0\linewidth}{ccccc}
\toprule
\multicolumn{3}{c}{Sample Permutations} & \multicolumn{2}{c}{$\mathcal{L}_{rec}$} \\
\cmidrule(r){1-3} \cmidrule(l){4-5}
Aligned    & View 1       & View 2       & Train              & Test               \\
\cmidrule(r){1-3} \cmidrule(l){4-5}
No         & Perm.     & Perm.    & 0.304              & 0.167              \\
Yes        & Perm.     & Perm.    & 0.148              & 0.082              \\
Yes        & Align    & Perm.    & 0.107              & 0.082              \\
Yes        & Align    & Align   & 0.072              & 0.082              \\ 
\bottomrule
\end{tabularx}
\vspace{-4mm}
\end{wraptable} 
Model alignment intuitively reduces training complexity by mapping all models to the same subspace. To evaluate its impact, we conduct training experiments with the same configuration on datasets with and without aligned models. In the dataset with aligned models, we use either the aligned form or 5 random permutations for the two views for both reconstruction and contrastive learning.
As shown in Table \ref{tab:scalable_hyper_reps:ablation_alignment}, the results show two effects. 
First, alignment through git re-basin simplifies the learning task and contributes to improved generalization, both training and test losses are reduced by more than 50\%. 
Second, anchoring at least one of the views to the aligned form does further reduce the training loss, but does not improve generalization.

\paragraph*{Window Size Ablation.}
The sequential decomposition of \ourmethod allows one to pretrain not on the full model sequence, but on subsequences. The choice of the length of the subsequence, the window size, is a critical parameter that balances computational load and context. We used a window of 256 for pretraining for most of our experiments. 

\begin{wrapfigure}{r}{0.45\linewidth}
\vspace{-2mm}
\centering
\includegraphics[trim=4mm 0mm 1.5mm 1.5mm, clip, width=0.85\linewidth]{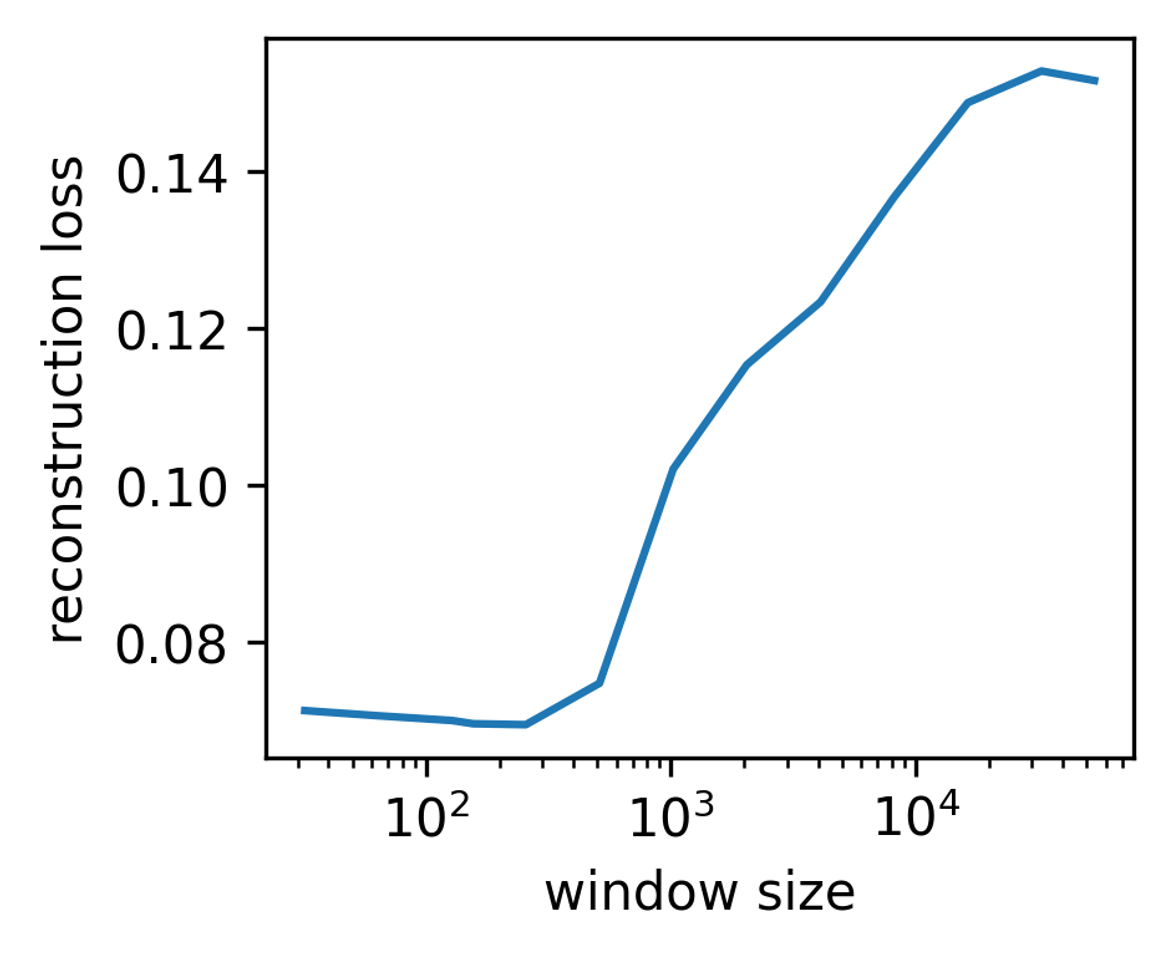}
\vspace{-4mm}
\captionof{figure}[Ablation Study of \ourmethod over Window Size]{\ourmethod reconstruction loss over number of tokens within a window. The loss is lowest around the training windowsize of 256 tokens, longer sequences up to the full models sequence length of 50k tokens cause interference and double the reconstruction error.
}
\vspace{-8mm}
\label{fig:scalable_hyper_reps:windowsize_ablation}    
\end{wrapfigure}
Here, we study the influence of the window size on reconstruction error, exploring values ranging from 32 to 2048. Our experiments did not reveal substantial impact of smaller windows on pretraining loss or sampling performance. This seems to suggest that a window size as large as 2048 may still be insufficient on ResNets to capture enough context. Alternatively, it may suggest that the underlying assumption that context matters may not entirely hold up.

However, we did observe an important impact on the relationship between training and inference window sizes. During inference, memory load is significantly lower. Inference allows much larger window sizes, up to the entire length of the ResNet sequence. However, departing from the training window size appears to introduce interference, which affects the reconstruction error (Figure \ref{fig:scalable_hyper_reps:windowsize_ablation}). 

\paragraph*{Halo and batch-norm conditioning.}
\begin{wraptable}{r}{0.37\linewidth}
\vspace{-8mm}
\captionof{table}[Ablation Study of \ourmethod over Haloing and BN Conditioning]{
Ablation of BN-conditioning and haloing.
}\label{tab:scalable_hyper_reps:ablation_halo_bn_cond}
\small
\setlength{\tabcolsep}{3pt}
\begin{tabularx}{1.0\linewidth}{clc}
\toprule
Ep.              & Method        & CIFAR-10           \\
\cmidrule(r){1-1} \cmidrule(lr){2-2} \cmidrule(l){3-3}
\multirow{5}{*}{0} & rand init     & $\sim$10 /\%       \\
                   & naïve         & 10$\pm$0.0            \\
\cmidrule(lr){2-2} \cmidrule(l){3-3}
                   
                   & Haloed        & 14.5$\pm$6.3          \\
                   & BN-cond       & 60.8$\pm$2.2          \\
                   & Haloed+BN-cond & \textbf{64.8$\pm$2.1} \\
\cmidrule(r){1-1} \cmidrule(lr){2-2} \cmidrule(l){3-3}
\multirow{5}{*}{5} & rand init     & 64.4$\pm$2.9          \\
                   & naïve         & 90.8$\pm$0.2          \\
\cmidrule(lr){2-2} \cmidrule(l){3-3}
                   & Haloed        & 90.9$\pm$0.1          \\
                   & BN-cond       & 90.7$\pm$0.2          \\
                   & Haloed+BN-cond & 90.9$\pm$0.2         
    \\ 
\bottomrule
\end{tabularx}
\vspace{-4mm}
\end{wraptable} 
Haloing and batch-norm conditioning aim at reducing noise in model sampling; see Section \ref{sec:scalable_hyper_reps:seq_hyper_reps}. 
To assess their impact on sampling performance, we conduct an in-domain experiment using \ourmethod trained on CIFAR-10 ResNet-18s, using prompt examples from the train set and fine-tuning on CIFAR-10. We compare with \emph{naïve} sampling without haloing and batch-norm conditioning.
The results in Table \ref{tab:scalable_hyper_reps:ablation_halo_bn_cond} show the significant improvements achieved by both haloing and batch-norm conditioning. From random guessing of naïve sampling, combining both improves to around 65\%. Since both methods aim at reducing noise for zero-shot sampling, their effect is largest then and diminishes somewhat during finetuning. Both methods not only improve zero-shot sampling per se but make the sampled models provide enough signal to facilitate subsampling or bootstrapping strategies.\looseness-1

\newpage
\section{\ourmethod Architecture Details}

\begin{wraptable}{r}{0.55\linewidth}
\vspace{-25pt}
\captionof{table}{
Architecture Details for \ourmethod
}
\label{tab:scalable_hyper_reps:architecture_details}
\small
\setlength{\tabcolsep}{12pt}
\begin{tabularx}{1.0\linewidth}{lcc}
\toprule
\multicolumn{1}{c}{Hyper-Parameter} & CNNs     & ResNet-18 \\ \midrule
tokensize                           & 289      & 288       \\
sequence lenght                     & $\sim$50 & $\sim$50k \\
window size                         & 32       & 256, 512  \\
d\_model                            & 1024     & 2048      \\
latent\_dim                         & 128      & 128       \\
transformer layers                  & 4        & 8         \\
transformer heads                   & 4,8      & 4,8       \\
\bottomrule
\end{tabularx}
\vspace{-40pt}
\end{wraptable} 
In Table \ref{tab:scalable_hyper_reps:architecture_details}, we provide additional information on the training hyper-parameters for \ourmethod on populations of small CNNs as well as ResNet18s. These values are the stable mean across all experiments, exact values can vary from population to population. Full experiment configurations are documented in the code.

\vspace{80pt}

\FloatBarrier
\section{\ourmethod Embedding Analysis - Additional Results}
\begin{figure}[h!]
\centering
\includegraphics[trim=4mm 0mm 1.5mm 1.5mm, width=0.6\linewidth]{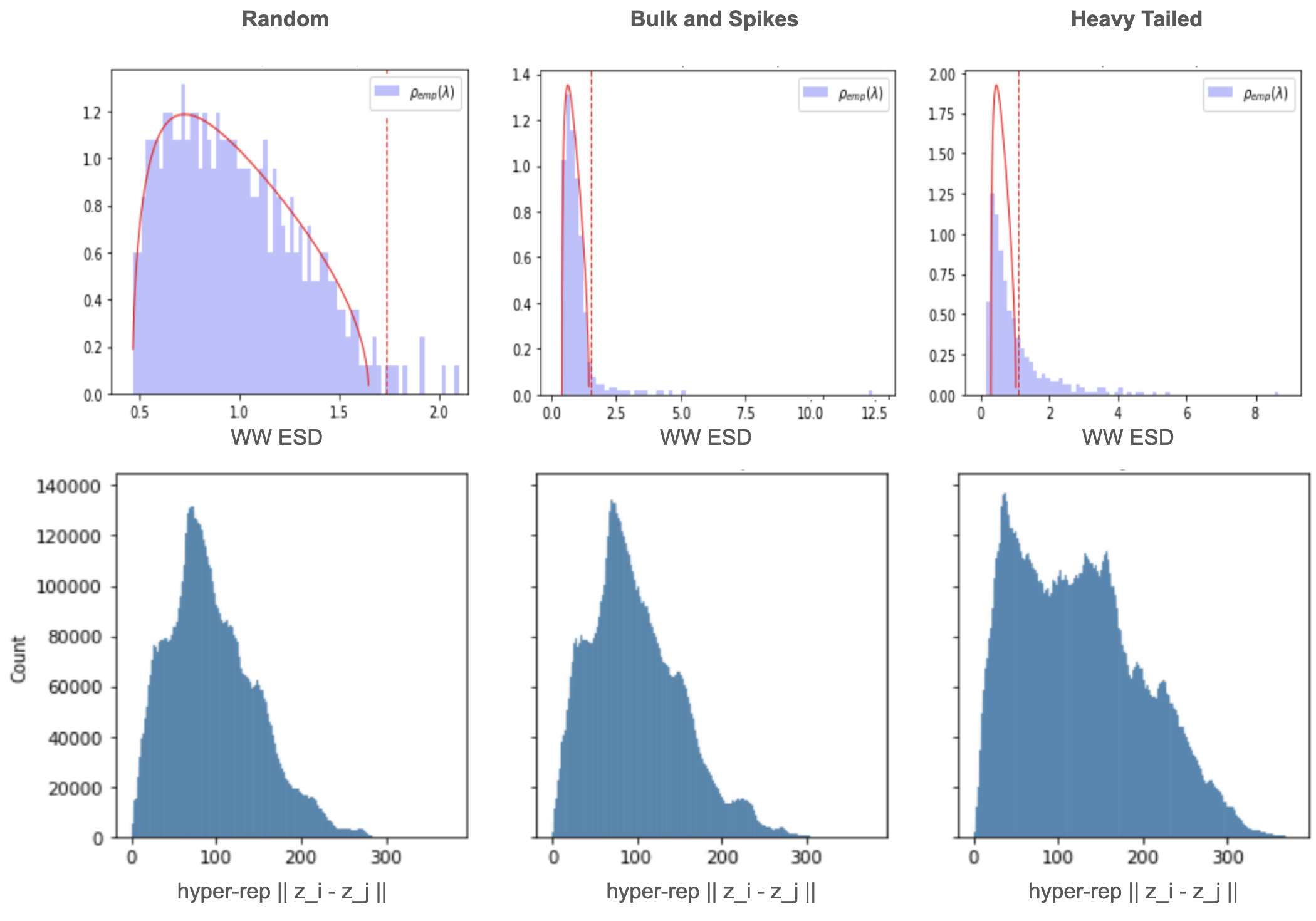}
\captionof{figure}[Comparison of Layer Wise Features at Different Training Phases]{Comparison between WeightWatcher features (top) and \ourmethod (bottom). \cite{martinTraditionalHeavyTailedSelf2019} identify different phases in the eigenvalue spectrum of trained weight matrices. We replicate the experiment setup and find ESDs similar to \textit{random} (top left), \textit{bulk and spikes} (top middle) and \textit{heavy tailed} (top right). We compare these against pairwise distances of \ourmethod embeddings of the same layer. While the distributions have a different shape, it appears to become more heavy tailed going from \textit{random} to \textit{heavy tailed}.
}
\label{fig:scalable_hyper_reps:ww_comparison_phases}    
\end{figure}

This section contains additional results on \ourmethod embedding analysis, in comparison with previous weight matrix analysis. 
In Figure \ref{fig:scalable_hyper_reps:ww_comparison_phases}, we compare the eigenvalue distribution for different models with \ourmethod embeddings. Replicating the experiment setup from \citep{martinTraditionalHeavyTailedSelf2019,martinImplicitSelfregularizationDeep2021}, we train MiniAlexNet models on CIFAR-10 varying only the batch size. With a smaller batch size and longer training duration, the eigenvalue distribution transitions from random with very few spikes, over a bulk with many spikes, to heavy-tailed. The embeddings of \ourmethod appear to also become more heavy-tailed, but do not seem to pick up on the change from few to many spikes. 
The results are suggestive, pointing to obvious follow-up work.

\begin{figure}[h!]
\centering
\includegraphics[trim=4mm 0mm 1.5mm 1.5mm, width=0.6\linewidth]{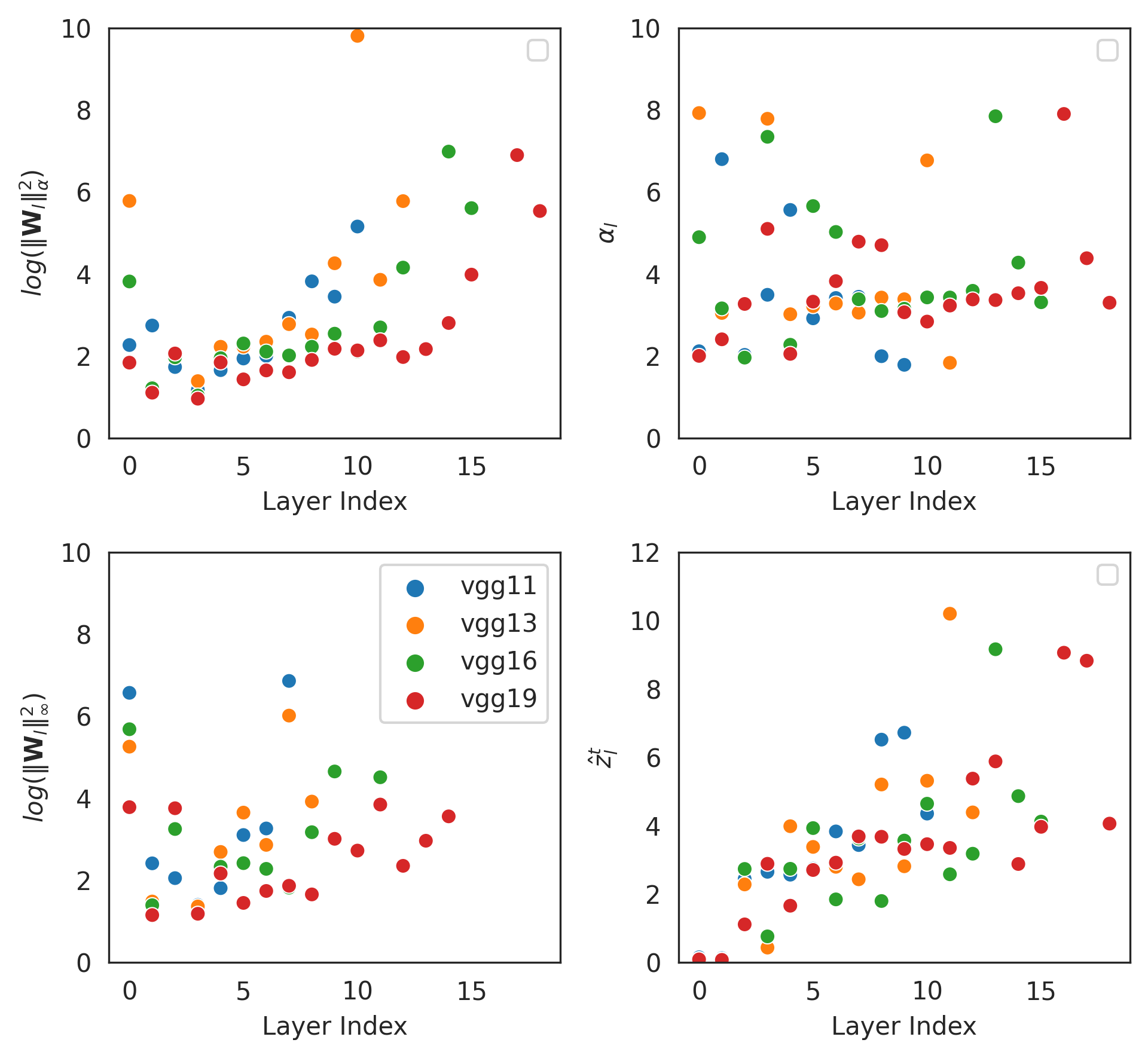}
\captionof{figure}[Comparison of Layer Wise features for VGGs]{Comparison between different WeightWatcher (WW) features (left) and \ourmethod (right). Features over layer index for VGGs from pytorchcv of different sizes. \ourmethod shows similar trends to WW, low values at early layers and a sharp increase at the end. }
\vspace{-2mm}
\label{fig:scalable_hyper_reps:ww_comparison_layers_vgg}
\end{figure}

\begin{figure}[h!]
\centering
\includegraphics[trim=4mm 0mm 1.5mm 1.5mm, width=0.6\linewidth]{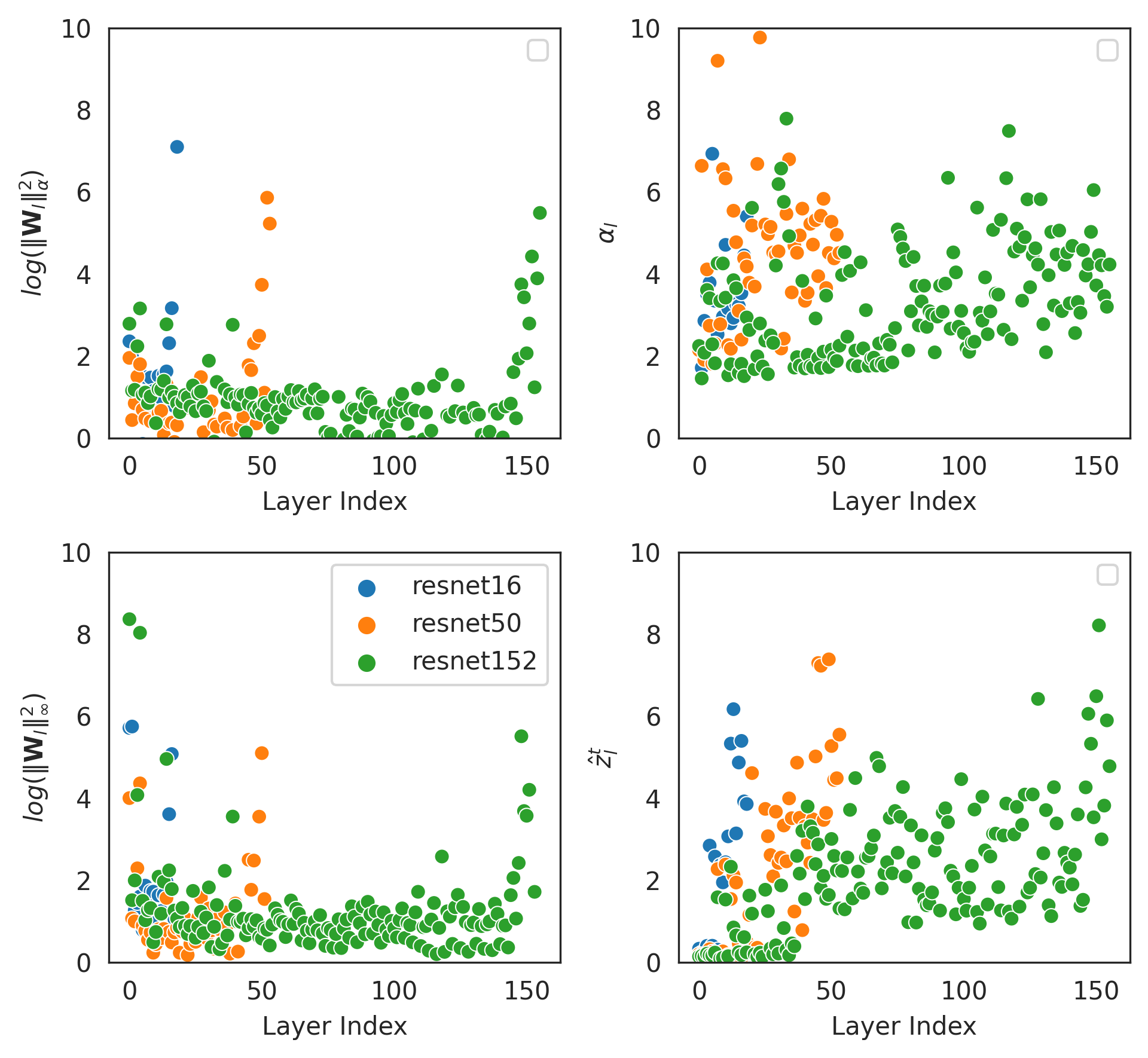}
\captionof{figure}[Comparison of Layer Wise features for ResNets]{Comparison between different WeightWatcher (WW) features (left) and \ourmethod (right). Features over layer index for Resnets from pytorchcv of different sizes. \ourmethod shows similar trends to WW, low values at early layers and a sharp increase at the end. }
\vspace{-2mm}
\label{fig:scalable_hyper_reps:ww_comparison_layers_resnets_2x2}
\end{figure}

Figures \ref{fig:scalable_hyper_reps:ww_comparison_layers_vgg} and \ref{fig:scalable_hyper_reps:ww_comparison_layers_resnets_2x2} compare \ourmethod with different WeightWatcher metrics on VGGs from pytorchcv \citep{semeryOsmrImgclsmob2024} and the ResNet-18 zoo from the modelzoo dataset \citep{schurholtModelZoosDataset2022}.

\begin{figure}[h!]
\centering
\includegraphics[trim=4mm 0mm 1.5mm 1.5mm, width=0.6\linewidth]{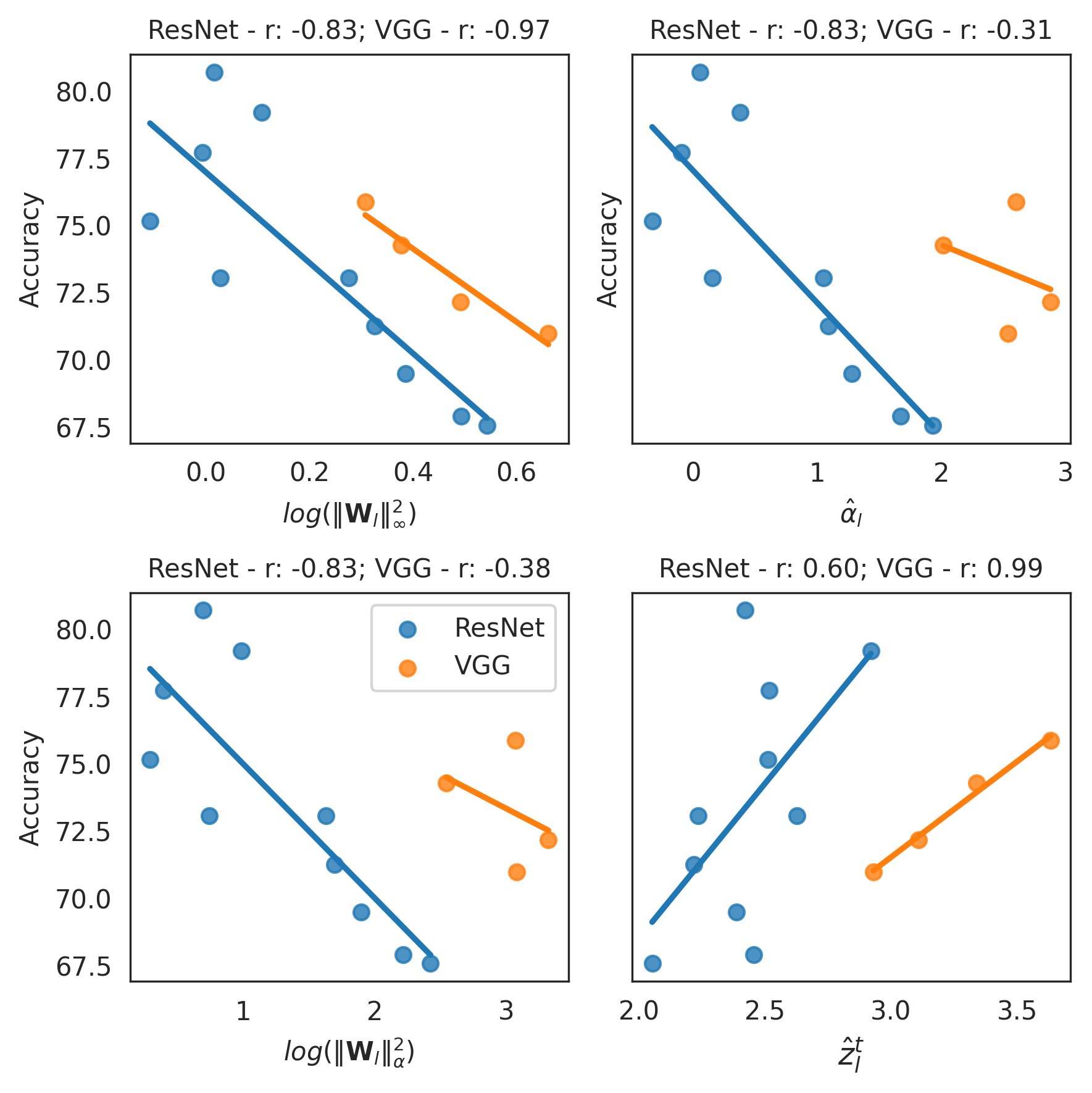}
\captionof{figure}[Comparison of Model Wise features for VGGs]{Comparison between WeightWatcher features (left) and \ourmethod (right). Accuracy over model features for Resnets and VGGs from pytorchcv of different sizes. \ourmethod shows similar trends to WW, low values at early layers and a sharp increase at the end. }
\vspace{-2mm}
\label{fig:scalable_hyper_reps:ww_comparison_accuracy_2x2}
\end{figure}
\begin{figure}[h!]
\centering
\includegraphics[trim=4mm 0mm 1.5mm 1.5mm, width=0.6\linewidth]{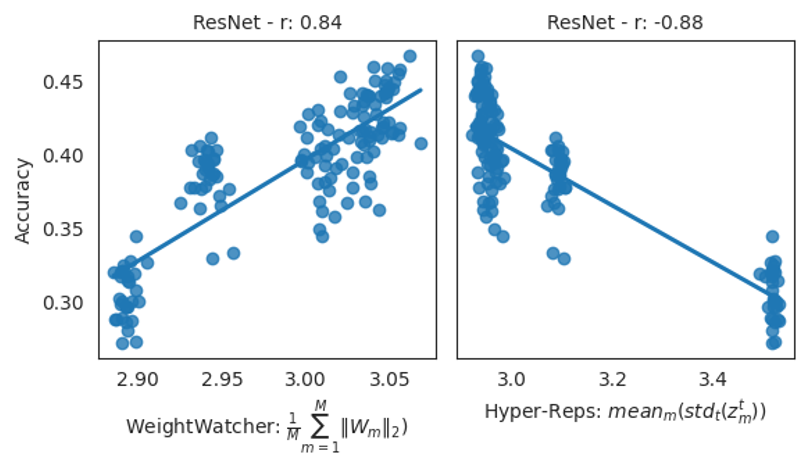}
\captionof{figure}[Comparison of Model Wise features for ResNets]{Comparison between WeightWatcher features (left) and \ourmethod (right). Accuracy over model features for ResNets from the ResNet model zoo. Although \ourmethod is pretrained in a self-supervised fashsion, it preserves the linear relation of a globally-aggregated embedding to model accuracy.}
\vspace{-2mm}
\label{fig:scalable_hyper_reps:ww_comparison_accuracy_our_zoo}    
\end{figure}

\FloatBarrier
\newpage
\section{Model Property Prediction - Additional Results}
In this section, we provide additional details for Section \ref{sec:scalable_hyper_reps:property_prediction}. Table \ref{tab:scalable_hyper_reps:discr_small_zoos_full} shows full results for populations of small CNNs. 
\begin{table}[h!]
\centering
\captionof{table}[Property Prediction Results for Small CNN Model Zoos. ]{
Property prediction on populations of small CNNs. 
}
\label{tab:scalable_hyper_reps:discr_small_zoos_full}
\scriptsize
\setlength{\tabcolsep}{5pt}
\begin{tabularx}{1.0\linewidth}{lccccccccccccccc}
\toprule
      & \multicolumn{3}{c}{MNIST}                                                      &  & \multicolumn{3}{c}{SVHN}                                                       &  & \multicolumn{3}{c}{CIFAR-10 (CNN)}                                             &  & \multicolumn{3}{c}{STL}                                                        \\ 
      \cmidrule(r){2-4} \cmidrule(lr){6-8} \cmidrule(lr){10-12} \cmidrule(l){14-16} 
      & \multicolumn{1}{c}{W} & \multicolumn{1}{c}{$s(W)$} & \multicolumn{1}{c}{\ourmethod} &  & \multicolumn{1}{c}{W} & \multicolumn{1}{c}{$s(W)$} & \multicolumn{1}{c}{\ourmethod} &  & \multicolumn{1}{c}{W} & \multicolumn{1}{c}{$s(W)$} & \multicolumn{1}{c}{\ourmethod} &  & \multicolumn{1}{c}{W} & \multicolumn{1}{c}{$s(W)$} & \multicolumn{1}{c}{\ourmethod} \\ 
      \cmidrule(r){1-1} \cmidrule(r){2-4} \cmidrule(lr){6-8} \cmidrule(lr){10-12} \cmidrule(l){14-16} 
ACC   & 0.965                 & \textbf{0.987}           & 0.978                       &  & 0.910                 & 0.985                    & \textbf{0.991}              &  & -7.580                & \textbf{0.965}           & 0.885                       &  & -18.818               & \textbf{0.919}           & 0.305                       \\
Epoch & 0.953                 & \textbf{0.974}           & 0.958                       &  & 0.833                 & \textbf{0.953}           & 0.930                       &  & 0.636                 & \textbf{0.923}           & 0.771                       &  & -1.926                & \textbf{0.977}           & 0.344                       \\
Ggap  & 0.246                 & 0.393                    & \textbf{0.402}              &  & 0.479                 & 0.711                    & \textbf{0.760}              &  & 0.324                 & \textbf{0.909}           & 0.772                       &  & -0.617                & \textbf{0.858}           & 0.307                       \\ \bottomrule
\end{tabularx}
\end{table} 

\subsection*{Comparison to Previous Work}
\label{app:scalable_hyper_reps:discr_comparison_previous_work}
Here, we compare \ourmethod with previous work to disseminate the information contained in model embeddings.
The experiment setup in this paper is designed around the ResNets, and therefore it uses sparse epochs for computational efficiency. For consistency, we use the same setup for the CNN zoos as well. The exact numbers are therefore not directly comparable to \citet{schurholtSelfSupervisedRepresentationLearning2021}.   
To provide as much context as possible, we approach the comparison from two angles:
\begin{itemize}
    \item [(1)]\textbf{Direct comparison to the published results:} to contextualize, we use the (deterministic) results of weight statistics $s(W)$ to adjust for the differences in setup. We mark the results for $s(W)$ from \citet{schurholtSelfSupervisedRepresentationLearning2021} as $s(W)_{pp}$ and compare to their $E_{c+}D$ where possible.
    \item [(2)] \textbf{Approximation of the effect of global embeddings:} previous work used global model embeddings, which we approximate by using the full model embedding sequence. We therefore compare \ourmethod + aggregated tokens (as proposed in the submission) to \ourmethod + full model sequence (similar to \citet{schurholtSelfSupervisedRepresentationLearning2021}).
\end{itemize}

The results in the Tables below allow the following conclusions:
\begin{itemize}
    \item [(1)] \textbf{\ourmethod matches the performance of previous work:} The only data available for direct comparison is the MNIST zoo. Here, both in direct comparison and in relation to s(W) cross-relating our results with published numbers, \ourmethod matches published performance of $E_c+D$. On other zoos, $E_c+D$ had similar performance to $s(W)$. We likewise find \ourmethod embeddings to have similar performance to $s(W)$ in our experiments.
    \item [(2)] \textbf{\ourmethod + full sequence improves downstream task performance over the \ourmethod + aggregated sequence:} That indicates that \ourmethod + full sequence contains more information for model prediction. However, both \citet{schurholtSelfSupervisedRepresentationLearning2021} and \ourmethod with full sequence have the disadvantage that they do not scale. With growing models, the representation learner of \citet{schurholtSelfSupervisedRepresentationLearning2021} and the input to the linear probe of \ourmethod + full sequence grow accordingly. \ourmethod + aggregated sequence does lose some information on small models, but scales gracefully to large models and remains competitive.
\end{itemize}

\begin{table}[h!]
\small
\centering
\caption[Property Prediction Details for MNIST-CNN(s) Zoo]{Property Prediction comparison to previous work on the MNIST-CNN model zoo. We compare our linear probing results from weights $W$, layer-wise quintiles $s(W)$, embeddings from \ourmethod either aggregated into one embedding or using the full sequence to results previously published in \citet{schurholtSelfSupervisedRepresentationLearning2021}. We mark their results for $s(W)$ as $s(W)_{pp}$. Since the experimental setup is not the same, the numbers of $s(W)$ do not match.}
\setlength{\tabcolsep}{7pt}
\begin{tabularx}{0.95\linewidth}{ccccccc}
\toprule
      & \multicolumn{1}{c}{W} & \multicolumn{1}{c}{$s(W)$} & \multicolumn{1}{c}{\ourmethod aggregated} & \ourmethod full sequence & \multicolumn{1}{c}{$s(W)_{pp}$} & \multicolumn{1}{c}{$E_{c+}D$} \\
\cmidrule(r){1-1} \cmidrule(lr){2-2} \cmidrule(lr){3-3} \cmidrule(lr){4-4} \cmidrule(lr){5-5} \cmidrule(lr){6-6} \cmidrule(l){7-7}  
ACC   & 0.965                 & \textbf{0.987}           & 0.978                               & \textbf{0.987}     & 0.977                            & 0.973                              \\
Epoch & 0.953                 & 0.974                    & 0.958                               & \textbf{0.975}     & 0.987                            & 0.989                              \\
Ggap  & 0.246                 & 0.393                    & 0.402                               & \textbf{0.461}     & 0.662                            & 0.667     \\
\bottomrule
\end{tabularx}
\label{tab:scalable_hyper_reps:discr_comparison_mnist}
\end{table}
\begin{table}[h!]
\small
\centering
\caption[Property Prediction Details for SVHN-CNN(s) Zoo]{Property Prediction comparison to previous work on the SVHN-CNN model zoo. We compare our linear probing results from weights $W$, layer-wise quintiles $s(W)$, to embeddings from \ourmethod either aggregated into one embedding or using the full sequence. For this zoo, previous results are not available.}
\setlength{\tabcolsep}{7pt}
\begin{tabularx}{0.95\linewidth}{ccccccc}
\toprule
      & \multicolumn{1}{c}{W} & \multicolumn{1}{c}{$s(W)$} & \multicolumn{1}{c}{\ourmethod aggregated} & \ourmethod full sequence & \multicolumn{1}{c}{$s(W)_{pp}$} & \multicolumn{1}{c}{$E_{c+}D$} \\
\cmidrule(r){1-1} \cmidrule(lr){2-2} \cmidrule(lr){3-3} \cmidrule(lr){4-4} \cmidrule(lr){5-5} \cmidrule(lr){6-6} \cmidrule(l){7-7}  
ACC   & 0.910                 & \textbf{0.985}           & 0.991                               & \textbf{0.993}     & \textit{n/a}         & \textit{n/a}         \\
Epoch & 0.833                 & \textbf{0.953}           & 0.930                               & \textbf{0.943}     & \textit{n/a}         & \textit{n/a}         \\
Ggap  & 0.479                 & 0.711                    & 0.760                               & \textbf{0.77}      & \textit{n/a}         & \textit{n/a}        \\
\bottomrule
\end{tabularx}
\label{tab:scalable_hyper_reps:discr_comparison_svhn}
\end{table}

\begin{table}[h!]
\small
\centering
\caption[Property Prediction Details for CIFAR-CNN(m) Zoo]{Property Prediction comparison to previous work on the CIFAR-CNN(m) model zoo. We compare our linear probing results from weights $W$, layer-wise quintiles $s(W)$, to embeddings from \ourmethod either aggregated into one embedding or using the full sequence. For this zoo, previous results are not available.}
\setlength{\tabcolsep}{7pt}
\begin{tabularx}{0.95\linewidth}{ccccccc}
\toprule
      & \multicolumn{1}{c}{W} & \multicolumn{1}{c}{$s(W)$} & \multicolumn{1}{c}{\ourmethod aggregated} & \ourmethod full sequence & \multicolumn{1}{c}{$s(W)_{pp}$} & \multicolumn{1}{c}{$E_{c+}D$} \\
\cmidrule(r){1-1} \cmidrule(lr){2-2} \cmidrule(lr){3-3} \cmidrule(lr){4-4} \cmidrule(lr){5-5} \cmidrule(lr){6-6} \cmidrule(l){7-7}  
ACC   & -7.580                & \textbf{0.965}           & 0.885                               & \textbf{0.947}     & \textit{n/a}         & \textit{n/a}         \\
Epoch & 0.636                 & \textbf{0.923}           & 0.771                               & \textbf{0.879}     & \textit{n/a}         & \textit{n/a}         \\
Ggap  & 0.324                 & \textbf{0.909}           & 0.772                               & \textbf{0.811}     & \textit{n/a}         & \textit{n/a}        \\
\bottomrule
\end{tabularx}
\label{tab:scalable_hyper_reps:discr_comparison_cifar}
\end{table}

\section{Model Generation - Additional Results}
This section contains additional results from model sampling experiments, extending Section \ref{sec:scalable_hyper_reps:generating_models}. 
In Table \ref{tab:scalable_hyper_reps:generative_cnn_zoos_transfer}, we show results on small CNNs transferring to a new task.
Similarly, Table \ref{tab:scalable_hyper_reps:generative_resnets_zoos_transfer} shows results on ResNet-18 models for task transfers.
Lastly, Tables \ref{tab:scalable_hyper_reps:generative_resnets_zoos_fewshot_resnet34}, \ref{tab:scalable_hyper_reps:generative_resnets_zoos_fewshot_ti_resnet34} and \ref{tab:scalable_hyper_reps:generative_resnets_zoos_fewshot_resnet34_one_prompt_example} contain additional results for transferring from ResNet-18 CIFAR-100 to ResNet34 and/or Tiny-Imagenet. 
\begin{table}[h!]
\small
\centering
\caption[Performance of CNN Models Sampled with \ourmethod for Transfer Learning]{Model generation on CNN model populations transfer learned on a new task. We compare sampled models at different epochs with models trained from scratch and models fine-tuned from the anchor samples.}
\setlength{\tabcolsep}{5pt}
\begin{tabularx}{1.0\linewidth}{ccccccc}
\toprule
  Method              & \multicolumn{3}{c}{SVHN to MNIST}                            & \multicolumn{3}{c}{CIFAR-10 to STL-10}                       \\
\cmidrule(r){1-1} \cmidrule(lr){2-4} \cmidrule(lr){5-7}
          & Epoch 0            & Epoch 1            & Epoch 25           & Epoch 0            & Epoch 1            & Epoch 25           \\
\cmidrule(r){1-1} \cmidrule(lr){2-2} \cmidrule(lr){3-3} \cmidrule(lr){4-4} \cmidrule(lr){5-5} \cmidrule(lr){6-6} \cmidrule(l){7-7}  
tr.fr.scratch   & $\sim$10 /\%       & 20.6+-1.6          & 83.3+-2.6          & $\sim$10 /\%       & 21.3+-1.6          & 44.0+-1.0          \\
pretrained      & 29.1+-7.2          & 84.1+-2.6          & 94.2+-0.7          & 16.2+-2.3          & 24.8+-0.8          & 49.0+-0.9          \\
$S_{KDE30}$      &  31.8+-5.6          &  86.9+-1.4          &  95.5+-0.4          & n/a          & n/a          & n/a          \\
\cmidrule(r){1-1} \cmidrule(lr){2-2} \cmidrule(lr){3-3} \cmidrule(lr){4-4} \cmidrule(lr){5-5} \cmidrule(lr){6-6} \cmidrule(l){7-7}  
\ourmethod$_{KDE30}$ & \textbf{40.2+-4.8} & 86.7+-1.6          & 94.8+-0.4          & 15.5+-2.3          & 24.9+-1.6          & 49.2+-0.5          \\
\ourmethod$_{SUB}$.  & 37.9+-2.8          & \textbf{88.2+-0.5} & \textbf{95.6+-0.3} & \textbf{17.4+-1.4} & \textbf{25.6+-1.7} & \textbf{49.8+-0.6} \\
\bottomrule
\end{tabularx}
\label{tab:scalable_hyper_reps:generative_cnn_zoos_transfer}
\end{table}

\begin{table}[]
\scriptsize
\centering
\caption[Performance of ResNet-18 Models Sampled with \ourmethod for Transfer Learning]{Model generation on ResNet-18 model populations transferred to a new task. We compare sampled models at different transfer learning epochs with models trained from scratch and models fine-tuned from the same anchor samples.}
\setlength{\tabcolsep}{6pt}
\begin{tabularx}{1.00\linewidth}{clccc}
\toprule
Epoch                  & \multicolumn{1}{c}{Method} & CIFAR-10 to CIFAR-100 & CIFAR-100 to Tiny-Imagenet & Tiny-Imagenet to CIFAR-100 \\
\cmidrule(r){1-1} \cmidrule(lr){2-2} \cmidrule(lr){3-3} \cmidrule(lr){4-4} \cmidrule(l){5-5} 
\multirow{5}{*}{0}     & tr. fr. scratch                  & $\sim$1 /\%                      & $\sim$0.5 /\%                        & $\sim$1 /\%                           \\
                       & Finetuned                  & 1.0+-0.3                     & 0.5+-0.0                             & 1.1+-0.2                              \\
                       & \ourmethod$_{KDE30}$                    & 1.0+-0.3                         & 0.5+-0.1                             & 1.0+-0.2                              \\
                       & \ourmethod$_{SUB}$                   & 1.0+-0.3                         & 0.6+-0.0                             & 1.1+-0.2                              \\
                       & \ourmethod$_{BOOT}$                  & 1.1+-0.2                         & 0.5+-0.0                             & 0.9+-0.2                              \\
\cmidrule(r){1-1} \cmidrule(lr){2-2} \cmidrule(lr){3-3} \cmidrule(lr){4-4} \cmidrule(l){5-5} 
\multirow{5}{*}{1}     & tr. fr. scratch                  & 17.5+-0.7                        & 13.8+-0.8                            & 17.5+-0.7                             \\
                       & Finetuned                  & 27.5+-1.3                     & 25.7+-0.5                            & 51.7+-0.5                             \\
                       & \ourmethod$_{KDE30}$                    & 26.8+-1.4                        & 21.5+-0.9                            & 40.2+-1.0                             \\
                       & \ourmethod$_{SUB}$                   & 26.4+-1.9                        & 21.5+-1.0                            & 40.63+-1.3                            \\
                       & \ourmethod$_{BOOT}$                  & 25.7\_01.9                       & 21.7+-1.0                            & 40.9+-0.8                             \\
\cmidrule(r){1-1} \cmidrule(lr){2-2} \cmidrule(lr){3-3} \cmidrule(lr){4-4} \cmidrule(l){5-5} 
\multirow{5}{*}{5}     & tr. fr. scratch                  & 36.5+-2.0                        & 31.1+-1.6                            & 36.5+-2.0                             \\
                       & Finetuned                  &          45.7+-1.0                        & 36.3+-2.5                            & 52.6+-1.3                             \\
                       & \ourmethod$_{KDE30}$                    & 44.5+-2.0                        & 36.3+-1.2                            & 47.2+-3.3                             \\
                       & \ourmethod$_{SUB}$                   & 45.6+-1.2                        & 35.8+-1.4                            & 49.8+-2.3                             \\
                       & \ourmethod$_{BOOT}$                  & 43.3+-2.4                        & \textbf{37.3+2.0}                    & 50.2+-3.4                             \\
\cmidrule(r){1-1} \cmidrule(lr){2-2} \cmidrule(lr){3-3} \cmidrule(lr){4-4} \cmidrule(l){5-5} 
\multirow{5}{*}{15}    & tr. fr. scratch                  & 53.3+-2.0                        & 38.5+-1.9                            & 53.3+-2.0                             \\
                       & Finetuned                  & 71.9+-0.1                     & 63.4+-0.2                            & \textbf{73.9+-0.3}                    \\
                       & \ourmethod$_{KDE30}$                    & 71.8+-0.3                        & \textbf{63.6+-0.2}                   & 73.4+-0.2                             \\
                       & \ourmethod$_{SUB}$                   & 72.0+-0.2                        & \textbf{63.6+-0.3}                   & 73.5+-0.2                             \\
                       & \ourmethod$_{BOOT}$                  & 71.9+-0.3                        & 63.4+-0.1                            & 73.7+-0.3                             \\
\cmidrule(r){1-1} \cmidrule(lr){2-2} \cmidrule(lr){3-3} \cmidrule(lr){4-4} \cmidrule(l){5-5} 
25 & tr. fr. scratch                  & 56.5+-2.0                        & 43.3+-1.9                            & 56.5+-2.0                             \\
50 & tr. fr. scratch                  & 70.7+-0.4                        & 57.3+-0.6                            & 70.7+-0.4                             \\
60 & tr. fr. scratch                  & 74.2+-0.3                        & 63.9+-0.5                            & 74.2+-0.3                         \\
\bottomrule
\end{tabularx}
\label{tab:scalable_hyper_reps:generative_resnets_zoos_transfer}
\end{table}

\begin{table}[h]
\vspace{-4mm}
\centering
\captionof{table}[Few-shot Model Generation to Larger ResNets]{
Few-shot model generation for a new task: Sampling ResNet-34 models for CIFAR-100. \ourmethod was pretrained on CIFAR-100 ResNet-18s, 5 samples are drawn using subsampling. To get prompt examples, we train 3 ResNet-34 models on CIFAR-100 for 2 epochs to a mean accuracy of 26 \%. }
\label{tab:scalable_hyper_reps:generative_resnets_zoos_fewshot_resnet34}
\small
\setlength{\tabcolsep}{12pt}
\begin{center}
\begin{tabularx}{0.65\linewidth}{clcc}
\toprule
\multicolumn{4}{c}{CIFAR100   ResNet-18 to ResNet-34}                 \\
\midrule
  Epoch   & \multicolumn{1}{c}{Method}  & 5 Epochs           & 15 Epochs          \\
\cmidrule(r){1-1}  \cmidrule(l){2-2} \cmidrule(l){3-4} 
0  & tr. fr. Scratch           & 1.0$\pm$0.1           & 1.0$\pm$0.1           \\
         & \ourmethod & \textbf{1.6$\pm$0.3}  & \textbf{1.6$\pm$0.3}  \\
\cmidrule(r){1-1}  \cmidrule(l){2-2} \cmidrule(l){3-4} 
 1  & tr. fr. Scratch           & 12.4$\pm$1.0          & 12.9$\pm$0.8          \\
         & \ourmethod & \textbf{16.8$\pm$0.7} & \textbf{23.1$\pm$0.3} \\
\cmidrule(r){1-1}  \cmidrule(l){2-2} \cmidrule(l){3-4} 
5  & tr. fr. Scratch           & 49.5$\pm$0.6          & 36.2$\pm$1.7          \\
         & \ourmethod & \textbf{51.9$\pm$0.6} & \textbf{37.8$\pm$1.4} \\
\cmidrule(r){1-1}  \cmidrule(l){2-2} \cmidrule(l){3-4} 
15 & tr. fr. scratch      &      & 68.8$\pm$0.4          \\
         & \ourmethod &                    & \textbf{69.3$\pm$0.3} \\
\cmidrule(r){1-1}  \cmidrule(l){2-2} \cmidrule(l){3-4} 
 & \ourmethod Ens.                & 53.5               & 71.3         \\
\bottomrule
\end{tabularx}
\end{center}
\vspace{-4mm}
\end{table} 
\begin{table}[h]
\vspace{-4mm}
\small
\captionof{table}[Few-shot Model Generation to Larger ResNets and New Task]{
Few-shot model generation for a new task and architecture: \ourmethod trained on CIFAR-100 ResNet-18s used to generate ResNet-34s for Tiny-Imagenet. 5 samples are drawn using subsampling. To get prompt examples, we train 3 ResNet-34 models on Tiny-Imagenet for 2 epochs to a mean accuracy of 28.5 \%. 
}
\label{tab:scalable_hyper_reps:generative_resnets_zoos_fewshot_ti_resnet34}
\centering
\begin{center}
\setlength{\tabcolsep}{12pt}
\begin{tabularx}{0.65\linewidth}{clcc}
\toprule
\multicolumn{4}{c}{ResNet-18   CIFAR100 to ResNet-34 Tiny-Imagenet}                           \\
\cmidrule(r){1-1}  \cmidrule(l){2-2} \cmidrule(l){3-4} 
Epoch               & Method                    & 5 epochs           & 15 epochs          \\
\cmidrule(r){1-1}  \cmidrule(l){2-2} \cmidrule(l){3-4} 
\multirow{2}{*}{0}  & tr. fr. Scratch           & 0.5$\pm$0.0           & 0.5$\pm$0.0           \\
                    & \ourmethod & 0.5$\pm$0.1           & 0.6$\pm$0.2           \\
\cmidrule(r){1-1}  \cmidrule(l){2-2} \cmidrule(l){3-4} 
\multirow{2}{*}{1}  & tr. fr. Scratch           & 10.5$\pm$1.4          & 11.9$\pm$1.9          \\
                    & \ourmethod & \textbf{13.3$\pm$0.5} & \textbf{18.5$\pm$0.7} \\
\cmidrule(r){1-1}  \cmidrule(l){2-2} \cmidrule(l){3-4} 
\multirow{2}{*}{5}  & tr. fr. Scratch           & 47.2$\pm$0.7          & 31.1$\pm$1.7          \\
                    & \ourmethod & \textbf{50.6$\pm$0.3} & \textbf{31.6$\pm$0.6} \\
\cmidrule(r){1-1}  \cmidrule(l){2-2} \cmidrule(l){3-4} 
\multirow{2}{*}{15} & \multicolumn{2}{l}{tr. fr. Scratch}            & 61.9$\pm$0.3          \\
                    & \ourmethod &                    & \textbf{62.7$\pm$0.3} \\
\cmidrule(r){1-1}  \cmidrule(l){2-2} \cmidrule(l){3-4} 
            & \ourmethod Ens.                           & 52                 & 65.1              
\\
\bottomrule
\end{tabularx}
\end{center}
\vspace{-4mm}
\end{table}

\subsection*{Diversity of sampled models}
\label{sec:scalable_hyper_reps:sampling_diversity}
An interesting question is whether sampling \ourmethod generates versions of the same model. To test that,  we evaluate the diversity of samples generated with only a few few-shot examples by combining the models to ensembles. 
The improvements of the ensembles over the individual models demonstrate their diversity.
This indicates that given very few, early-stage prompt examples, sampling hyper-representations improves learning speed and performance in otherwise equal settings. 
Additionally, we conducted experiments with varying numbers of prompt examples, revealing that increasing the number of prompt examples enhances both performance and diversity. 
Nonetheless, even a single prompt example trained for just 2 epochs contains sufficient information to generate model samples that surpass those derived from random initialization; see Table \ref{tab:scalable_hyper_reps:generative_resnets_zoos_fewshot_resnet34_one_prompt_example}.

\begin{table}[h]
\centering
\captionof{table}[One-shot Model Generation to New Model and Task]{
Sampling ResNet-34 models for CIFAR-100. \ourmethod was pretrained on CIFAR-100 ResNet-18s, 5 samples are drawn using subsampling. To get prompt examples, we train a single ResNet-34 model on CIFAR-100 for 2 epochs to an accuracy of 26 \%. 
}
\label{tab:scalable_hyper_reps:generative_resnets_zoos_fewshot_resnet34_one_prompt_example}
\small
\begin{center}
\setlength{\tabcolsep}{7pt}
\begin{tabularx}{0.7\linewidth}{clcc}
\toprule
\multicolumn{4}{c}{CIFAR100   ResNet-18 to ResNet-34}                 \\
\cmidrule(r){1-1}  \cmidrule(l){2-2} \cmidrule(l){3-4} 
  Epoch   & \multicolumn{1}{c}{Method}  & 5 Epochs           & 15 Epochs          \\
\cmidrule(r){1-1}  \cmidrule(l){2-2} \cmidrule(l){3-4} 
0  & tr. fr. Scratch            & 1.0+-0.1           & 1.0+-0.1           \\
         & \ourmethod           & \textbf{1.5+-0.2}  & \textbf{1.6+-0.1}  \\
\cmidrule(r){1-1}  \cmidrule(l){2-2} \cmidrule(l){3-4} 
 1  & tr. fr. Scratch           & 12.4+-1.0          & 12.9+-0.8          \\
         & \ourmethod           & \textbf{16.9+-0.7} & \textbf{19.4+-0.2} \\
\cmidrule(r){1-1}  \cmidrule(l){2-2} \cmidrule(l){3-4} 
5  & tr. fr. Scratch            & 49.5+-0.6          & 36.2+-1.7          \\
         & \ourmethod           & \textbf{51.5+-0.3} & \textbf{38.6+-1.6} \\
\cmidrule(r){1-1}  \cmidrule(l){2-2} \cmidrule(l){3-4} 
15 & tr. fr. scratch            &      &             68.8+-0.4          \\
         & \ourmethod &                              & \textbf{69.1+-0.1} \\
\cmidrule(r){1-1}  \cmidrule(l){2-2} \cmidrule(l){3-4} 
Ensemble & \ourmethod                & 51.8          & 70.2              
\\
\bottomrule
\end{tabularx}
\end{center}
\end{table} 

\end{subappendices}
\newpage
\FloatBarrier





\newpage
\renewcommand{\rightmark}{\oldrightmark}
\renewcommand{\leftmark}{\oldleftmark}

\let\clearpage\relax
\bibliographystyle{plainnat}
{\footnotesize \footnotesize
\bibliography{./bib_auto.bib,./bib_manual.bib}

\pagebreak
\addtocontents{toc}{\vspace{2cm}} 
\renewcommand{\thechapter}{\Alph{chapter}}%
\appendix 



\FloatBarrier
\newpage
\newpage
\chapter*{Applied AI Software}

Throughout the preparation of this thesis, the \textit{Generative Pretrained Transformer (GPT)} 4 models by OpenAI~\citep{openaiChatGPT2024}, specifically version `gpt-4-1106-preview' with 128k tokens context and training data up to April 2023~\citep{openaiChatGPTModelVersions2024}, served as an auxiliary tool for spell checking, and reviewing language and grammar. Additionally, the \textit{Grammarly} software~\citep{grammarlyinc.GrammarlyFreeAI2024} was used to assist in the proofreading and refinement of the thesis. The inclusion of \textit{GPT-4} and \textit{Grammarly} was used for editorial purposes only, to improve the presentation and clarity of the manuscript.

\newpage
\includepdf[pages=-, pagecommand={\thispagestyle{emptyempty}}]{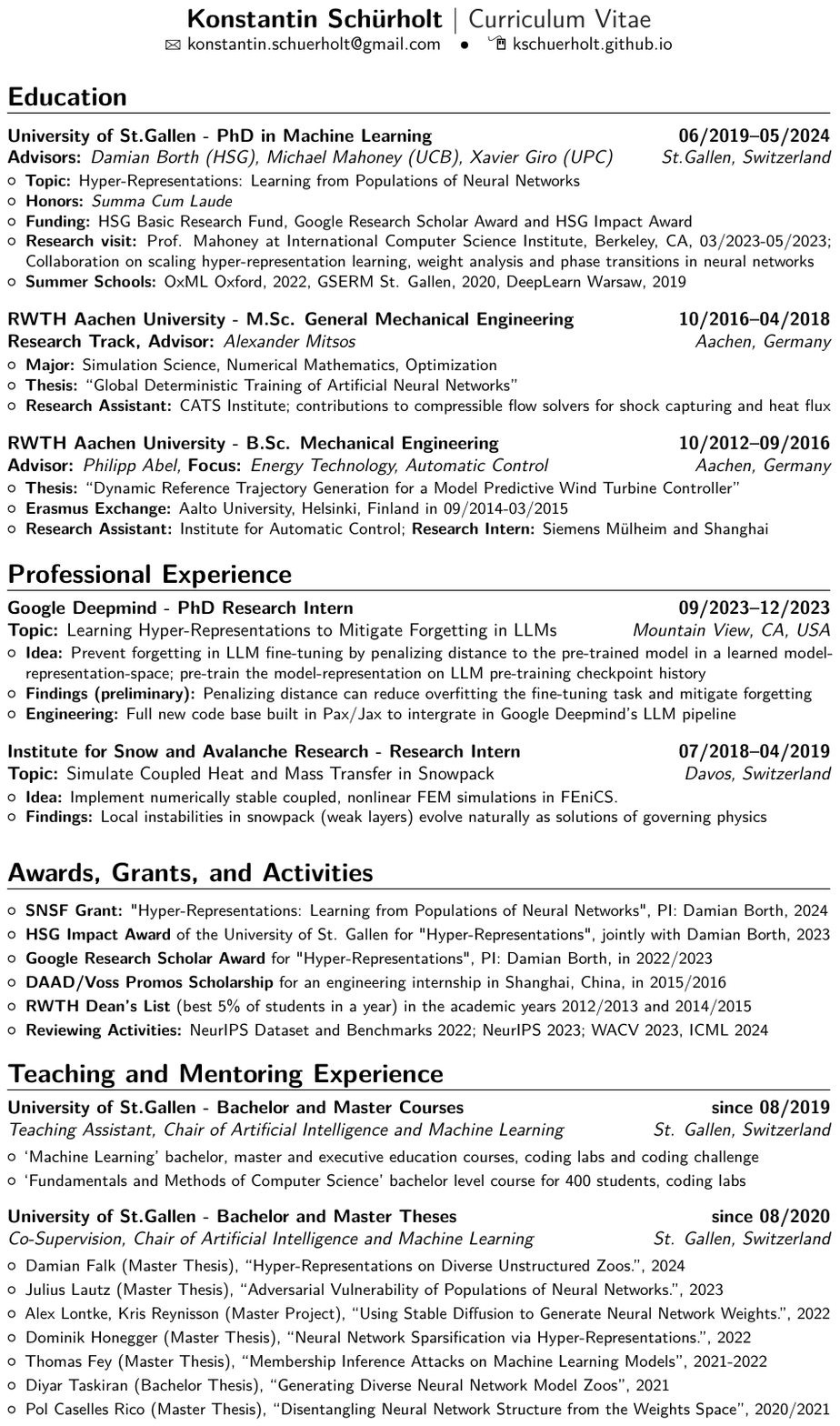}

\newpage
\thispagestyle{emptyempty}
\mbox{}
\newpage

\end{document}